%% file: main.tex
\documentclass{article}

\usepackage{microtype}
\usepackage{graphicx}
\usepackage{subfig} 
\usepackage{booktabs} 
\usepackage{multirow} 

\usepackage{hyperref}

\usepackage[accepted]{icml2025}

\usepackage{amsmath}
\usepackage{amssymb}
\usepackage{mathtools}
\usepackage{amsthm}

\usepackage[capitalize,noabbrev]{cleveref}

\theoremstyle{plain}

\theoremstyle{definition}

\theoremstyle{remark}

\usepackage[textsize=tiny]{todonotes}

\icmltitlerunning{Model Collapse and Spurious Shift Performance Prediction from Training on Uncurated Text Embeddings}

\begin{document}

\twocolumn[
\icmltitle{Data Curation Matters: Model Collapse and Spurious Shift Performance \\ Prediction from Training on Uncurated Text Embeddings}



\icmlsetsymbol{equal}{*}

\begin{icmlauthorlist}
\icmlauthor{Lucas Mattioli}{equal,yyy}
\icmlauthor{Youness Ait-Hadichou}{equal,yyy}
\icmlauthor{Sabrina Chaouche}{yyy}
\icmlauthor{Martin Gonzalez}{yyy}
\end{icmlauthorlist}

\icmlaffiliation{yyy}{IRT SystemX, France}

\icmlcorrespondingauthor{Martin Gonzalez}{martin.gonzalez@irt-systemx.fr}

\icmlkeywords{Machine Learning, ICML}

\vskip 0.3in
]



\printAffiliationsAndNotice{\icmlEqualContribution} 

\begin{abstract}
Training models on uncurated Text Embeddings (TEs) derived from raw tabular data can lead to a severe failure mode known as \underline{model collapse}, where predictions converge to a single class regardless of input. By comparing models trained with identical hyper-parameter configurations on both raw tabular data and their TE-derived counterparts, we find that collapse is a consistent failure mode in the latter setting. We introduce a set of metrics that capture the extent of model collapse, offering a new perspective on TE quality as a proxy for data curation. Our results reveal that TE alone does not effectively function as a curation layer—and that their quality significantly influences downstream learning. More insidiously, we observe that the presence of model collapse can yield artificially inflated and \underline{spurious Accuracy-on-the-Line correlation}. These findings highlight the need for more nuanced curation and evaluation of embedding-based representations, particularly in out-of-distribution settings.
\end{abstract}

\input{1-introduction}

\input{2-background}
\input{3-method}

\input{4-experiments.tex}

\input{9-conclusion}

\bibliography{biblio}
\bibliographystyle{icml2025}

\newpage
\appendix
\onecolumn

\input{a1-exp-details}

\input{a2-suppl-figs}


\end{document}

%% file: 1-introduction.tex
\section{Introduction} 
\label{sec:intro}

\begin{figure}[t!]
\centering
\includegraphics[width=\columnwidth]{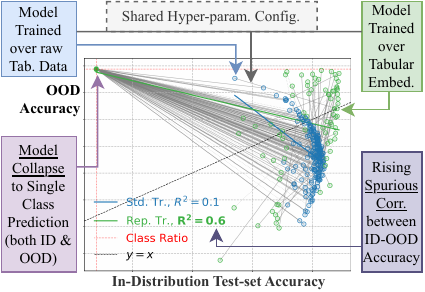}
\caption{Illustration of our results. We train models with shared hyper-parameter configuration on raw tabular data and their LLM-based embedded counterparts. We exhibit model collapse to single class prediction both in ID \& OOD test-sets and observe a rise of spurious correlation between ID-OOD accuracy (i.e. Accuracy-on-the-Line), as a consequence of model collapse severity.}
\label{fig:global-approach}
\end{figure}

Tabular data is a central modality to critical domains like finance, healthcare, and the social sciences \cite{sirignano2019universal, johnson2016mimic, athey2017beyond, shwartz2022tabular}, encoding structured, heterogeneous information vital for decision-making. Despite the rise of deep learning across other modalities, gradient-boosted decision trees (GBDTs) \cite{friedman2001greedy, friedman2002stochastic, chen2016xgboost, ke2017lightgbm} remain state-of-the-art for tabular data, both in prediction accuracy \cite{arik2021tabnet, huang2020tabtransformer, kadra2021well, katzir2020net}, and robustness under distribution shifts \cite{gardner2022subgroup, liu2023need, gorishniy2021revisiting, shwartz2022tabular}. However, they face limitations in low-data regimes and struggle with high-dimensional or unstructured inputs like text \cite{ke2017lightgbm}, often requiring external embeddings. They also perform poorly under conditional distribution shifts of $P(Y|X)$, which we denote $Y|X$-shifts \cite{gardner2023tableshift}.

To address these issues, large language models (LLMs) and their text embeddings (TEs) have emerged as a complementary modality. By converting mixed-type tabular fields into dense representations, TEs act as a form of data curation—organizing and filtering inputs to create a more model consumable format. Recent work has even shown some positive results of this approach for out-of-distribution (OOD) robustness against $Y|X$-shifts.

We investigate the use of text embeddings as a form of data curation in tabular classification. A central theme of our study concerns Accuracy-on-the-Line (ACL): a striking phenomenon in ML where in-distribution (ID) accuracy positively correlates with OOD accuracy across varying model configurations.

Our primary contribution is the identification of a critical and previously overlooked failure mode: a new kind of \underline{model collapse}, where models trained on uncurated text embeddings (TEs), degenerate their predictions to a single class regardless of inputs. This not only compromises performance but can also \underline{spuriously inflate the ACL} correlation, naively giving a false signal of OOD performance prediction. We introduce a set of dedicated metrics to quantify collapse severity and evaluate on 4615 model configurations, tested on numerous domain-shift scenarios, picked to exemplify our findings rather than to propose exhaustive numerical experiments. In doing so, we also uncover notable inconsistencies between \textsc{MTEB} leaderboard rankings \cite{muennighoff2023mteb} and performance in tabular classification tasks—highlighting the limitations of benchmark-driven model selection. Shifting from traditional tabular training to LLM-based embedding pipelines introduces domain-specific challenges in representation and alignment, which our work addresses through generalizable metrics for evaluating embedding quality and model robustness. 
Moreover, these findings emphasize the need for human oversight on fully automated pipelines, as relying on metrics like $R^2$ can fail silently. Without human intervention, uncurated TE may inject bias in a worst-case fashion which further highlights the ethical dimensions of representation learning, where embedding quality, fairness, and robustness intersect.

%% file: 2-background.tex
\section{Background and Related Work}
\label{sec:background}

\paragraph{Robust OOD generalization.} Generalizing under OOD conditions remains a central challenge in ML. Benchmarks — especially in vision often focus on covariate shift ($X$-shifts), offering a narrow view of robustness ~\cite{HendrycksDi19, TaoriDaShCaReSc20}. Yet, in many real-world scenarios, particularly with tabular data, $Y|X$-shifts are far more prevalent and problematic, arising from missing variables, hidden confounders, or evolving social-economic contexts \cite{zeng2024llm, gardner2023tableshift, alvarezmelis2020robustness}.

Methods to address distributional shifts include distributionally robust optimization (DRO) \cite{duchi2021statistics}, balancing techniques, and invariant learning. DRO constructs worst-case distributions within uncertainty sets using metrics like $\chi^2$ divergence \cite{duchi2021statistics}, KL divergence \cite{hu2013kullback}, and Wasserstein distance  \cite{blanchet2019data, blanchet2023unifying}. However, these approaches often underperform on real-world tabular data \cite{liu2023need}. Invariant learning \cite{arjovsky2019invariant, koyama2020invariance, peters2016causal} assumes access to multiple training environments, which limits practical applicability. While some theoretical results exist for linear transfer models \cite{li2022transfer, tian2023transfer}, they rarely extend to complex models like GBDTs or deep networks. Additionally, most domain adaptation studies \cite{iwasawa2021test, chen2023improved, liang2023ttasurvey} focus on covariate $X$-shifts in images, whereas we address conditional $Y|X$-shifts in tabular contexts.

Inference-time adaptation methods like TENT \cite{wang2021tent}, struggle in tabular settings, while recent approaches like FTAT are specifically designed to handle both $X$ and $Y|X$-shifts effectively \cite{zhou2025fully}.
New metrics such as DISDE help decompose performance drops into $X$ and $Y|X$-shift contributions  \cite{CaiNaYa23}, while Accuracy-on-the-Line (ACL) \cite{MillerTaRaSaKoShLiCaSc21} measures how ID vs. OOD performances correlate, and Agreement-on-the-Line \cite{baek2022agreement}, an analog phenomenon of ACL which, if valid, allows to induce ACL without the need of labeled test-set data.

\paragraph{LLM Embeddings.} LLMs have recently been applied to tabular classification in two main ways. One strategy treats classification as a text generation task using decoder-based models with prompts like “Is this patient diabetic? Yes/No,” though this approach is sensitive to prompt design and yields variable results \cite{hegselmann2023tabllm, slack2023tablet}. Recent work \cite{zeng2024llm} shows that LLM embeddings derived from tabular data can improve OOD generalization. However, their setting is established using embeddings from a single LLM, leaving open the question of whether such generalization holds across model families and LLM variants.
Several recent LLM variants aim to enhance  embeddings quality through fine-tuning or architectural enhancements: \texttt{Linq-Embed-Mistral} \cite{choi2024linq}, is fine-tuned specifically to focus on hard negatives, \texttt{SFR-Embedding-Mistral} \cite{meng2024sfr} leverages transfer learning to improve generalization across domains, \texttt{e5-Mistral-7B-Instruct} \cite{wang2024e5}  serves as a general-purpose model,  while \texttt{Zeta-Alpha-E5-Mistral} \cite{camara2024zeta} uses LoRA adapters for efficiency under resource-constrained conditions. While these approaches diversify embedding strategies, their robustness under distribution shift in tabular settings is still poorly understood.

\paragraph{Benchmark-Based Assumptions.} Benchmarks like \textsc{MTEB} \cite{muennighoff2023mteb} have become standard tools for assessing LLM performance across NLP tasks, offering scalable comparisons in fast-evolving model ecosystems. However, a growing reliance on rankings only raises concerns about their performance on domain-specific applications—such as tabular data under distribution shift. Recent work has highlighted that spurious correlation benchmarks often disagree in their evaluations, underscoring the need to revisit benchmark assumptions and develop task-sensitive evaluation protocols \cite{bell2024reassessing}.

\paragraph{Model collapse of Embedding-Based Models.} 
Recent studies have uncovered diverse failure modes across ML settings. \cite{herman2024language} show that indiscriminate use of LLM-generated content during training leads to irreversible model collapse, where tails of the original data distribution disappear. Modality collapse in multimodal models occurs when noisy features from one modality overshadow useful signals from others; the authors propose Explicit Basis Reallocation (EBR) to mitigate this by disentangling shared representations \cite{chaudhuri2025ebr}. Strong collapse from synthetic data emerges when models overfit even small amounts of synthetic inputs, degrading generalization—particularly in large models—despite increased training size \cite{dohmatob2024strong}. Finally, non-iterative model collapse in LLMs arises from single-phase exposure to synthetic data, narrowing coverage and inflating frequent patterns; the authors propose ToEdit, a token-level data editing method, to alleviate this \cite{zhu2024toedit}. 

%% file: 3-method.tex
\section{Evaluation Methodology}
\label{sec:method}

To define and evaluate model collapse, we build on two key ideas: using LLMs to generate embeddings as a data curation layer for downstream models, and analyzing ID vs. OOD performance across model families to uncover hidden patterns. This section first outlines the two underlying procedures, then introduces our notion of model collapse along with metrics to categorize and quantify its severity.

\subsection{Text Embeddings as Data Curation}\label{sec:te}

Our method begins by collecting tabular rows from both a source and a target
domain. These rows are converted into natural language through a serialization
process (Tab2Text), which reformulates each row into a sentence, while a
task-specific instruction is appended. This serialized input is then passed to
a pre-trained LLM that serves as an encoder. It will generate vector
representations, called embeddings. More specifically, let \texttt{LLM} be
an LLM encoder (such as \texttt{e5-Mistral-7b-Instruct}) and let $S$ be a
tabular dataset The overall procedure for transforming $S$ into a set, denoted
\texttt{LLM}$|S$, of LLM embeddings of $S$ based on \texttt{LLM} has been
described in detail in {\cite{zeng2024llm}}. In short, it consists on
instructing an LLM to render plain text $T_n$ out of the information of the
$n$th row $s_n$ of the table $S$, incorporating a task specific prompt $I$ and
use both $(T_n, I)$ as input for \texttt{LLM} to produce a vector encoding
$\Phi (s_n)$ of $s_n$. When repeating this procedure for the whole dataset $S$
we obtain a vector set of LLM embeddings that we will denote
\texttt{LLM}$|S$.
We choose LLMs with same embedding vector dimension (4096) to rule out
possible biases or impacts which might directly be associated to vector
dimensionality and we do not make any changes to the chosen off-the-shelf
LLMs. In other words, all LLMs are used as a form of automated data curation
strategy.

\subsection{On-the-line Analysis \& Model Collapse}\label{sec:otl}

According to the data-centric approach of robustness, well-curated data
representation should exhibit a certain degree of resilience to distribution
shifts. Although the usual way of evaluating robustness is simply by testing
against a specific dataset, a less known line of work is to analyze the
behavior of models both on In-Distribution (ID) and OOD test sets as a way to
understand the shifted data in question. In this second line of work, many
examples of distribution shifts arising from Computer Vision have exhibited a
specific linear correlation pattern arises for such pairs of test sets, a
phenomenon coined as Accuracy-on-the-line (ACL), and which, if
present, can serve to predict the OOD performance of a model if one knows its
ID performance and the parameters of the linear fit arising from ACL.
Determining validity and strength of ACL is usually deduced by the value of
the $R^2$ coefficient of the linear fit.
Our method pics a set of training Hyper-Parameter Configurations (HPCs) for different models, which are. trained on a source domain of tabular
data. Let $f_c$ be a model trained with HPC $c$ and
let $S$ be a test-set. We denote $\mathcal{A}^S (f_c)$ for the prediction
accuracy of $f_c$ on $S$. Given ID and OOD test sets $S$ and $Q$, and a
HPC set $\mathrm{HP}$ of cardinality $N$, we plot all
$N$ models $f_c$ in what we call the \underline{ACL-plane}, for which each
model $f_c$ has coordinates $(\mathcal{A}^S (f_c), \mathcal{A}^Q (f_c))$.
Recognizing that shifts often co-occur with class imbalance, and that accuracy
may be insufficiently sensitive to such cases, we call \underline{F1L-plane}
the plane where where each $f_c$ has coordinates its corresponding macro F1
scores over $S$ and $Q$ respectively.

\subsection{Binding Models across Modalities}\label{sec:bind}

Once we have a specified LLM encoder \texttt{LLM}, we bind model couples
($f_{c, 1}, f_{c, 2}$) with shared hyper-parameter configuration where $f_{c,
1}$ has been trained over a set $D_{\mathrm{train}}$ of raw tabular data and
$f_{c, 2}$ was trained over the curated data $\text{\texttt{LLM}} |
D_{\mathrm{train}}$ by means of the procedure described in Subsection
\ref{sec:te}. Then, given two tabular test sets $S$ and $Q$, and we plot the
coordinates of for each model $f_{c, 1}$ both in the \underline{ACL-plane} and
the F1L-plane with respect to $S$ and $Q$ according to the procedure described
in Subsection \ref{sec:otl}. Next, \textit{we use the same LLM to encode
both $S$ and $Q$} and evaluate $f_{c, 2}$ accordingly. Namely, for $f_{c, 2}$,
we plot a point with coordinates $\left( \mathcal{A}^{\text{\texttt{LLM}}
| S} (f_{c, 2}), \mathcal{A}^{\text{\texttt{LLM}} | Q} (f_{c, 2}) \right)$
in the \underline{ACL-plane}. Now that both $f_{c, 1}$ and $f_{c, 2}$ have
their respective coordinates in the \underline{ACL-plane} and the
\underline{F1L-plane}, we bind the points corresponding to $f_{c, 1}$ and
$f_{c, 2}$ by means of a straight line attaching them.
We highlight the fact that our method is primarily focused on examining the
patterns of the bindings of all pairs of models with shared hyper-parameter
configuration, and incidentally then focused on the particular patterns of the
set of models trained on each of the two modalities (tabular and embeddings).
Nevertheless, one should not be overlooked the fact that the coordinates of
$f_{c, 1}$ and $f_{c, 2}$ are obtained from $S$ and $Q$ and from
$\text{\texttt{LLM}} | S$ and $\text{\texttt{LLM}} | Q$ respectively.
Rigorously, the straight line representing the bind between $f_{c, 1}$ and
$f_{c, 2}$ is actually a line binding two different points of two different
planes that we have overlapped. As a consequence, we report the $R^2$
coefficient associated to each of these two sets of models separately.
Our overall method can be illustrated in Fig \ref{fig:global-approach}, where
we can already see what a model collapse looks like : numerous models trained
on the tabular modality are bound to what seems to be a ``single'' collapse
point un the upper-left part of the ACL-plane. Measuring the severity of model
collapses along different pairs of source and target domains, as well as their
different kinds, brings us to propose several model collapse metrics.

\subsection{Model collapse metrics on binary classification}

In this subsection, we define model collapse metrics that we will use to
distinguish between different kinds of collapses as well as to measure their
severity. Our approach consists on considering, for a given model $f$ (such as
MLP, LR, GBM, XGB, etc.), a set $\mathrm{HP}_f$ of training Hyper-Parameter
combinations for which models trained on raw tabular datasets provide no
single-class predictors, and then ask how many of those incur into
single-class predictors when trained under same HP configurations for over LLM
embeddings.

Let $f_c$ be a model trained with hyper-parameter configuration $c$ and let
$S$ be a test-set. We denote $\mathrm{PP}^S (f_c)$ (resp. $\mathrm{PN}^S
(f_c)$) for the number of predicted positives (resp. predicted negatives) of
$f_c$ on $S$. We say that the model $f_c$ is collapsed over $S$ if
$\mathrm{PN}^S (f_c) = 0$ or $\mathrm{PP}^S (f_c) = 0$.

Let $f$ be a method, $D$ be a domain with training (resp. test) tabular data
$D_{\mathrm{train}}$ (resp. $D_{\mathrm{test}}$) and a test-set $S$. We will say
that a hyper-parameter set $\mathrm{HP}_f$ is \textit{well-defined} w.r.t.
$S$, and we denote it $\mathrm{HP}_{f, S}$, if for all hyper-parameter
combinations $c \in \mathrm{HP}_{f, S}$, the trained model $f_c$ on
$D_{\mathrm{train}}$ does not collapse over $D_{\mathrm{test}}$ and $S$. For simplicity, we denote it $\mathrm{HP}_f$ in the special case when $S=D_{\mathrm{test}}$.

Let \texttt{LLM} be an LLM Encoder. For a model $f_{c, D}$ trained over a
tabular dataset $D_{\mathrm{train}}$, we denote $\text{\texttt{LLM}} | f_{c,
D}$ for the model $f$ trained with same HP configuration $c$ but over the
associated embeddings $\text{\texttt{LLM}} | D_{\mathrm{train}}$ obtained
according to procedure \ref{sec:te}. While the context is clear, we will simply denote it $f_{c}$.

Given a well-defined set $\mathrm{HP}_f$, we define metrics to measure model
collapses arising from switching from standard tabular training on
$D_{\mathrm{train}}$ to representation training over LLM embeddings
$\text{\texttt{LLM}} | D_{\mathrm{train}}$. In other words, we present
metrics that will allow to measure how the set
$\mathrm{HP}_{\text{\texttt{LLM}} | f}$ is far from being well-defined
over $\text{\texttt{LLM}} | S$. The rationale is that well-curated
training data do not incur into single-class predictors on binary
classification tasks, so that the ratio of combinations in
$\mathrm{HP}_{\text{\texttt{LLM}} | f}$ incurring into model collapse over
different test-sets $\text{\texttt{LLM}} | S$, is an informative indicator
on the quality of the text embeddings in their function as a data curation
layer.

In what follows, we fix a method $f$, we simply denote $\mathrm{HP}$ for a
well-defined set $\mathrm{HP}_{f, D}$ as long as the context is clear.
Let $S$ be a test-set. We define the Collapse Ratio of $\mathrm{HP}$ over $S$
as the ratio of hyper-parameter combinations resulting in models collapsed
into single class prediction on $S$ as
\[ \mathrm{CR}^S_{\mathrm{HP}} = \frac{1}{| \mathrm{HP} |}  \sum_c
   \mathbb{I}_{\{ \mathrm{Cond}^S_T (c)\}} +\mathbb{I}_{\{ \mathrm{Cond}^S_F
   (c)\}}, \]
where $\mathrm{Cond}^S_T (c) = (\mathrm{PN}^S (f_c) = 0)$ and
$\mathrm{Cond}^S_F (c) = (\mathrm{PP}^S (f_c) = 0)$. Notice that condition
$\mathrm{PN}^S (f_c) = 0$ implies $\mathcal{A}^S (f_c) = \mathrm{TR}^S$ and
condition $\mathrm{PP}^S (f_c) = 0$ implies $\mathcal{A}^S (f_c) =
\mathrm{FR}^S$, where
\[ \mathrm{FR}^S = \frac{| \mathrm{cl} (\mathrm{FALSE}) |}{\left| \text{S}
   \right|}, \qquad \mathrm{TR}^S = \frac{| \mathrm{cl} (\mathrm{TRUE})
   |}{\left| \text{S} \right|} \]
are the proportions of each of the two classes $\{ \mathrm{FALSE},
\mathrm{TRUE} \}$ in $S$. Also, as we will see in short, models not always
collapse to the same class so $\mathrm{CR}^S_{\mathrm{HP}}$ computes de ratio
of configurations inducing model collapse on either one of the two classes.
Let $S$ and $Q$ be two test sets. We define the pair-wise \textit{strong}
collapse ratio $\mathrm{CR}^{S, Q}_{s, \mathrm{HP}}$ of $\mathrm{HP}$ over $S$
and $Q$ as the ratio of hyper-parameter combinations resulting into model
collapse simultaneously for both test sets. For simplicity, in what follows we
assume that our procedure exhibits model collapses on the positive classes in
$S$ and $Q$, then $\mathrm{CR}^{S, Q}_{s, \mathrm{HP}}$ is formalized as
\[ \mathrm{CR}^{S, Q}_{s, \mathrm{HP}} = \frac{1}{| \mathrm{HP} |}  \sum_c
   \mathbb{I}_{\{ \mathrm{Cond}^S_T (c) \cap \mathrm{Cond}^Q_T (c)\}} . \]
The pair-wise projection ratio as the ratio of hyper-parameter combinations
resulting into model collapse in either one of the test-sets
\[ \mathrm{CR}^{S, Q}_{p, \mathrm{HP}} = \frac{1}{| \mathrm{HP} |}  \sum_c
   \mathbb{I}_{\{ \mathrm{Cond}^S_T (c) \cup \mathrm{Cond}^Q_T (c)\}} . \]
When the context is clear, we will denote this metrics simply $\mathrm{CR}_{s}$ and $\mathrm{CR}_{p}$.

Finally, we define a metric to quantify near-collapse phenomena. Experiments
show that there are HP configurations entailing models that are almost
collapsed but not completely, as they have a small fraction of false or true
negatives. On the one hand, the above metrics are too rigid to account such
near-collapses. On the other hand, such predicted negatives can only be
considered to be negligible relative to the size of their associated test-set
and the proportion of ground-truth negatives among the latter. In what
follows, all collapse ratios will include
near-collapses and we defer the formal definition of this more involved
metrics to Subsection \ref{app:col-metrics} in the appendix.

%% file: 4-experiments.tex
\section{Experiments}
\label{sec:exp}

We make experiments according to the following
questions:
\begin{enumerate}
  \item How prevalent is model collapse across different LLMs, 
  and what impact do they have on the validity and strength of
  the ACL phenomenon?
  
  \item For a fixed model family and source and target domains, how does model
  collapse vary in kind and intensity over the ACL \& F1L planes across
  different LLMs?
  
  \item Does valid ACL for models trained on tabular data imply valid ACL for
  their representation learning counterparts? Do current Test-Time
  Adaptation methods for domain shifts remedy any lack of ACL validity?
  
  \item Is there a correlation between LLM-based Text Embedding quality
  benchmarking as present in \textsc{MTEB}
  and their quality as data curation layers for real-world
  domain shifts?
  
  \item Are LLM embeddings transferable between LLMs? More specifically, how
  do models trained on embeddings from one LLM behave when tested on test sets
  encoded by other LLMs?
\end{enumerate}
To the best of our knowledge, this is the first evaluation on the model
collapse phenomenon as framed in this work. It consists of an initial
investigation of the concept and we prioritize qualitative insights as primary
findings, hoping to open the door to further benchmarking and systematic
analysis such as the one performed in \cite{zeng2024llm}. Nevertheless, we
include a re-implementation and extension of part of their experiments
as a reference point. Where appropriate, we include
results using the same LLM and MLP configuration they use,
to allow direct comparison with their findings.

\begin{figure*}[ht!]
\centering
\subfloat[Geometrical Pattern]{\includegraphics[width=0.33\textwidth]{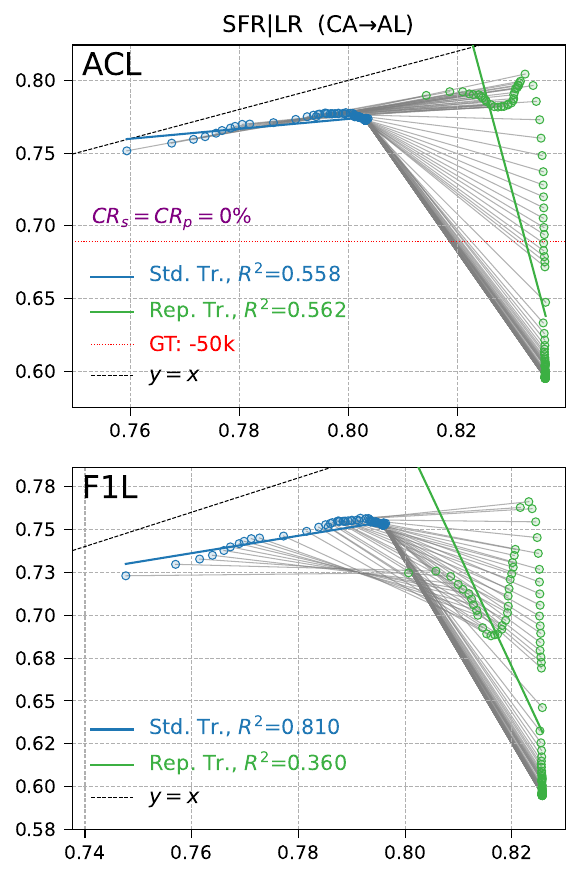}\label{fig:SFR-2-AL-3-main}}
\subfloat[Strong Collapse Pattern]{\includegraphics[width=0.33\textwidth]{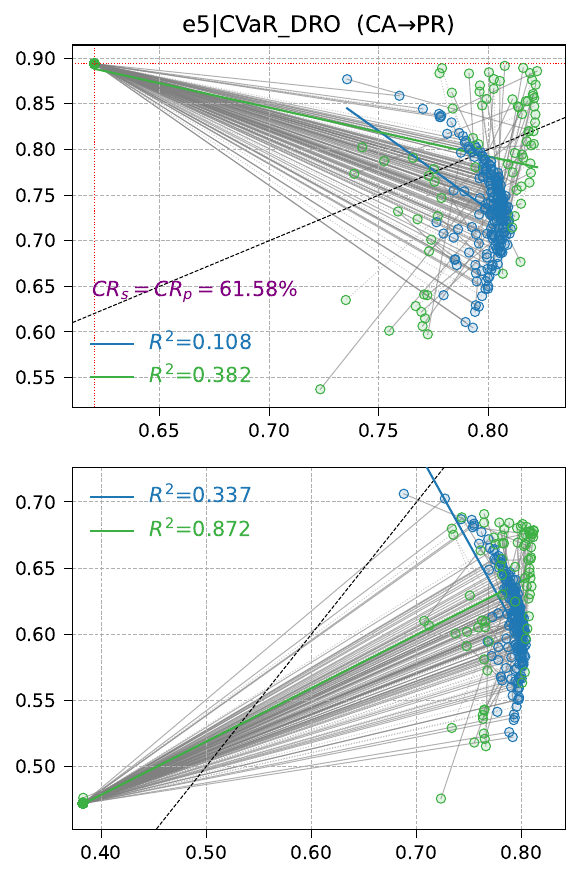}
\label{fig:pr-cvar-e5-main}}
\subfloat[Projection Colapse Pattern]{\includegraphics[width=0.33\textwidth]{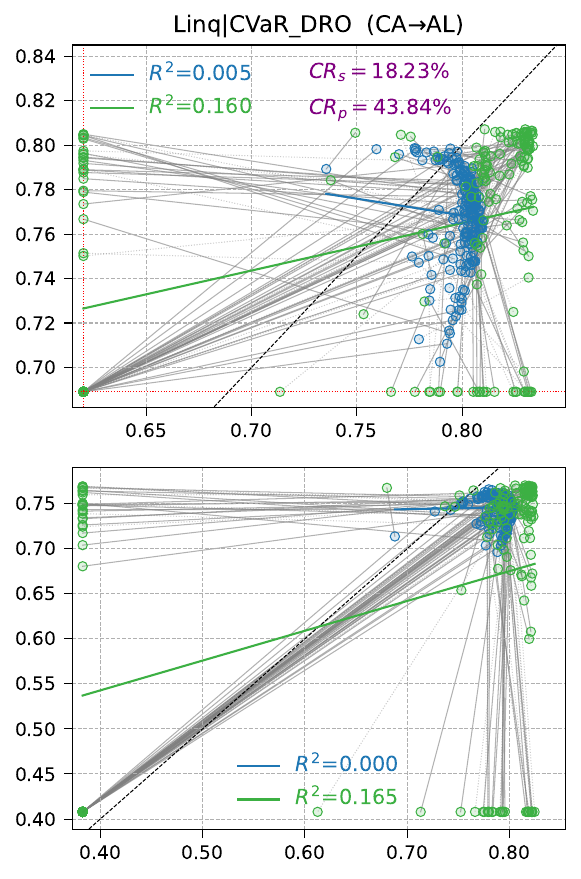}\label{fig:al-cvar-linq-main}}
\caption{Different patterns as exemplified by different combinations of LLMs, models and target states.}
\label{fig:patterns-main}
\end{figure*}

We focus on a binary classification task to predict whether US working adults'
yearly income exceeds \$50,000 in 2018, based on the \texttt{ACS Income}
dataset {\cite{DingHaMoSc21}}. We take into consideration six states: Alabama
(AL), Alaska (AK), Arizona (AZ), Arkansas (AR), California (CA), Puerto
Rico
(PR). As we will already see, this very restricted context already provides
many insightful answers to the above questions. In all our experiments, the
``$<$50k'' class is going to be the positive class.

Four LLMs are identified and chosen from the \textsc{MTEB} Leaderboard, which ranks LLMs based on the performance of the
text embeddings they generate on a variety of tasks. The selection accounted
for the LLM used in all experiments reported in \cite{zeng2024llm} and, for
the remaining LLMs, they were restricted only to models with a fixed embedding
dimensionality of 4096 to improve the rigor of our comparative analysis. Tables \ref{tab:model-configs} and \ref{tab:llm-ranking} show the number of
model configurations and \textsc{MTEB} ranks, as consulted on March 1, 2025. 
Notice that 3 LLMs were
chosen out of the top-10 best ranked models while we chose one LLM ranked way
below the 100th place.

\begin{table}[ht]
\centering
\caption{Name \& Ranking of LLMs according to \textsc{MTEB}.}\label{tab:llm-ranking}
\begin{tabular}{rlc}
Name & LLM Full Name & Rank \\
\midrule
\texttt{Linq} & \texttt{Linq-Embed-Mistral} & 2 \\
\texttt{SFR} & \texttt{SFR-Embedding-Mistral} & 5 \\
\texttt{e5} & \texttt{e5-Mistral-7B-Instruct} & 9 \\
\texttt{Zeta}& \texttt{Zeta-Alpha-E5-Mistral} & 177 \\
\end{tabular}
\end{table}

We organized our results into two sets of experiments: the first addressed the initial
three questions, while the second provided evidence relevant to the final two.

\subsection{Model Collapse arising from On-the-Line Analysis}\label{sec:mc}

\paragraph{Experiment setup.}We fix CA as our training domain for this set of
experiments and the remaining states as test domains. We select 7 families of
models with varying number of training hyper-parameter configurations as shown
in Table \ref{tab:model-configs} providing a total of 934 model
configurations.

{\setlength{\tabcolsep}{2pt}
\begin{table}[ht]
\centering
\caption{Number of configurations per model family.}\label{tab:model-configs}
\vspace{0.1cm}
\begin{tabular}{lccccccc}
\toprule
\textbf{Family} & MLP & CVaR-DRO & LR & XGB & GBM & RF & SVM \\
\midrule
\textbf{Configs} & 96 & 203 & 100 & 200 & 200 & 101 & 34 \\
\bottomrule
\end{tabular}
\end{table}}

The configurations match those reported in \cite{zeng2024llm}, with the
exception of LR, where we add additional configurations to obtain a total of
100 models and RF, which is new. All model
configurations are then trained on the tabular data $\mathrm{CA}$ and on
each of the four LLM embeddings of $\mathrm{CA}$. We highlight the fact that the sets $S$ and $\text{\texttt{LLM}} | S$ have different modalities, but have the same positive class ratio.

\paragraph{Model collapse is ubiquitous over all tested LLM Embeddings and
almost all model families.}We illustrate our findings directly by providing
the 140 pairs of ACL and F1L planes with plotted bindings among models with
shared HP configurations, for each combination of test domain, model family
and LLM encoder. Notice that in this setup, the LLM used to encode the
training set from which models are trained is the same LLM used to encode the
test sets. All figures are reported in Subsection \ref{app:figs} in the
Appendix. All tested LLMs induce model collapse and, while some combinations
of LLMs, model family and (source, target) states show no model collapse,
there is no model family not exhibiting model collapse for all tested states
for any tested LLM. Moreover, such collapses do not only happen on states with
strong $Y|X$-Shifts but are rather systematically appearing on all the tested
target states. Additionally, there are configurations inducing model collapse
simultaneously on all states. For instance, in Figure \ref{fig:pr-mlp} one
exhibits that, out of the 96 \texttt{e5}$| \mathrm{MLP}$ configs, there are
34 are collapsed on \texttt{e5}$| \mathrm{CA}$ and \texttt{e5}$|
\mathrm{PR}$ (these are the pair-wise strong collapses). It turns out that
there are 12, 33, and 31 configs that induce strong model collapse for
\texttt{Linq}, \texttt{SFR} and \texttt{Zeta} respectively, and
that there are 7 that induce strong collapses for the pair
$(\mathrm{CA}, \mathrm{PR})$ for all LLMs (see Table \ref{table:4} for details).

\paragraph{Model collapse varies in kind and severity across LLMs.}An
attentive look of all figures in Subsection \ref{app:figs} in the Appendix
reveals that there are several kinds of patterns arising from analyzing the
bindings in the ACL and F1L planes. We showcase 3 such distinguished patterns
that we report in Figure \ref{fig:patterns-main}. Detailed calculations of all
metrics presented here can be found in the appendix.

\begin{figure*}[ht!]
\centering
\subfloat[Valid ACL\&F1L]{\includegraphics[width=0.25\textwidth]{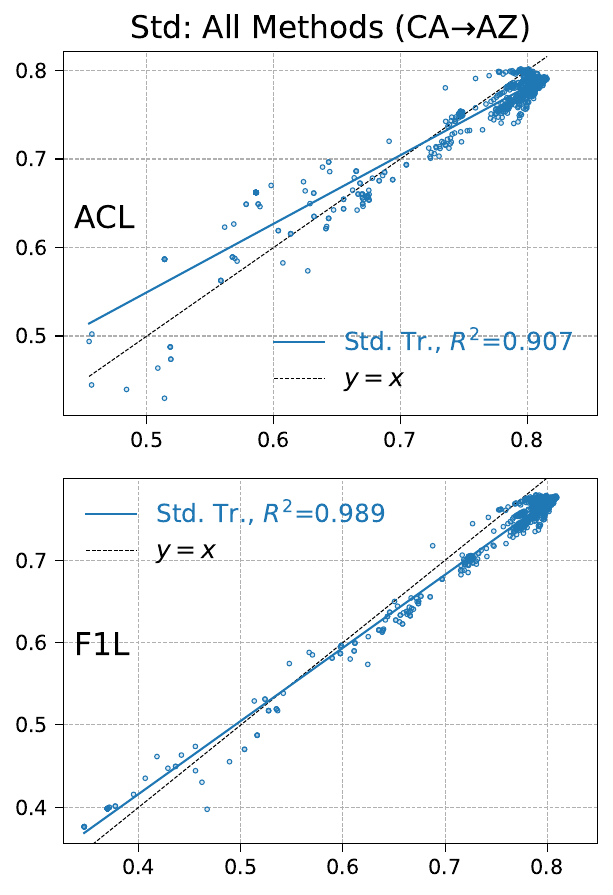}
\label{fig:acl-ok}}
\subfloat[Finetuning]{\includegraphics[width=0.25\textwidth]{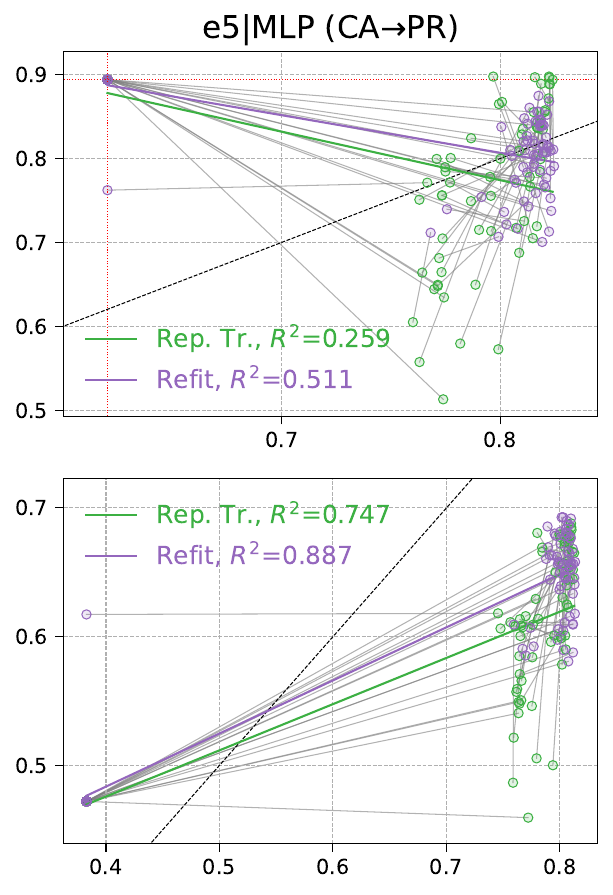}\label{fig:ft}}
\subfloat[\textsc{TENT}]{\includegraphics[width=0.25\textwidth]{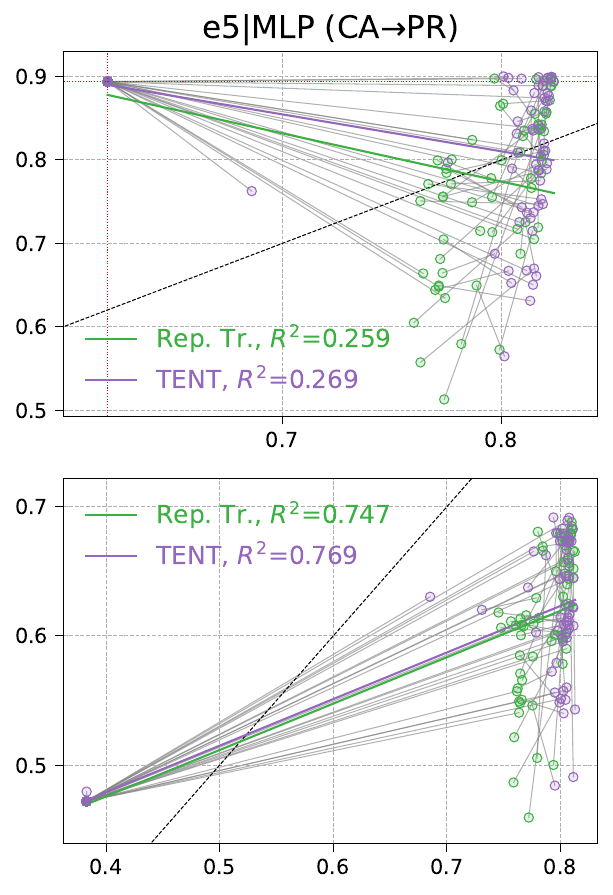}\label{fig:tent}}
\subfloat[\textsc{FtaT}]{\includegraphics[width=0.25\textwidth]{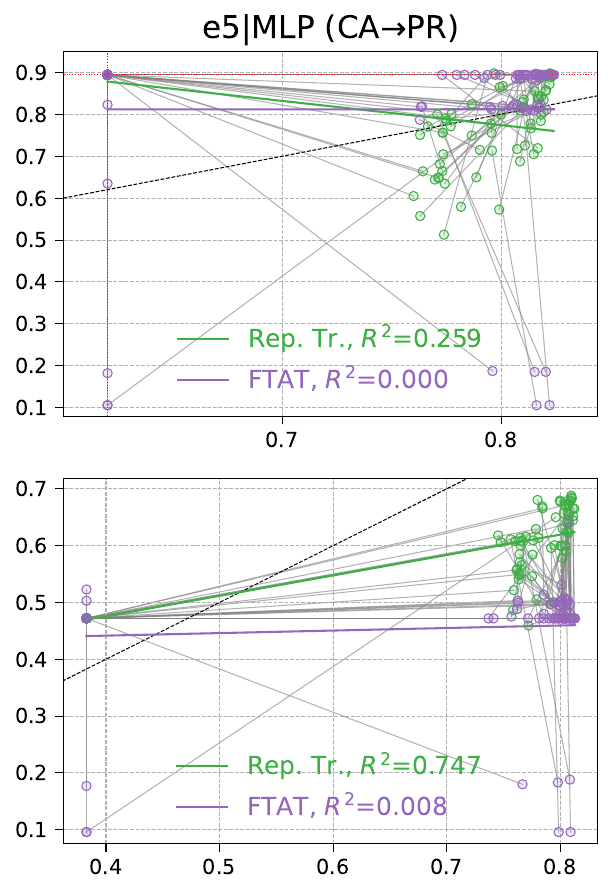}\label{fig:ftat}}
\caption{(a) An example of a valid ACL and F1L; (b,c,d) Test-Time adaptation methods fail to strengthen ACL or F1L.  }
\label{fig:patterns-main2}
\end{figure*}

First, Figure \ref{fig:SFR-2-AL-3-main} shows both the ACL and F1L planes for
the LR model family, with target domain AL and with base LLM encoder
\texttt{LLM=SFR}. Blue points in the ACL plane correspond
to models $f_{c, 1}$, for $c \in \mathrm{HP}_{\mathrm{LR}}$, with coordinates
$(\mathcal{A}^{\mathrm{CA}} (f_{c, 1}), \mathcal{A}^{\mathrm{AL}} (f_{c, 1}))$
while green points correspond to models $f_{c, 2}$ with coordinates $\left(
\mathcal{A}^{\text{\texttt{LLM}} | \mathrm{CA}} (f_{c, 2}),
\mathcal{A}^{\text{\texttt{LLM}} | \mathrm{AL}} (f_{c, 2}) \right)$.
Analyzing the bindings show that there is no model collapse in this scenario.
This fact is confirmed by computing the confusion matrix of each point, and in
particular showing that the green points close to the ratio of the class
``$<$50k'' (i.e. the positive class ratio) are no near collapses. More
importantly, the bindings between points seems to follow some logic of
geometric nature, a fact that is only observed for the LR family but that is
present in all pairs of tested states and all 4 LLMs.

Second, Figure \ref{fig:pr-cvar-e5-main} exhibits the frequent kind of model
collapse, where a high number of blue points are bound to multiple overlapping
green points exactly at the intersection of the positive class ratio of both
the $\text{\texttt{LLM}} | \mathrm{CA}$ and $\text{\texttt{LLM}} |
\mathrm{PR}$ (which also equals the intersection of the positive class ratio of
the corresponding $\mathrm{CA}$ and $\mathrm{PR}$ tabular test-sets). The specific
values of such collapsed models correspond \textit{exactly} to the
population statistics found in Table \ref{tab:states-stats}. Analyzing the
confusion metric of the rest of the green points close to the positive class
ratio of $\mathrm{CA}$ confirms that there are \textit{only} pair-wise strong
collapses in this scenario, so that $\mathrm{CR}_s = \mathrm{CR}_p = 61, 58
\text{\%}$.

Third, Figure \ref{fig:al-cvar-linq-main} exhibits another frequent kind of
model collapse, where blue points are bound to green points which are
projected into at least one of the two corresponding positive class ratios for
$\mathrm{CA}$ and $\mathrm{AL}$. In this case, there are both pair-wise strong
collapses as well as projection collapses so that $\mathrm{CR}_s = 18, 23
\text{\%}$ and $\mathrm{CR}_p = 43, 84 \text{\%}$.

Last, but not least, all figures for RF in the appendix show that the set
$\mathrm{HP}_{\mathrm{RF}}$ is not well-defined w.r.t. CA. in the sense of
this article. In other words, there are already models collapsing over tabular
data under our choice of HP configurations. Notice that collapsed models in
tabular training do not imply collapsed models in rep training. In the case of
the scenario of Figure \ref{fig:pr-rf-sfr}, we have 26 strong collapses of
models trained on tabular data, 11 such collapses on the corresponding
\texttt{SFR} embeddings and only 3 bindings between pairs of collapsed
models. In other words, collapses in the tabular modality do not imply
collapses in the embedding modality, which is why we restrict our attention in
this article on model collapses arising in \textit{well-defined} HP
configuration sets (i.e. those that do not present model collapse on the
tabular modality).

\paragraph{Strong Model collapse induces spurious ACL.}When analyzing both the
ACL and F1L planes in Figure \ref{fig:pr-cvar-e5-main}, one can notice that
the $R^2$ coefficient of the green points is significantly higher than the one
for the blue points. In appearance, one would be tempted to say that LLM
Embeddings might improve the correlation between ID and OOD performance of
models but this is quickly shown to be due to the fact that the strong
collapse ratio is high, which means that the fact that many green points are
overlapping creates a misleading increase of the $R^2$ parameter while it is
clear that the green points that are not collapsed are nowhere near forming
some linear correlation both on the ACL and the F1L planes. Worse yet, because
of the high class imbalance presented in $\mathrm{PR}$, single prediction models
are in appearance among the best performing in terms of accuracy while they
are systematically predicting the same class both in CA and PR. Hence the
importance of the F1L plane, which resolves part of the issue but still shows
spurious correlations and cannot in any way be used for OOD performance
prediction.

\paragraph{Delimiting the Validity and Correlation Perimeter of
Accuracy/F1-on-the-Line Phenomena.} Although we have used the ID vs OOD
F1-plane to get a more nuanced view on how LLMs behave, it is interesting to
hint that there is a scope of validity when ACL and F1L are correlated when
they are valid. When observed, as shown in Figure \ref{fig:acl-ok}, this is
analog to the experimentally exhibited correlation between ACL and the
so-called Agreement-on-the-line (AGL) phenomenon. Now, AGL is particularly
useful to be able to predict OOD performance without then need of labeled OOD
data because the linear fit has shown to match the one obtained for ACL.
Although we do not expect that the linear fits match between ACL and F1L, our
findings suggest that on-the-line phenomena and their scope of validity should
be studied with different metrics of interest beyond accuracy. In the original
ACL paper, the F1-score was used in one experimental setup but it was not tied
to a reflection about imbalanced data but rather as the standard metric
associated to that specific task. We stress out that studying on-the-line
phenomena for different metrics induced by data-centric approaches may yield
novel insights about OOD Robustness.

Finally, Figure \ref{fig:on-the-line-app} in the
Appendix shows that when ACL is valid on the basis of standard training,
\textbf{it does not imply ACL on the basis of LLM Embeddings}. As such, our
experiments exhibit a case of a correlation between valid ACL and valid F1L
and a counter-example of a valid ACL at the standard training level that fails
to correlate with a valid ACL on the representation training counterpart.

\paragraph{Available TTA methods do not induce stronger ACL.}

The paper {\cite{kimtest}}, the authors reveals that Test-Time Adaptation
(TTA) methods not only improve model performance on out-of-distribution data
but also significantly strengthen the linear correlation between
in-distribution and out-of-distribution metrics. Inspired by this approach, we
employ TTA methods for MLP only, and we select Representation
Finetuning as described in {\cite{zeng2024llm}} (F.2), TENT
{\cite{wang2021tent}}, and FTAT {\cite{zhou2025fully}} giving a total of 288
adapted models. Our findings in Figures \ref{fig:ft}, \ref{fig:tent} and \ref{fig:ftat} show that current test-time adaptation methods,
including \textsc{TENT}  or its more recent analog
designed for tabular data \textsc{FtaT}, fail to
provide any help if one seeks to use these methods to widen the scope of
validity of ACL even when dealing with $X$-shifts. Concerning \textsc{FtaT},
a new pattern arises as a second horizontal line emerges in both the ACL and
F1L lines, and which does not match either class ratio.

\paragraph{LLM Embeddings Push Toward ID/OOD Generalization upper bounds.}

\begin{figure}[ht!]
\centering
\subfloat[ACL Standard\label{fig:acl-std}]{\includegraphics[width=0.5\columnwidth]{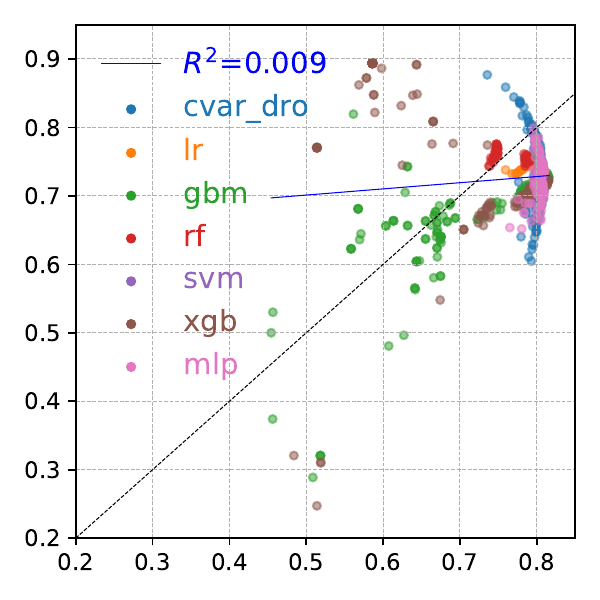}}
\subfloat[ACL \texttt{e5}\label{fig:acl-e5}]{\includegraphics[width=0.5\columnwidth]{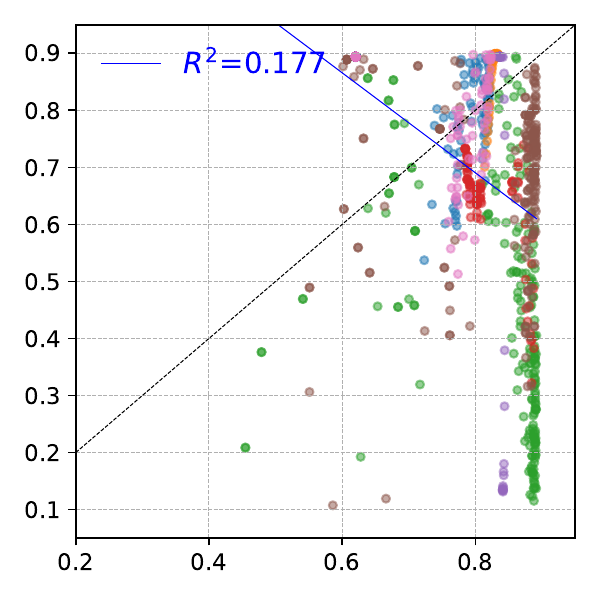}}
\caption{LLM Embeddings In-Distribution and OOD Generalization Upper-Bounds.}
\label{fig:ontheline}
\end{figure}

In Figure \ref{fig:acl-std} we report all 923 model combinations along all
families of models in the ACL plane for standard tabular
training. This is a partial reproduction of the ACL plane reported
in {\cite{liu2023need}}. Figure \ref{fig:acl-e5} reports the
combinations on the basis of training over
\texttt{e5} embeddings. Although the collapses are
not directly visible in the figure, one can see that the overall movement of
models with same configurations as those used to plot Figure
\ref{fig:acl-std}, tends to approach an upper bound in terms of
in-distribution accuracy , with heavily varying levels of accuracy on the
target domain, reaching a generalization upper-bound.

\subsection{Benchmarking of LLM Embeddings' Robustness}\label{sec:bench}

\paragraph{Experiment setup.} In this set of experiments we choose two MLP
configurations: the first, $c_1$, being one that we know
results in a collapse for \texttt{LLM=e5} and the
$(\mathrm{CA} \rightarrow \mathrm{PR})$ pair; the second, denoted $c_2$, being the
optimal HP configuration found in {\cite{zeng2024llm}}, where such
configuration was the one giving the best average macro F1 score among MLPs
for all pairs of US states (including PR).

\begin{figure}[ht!]
\centering
\subfloat[Accuracy]{\includegraphics[width=0.49\textwidth]{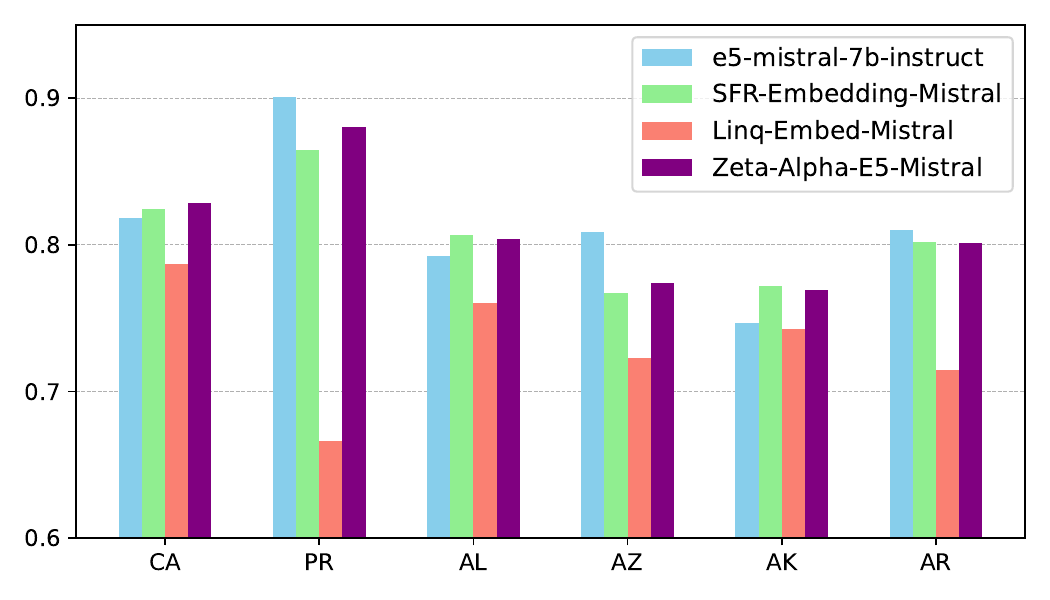}}
\label{fig:acc-per-state}
\subfloat[F1-Score]{\includegraphics[width=0.49\textwidth]{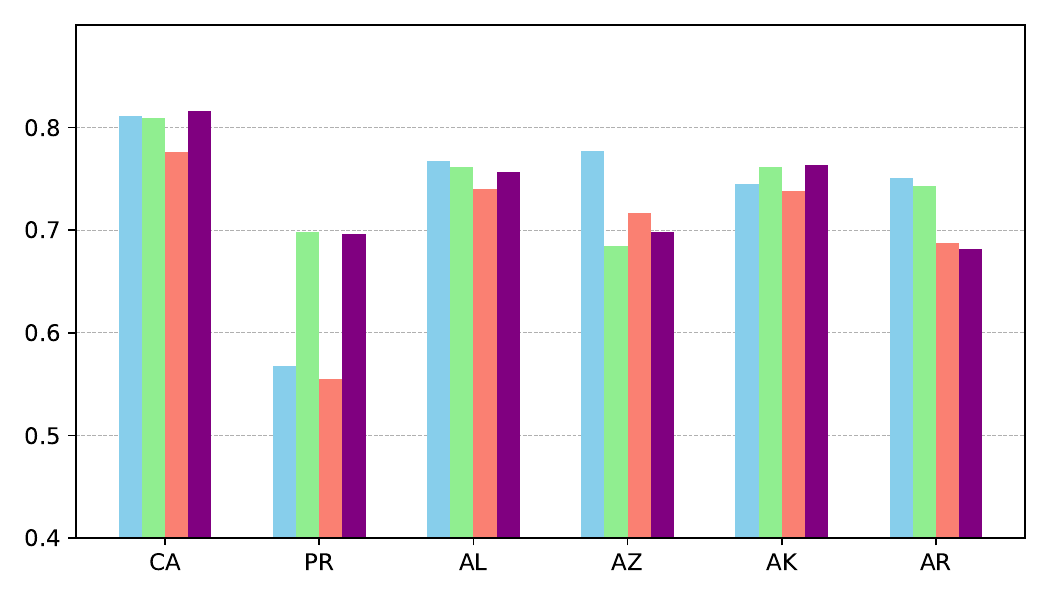}}\label{fig:f1-per-state}
\caption{Pattern behavior across different LLMs for MLP.}
\label{fig:acc-f1-per-state}
\end{figure}

\begin{figure*}[ht!]
\centering
\subfloat[Single Model Strong Collapse]{\includegraphics[width=0.33\textwidth]{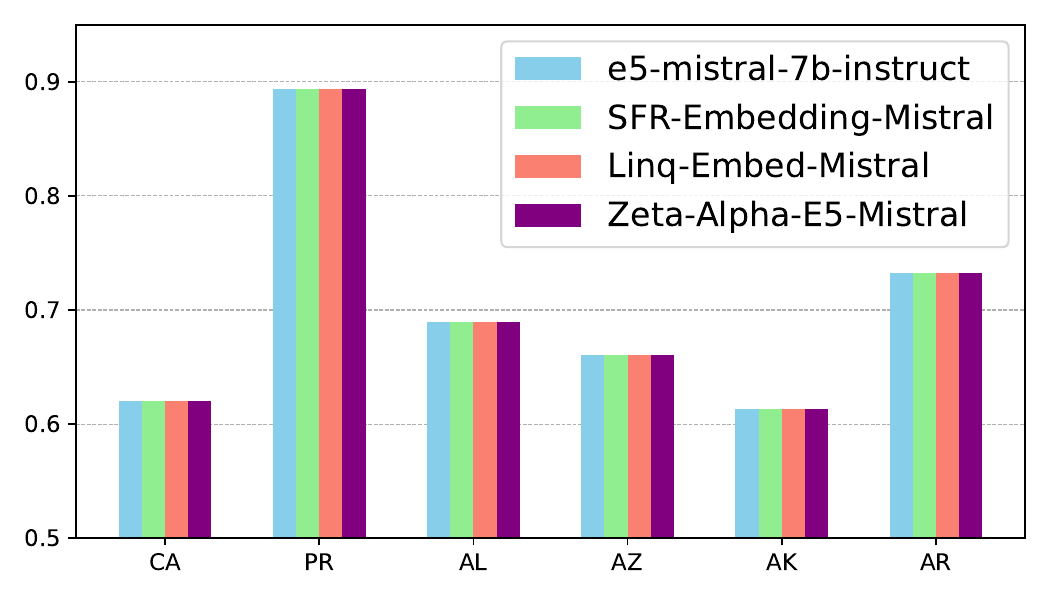}
\label{fig:accuracy-mlp-config2-e5}}
\subfloat[Optimal \texttt{e5} HP (Acc.)]{\includegraphics[width=0.33\textwidth]{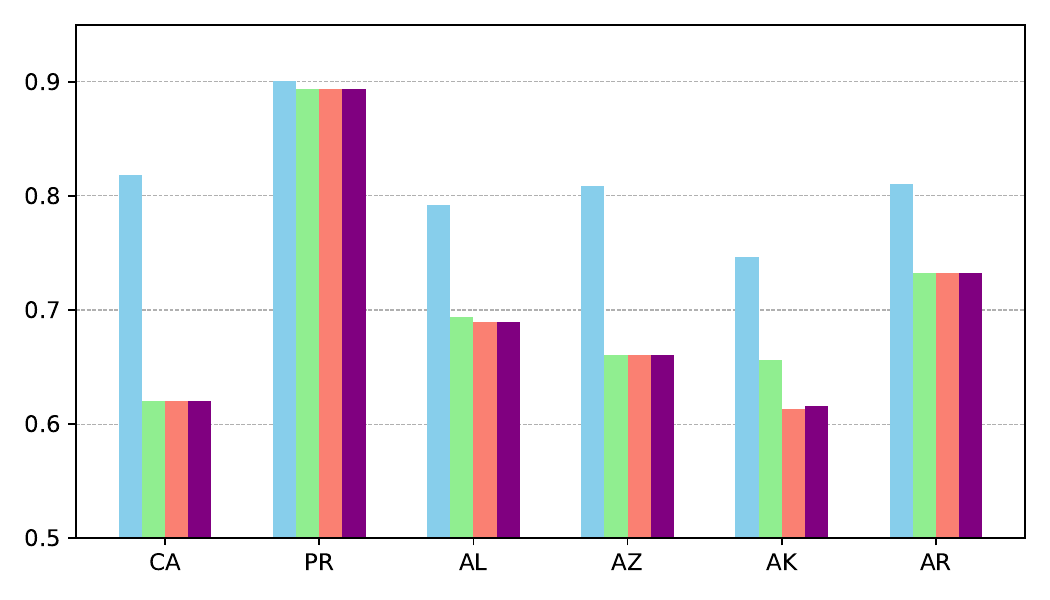}\label{fig:accuracy-mlp-config1-e5}}
\subfloat[Optimal \texttt{e5} HP (F1)]{\includegraphics[width=0.33\textwidth]{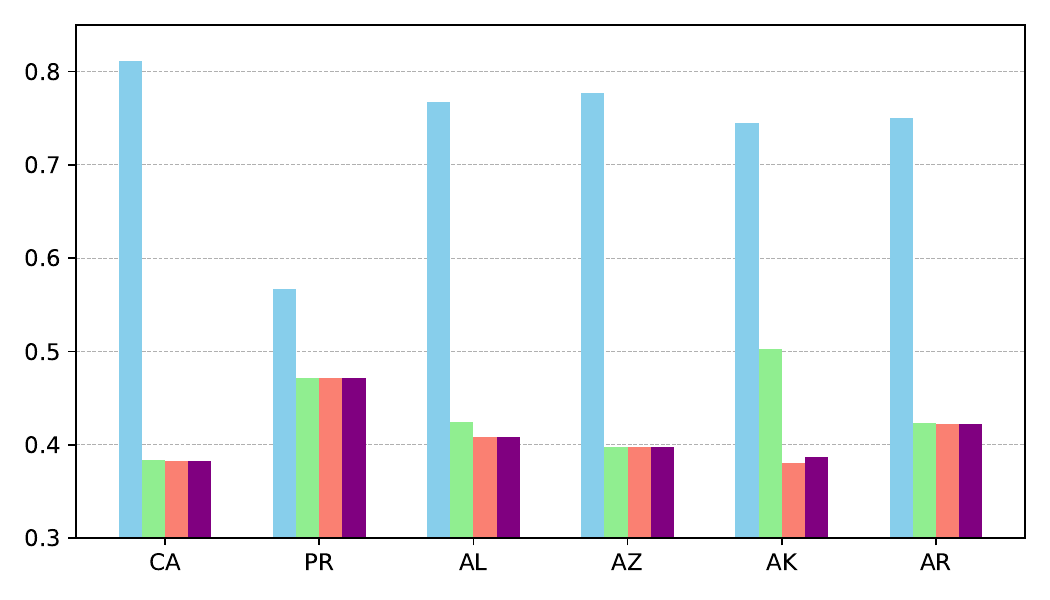}\label{fig:f1_score-mlp-config1-e5}}
\caption{Model behavior across LLM Embeddings. (a) Results for $f_{c_1}$; b) (resp. c)) Accuracy (resp. F1-score) for $f_{c_2}$.}
\label{fig:patterns-e5-fixed}
\end{figure*}

\paragraph{The \textsc{MTEB} ranking of LLMs is not correlated with their OOD
Generalization capabilities.} To address Question 4, we train MLPs $f_{c_2}$ for the optimal
reported configuration from \cite{zeng2024llm} over the 4 LLM embeddings of CA
separately. We test each of the four models obtained on
all states encoded by the same LLM used to train each
model. As such, we evaluate the performance of each LLM by means of how
each model trained along a choice of LLM embeddings performs when tested on
embeddings of other domains but encoded by the same LLM. This could be
reframed as a particular evaluative task for benchmarking LLM embeddings
consisting on OOD robustness beyond pure performance on classification tasks.
The results are reported in Figure \ref{fig:acc-f1-per-state}. 

Surprisingly, \texttt{Linq},
the second highest ranked LLM for \textsc{MTEB}, is never in
the first position. But
most importantly, \texttt{Zeta} (ranked 177th in \textsc{MTEB}) shows surprisingly superior
performance on this task in comparison to all other LLMs. 
In order to make an
extended assessment of our insight, we use the FractionBest
introduced in \cite{zeng2024llm}, and we conducted an exhaustive pairwise
comparison across all LLMs, hyper-parameter configurations for all families,
and all combinations of train and target domains which we report see Figure
\ref{fig:fractionbest-delta0-app}. We find that
\texttt{Zeta} outperforms competing LLMs in 31\% of the
pairwise comparisons based on accuracy, and in 22\% based on F1-score. We
stress that this experiment takes into account models trained over all
different states and not just the CA state.

Second, the ranking agreement between our 4 LLMs across states is not
universally valid. The ranking is consistent for the couples (PR,AZ) and (AL,AK) w.r.t accuracy, while presenting different degrees
of variability across other states. Also, when testing on
CA, Accuracy and F1-score present a slight ranking switch between
\texttt{e5} and \texttt{SFR}.

Third, different states present different metric variability:
on the source domain CA, Accuracy and F1-score values are fairly comparable
for all 4 LLMs but on the target domain PR, there is a much
more pronounced metric variability across all LLMs.
Lastly, domain gap w.r.t. accuracy does not universally penalize models
on target domains due to class imbalance: 3
out of 4 LLMs provide improved performance in PR compared to
the CA, with only \texttt{Linq} actually performing
worse. Nevertheless, domain gap w.r.t. F1-score does
penalize all models on all states.

\paragraph{Train/Test LLM mismatch significantly increases Model Collapse.}
In order to address question 5, we proceed to train two MLPs along
configurations $c_1$ and $c_2$ for \texttt{e5}
embeddings only, which we denote $f_{c_1}$ and $f_{c_2}$. We then test both models on all 4 LLM embeddings of all states.
The rationale here is to study how a model that has shown to collapse (resp.
has good performance) over states (e.g. $f_{c_1}$, resp. $f_{c_2}$) behaves on
those same states but encoded by other LLMs. The results are reported in
Figure \ref{fig:patterns-e5-fixed}.

On the one hand, Figure \ref{fig:accuracy-mlp-config2-e5} shows that the model
$f_{c_1}$ which we knew was collapsed over the \texttt{e5}$| \mathrm{CA}$
and \texttt{e5}$| \mathrm{PR}$ domains, remains collapsed along all
\texttt{e5} embeddings of the rest of the states. More strikingly,
$f_{c_1}$ collapses on the embeddings of all states for all 4
LLMs and not just the one used to train it. This suggests that model
collapses transfer across LLM embeddings for different LLMs than the one used
to train the model. In other words, changing the LLM does not induce a
positive behavior on collapsed models.

On the other hand, Figure \ref{fig:accuracy-mlp-config1-e5} shows that performance
of the model $f_{c_2}$ for the optimal configuration from {\cite{zeng2024llm}}
not only does not maintain its performance across embeddings, but it even
often results in model collapse over most states encoded by a different base
LLM encoder. Notice that this model, when evaluated on the all \texttt{Linq}
embeddings of all states, exhibits collapse. We report in Table
\ref{tab:conf-matrix-opt-e5} the confusion matrix values for each model. For
$f_{c_2}$, we observe:strict collapse on \texttt{Linq}$| \mathrm{PR}$, \texttt{SFR}$|
  \mathrm{PR}$, \texttt{Zeta}$| \mathrm{PR}$, \texttt{Zeta}$|
  \mathrm{CA}$, \texttt{Zeta}$| \mathrm{AL}$, \texttt{Zeta}$|
  \mathrm{AZ}$, \texttt{Zeta}$| \mathrm{AR}$; near collapse on \texttt{SFR}$| \mathrm{CA}$, \texttt{SFR}$|
  \mathrm{AR}$, \texttt{SFR}$| \mathrm{AZ}$, \texttt{SFR}$| \mathrm{AL}$,
  \texttt{Zeta}$| \mathrm{AK}$.
Finally, $f_c$ is \textit{not} a near collapse on \texttt{SFR}$|
\mathrm{AK}$: we have $\mathrm{TNR} \left( f_c, \text{\texttt{SFR}} |
\mathrm{AK}) + \mathrm{FNR} \left( f_c, \text{\texttt{SFR}} | \mathrm{AK}) =
0, 142 > \varepsilon \right. \right.$. Additionally, the model trained
on the \texttt{linq} embeddings with these same hyper-parameters
\textit{does not} collapse in any of the \texttt{linq} embeddings of the 6
states. 
Last, but not least, \textbf{models already collapse on the source domain for different LLM
embeddings}. Figure \ref{fig:on-the-line-CA} exhibits model collapse even when the target test set is CA but encoded by different LLMs. 
This suggest that the conclusions of \cite{zeng2024llm} may differ across LLMs.

%% file: 9-conclusion.tex
\paragraph*{Conclusion. } 
Training on uncurated text embeddings (TEs) from raw tabular data can lead to model collapse, where models predict a single class regardless of input. We show this is a consistent failure mode in TE-based models, unlike those trained on raw data. To capture this effect, we introduce metrics that assess collapse severity and reveal TE quality as a key factor in downstream performance. Notably, strong model collapse can induce spurious ACL correlations, distorting evaluation and misleading OOD performance prediction. We also find no correlation between \textsc{MTEB} rankings and an LLM’s effectiveness as a data curation layer. We hope  that our experimental setup, far from being comprehensive, is already rich enough to offer a foundation for more robust OOD evaluation for text embedding benchmarks.

\paragraph*{Acknowledgments. }
This work has been supported by the French government 
under the ”France 2030” program, as part of the SystemX Technological Research 
Institute. M.G. thanks Martin Royer for his several insightful comments, which greatly improved the quality of this paper and to Jiashuo Liu for many fruitful conversations about methodologies for modelling distribution shifts.

%% file: a1-exp-details.tex
\section{Appendix}

\subsection{Auxiliary findings}
\label{app:aux-find}

This subsection consists of supporting auxiliary findings that are mentioned in the main part of this paper but whose details are delegated due to space constraints.

\subsubsection{FractionBest Metric for different LLMs}

We examine the FractionBest metric introduced in \cite{zeng2024llm} (both with respect to accuracy and macro F1-score) to compare LLM\textbar$f$ for $f$ one of our chosen model families,  using four different encoders (\texttt{e5},
\texttt{SFR}, \texttt{Linq} and \texttt{Zeta}) against typical methods on
tabular features. This will help us understand how different embedding models
perform across all possible combinations of our distribution shift scenarios.

\begin{figure}[h!]
\centering
\subfloat[\texttt{Accuracy}]{\includegraphics[width=0.4\textwidth]{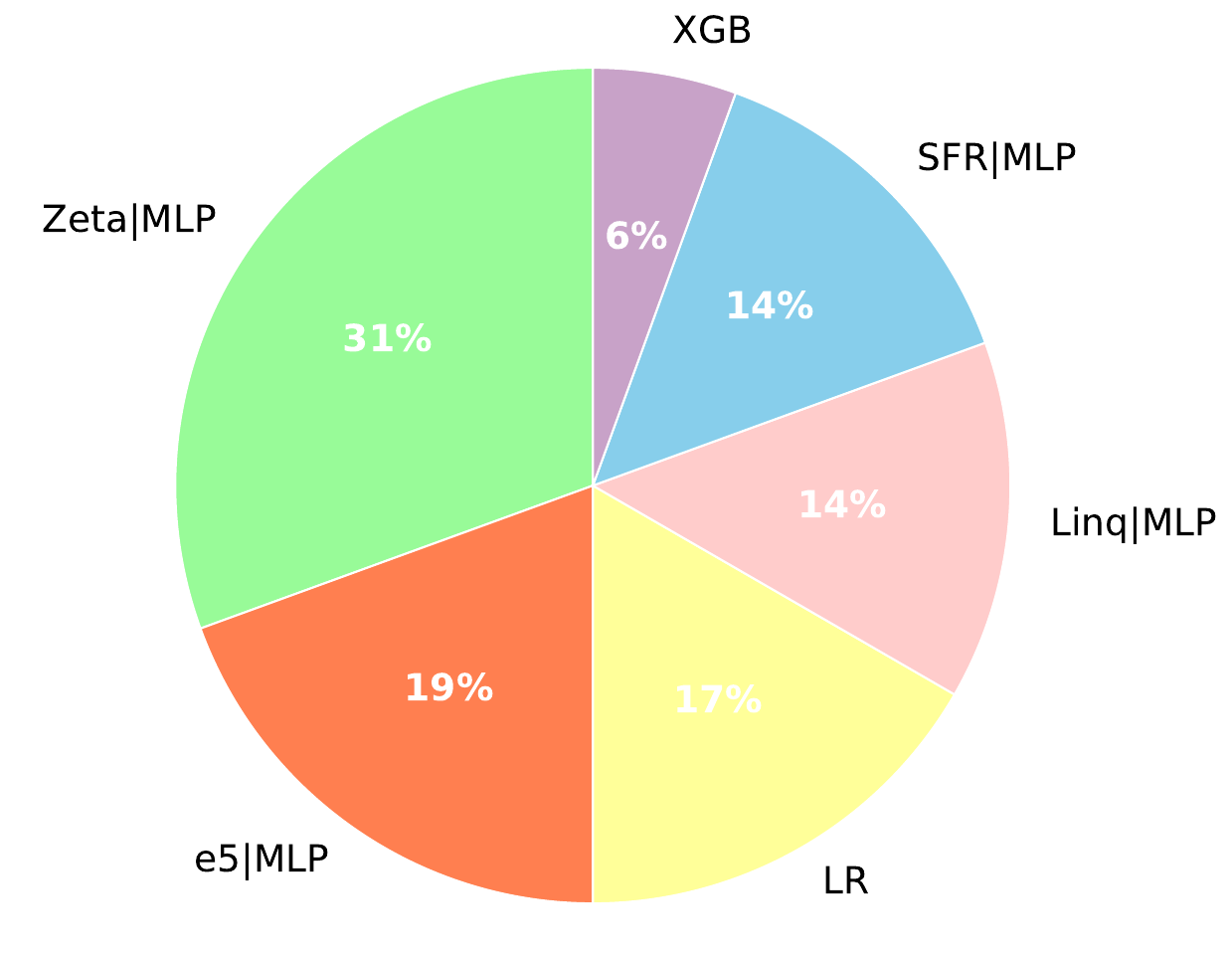}
\label{fig:fractionbest-delta0-accuracy}}
\subfloat[\texttt{F1-Score}]{\includegraphics[width=0.4\textwidth]{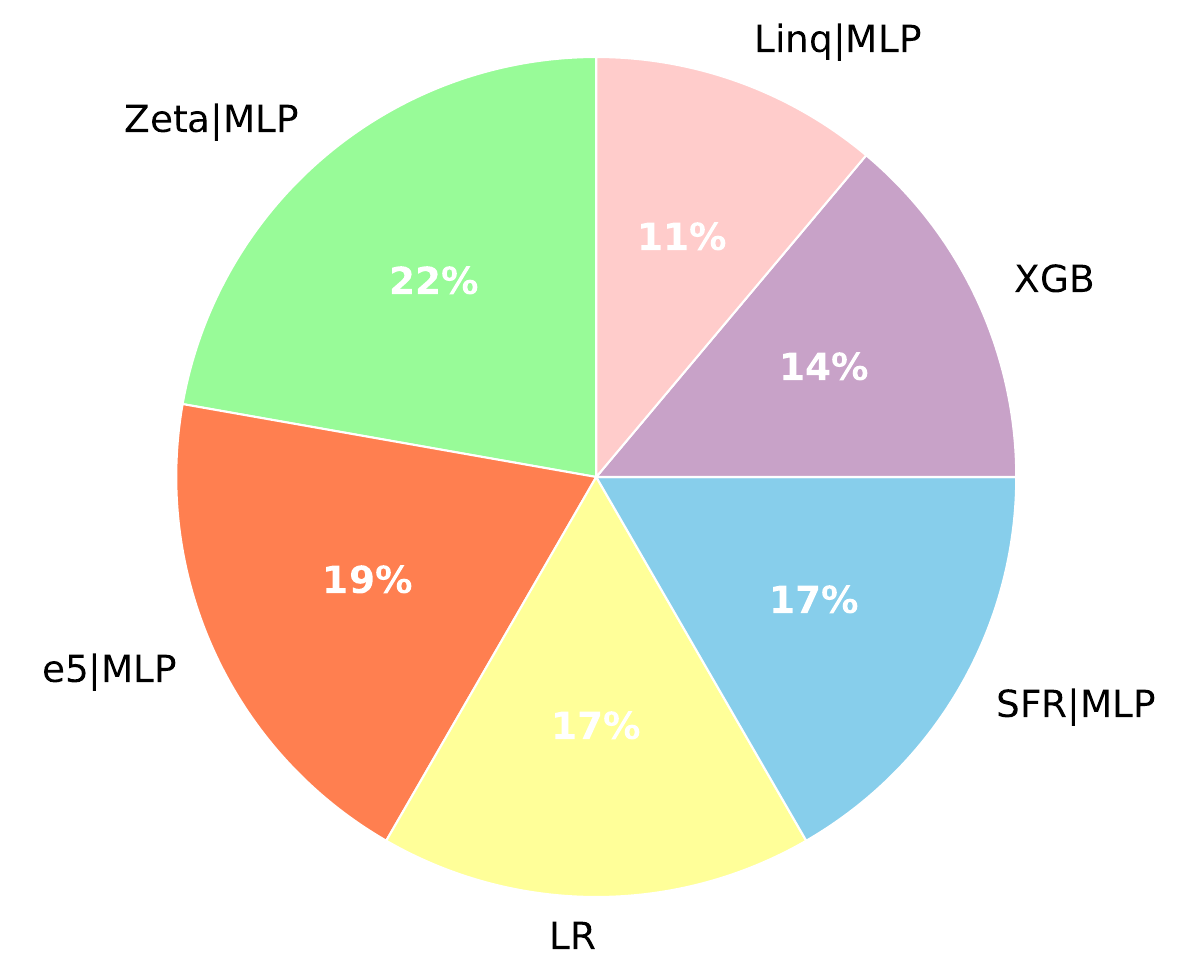}\label{fig:fractionbest-delta0-f1}}
\caption{Fraction Best across all states with $\Delta$ = 0\%.}
\label{fig:fractionbest-delta0-app}
\end{figure}

The results of FractionBest w.r.t. accuracy reveal striking differences in performance across
encoders. All embedding based models outperform traditional methods like
XGBoost (6\%), while traditional methods like Logistic Regression (17\%)
remain competitive. The results of FractionBest w.r.t. macro F1-score  consolidate \texttt{Zeta} as the best performing LLM in terms of providing good quality text embeddings while \texttt{Linq} shows to fall below all embedding-based methods and even tabular training methods such as LR and XGBoost.

\subsubsection{Model Collapse on the CA domain for different LLM Embeddings}

\begin{figure}[h!]
\centering
\subfloat[CA (e5 vs linq)]{\includegraphics[width=0.30\textwidth]{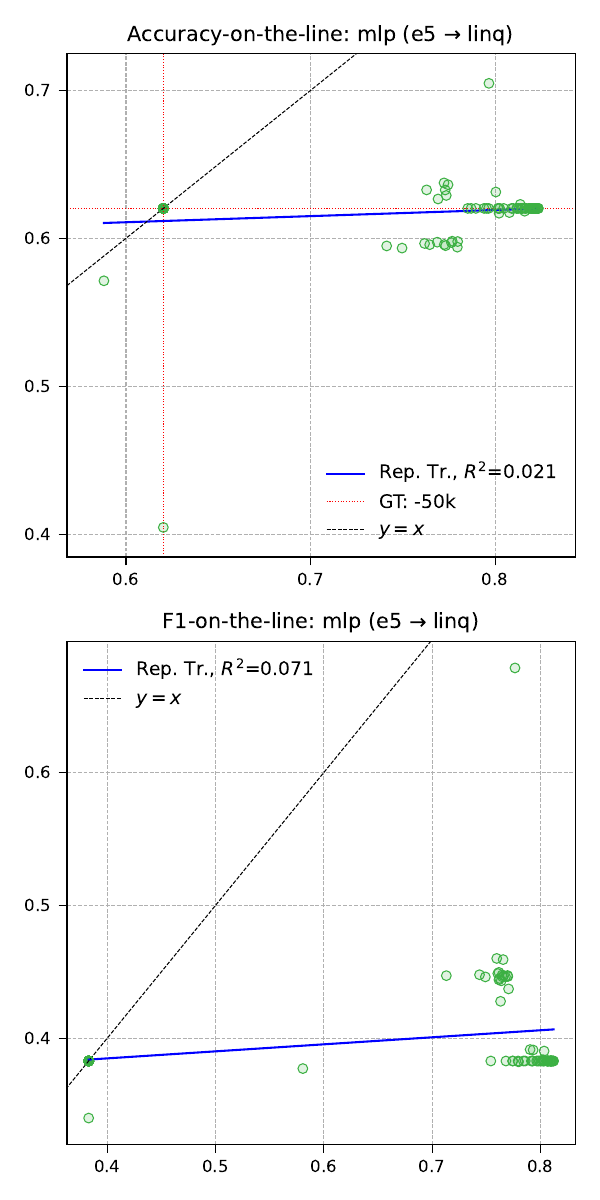}
\label{fig:e5_vs_linq_mlp_CA}}
\subfloat[CA (e5 vs sfr)]{\includegraphics[width=0.30\textwidth]{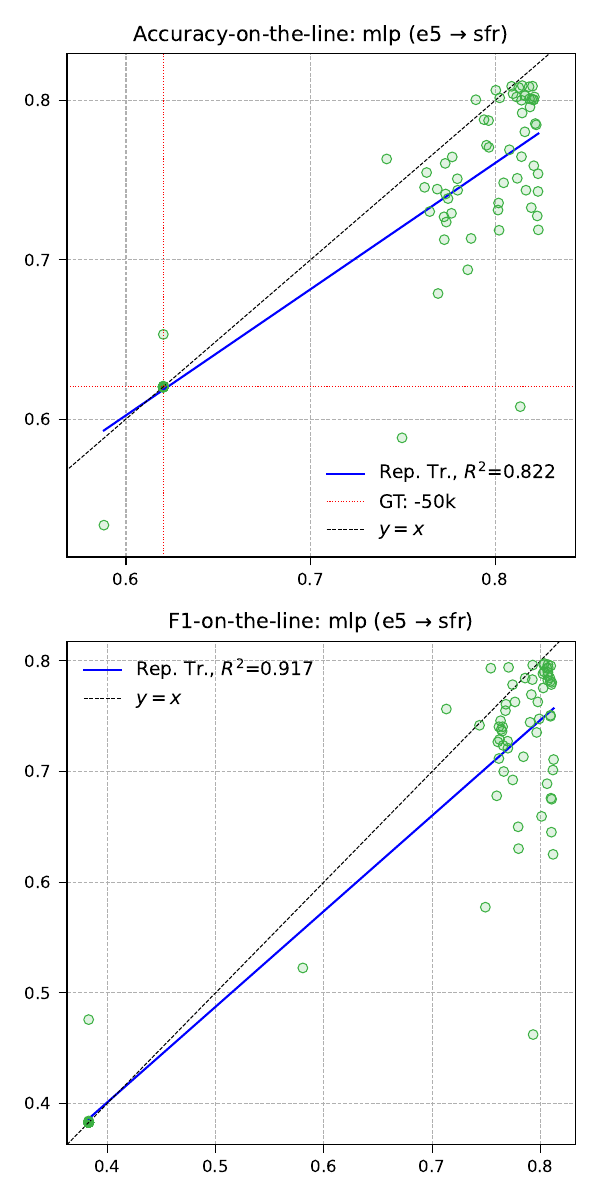}\label{fig:e5_vs_sfr_mlp_CA}}
\subfloat[CA (e5 vs zeta)]{\includegraphics[width=0.30\textwidth]{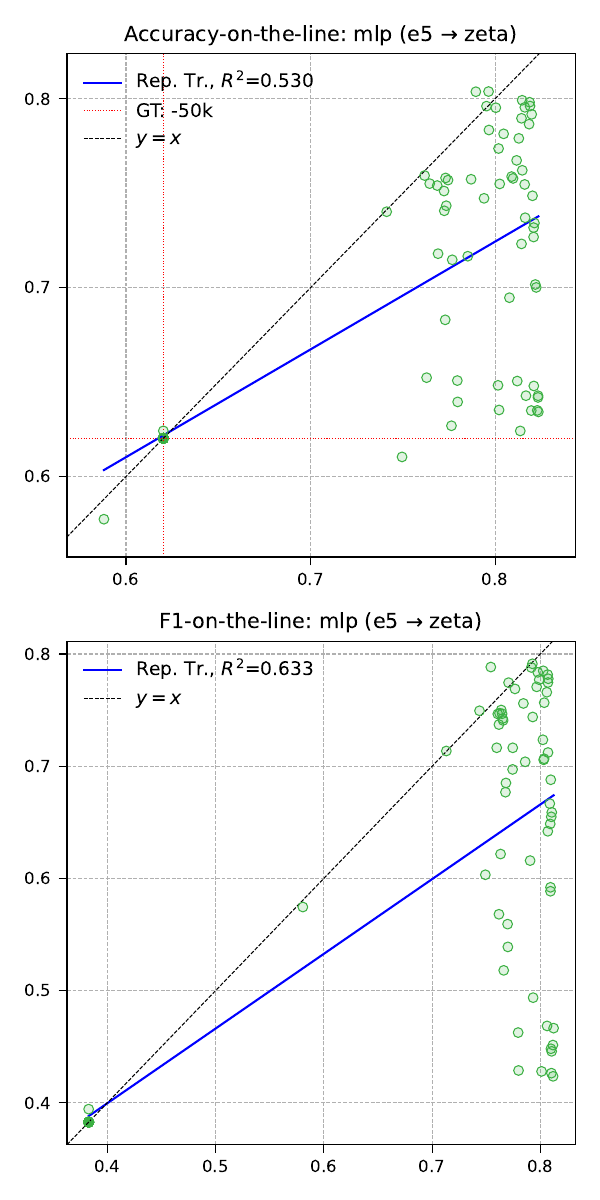}\label{fig:e5_vs_zeta_mlp_CA}}
\caption{Pattern behavior across different LLMs Embeddings for CA on the MLP HP Configuration Set.}
\label{fig:on-the-line-CA}
\end{figure}

Figure \ref{fig:on-the-line-CA} shows the pattern behavior of different LLMs embeddings of the same state CA for the ACL and the F1L planes.
In this experiment, we choose a test set from the state of California (CA) as our source domain, we plot on the $x$-axis accuracy and macro F1-score for the \texttt{e5-Mistral-7B-Instruct} embeddings and in $y$-axis we have the same metrics but for the CA embeddings obtained by the other 3 LLMs (namely \texttt{Linq}, \texttt{SFR}, and \texttt{Zeta}).

\subsubsection{Accuracy and F1 on-the-line do not hold for Representation training}

\begin{figure}[h!]
\centering
\subfloat[\texttt{Standard}]{\includegraphics[width=0.25\textwidth]{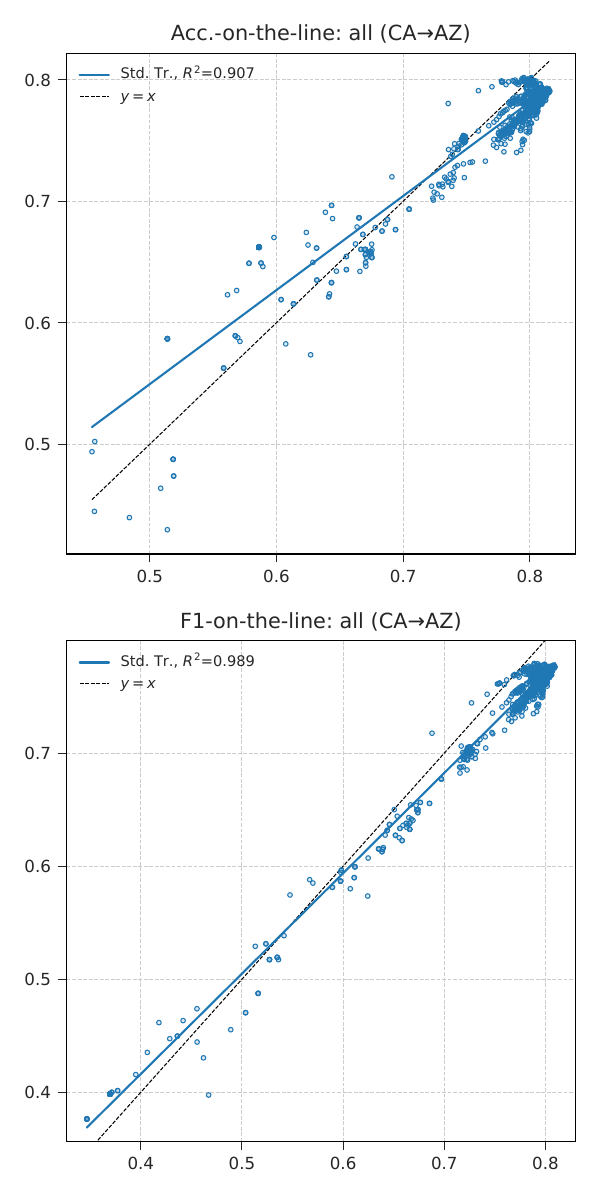}
\label{fig:std-all}}
\subfloat[\texttt{e5}]{\includegraphics[width=0.25\textwidth]{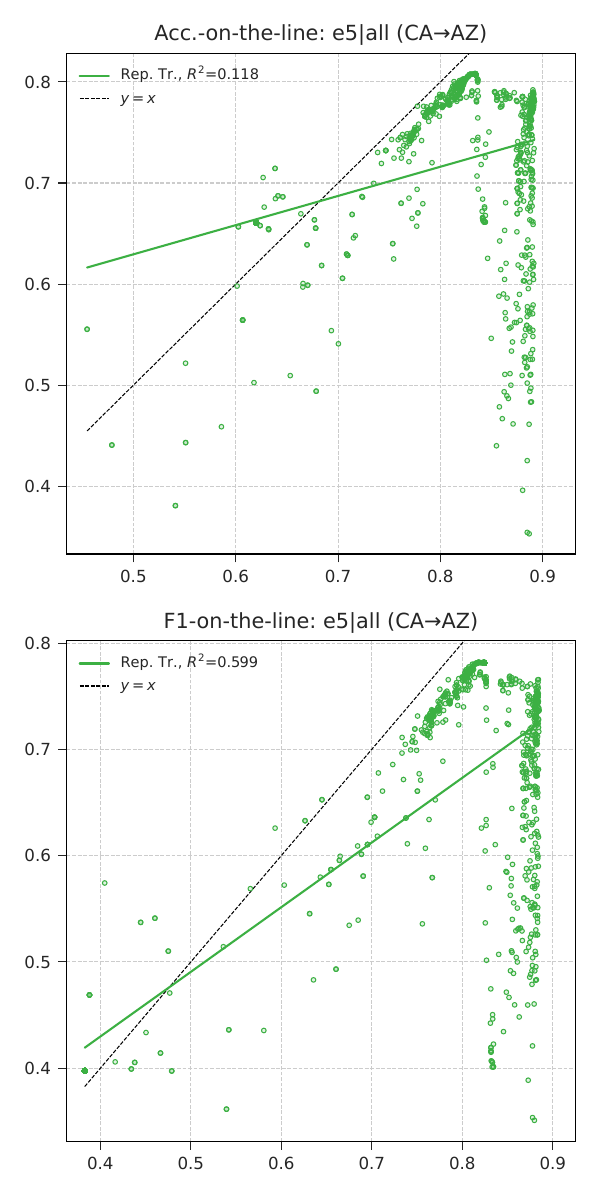}\label{fig:e5-all}}
\subfloat[\texttt{Linq}]{\includegraphics[width=0.25\textwidth]{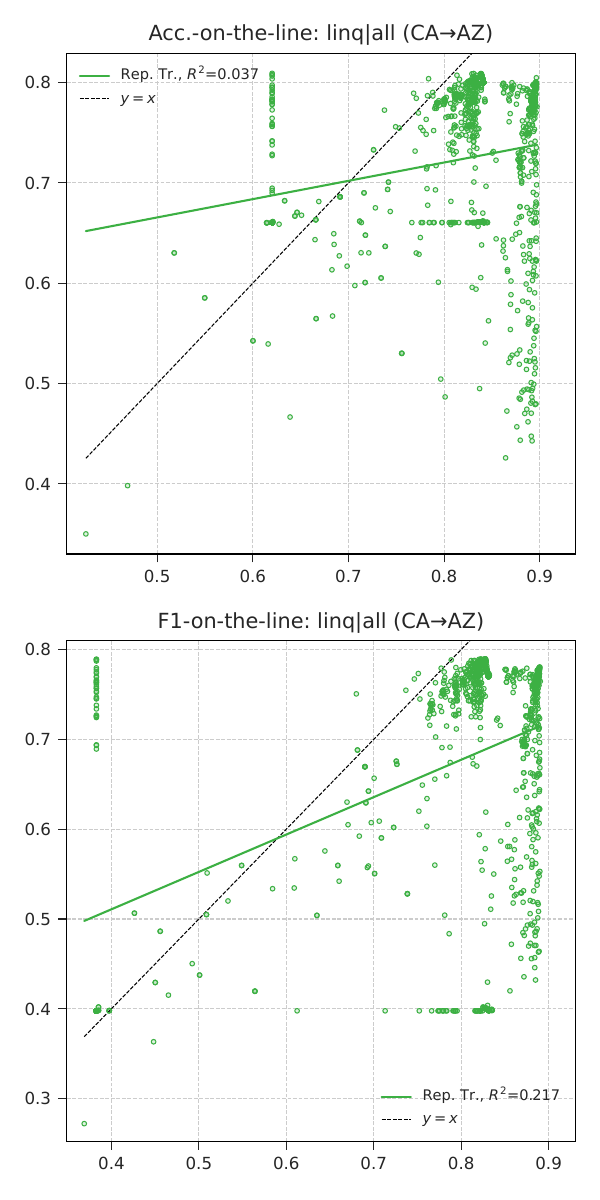}\label{fig:linq-all}}
\subfloat[\texttt{SFR}]{\includegraphics[width=0.25\textwidth]{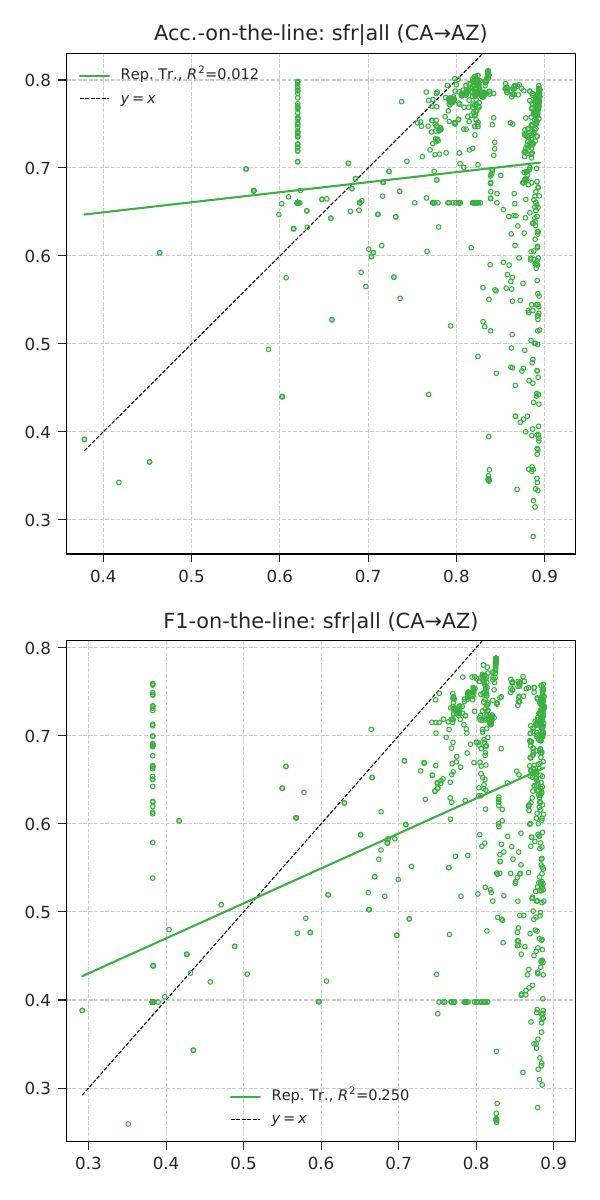}\label{fig:sfr-all}}
\caption{Pattern behavior across different LLMs for (CA,AZ). ACL and F1L are valid in tabular training but do not transfer into valid ACL or F1L in LLM-based representation training.}
\label{fig:on-the-line-app}
\end{figure}

Figure \ref{fig:on-the-line-app} shows the different patterns behaviors of the embeddings stemming from the four models for the pair (California (CA), Arizona (AZ)) in contrast to the behavior of a standard MLP trained on the same pair.
In this setting, we trained the MLP on the embeddings of the source (CA) for the four LLMs and evaluated its performance embeddings of the target (AZ).
We report as before for the two phenomena: accuracy on the line and F1 on the line.

\subsubsection{Configurations inducing collapse on all states (for fixed base LLM encoder)}

Table \ref{table:4} displays the different configurations we have selected for the MLP that lead to a systematic collapse across all the LLM-based embeddings.
\newline
For each configuration, the MLP was trained on the embeddings produced by the LLMs on the source CA then tested on the other states; for instance, for cfg-1, the MLP with a hidden size of $16$ was trained for $200$ epochs on the embeddings of CA with a $lr=0.01$ and a dropout ratio of $0.1$ then was tested on the embeddings of the remaining states.

\begin{table}[h!]
  \centering
  \caption{Shared MLP configurations that systematically lead to collapse across all LLM-based embeddings.}
  \begin{tabular}{lccccccc}
    \textbf{Collapse Configs} & \texttt{cfg-1} & \texttt{cfg-2} & \texttt{cfg-3} & \texttt{cfg-4} & \texttt{cfg-5} & \texttt{cfg-6} & \texttt{cfg-7} \\
    \midrule
    \textbf{lr} & 0.01 & 0.01 & 0.01 & 0.01 & 0.01 & 0.01 & 0.01 \\
    \textbf{hidden\_size} & 16 & 16 & 32 & 32 & 64 & 128 & 128 \\
    \textbf{dropout\_ratio} & 0.1 & 0.1 & 0.1 & 0.1 & 0.1 & 0.1 & 0.1 \\
    \textbf{train\_epochs} & 200 & 100 & 200 & 50 & 200 & 200 & 100 \\
  \end{tabular}
  \label{table:4}
\end{table}

\subsection{Detailed Model Collapse Metrics}
\label{app:col-metrics}

Let $S$ and $Q$ be two test-sets. We define the pair-wise \textit{strong}
collapse ratio as the ratio of hyper-parameter combinations resulting into
model collapse both test-sets
\[ \mathrm{CR}^{S, Q}_{\mathrm{strong}, \mathrm{HP}} = \frac{1}{| \mathrm{HP} |}
   \underset{c}{\sum} \mathbb{I}_{\{ \mathrm{Cond}^S_{K_1} (c) \cap
   \mathrm{Cond}^Q_{K_2} (c) \}}, \text{ for } K_1, K_2 \in \{ T, F \} \]
\[ = \frac{1}{| \mathrm{HP} |} \underset{c}{\sum} \mathbb{I}_{\{ \mathrm{Cond}^S_T
   (c) \cap \mathrm{Cond}^Q_T (c) \}} +\mathbb{I}_{\{ \mathrm{Cond}^S_F (c) \cap
   \mathrm{Cond}^Q_F (c) \}} +\mathbb{I}_{\{ \mathrm{Cond}^S_T (c) \cap
   \mathrm{Cond}^Q_F (c) \}} +\mathbb{I}_{\{ \mathrm{Cond}^S_F (c) \cap
   \mathrm{Cond}^Q_T (c) \}} . \]
Let us denote $\mathrm{TNR}^S (f_c)$ (resp. $\mathrm{FNR}^S (f_c)$) for the true
negative (resp. false negative) rate of $f_c$ on $S$.

Let $\varepsilon > 0$. We define the positive $\varepsilon$-near collapse
ratio of $\mathrm{HP}$ w.r.t. $S$ is defined as
\[ \mathrm{PCR}^S_{\mathrm{HP}} (\varepsilon) = \frac{1}{| \mathrm{HP} |}
   \underset{c}{\sum} \mathbb{I}_{\{ \mathrm{Cond}^S_T (c, \varepsilon) \}},
   \text{ where } \]
$\mathrm{Cond}^S_T (c, \varepsilon) = [(\mathcal{A}^S (f_c) \simeq \mathrm{TR}^S)
\cap (\mathrm{TNR}^S (f_c) + \mathrm{FNR}^S (f_c) < \varepsilon)]$. Here the
$\simeq$ symbol refers to the resulting approximation of $\mathrm{TR}^S$
depending on the $\mathrm{TNR}^S (f_c)$ and $\mathrm{FNR}^S (f_c)$ values. The
value of $\varepsilon$ varies from one experiment to the next, and in
particular should be chosen with consideration to the overall shape of the ID
vs. OOD scatter of $f_c$ performance variations for all $c \in \mathrm{HP}$. The
definition of negative $\varepsilon$-near collapse ratio is analogously
defined. Finally, let $\mathbf{\varepsilon} = (\varepsilon_1,
\varepsilon_2)$. The strong positive \textbf{$\varepsilon$}-near collapse
ratio of $\mathrm{HP}$ w.r.t. $S$ is defined as
\[ \mathrm{PCR}^{S, Q}_{s, \mathrm{HP}} (\mathbf{\varepsilon}) = \frac{1}{|
   \mathrm{HP} |} \underset{c}{\sum} \mathbb{I}_{\{ \mathrm{Cond}^S_T (c,
   \varepsilon_1) \cap \mathrm{Cond}^Q_T (c, \varepsilon_2) \}} . \]
The definition of the remaining combinations of the above metrics are not used
in this article, their formalization is left to the reader.

\begin{table}[h]
\centering
\caption{Selection of US States, with their respective test-set size and
  class balance.}\label{tab:states-stats}
  \begin{tabular}{cccccc}
    State & Test-set size & cl($< 50$k) size & cl($> 50$k) size & cl($<
    50$k)\% & cl($> 50$k)\%\\
    \hline
    CA & 50000 & 31011 & 18989 & 62.022\% & 37.978\%\\
    PR & 9071 & 8109 & 962 & 89.395\% & 10.605\%\\
    AZ & 33277 & 21977 & 11300 & 66.043\% & 33.957\%\\
    AR & 13929 & 10196 & 3733 & 73.200\% & 26.800\%\\
    AK & 3546 & 2175 & 1371 & 61.337\% & 38.663\%\\
    AL & 22268 & 15344 & 6924 & 68.906\% & 31.094\%
  \end{tabular}
\end{table}
Let us exemplify the implementation of these metrics on some basic examples. First of all, in Table \ref{tab:states-stats}, we report the population statistics of the test sets in our experiments. These will be the base reference to compute our model collapse metrics.

\paragraph{Detailed metric computation for Figure \ref{fig:pr-cvar-e5-main}.}

Let $f$ be $\texttt{e5}{\mid}\mathrm{CVaR-DRO}$, $S$ be
$\texttt{e5}{\mid}\mathrm{CA}_{\mathrm{test}}$ and $Q$ be
$\texttt{e5}{\mid}\mathrm{PR}_{\mathrm{test}}$. We have
$\mathrm{HP}_{\texttt{e5}{\mid}\mathrm{CVaR-DRO}} = 203$. There are 123
configurations $c \in \mathrm{HP}_{\texttt{e5}{\mid}\mathrm{CVaR-DRO}}$
satisfying $\mathrm{PN}^S (f_c) = 0$ and there are two configurations inducing
near collapses. As such, we have
\[ \mathrm{PCR}^{S, Q}_{s, \mathrm{HP}} (\mathbf{\varepsilon}) =
   \mathrm{PCR}^{\texttt{e5}{\mid}\mathrm{CA}_{\mathrm{test}},\texttt{e5}{\mid}\mathrm{PR}_{\mathrm{test}}}_{s,
   \mathrm{HP}_{\texttt{e5}{\mid}\mathrm{CVaR-DRO}}} (0.1, 0.1) =
   \frac{125}{203} = 61, 58 \text{\%} \]

An important subtle point about our metrics for near collapse is that the
underlying configurations that induce them might contribute to different
metrics and need to be handled with care. For instance, if we denote $c_1$ and
$c_2$ for the above near collapse configurations, we have,
\begin{eqnarray*}
  \mathrm{PN}^S (f_{c_1}) = 0 & ; & \mathrm{PN}^Q (f_{c_1}) = 1\\
  \mathrm{PN}^S (f_{c_2}) = 2 & ; & \mathrm{PN}^Q (f_{c_2}) = 8\\
  \mathrm{TNR}^S (f_{c_1}) + \mathrm{FNR}^S (f_{c_1}) = 0.00187 & < & \varepsilon
  = 0.1\\
  \mathrm{TNR}^S (f_{c_2}) + \mathrm{FNR}^S (f_{c_2}) = 0.02314 & < & \varepsilon
  = 0.1
\end{eqnarray*}
In particular, configuration $c_1 \in \mathrm{HP}$ induces a model $f_{c_1}$
that is both a strong positive $\mathbf{\varepsilon}$-near collapse with
respect to $S$ and $Q$ (i.e. it contributes to $\mathrm{PCR}^{S, Q}_{s,
\mathrm{HP}} (\mathbf{\varepsilon})$) but is also a projection (strict)
collapse with respect to $S$ (i.e. it contributes to
$\mathrm{CR}^S_{\mathrm{HP}}$).

\paragraph{Detailed metric computation for Figure \ref{fig:al-cvar-linq-main}.}

Let $f$ be $\texttt{Linq}{\mid}\mathrm{CVaR-DRO}$, $S$ be
$\texttt{Linq}{\mid}\mathrm{CA}_{\mathrm{test}}$ and $Q$ be
$\texttt{Linq}{\mid}\mathrm{AL}_{\mathrm{test}}$. We have
$\mathrm{HP}_{\texttt{Linq}{\mid}\mathrm{CVaR-DRO}} = 203$, and in this
particular case, there are no near collapses but only strict collapses:
\begin{eqnarray*}
  \mathrm{PCR}^S_{\mathrm{HP}} = \mathrm{PCR}^S_{\mathrm{HP}} (\varepsilon) =
  \mathrm{PCR}^{\texttt{Linq}{\mid}\mathrm{CA}_{\mathrm{test}}}_{\mathrm{HP}_{\texttt{Linq}{\mid}\mathrm{CVaR-DRO}}}
  (0.1) & = & \frac{61}{203} \simeq 30, 05 \text{\%}\\
  \mathrm{PCR}^Q_{\mathrm{HP}} = \mathrm{PCR}^Q_{\mathrm{HP}} (\varepsilon) =
  \mathrm{PCR}^{\texttt{Linq}{\mid}\mathrm{AL}_{\mathrm{test}}}_{\mathrm{HP}_{\texttt{Linq}{\mid}\mathrm{CVaR-DRO}}}
  (0.1) & = & \frac{65}{203} \simeq 32, 02 \text{\%}\\
  \mathrm{PCR}^{S, Q}_{s, \mathrm{HP}} (\mathbf{\varepsilon}) =
  \mathrm{PCR}^{\texttt{Linq}{\mid}\mathrm{CA}_{\mathrm{test}},
  \texttt{Linq}{\mid}\mathrm{AL}_{\mathrm{test}}}_{s,
  \mathrm{HP}_{\texttt{Linq}{\mid}\mathrm{CVaR-DRO}}} & = & \frac{37}{203}
  \simeq 18, 23 \text{\%}\\
  \mathrm{CR}^{S, Q}_{p, \mathrm{HP}} (\mathbf{\varepsilon}) =
  \mathrm{PCR}^{\texttt{Linq}{\mid}\mathrm{CA}_{\mathrm{test}},
  \texttt{Linq}{\mid}\mathrm{AL}_{\mathrm{test}}}_{p,
  \mathrm{HP}_{\texttt{Linq}{\mid}\mathrm{CVaR-DRO}}} & = & \frac{89}{203}
  \simeq 43, 84 \text{\%}
\end{eqnarray*}
\paragraph{On models which are not near collapses but have an accuracy close
to a class ratio.}In many of the experiments of this paper, there are examples
of points that \textit{seem} to be near collapses by the fact that their
accuracy is very close to a certain class ratio but where the sum
$\mathrm{TNR}^S (f_c) + \mathrm{FNR}^S (f_c)$ is far bigger than 0.1. 

Next, the following table reports the number of non zero predicted negatives corresponding to results in Figure \ref{fig:accuracy-mlp-config1-e5}.

\begin{table}[ht]
\centering
\caption{True Negatives (TN) and False Negatives (FN) by State for Optimal \texttt{e5} HP. We only report for LLMs inducing non-zero Predicted Negatives.}\label{tab:conf-matrix-opt-e5}
\begin{tabular}{lcc}

 & \textbf{SFR} & \textbf{Zeta} \\
\textbf{State} & TN$|$ FN & TN$|$ FN \\
\midrule
CA  & 21 $|$  1  & - \\
AK  & 179 $|$  27  &  9 $|$   0 \\
AR  & 1 $|$  0  &  - \\
AL  & 107 $|$  7  & -  \\
AZ  & 2 $|$  0  &  - \\
\end{tabular}
\end{table}
Together with the class statistics from Table \ref{tab:states-stats}, we readily check that, for $f_c$ the MLP trained over the optimal configuration lr$=$0.001, hidden\_size$=$32, dropout\_ratio$=$0, and train\_epochs$=$500 only on \texttt{e5} embeddings of CA, and tested on the rest of embeddings of all states, we have:
\begin{itemize}
    \item $\mathrm{TNR} \left( f_c, \text{\texttt{SFR}} |
\mathrm{AL}) + \mathrm{FNR} \left( f_c, \text{\texttt{SFR}} | \mathrm{AL}) =
0,0004562 + 0,0154535 < \varepsilon \right. \right.$;
    \item $\mathrm{TNR} \left( f_c, \text{\texttt{Zeta}} |
\mathrm{AK}) + \mathrm{FNR} \left( f_c, \text{\texttt{Zeta}} | \mathrm{AK}) =
0 + 0,00656455 < \varepsilon \right. \right.$;
    \item $\mathrm{TNR} \left( f_c, \text{\texttt{SFR}} |
\mathrm{AK}) + \mathrm{FNR} \left( f_c, \text{\texttt{SFR}} | \mathrm{AK}) =
0,01241379 + 0,13056163 > \varepsilon \right. \right.$.
\end{itemize}

\newpage

\subsection{Shared Hyperparameter Configuration Sets for Experiment \ref{sec:mc}}
\label{sec: appendix hps}

For each model, we report the associated set of hyperparameters as shown in 
Table~\ref{table: training HPs}, along the same logic from the HP grids reported in \cite{liu2023need} and \cite{zeng2024llm}.

\begin{table}[ht!]
  \caption{Sets of shared hyper-parameter combinations used in all experiments. $^\diamond$for methods with the total grid size above 200, we randomly sample 200 configurations for fair comparisons. The column "Well-defined" refers to empirical observation that the corresponding HP sets do not collapse for the tabular modality in our setting.}
  \label{table: training HPs}
  \vspace{0.1in}
  \resizebox{\textwidth}{!}{%
  \begin{tabular}{@{}ccccc@{}}   
    \toprule
    Model & \# of HPs & Well-Defined & Hyperparameter & Value Range  \\ \midrule

    \multirow{4}{*}{SVM} & \multirow{4}{*}{96} & \multirow{4}{*}{Yes} & C &
      {\footnotesize$\{1e^{-2},1e^{-1},1,1e^{1},1e^{2},1e^{3}\}$} \\[2pt]
   & & & Kernel          & $\{\text{linear},\text{RBF}\}$ \\
   & & & Loss            & Squared-Hinge \\
   & & & $\gamma$        &
      {\footnotesize$\{0.1,0.3,0.5,1,1.5,2,\text{scale},\text{auto}\}$}  \\ \midrule

    \multirow{4}{*}{LR} & \multirow{4}{*}{100} & \multirow{4}{*}{Yes} & \multirow{4}{*}{$L_2$ penalty} &
      $\{1e^{-4},2e^{-4},\dots,1e^{-3},1.5e^{-3},\dots,1e^{-2},$ \\
  &  & & & $1.5e^{-2},\dots,1e^{-1},1.5e^{-1},\dots,1,1.3,1.5,1.7,2,2.5,\dots,5,6,$ \\
   & & & & $1e^{1},1.5e^{1},2e^{1},3e^{1},\dots,1e^{2},2e^{2},3e^{2},5e^{2},7e^{2},$ \\
   & & & & $1e^{3},2e^{3},3e^{3},5e^{3},1e^{4},5e^{4}\}$  \\ \midrule

    \multirow{4}{*}{MLP} & \multirow{4}{*}{96} & \multirow{4}{*}{Yes} & Learning Rate &
      $\{0.001,0.003,0.005,0.01\}$ \\
   & & & Hidden Layer Dim & $\{16,32,64,128\}$ \\
   & & & Dropout Ratio    & $\{0,0.1\}$ \\
   & & & Train Epoch      & $\{50,100,200\}$  \\ \midrule

    \multirow{4}{*}{GBM} & \multirow{4}{*}{1680$^\diamond$} & \multirow{4}{*}{Yes} & Learning Rate  &
      $\{1e^{-2},1e^{-1},5e^{-1},1\}$ \\
   & & & Num. Estimators  & $\{32,64,128,256\}$ \\
   & & & Max Depth        & $\{2,4,8,16\}$ \\
   & & & Min. Child Samples & $\{1,2,4,8\}$  \\ \midrule

    \multirow{7}{*}{XGB} & \multirow{7}{*}{1944$^\diamond$} & \multirow{7}{*}{No} & Learning Rate &
      $\{0.1,0.3,1,2\}$ \\
   & & & Min. Split Loss & $\{0,0.1,0.5\}$ \\
   & & & Max Depth       & $\{4,6,8\}$ \\
   & & & Column Subsample (tree) & $\{0.7,0.9,1\}$ \\
   & & & Column Subsample (level)& $\{0.7,0.9,1\}$ \\
   & & & Max Bins        & $\{128,256,512\}$ \\
   & & & Growth Policy   & $\{\text{Depthwise},\text{Loss\,Guide}\}$  \\ \midrule

    \multirow{5}{*}{RF} & \multirow{5}{*}{640$^\diamond$} & \multirow{5}{*}{No} & Num. Estimators &
      $\{32,64,128,256,512\}$ \\
   & & & Min. Samples Split & $\{2,4,8,16\}$ \\
   & & & Min. Samples Leaf  & $\{1,2,4,8\}$ \\
   & & & Max. Features      & $\{\text{sqrt},\text{log2}\}$ \\
   & & & CCP Alpha          & $\{0,0.001,0.01,0.1\}$  \\ \midrule

    \multirow{2}{*}{CVaR-DRO} & \multirow{2}{*}{1620$^\diamond$} & \multirow{2}{*}{Yes} &
      Worst-case Ratio $\alpha$ & $\{0.01,0.1,0.2,0.3,0.5,1\}$ \\
   & & & Underlying Model Class & MLP  \\

    \bottomrule
  \end{tabular}}
\end{table}

\newpage

\subsection{Supporting Tables for Experiment \ref{sec:bench}}
\label{app:res}

\subsubsection{Details for Figure \ref{fig:acc-f1-per-state}}

Table \ref{tab:exp5} presents the accuracy and F1 scores for LLM$|$MLP model evaluated under these parameters: lr$=$0.001, hidden\_size$=$32, dropout\_ratio$=$0, and train\_epochs$=$500.

\begin{table}[ht!]
  \centering
  \caption{Detailed Results for Figure \ref{fig:acc-f1-per-state}.}\label{tab:exp5}
  \begin{tabular}{c|cccccc|cccccc}
    \multicolumn{1}{c|}{} & \multicolumn{6}{c|}{\textbf{Accuracy}} & \multicolumn{6}{c}{\textbf{F1-macro}} \\
    \textbf{LLM\textbackslash State} & CA & PR & AL & AZ & AK & AR & CA & PR & AL & AZ & AK & AR \\
    \hline
    \texttt{e5}   & 81.85 & 90.07 & 79.23 & 80.88 & 74.64 & 81.02 & 81.13 & 56.74 & 76.71 & 77.73 & 74.52 & 75.06 \\
    \texttt{linq} & 78.71 & 66.59 & 76.04 & 72.29 & 74.25 & 71.44 & 77.60 & 55.54 & 73.96 & 71.63 & 73.84 & 68.75 \\
    \texttt{sfr}  & 82.41 & 86.49 & 80.68 & 76.70 & 77.15 & 80.15 & 80.95 & 69.80 & 76.16 & 68.44 & 76.18 & 74.28 \\
    \texttt{zeta} & 82.84 & 88.00 & 80.40 & 77.38 & 76.90 & 80.11 & 81.61 & 69.64 & 75.68 & 69.84 & 76.29 & 68.16 \\
  \end{tabular}
\end{table}

\subsubsection{Details for Figure \ref{fig:accuracy-mlp-config1-e5}}

\begin{table}[ht!]
  \centering
  \caption{Accuracy of an MLP trained on CA and evaluated on each state with different LLM embeddings (Fig. \ref{fig:accuracy-mlp-config1-e5}).}
  \begin{tabular}{ccccccc}
     LLM\textbackslash State & CA & PR & AL & AZ & AK & AR\\
     \hline
     \texttt{e5}  & 81.85 & 90.07 & 79.23 & 80.88 & 74.64 & 81.02\\
     \texttt{linq} & 62.02 & 89.39 & 68.90 & 66.042 & 61.33 & 73.19\\
     \texttt{sfr} & 62.06 & 89.39 & 69.35 & 66.048 & 65.62 & 73.20\\
     \texttt{zeta} & 62.02 & 89.39 & 68.90 & 66.042 & 61.59 & 73.19
   \end{tabular}
\end{table}

%% file: a2-suppl-figs.tex

\subsection{Supplementary Figures for Experiment \ref{sec:mc}}
\label{app:figs}


\paragraph{Target State: Puerto Rico} 

\begin{figure}[bp!]
\centering
\subfloat[\texttt{e5}]{\includegraphics[width=0.25\textwidth]{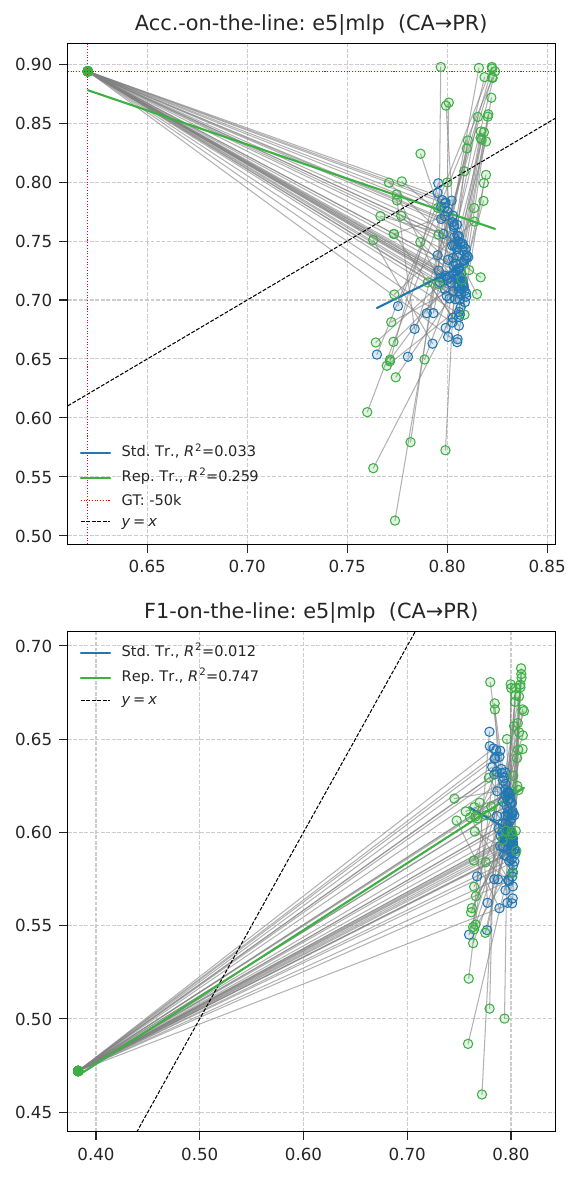}
\label{fig:pr-mlp-e5}}
\subfloat[\texttt{Linq}]{\includegraphics[width=0.25\textwidth]{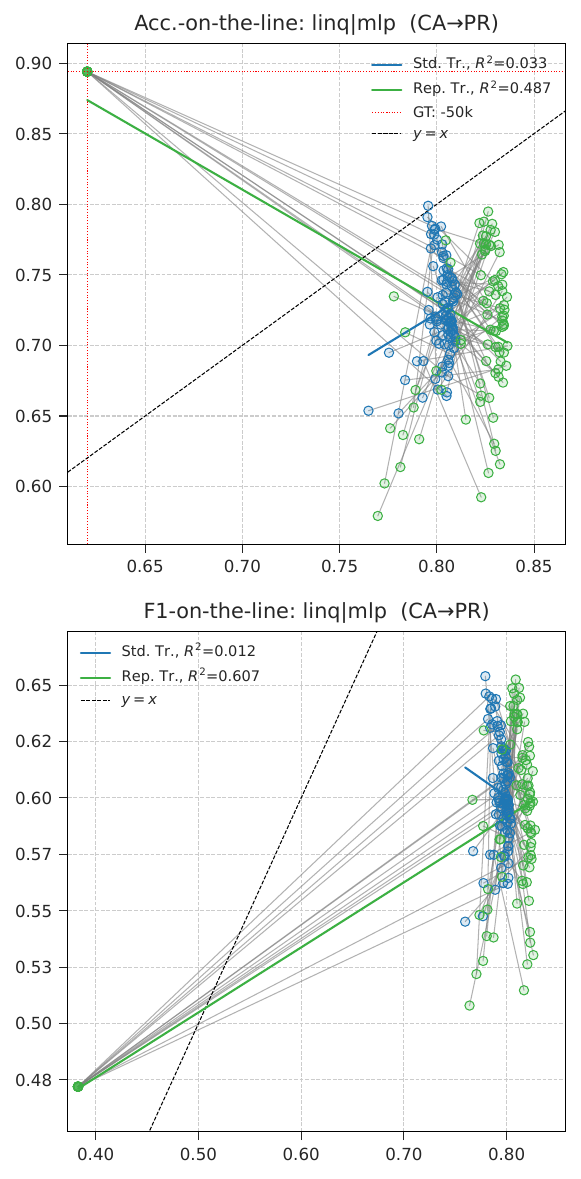}\label{fig:pr-mlp-linq}}
\subfloat[\texttt{SFR}]{\includegraphics[width=0.25\textwidth]{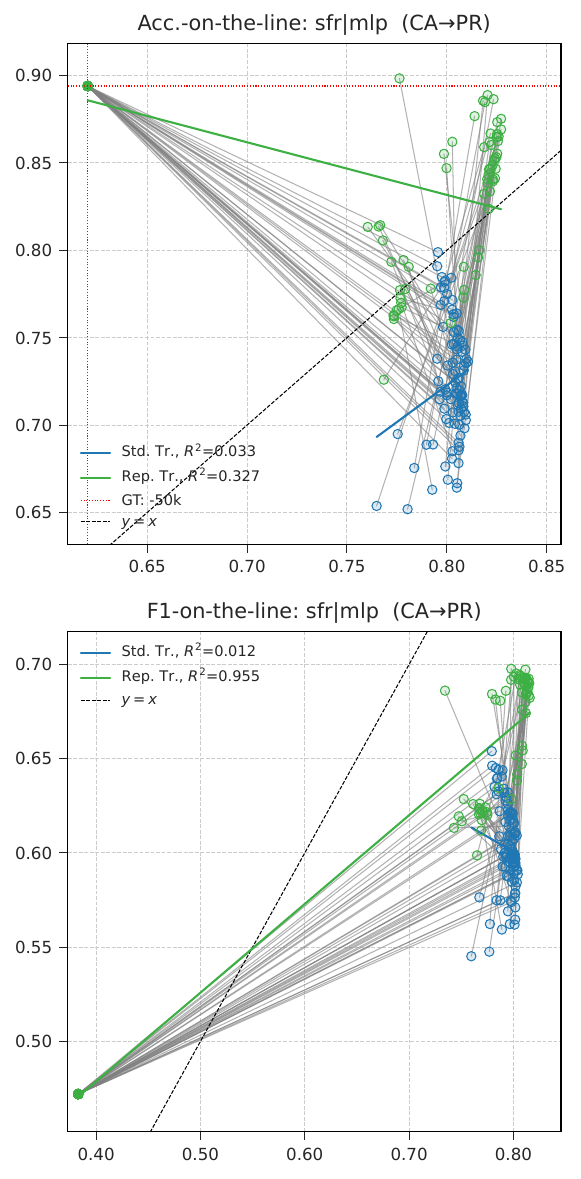}\label{fig:pr-mlp-sfr}}
\subfloat[\texttt{Zeta}]{\includegraphics[width=0.25\textwidth]{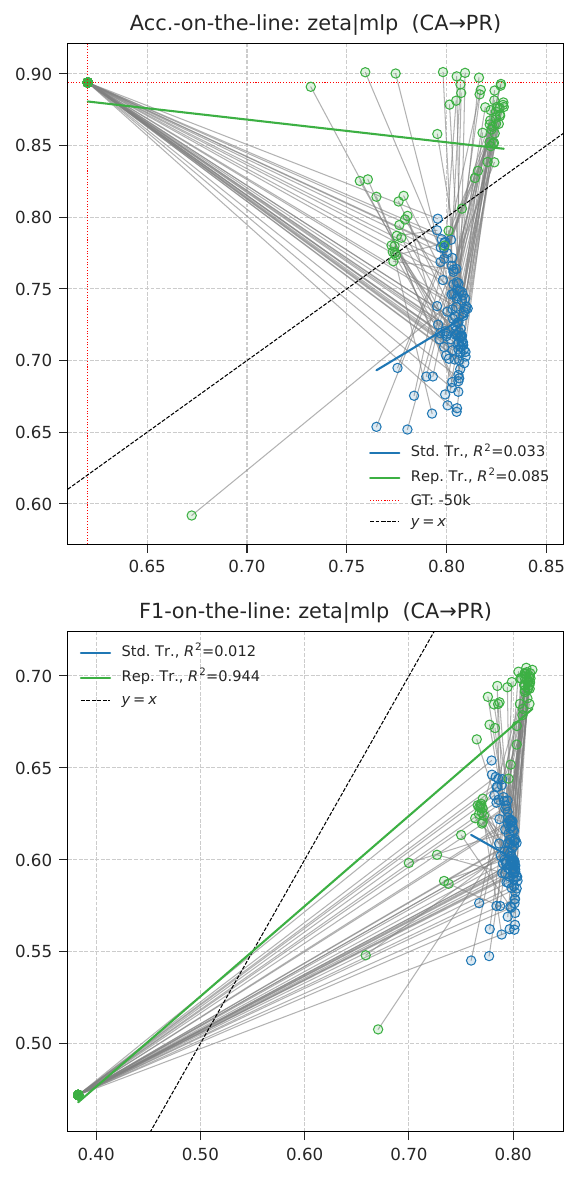}\label{fig:pr-mlp-zeta}}
\caption{Pattern behaviour across different LLMs for MLP.}
\label{fig:pr-mlp}
\end{figure}

\begin{figure}[bp!]
\centering
\subfloat[\texttt{e5}]{\includegraphics[width=0.25\textwidth]{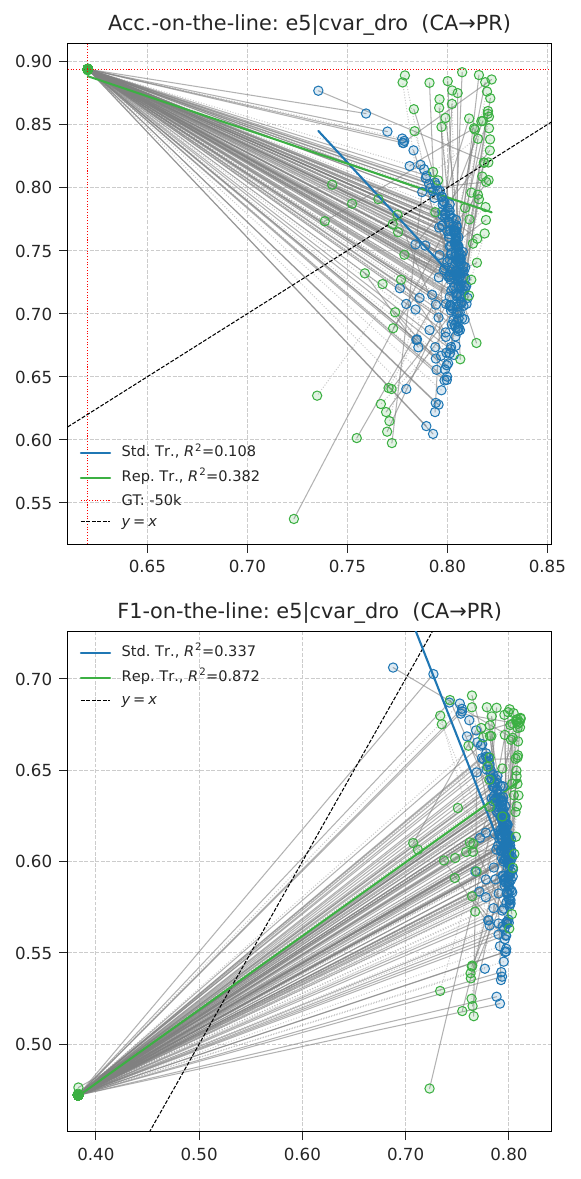}
\label{fig:pr-cvar_dro-e5}}
\subfloat[\texttt{Linq}]{\includegraphics[width=0.25\textwidth]{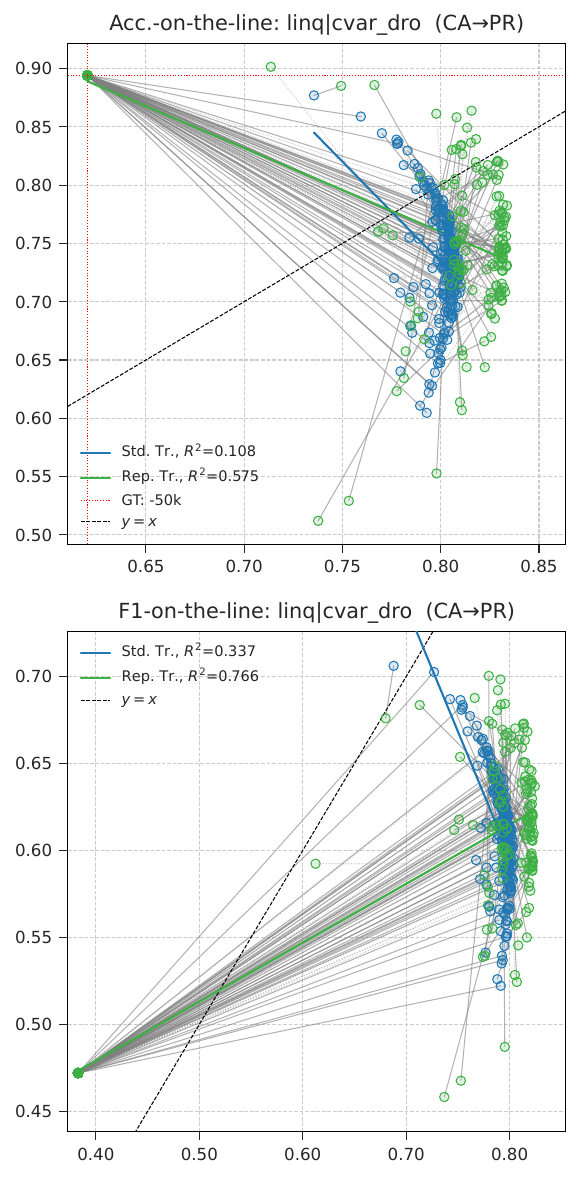}\label{fig:pr-cvar_dro-linq}}
\subfloat[\texttt{SFR}]{\includegraphics[width=0.25\textwidth]{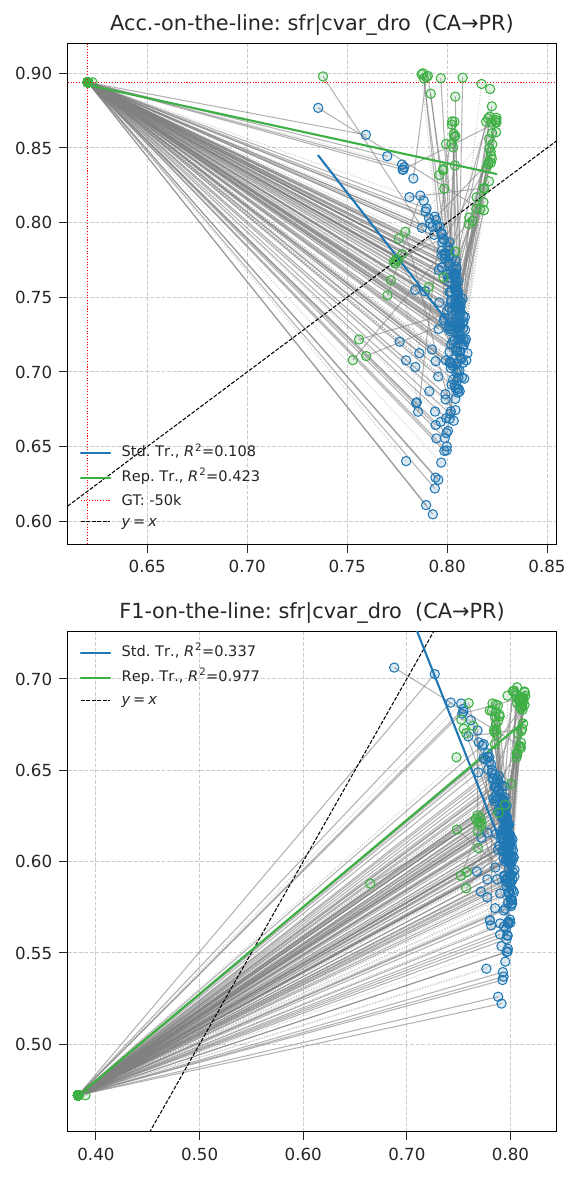}\label{fig:pr-cvar_dro-sfr}}
\subfloat[\texttt{Zeta}]{\includegraphics[width=0.25\textwidth]{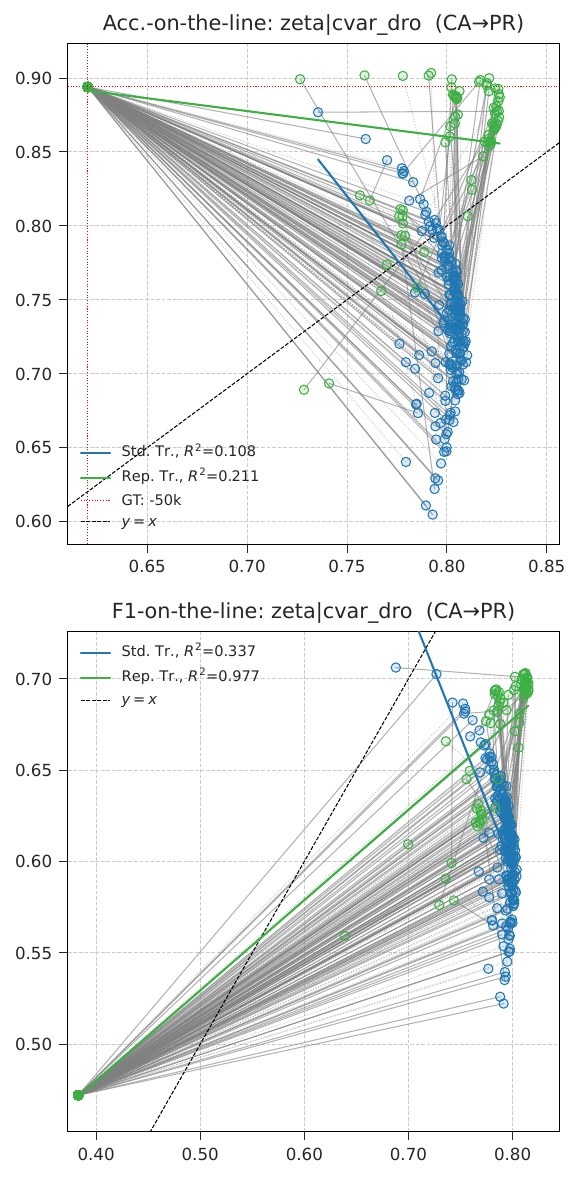}\label{fig:pr-cvar_dro-zeta}}
\caption{Pattern behaviour across different LLMs for CVaR-DRO.}
\label{fig:pr-cvar_dro}
\end{figure}


\begin{figure}[bp]
\centering
\subfloat[\texttt{e5}]{\includegraphics[width=0.25\textwidth]{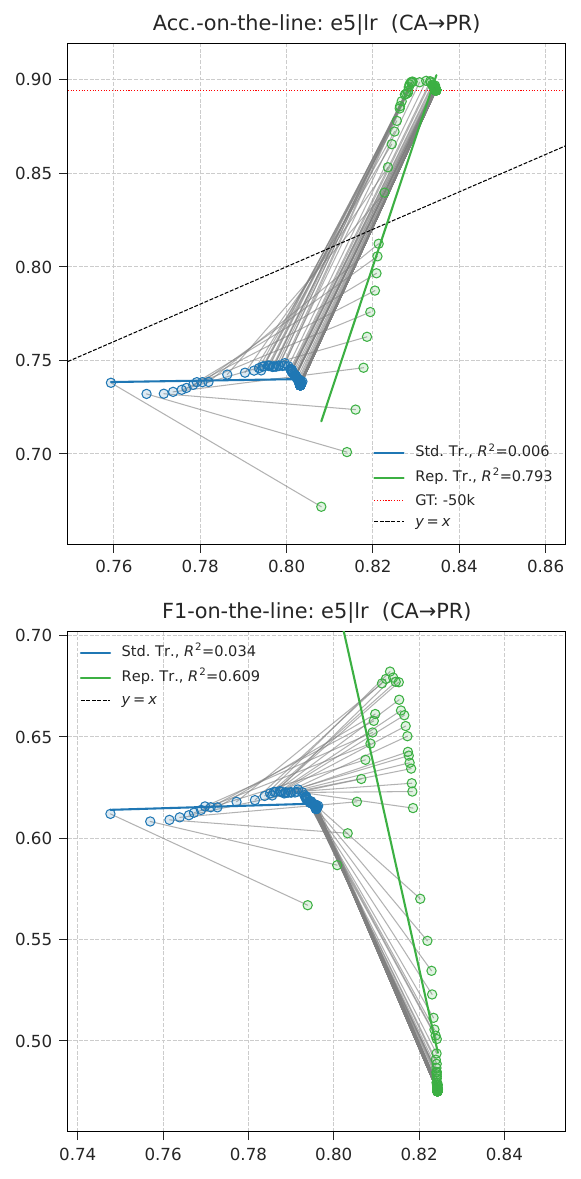}
\label{fig:pr-lr-e5}}
\subfloat[\texttt{Linq}]{\includegraphics[width=0.25\textwidth]{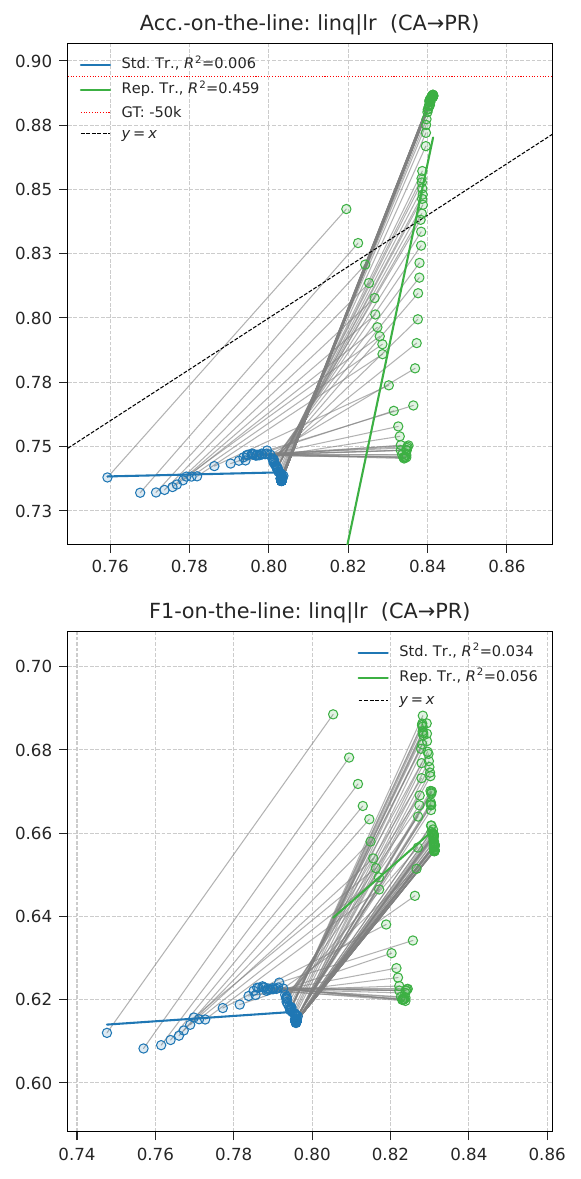}\label{fig:pr-lr-linq}}
\subfloat[\texttt{SFR}]{\includegraphics[width=0.25\textwidth]{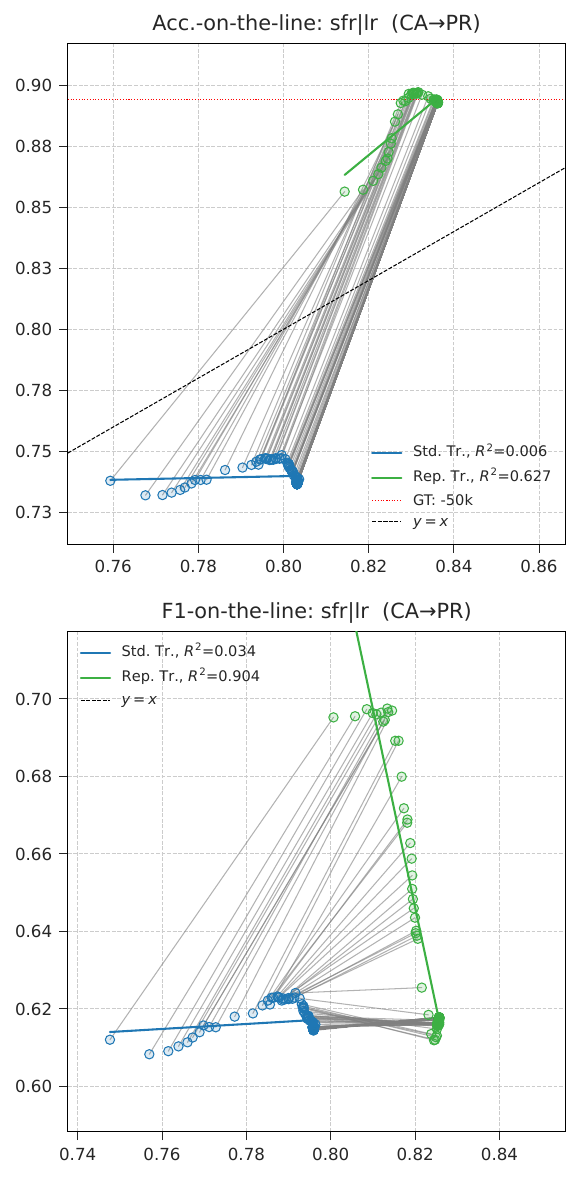}\label{fig:pr-lr-sfr}}
\subfloat[\texttt{Zeta}]{\includegraphics[width=0.25\textwidth]{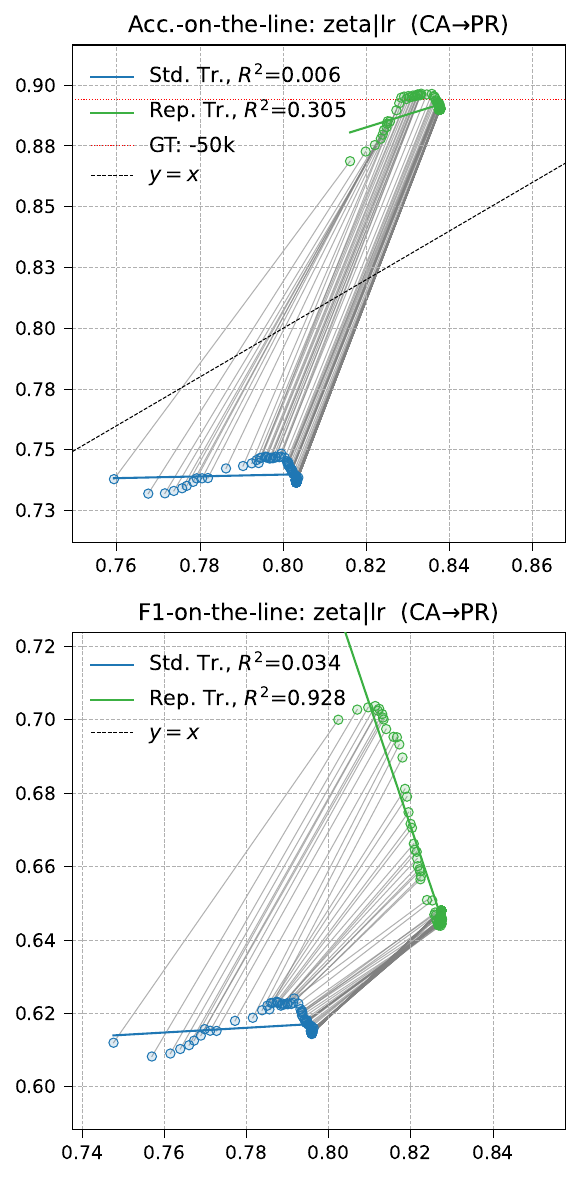}\label{fig:pr-lr-zeta}}
\caption{Pattern behaviour across different LLMs for LR.}
\label{fig:pr-lr}
\end{figure}

\begin{figure}[bp]
\centering
\subfloat[\texttt{e5}]{\includegraphics[width=0.25\textwidth]{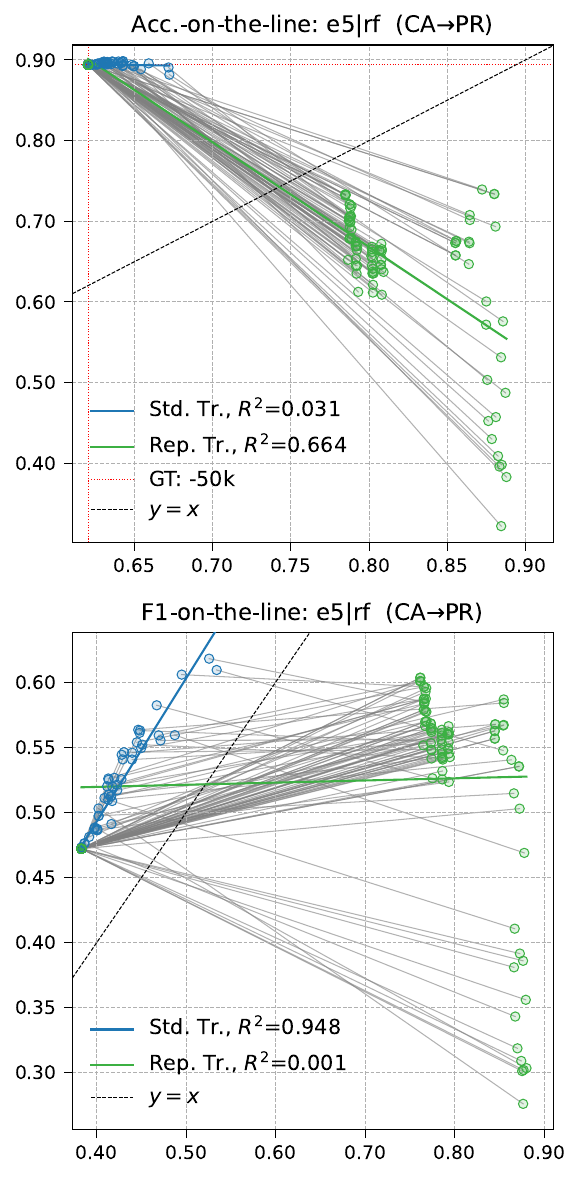}
\label{fig:pr-rf-e5}}
\subfloat[\texttt{Linq}]{\includegraphics[width=0.25\textwidth]{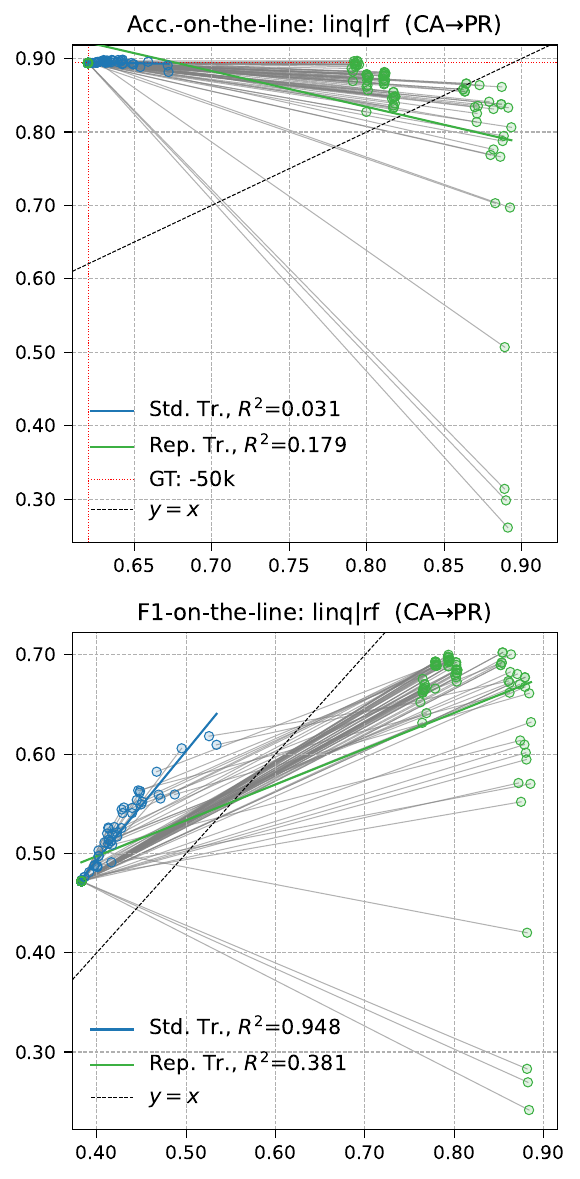}\label{fig:pr-rf-linq}}
\subfloat[\texttt{SFR}]{\includegraphics[width=0.25\textwidth]{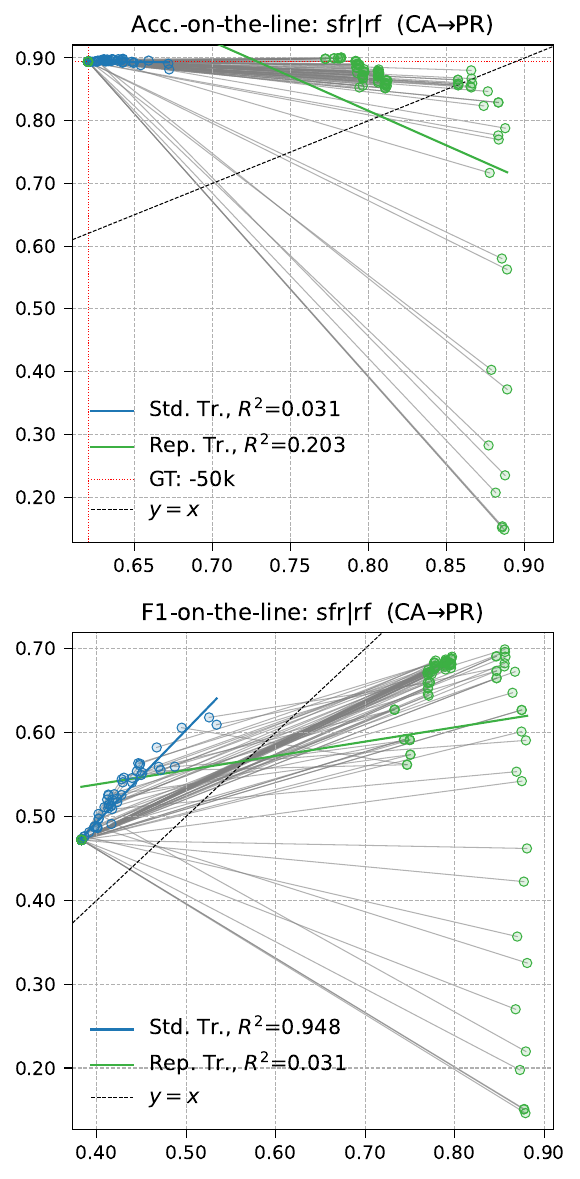}\label{fig:pr-rf-sfr}}
\subfloat[\texttt{Zeta}]{\includegraphics[width=0.25\textwidth]{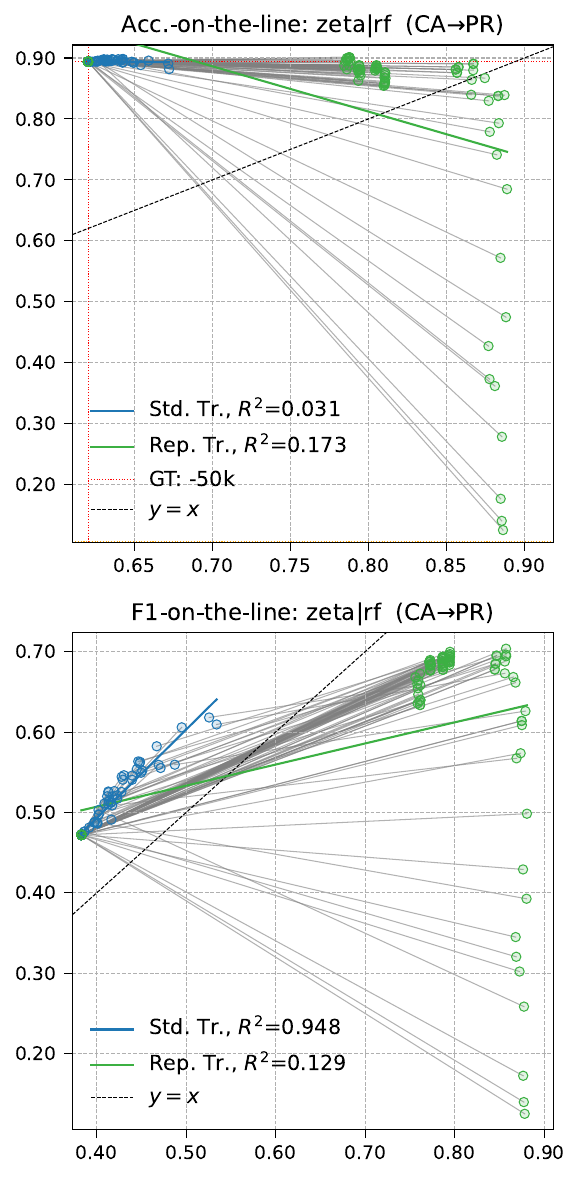}\label{fig:pr-rf-zeta}}
\caption{Pattern behaviour across different LLMs for RF.}
\label{fig:pr-rf}
\end{figure}


\begin{figure}[bp]
\centering
\subfloat[\texttt{e5}]{\includegraphics[width=0.25\textwidth]{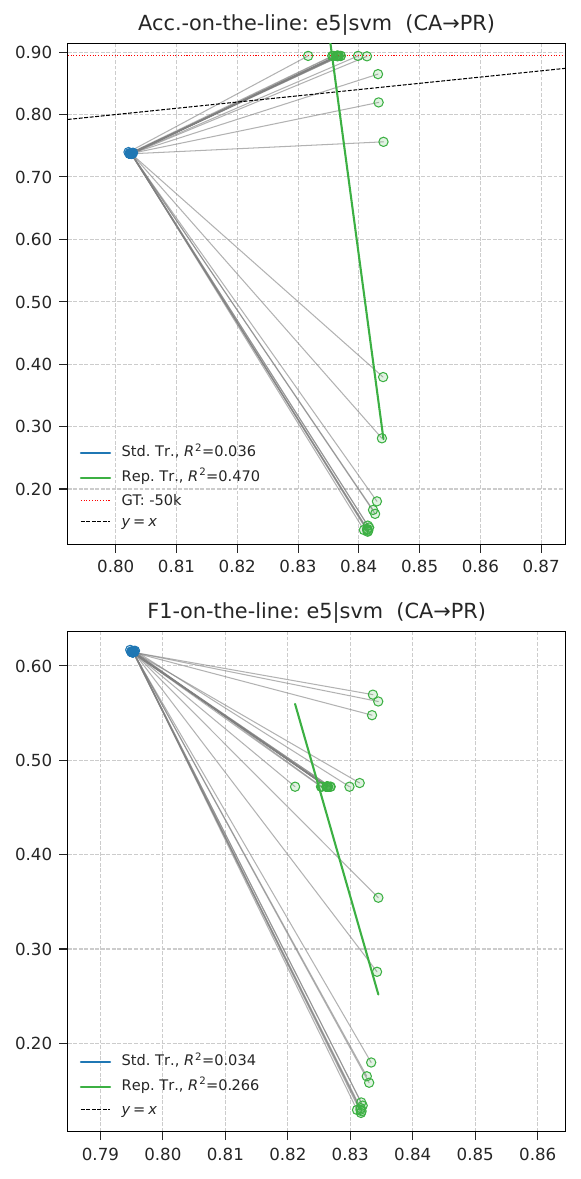}
\label{fig:pr-svm-e5}}
\subfloat[\texttt{Linq}]{\includegraphics[width=0.25\textwidth]{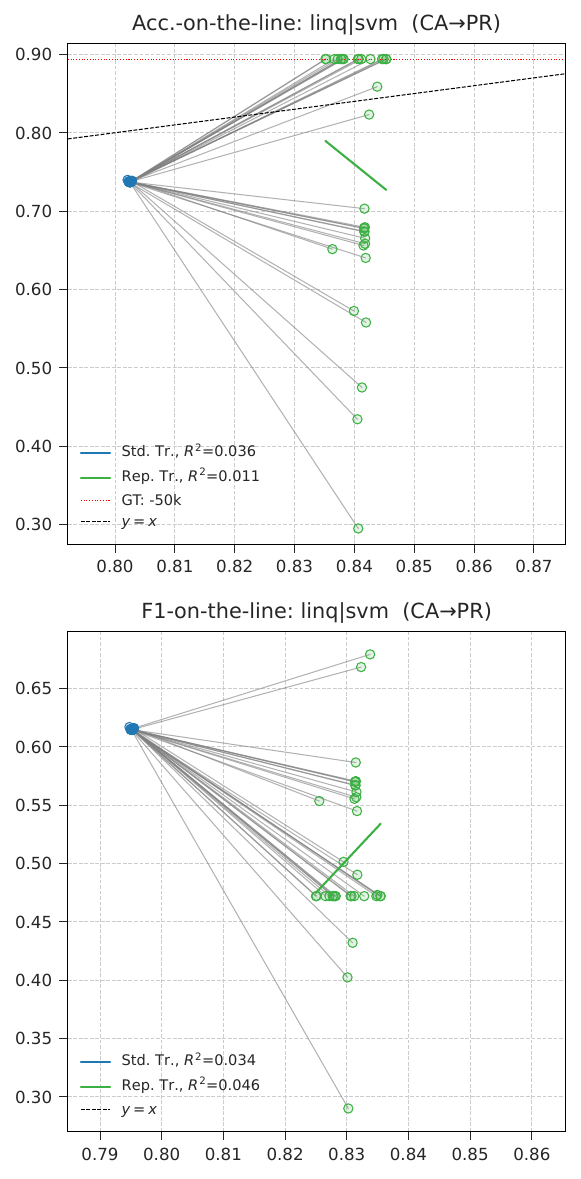}\label{fig:pr-svm-linq}}
\subfloat[\texttt{SFR}]{\includegraphics[width=0.25\textwidth]{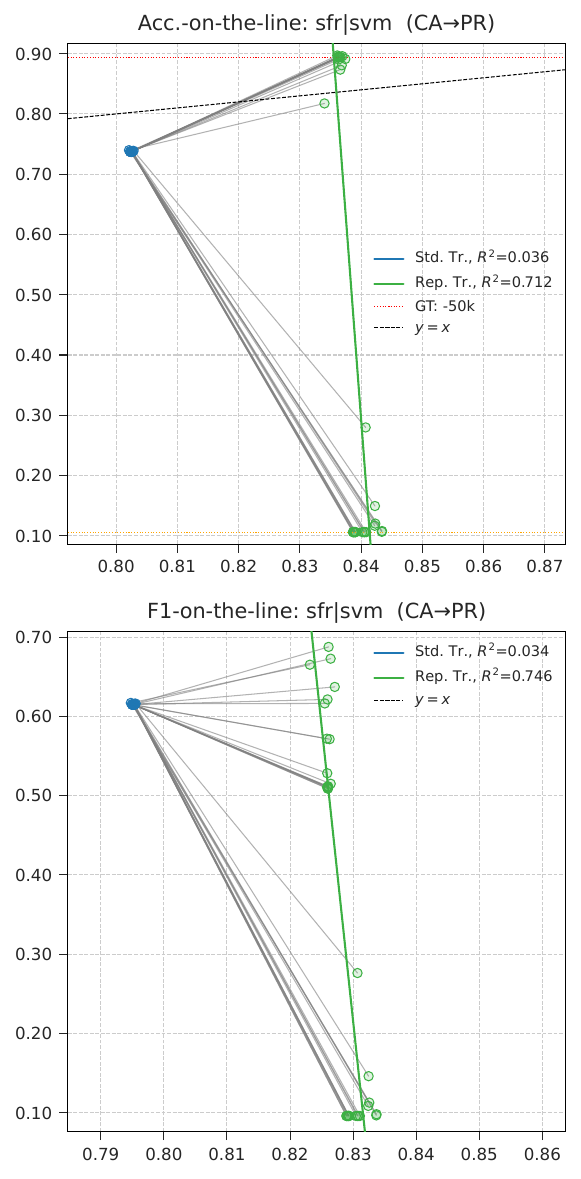}\label{fig:pr-svm-sfr}}
\subfloat[\texttt{Zeta}]{\includegraphics[width=0.25\textwidth]{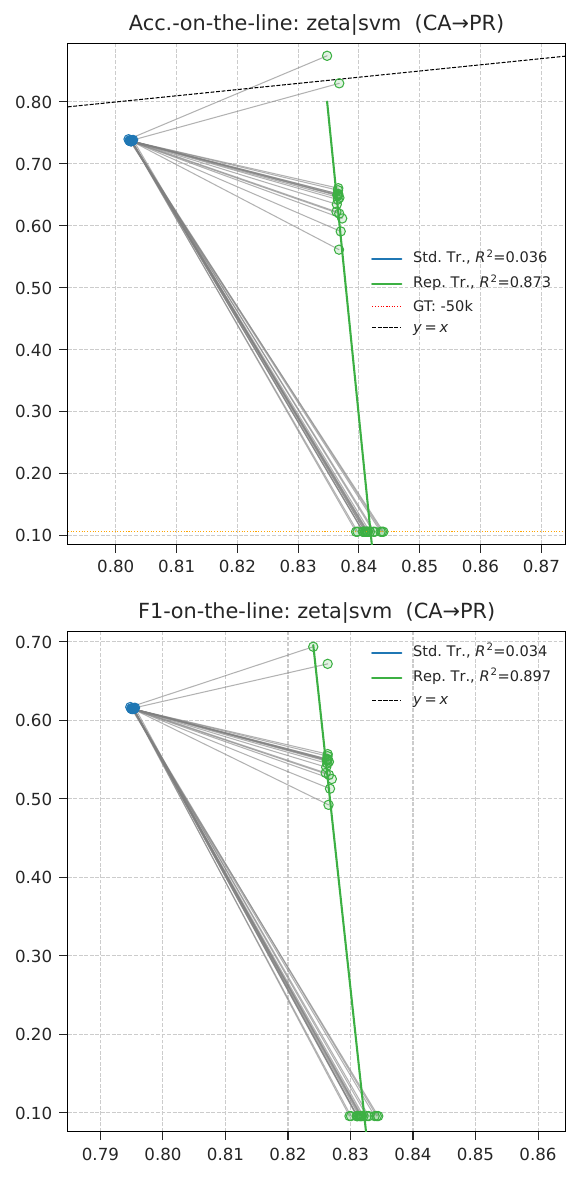}\label{fig:pr-svm-zeta}}
\caption{Pattern behaviour across different LLMs for SVM.}
\label{fig:pr-svm}
\end{figure}

\begin{figure}[bp]
\centering
\subfloat[\texttt{e5}]{\includegraphics[width=0.25\textwidth]{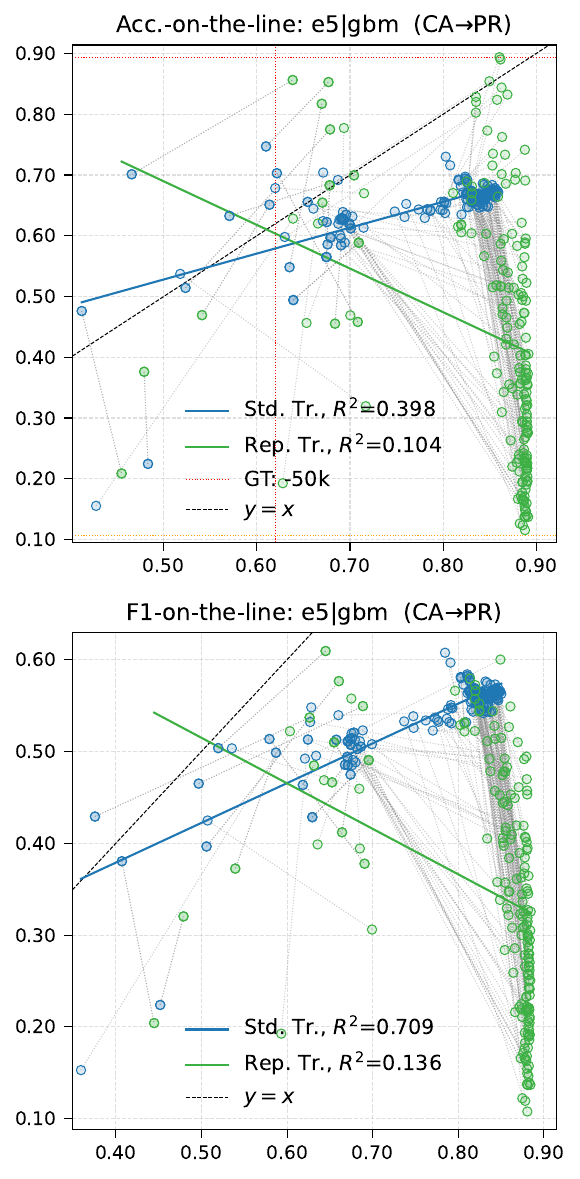}
\label{fig:pr-gbm-e5}}
\subfloat[\texttt{Linq}]{\includegraphics[width=0.25\textwidth]{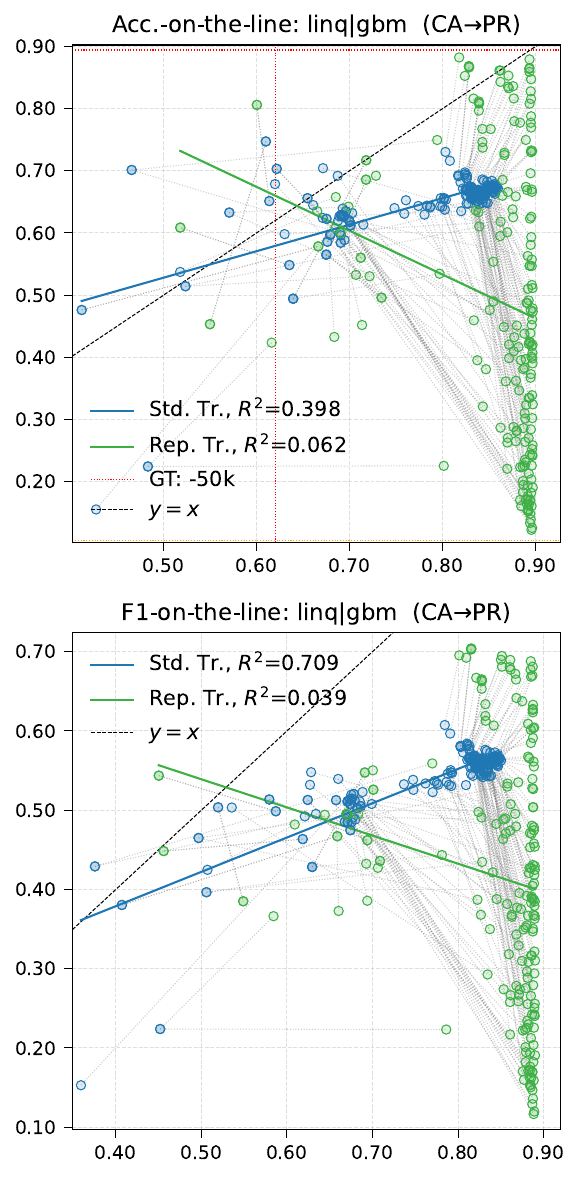}\label{fig:pr-gbm-linq}}
\subfloat[\texttt{SFR}]{\includegraphics[width=0.25\textwidth]{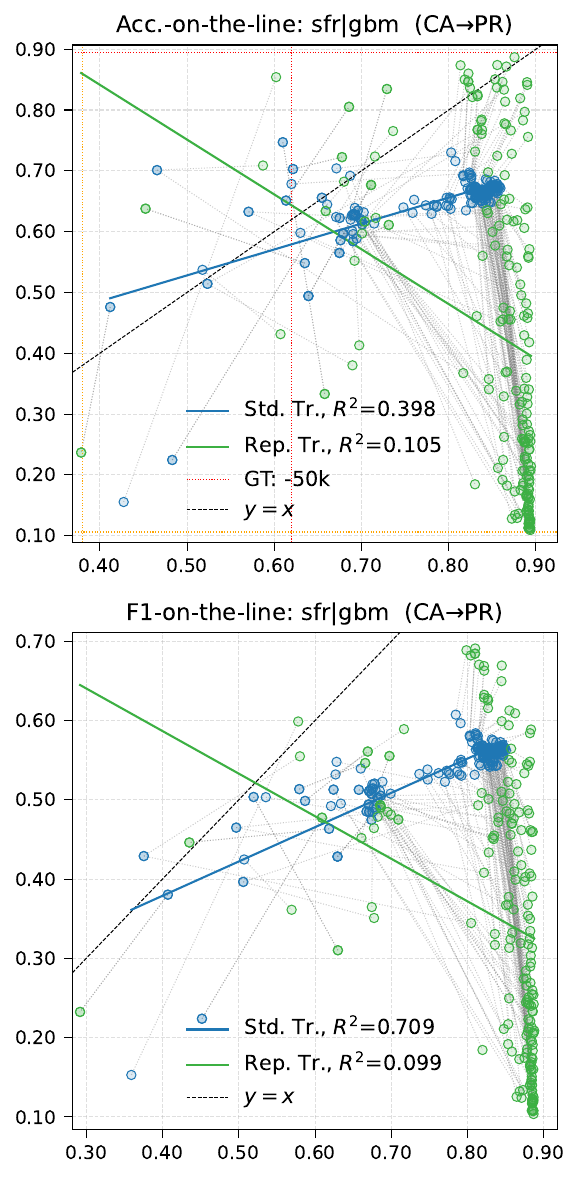}\label{fig:pr-gbm-sfr}}
\subfloat[\texttt{Zeta}]{\includegraphics[width=0.25\textwidth]{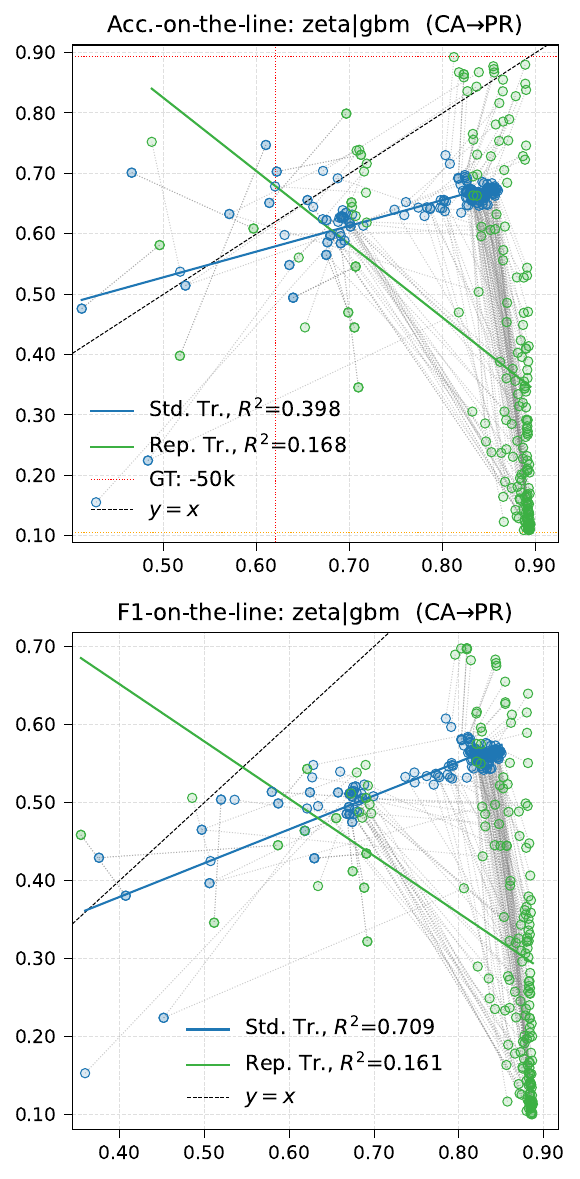}\label{fig:pr-gbm-zeta}}
\caption{Pattern behaviour across different LLMs for GBM.}
\label{fig:pr-gbm}
\end{figure}


\begin{figure}[bp]
\centering
\subfloat[\texttt{e5}]{\includegraphics[width=0.25\textwidth]{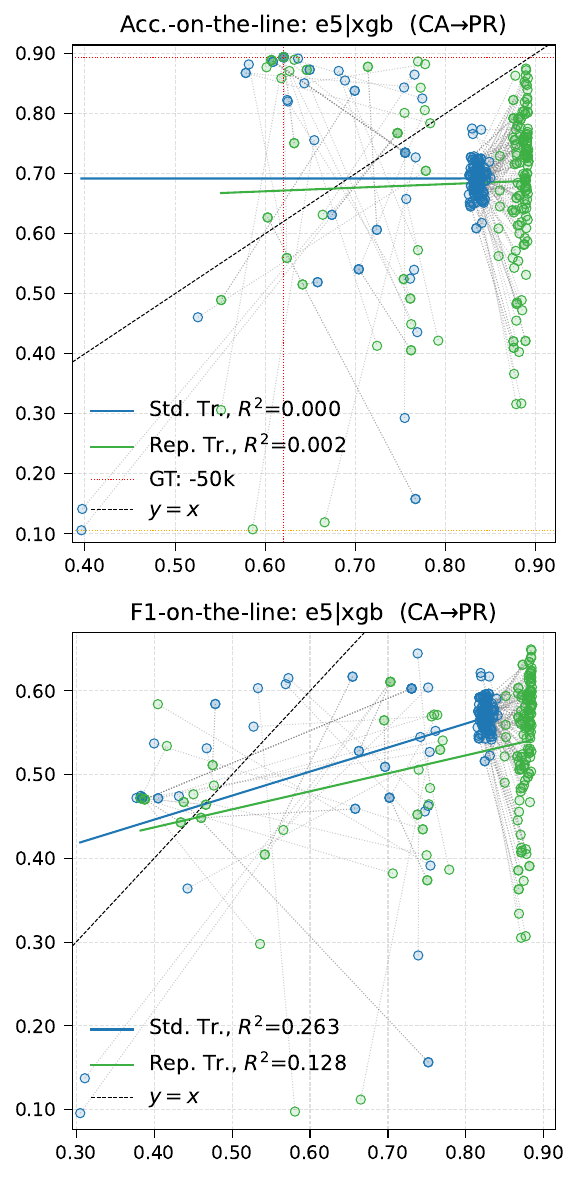}
\label{fig:pr-xgb-e5}}
\subfloat[\texttt{Linq}]{\includegraphics[width=0.25\textwidth]{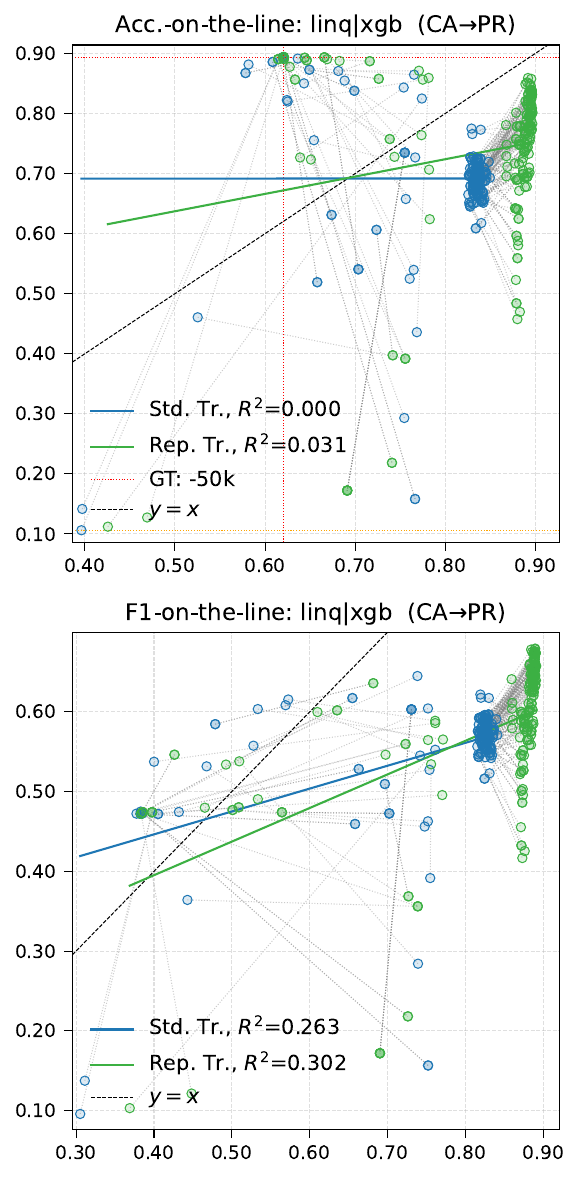}\label{fig:pr-xgb-linq}}
\subfloat[\texttt{SFR}]{\includegraphics[width=0.25\textwidth]{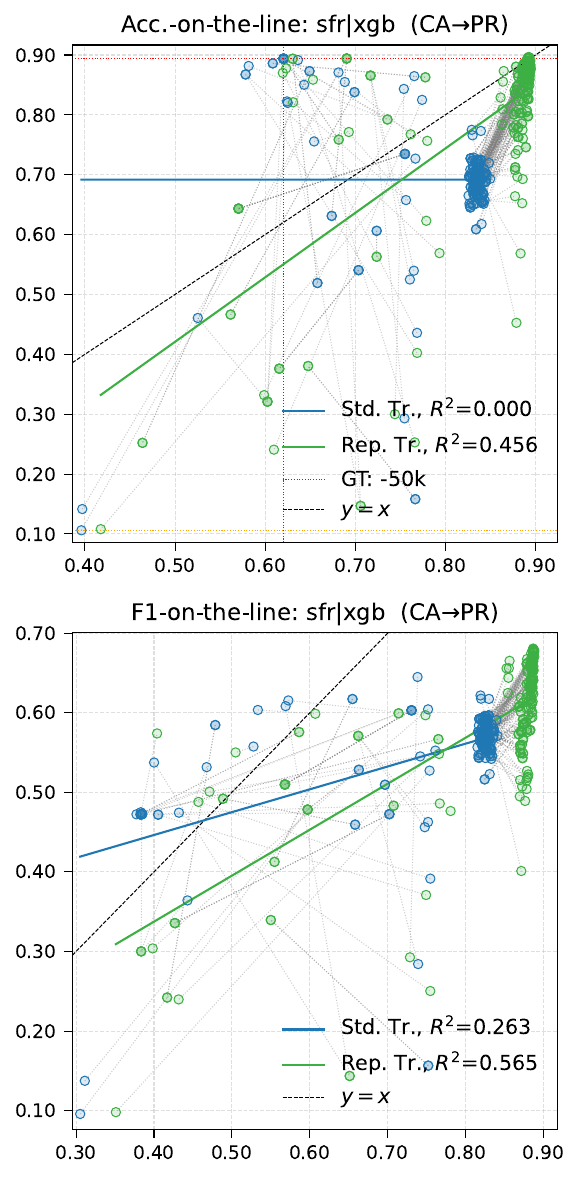}\label{fig:pr-xgb-sfr}}
\subfloat[\texttt{Zeta}]{\includegraphics[width=0.25\textwidth]{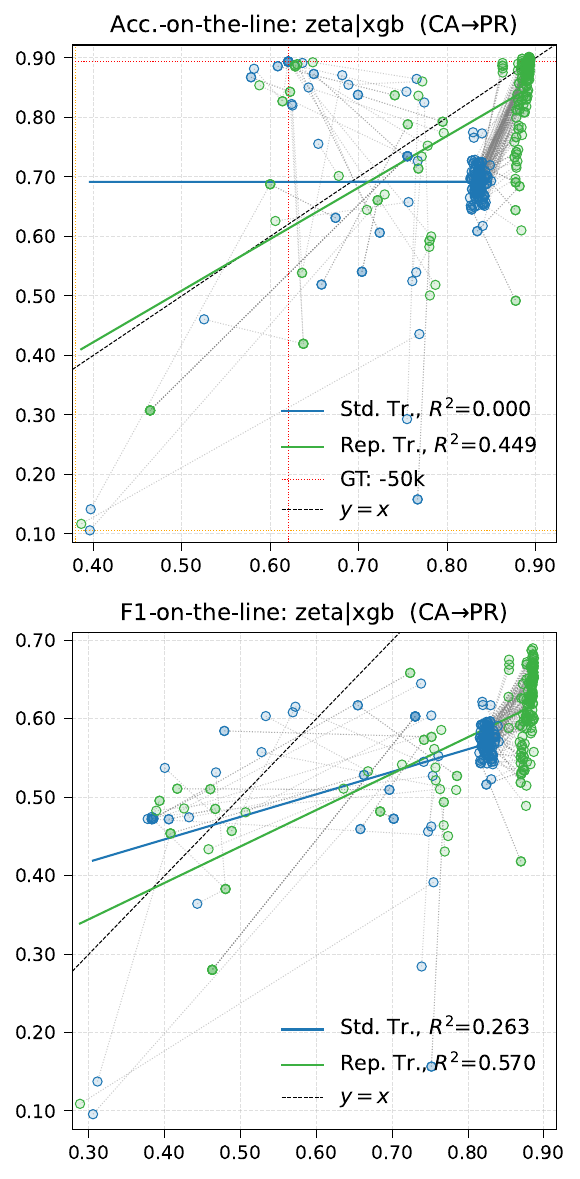}\label{fig:pr-xgb-zeta}}
\caption{Pattern behaviour across different LLMs for XGB.}
\label{fig:pr-xgb}
\end{figure}

\clearpage

\paragraph{Target State: Alabama} 

\begin{figure}[bp!]
\centering
\subfloat[\texttt{e5}]{\includegraphics[width=0.25\textwidth]{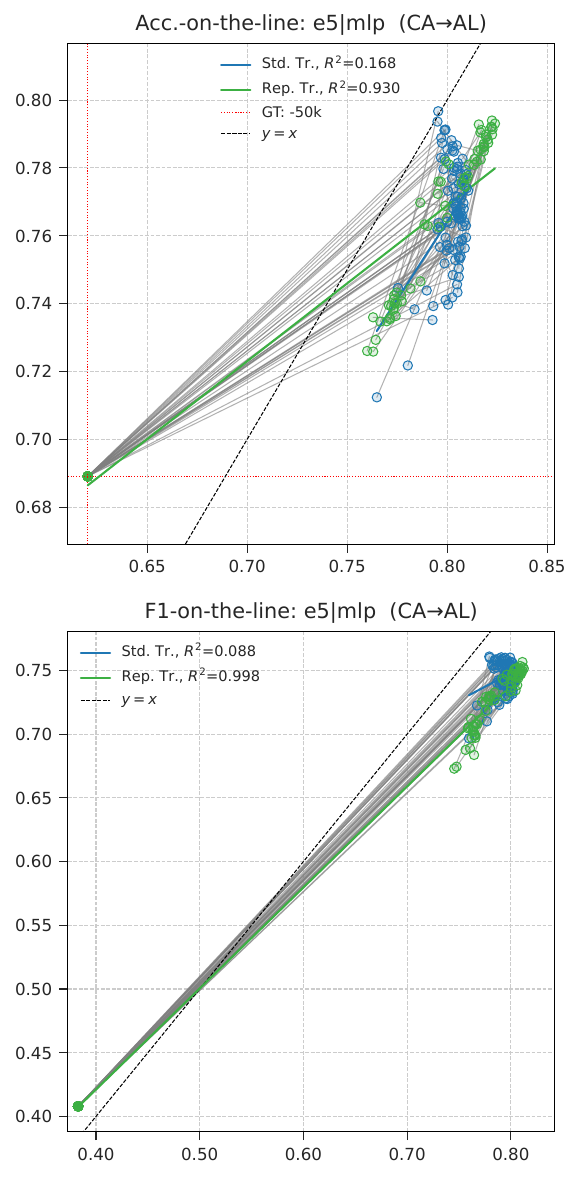}
\label{fig:al-mlp-e5}}
\subfloat[\texttt{Linq}]{\includegraphics[width=0.25\textwidth]{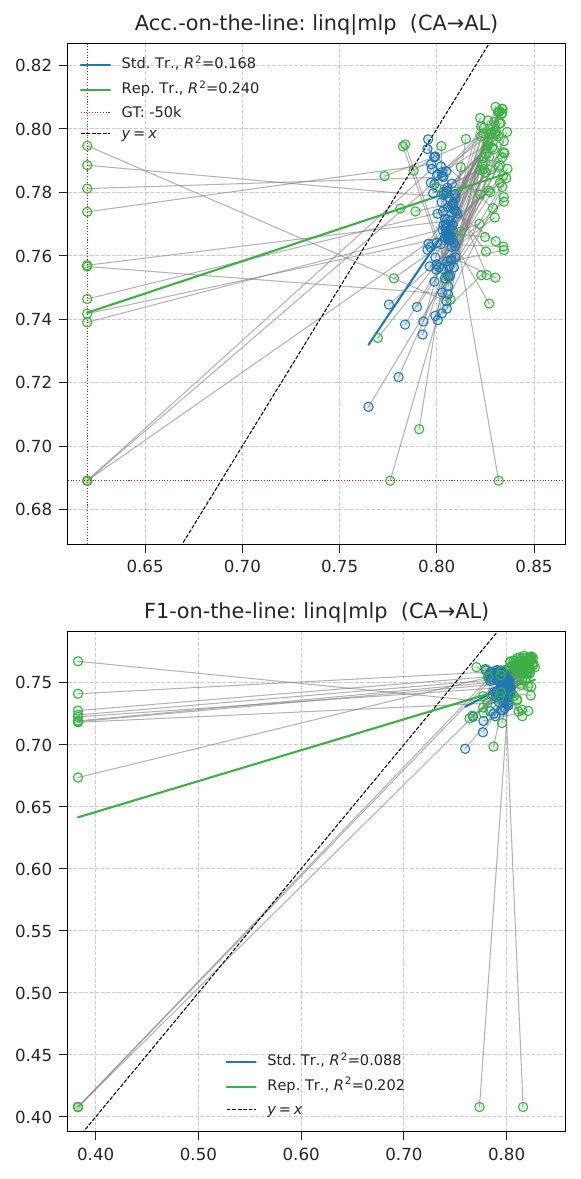}\label{fig:al-mlp-linq}}
\subfloat[\texttt{SFR}]{\includegraphics[width=0.25\textwidth]{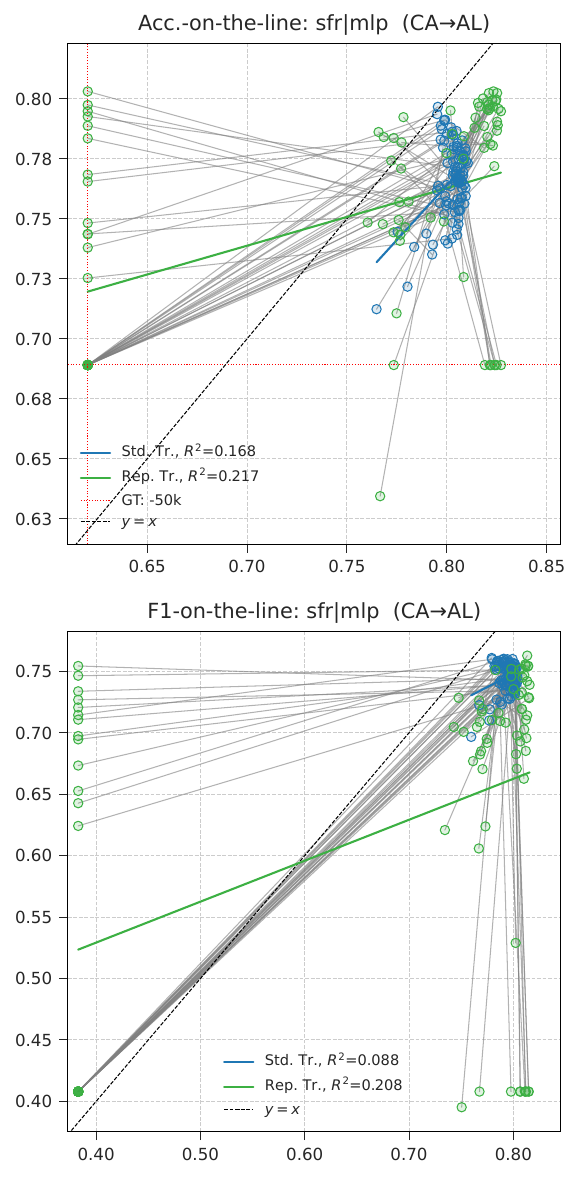}\label{fig:al-mlp-sfr}}
\subfloat[\texttt{Zeta}]{\includegraphics[width=0.25\textwidth]{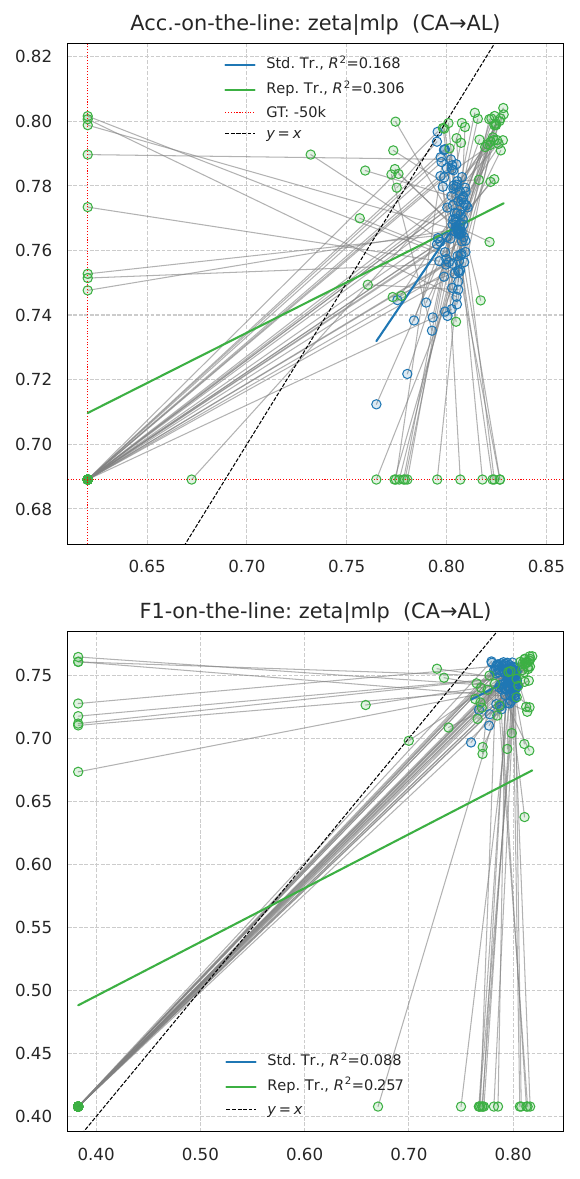}\label{fig:al-mlp-zeta}}
\caption{Pattern behaviour across different LLMs for MLP.}
\label{fig:al-mlp}
\end{figure}

\begin{figure}[bp!]
\centering
\subfloat[\texttt{e5}]{\includegraphics[width=0.25\textwidth]{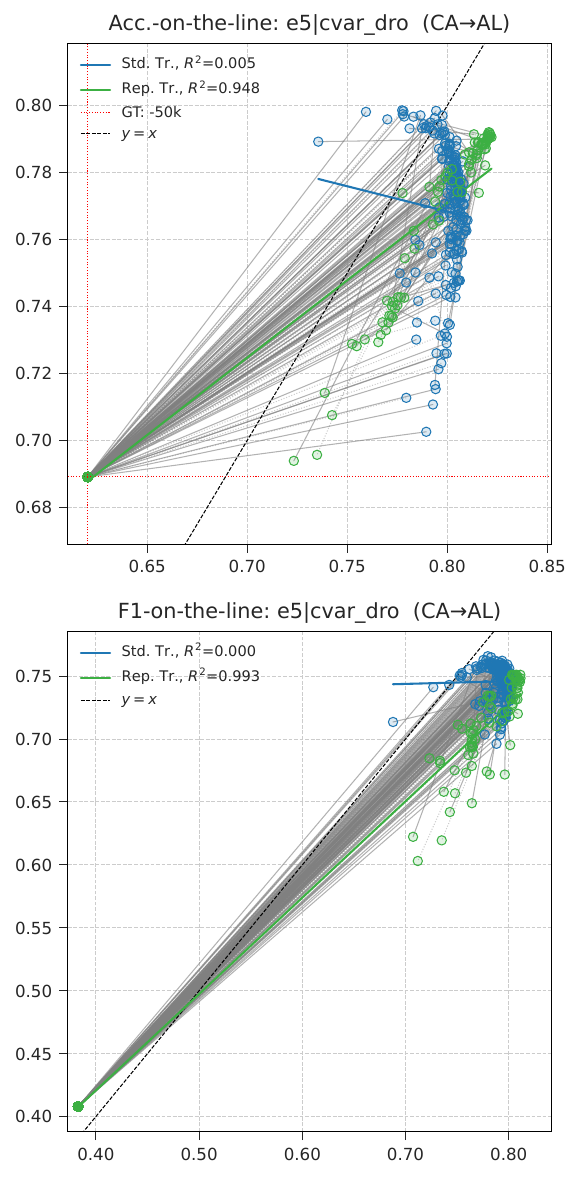}
\label{fig:al-cvar_dro-e5}}
\subfloat[\texttt{Linq}]{\includegraphics[width=0.25\textwidth]{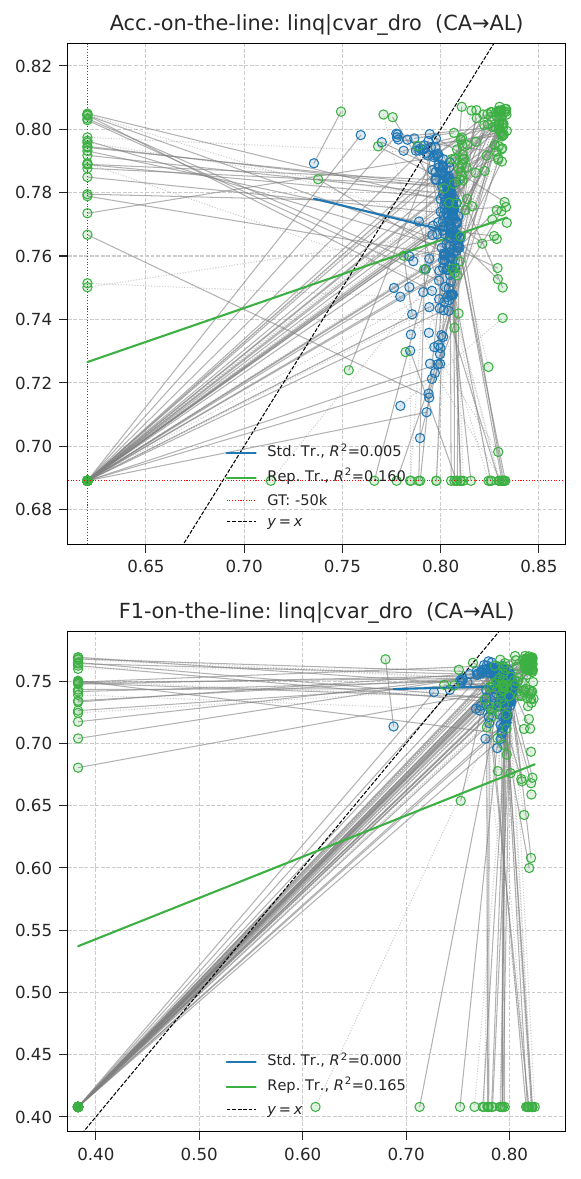}\label{fig:al-cvar_dro-linq}}
\subfloat[\texttt{SFR}]{\includegraphics[width=0.25\textwidth]{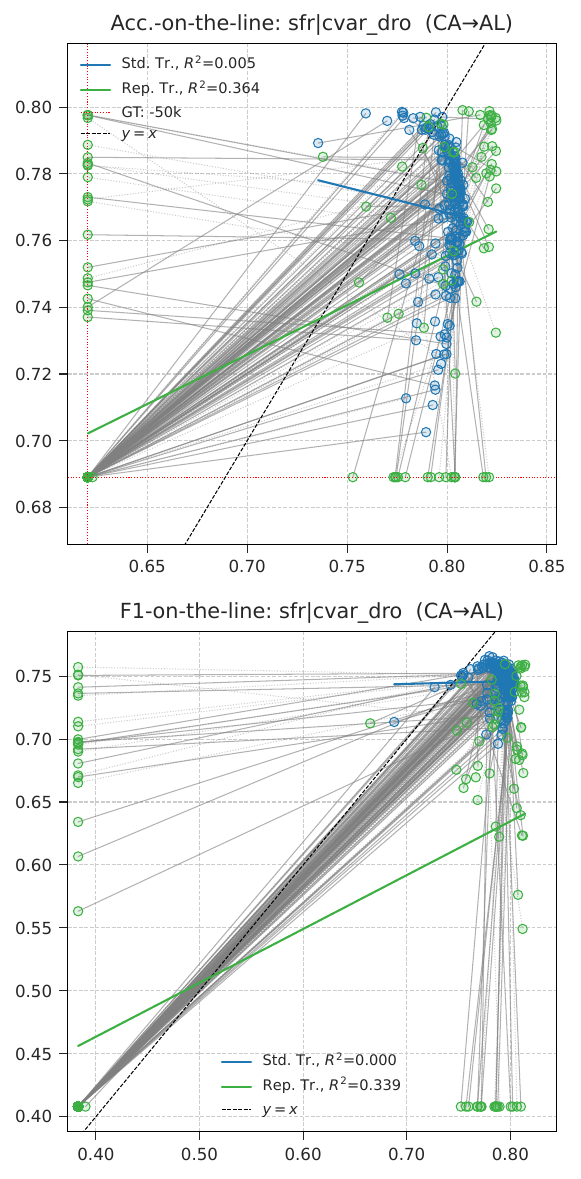}\label{fig:al-cvar_dro-sfr}}
\subfloat[\texttt{Zeta}]{\includegraphics[width=0.25\textwidth]{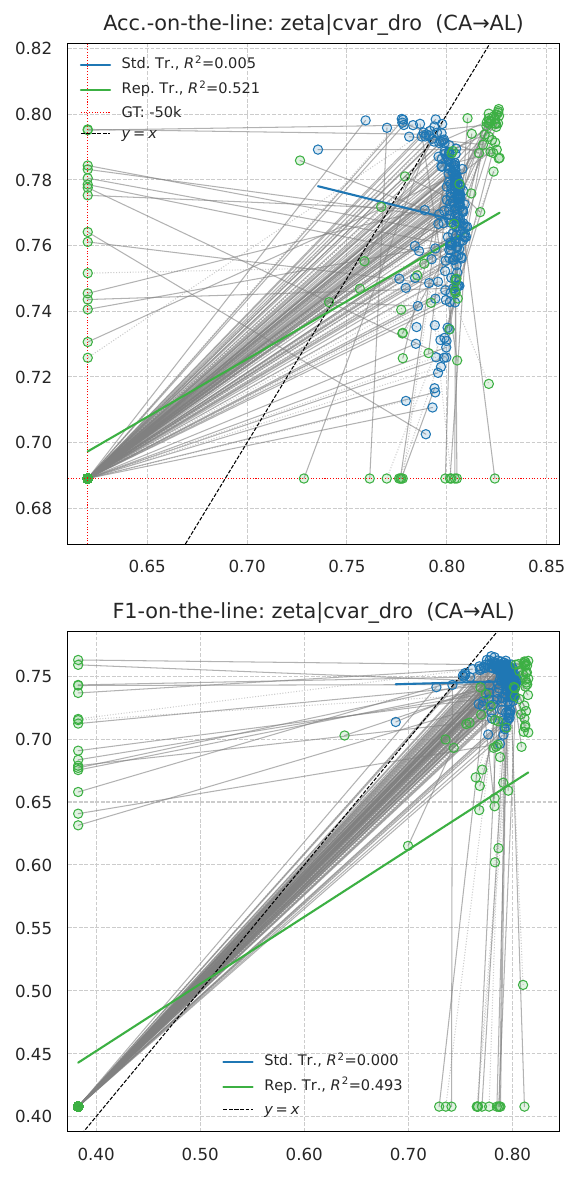}\label{fig:al-cvar_dro-zeta}}
\caption{Pattern behaviour across different LLMs for CVaR-DRO.}
\label{fig:al-cvar_dro}
\end{figure}

\begin{figure}[bp]
\centering
\subfloat[\texttt{e5}]{\includegraphics[width=0.25\textwidth]{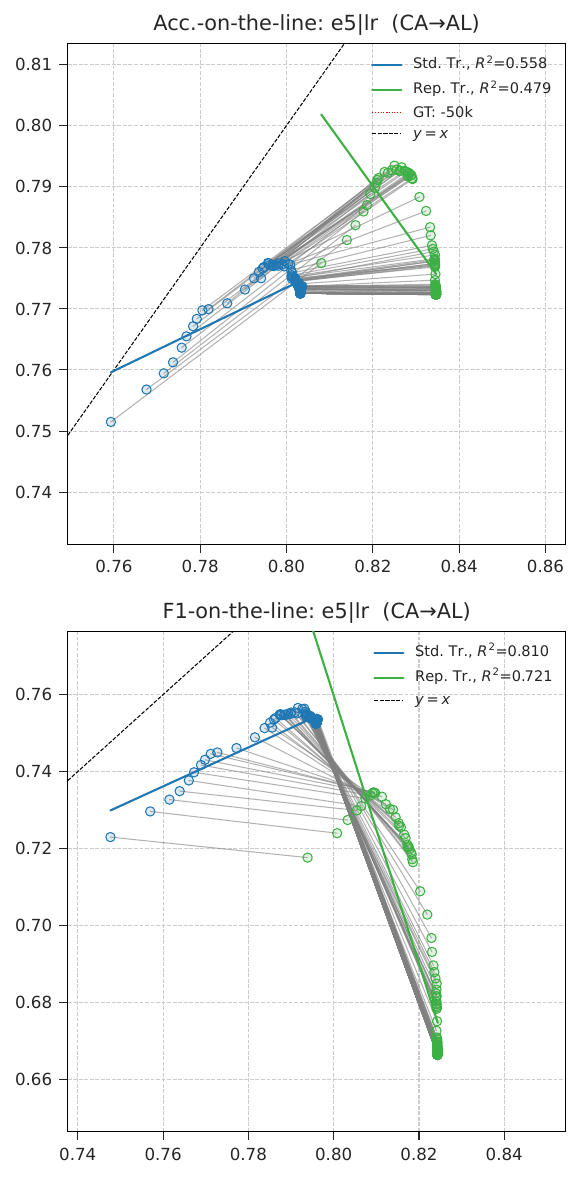}
\label{fig:al-lr-e5}}
\subfloat[\texttt{Linq}]{\includegraphics[width=0.25\textwidth]{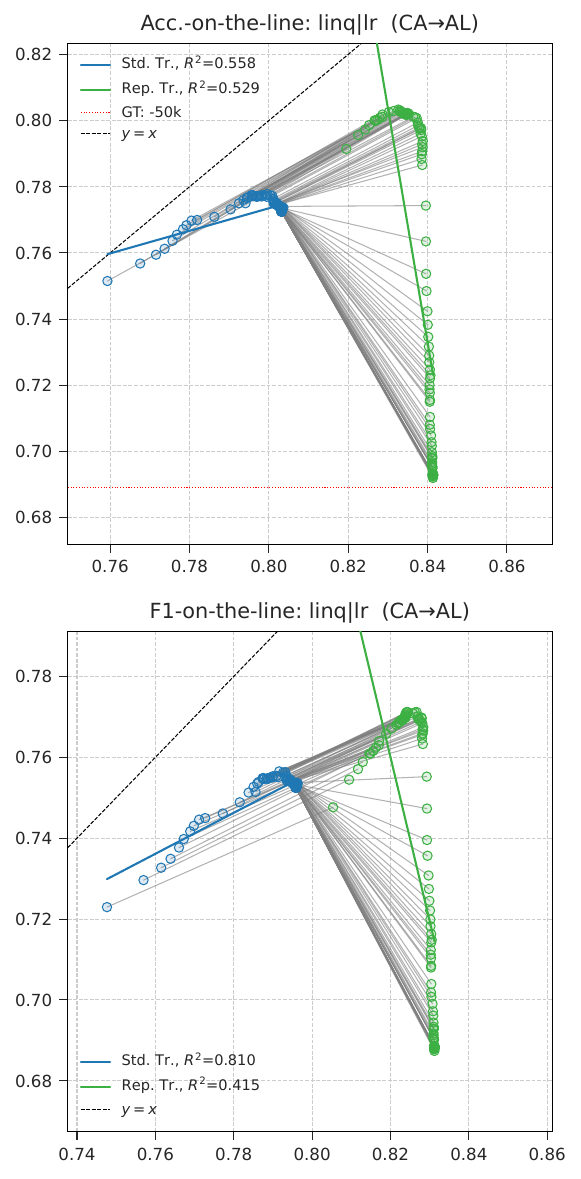}\label{fig:al-lr-linq}}
\subfloat[\texttt{SFR}]{\includegraphics[width=0.25\textwidth]{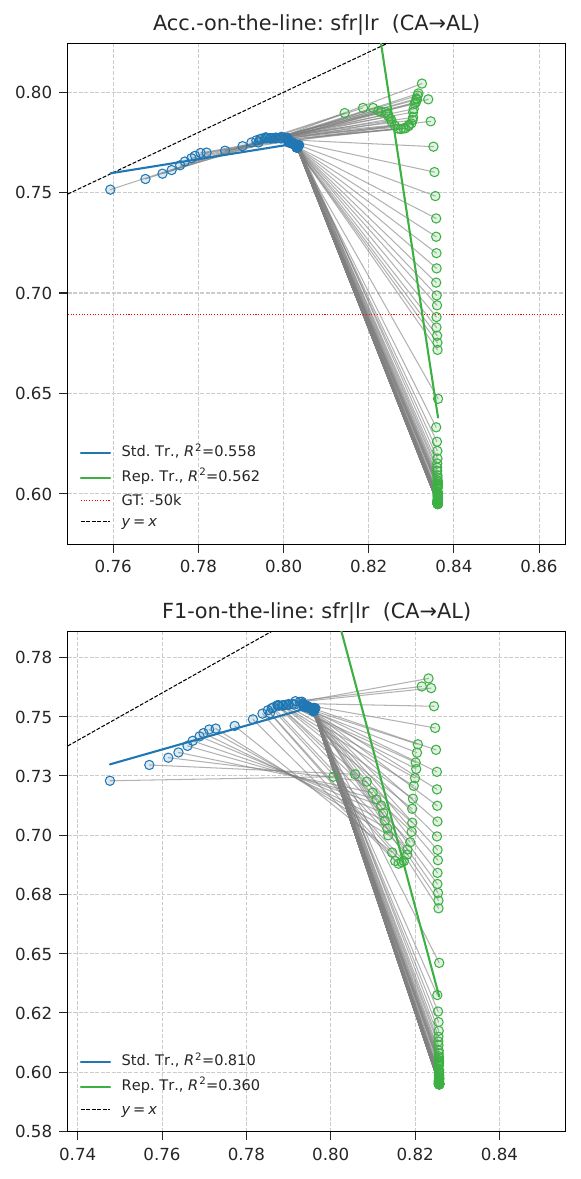}\label{fig:al-lr-sfr}}
\subfloat[\texttt{Zeta}]{\includegraphics[width=0.25\textwidth]{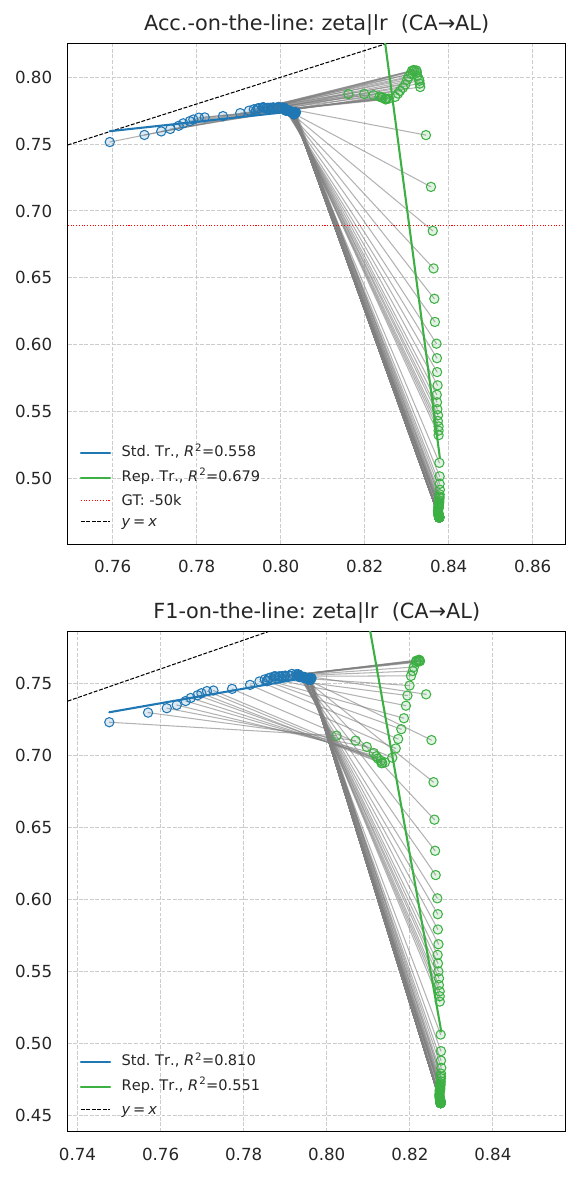}}
\caption{Pattern behaviour across different LLMs for LR.}
\label{fig:al-lr}
\end{figure}

\begin{figure}[bp]
\centering
\subfloat[\texttt{e5}]{\includegraphics[width=0.25\textwidth]{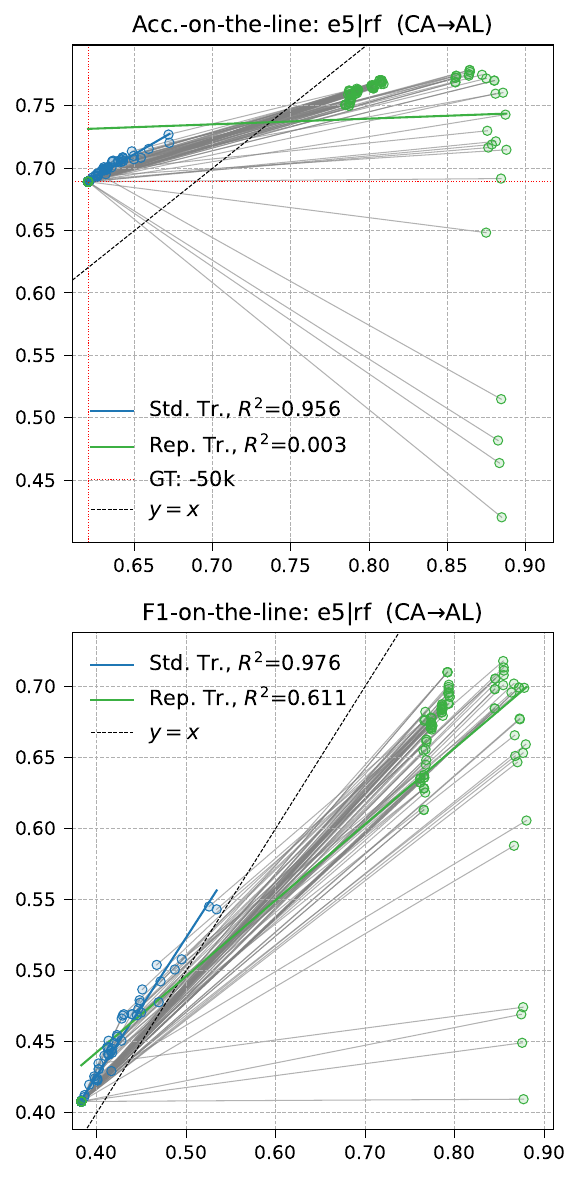}
\label{fig:al-rf-e5}}
\subfloat[\texttt{Linq}]{\includegraphics[width=0.25\textwidth]{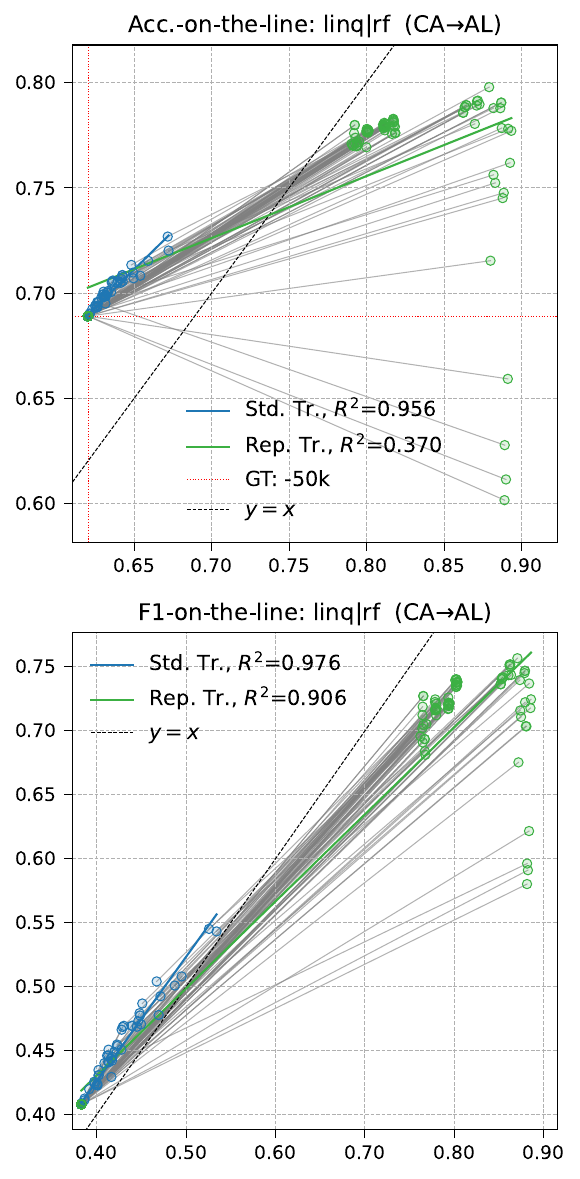}\label{fig:al-rf-linq}}
\subfloat[\texttt{SFR}]{\includegraphics[width=0.25\textwidth]{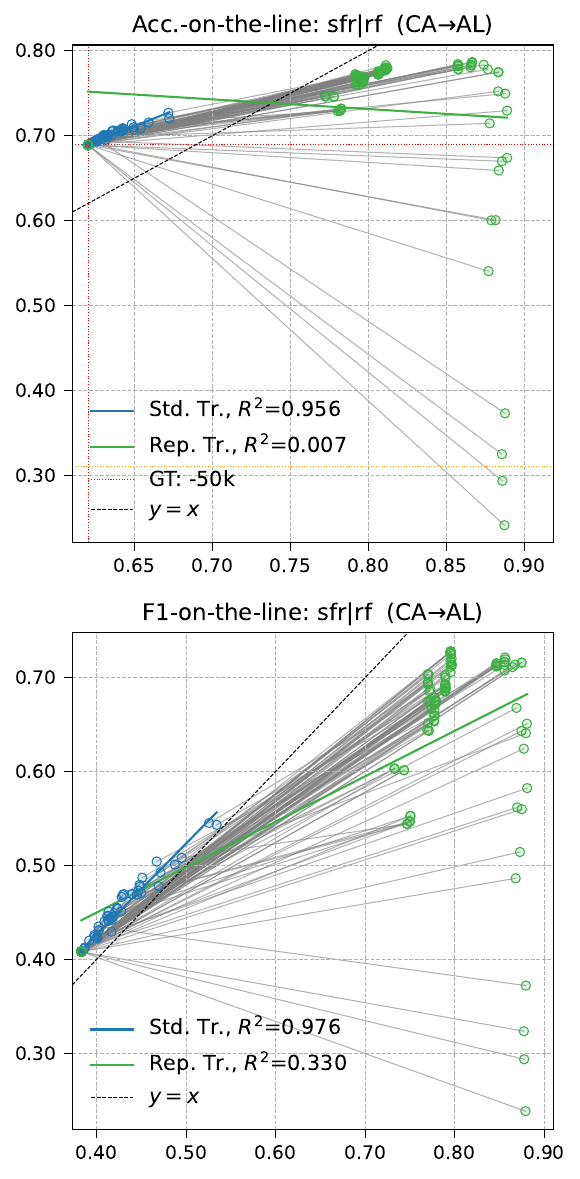}\label{fig:al-rf-sfr}}
\subfloat[\texttt{Zeta}]{\includegraphics[width=0.25\textwidth]{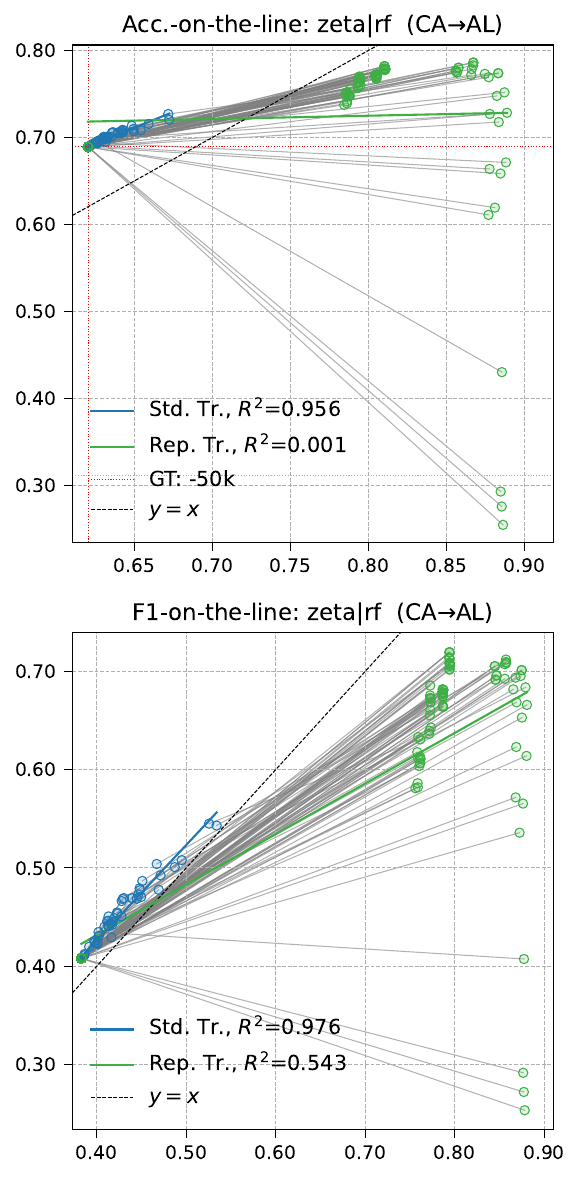}\label{fig:al-rf-zeta}}
\caption{Pattern behaviour across different LLMs for RF.}
\label{fig:al-rf}
\end{figure}

\begin{figure}[bp]
\centering
\subfloat[\texttt{e5}]{\includegraphics[width=0.25\textwidth]{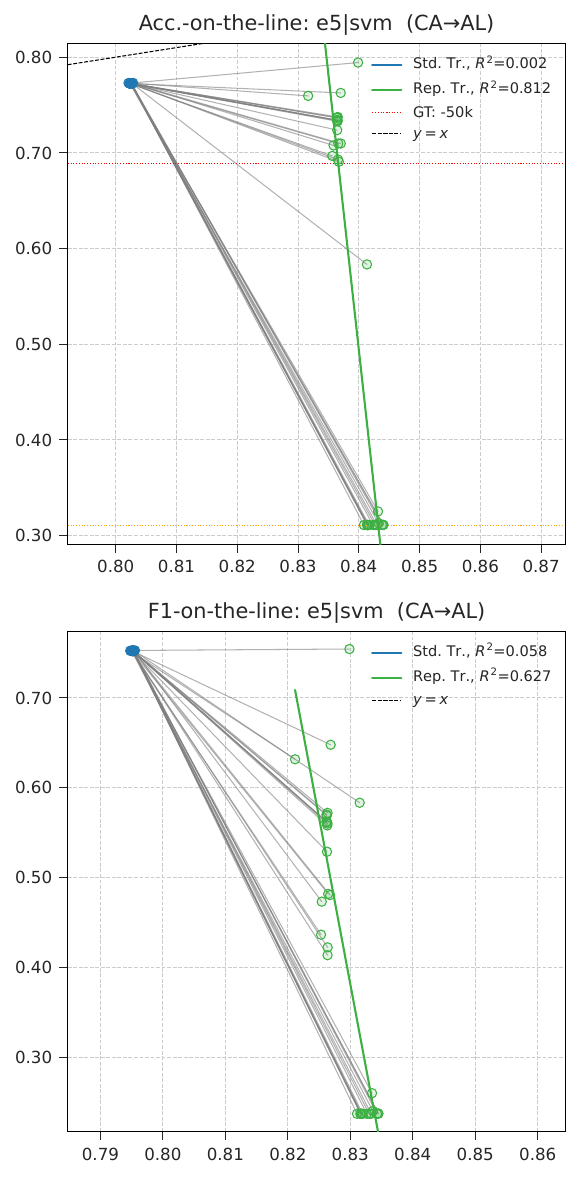}
\label{fig:al-svm-e5}}
\subfloat[\texttt{Linq}]{\includegraphics[width=0.25\textwidth]{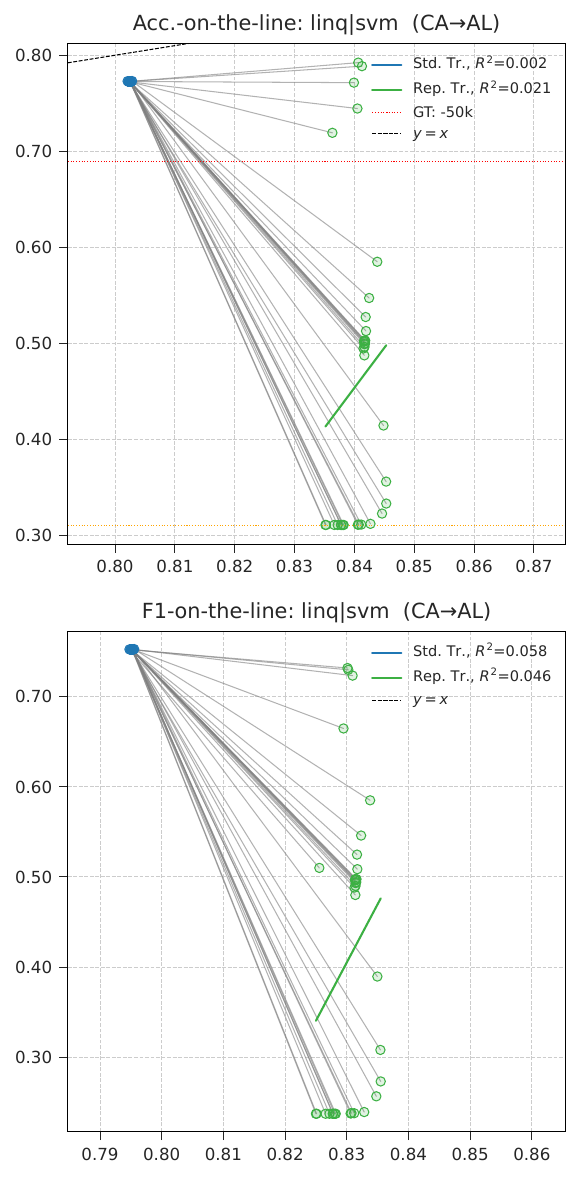}\label{fig:al-svm-linq}}
\subfloat[\texttt{SFR}]{\includegraphics[width=0.25\textwidth]{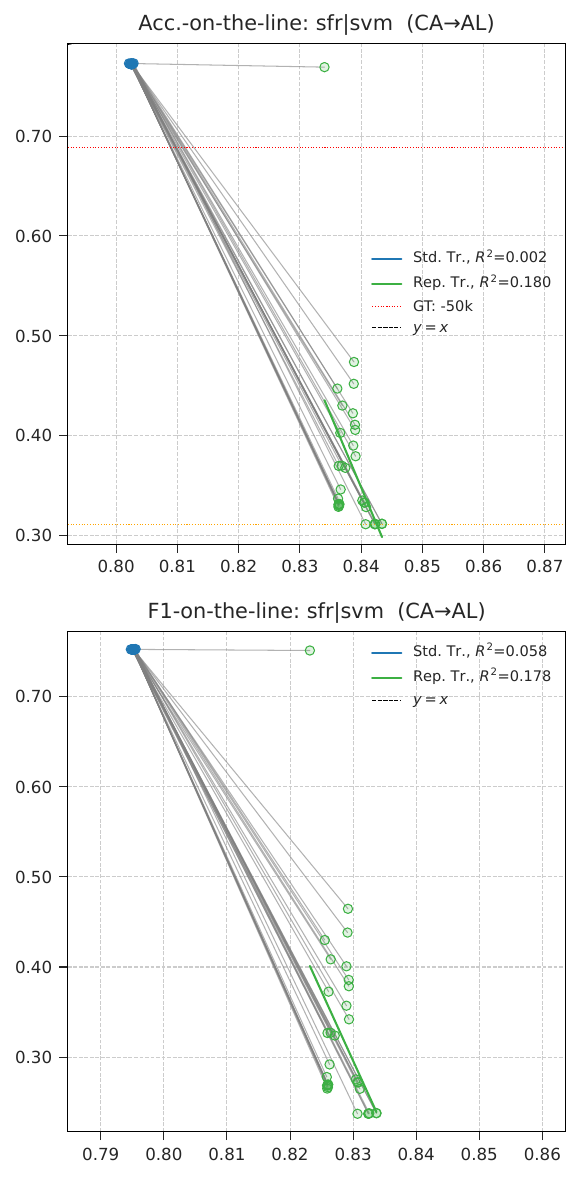}\label{fig:al-svm-sfr}}
\subfloat[\texttt{Zeta}]{\includegraphics[width=0.25\textwidth]{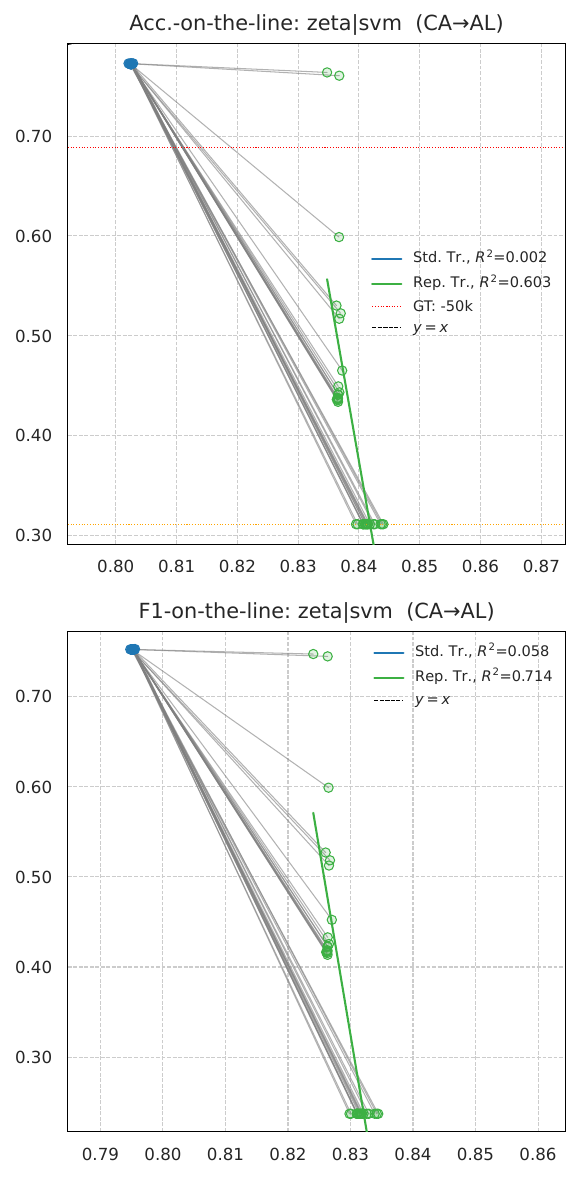}\label{fig:al-svm-zeta}}
\caption{Pattern behaviour across different LLMs for SVM.}
\label{fig:al-svm}
\end{figure}

\begin{figure}[bp]
\centering
\subfloat[\texttt{e5}]{\includegraphics[width=0.25\textwidth]{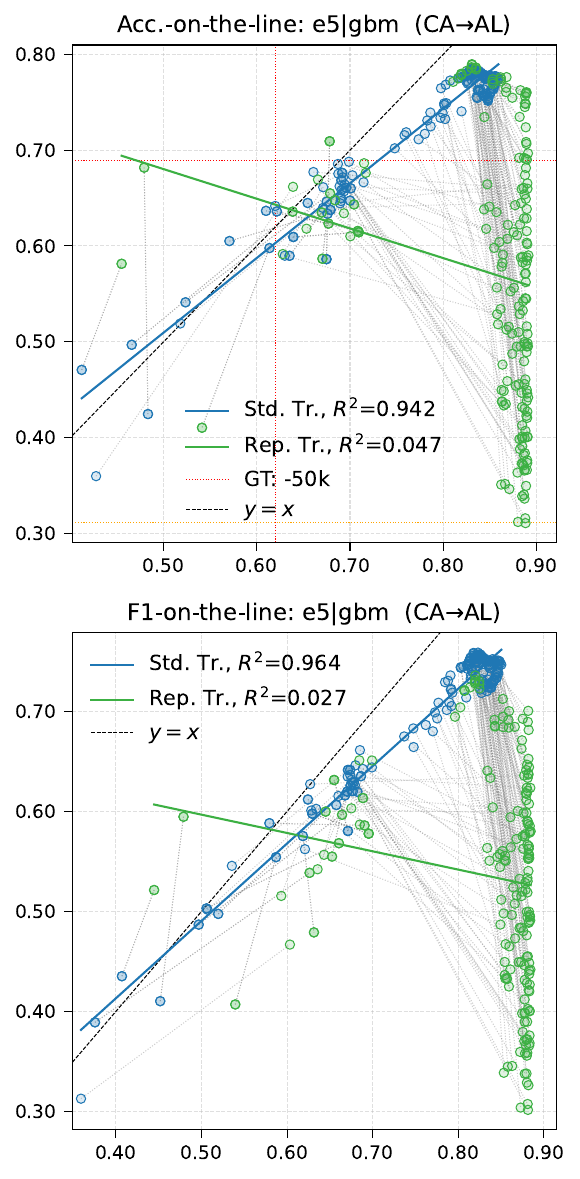}
\label{fig:al-gbm-e5}}
\subfloat[\texttt{Linq}]{\includegraphics[width=0.25\textwidth]{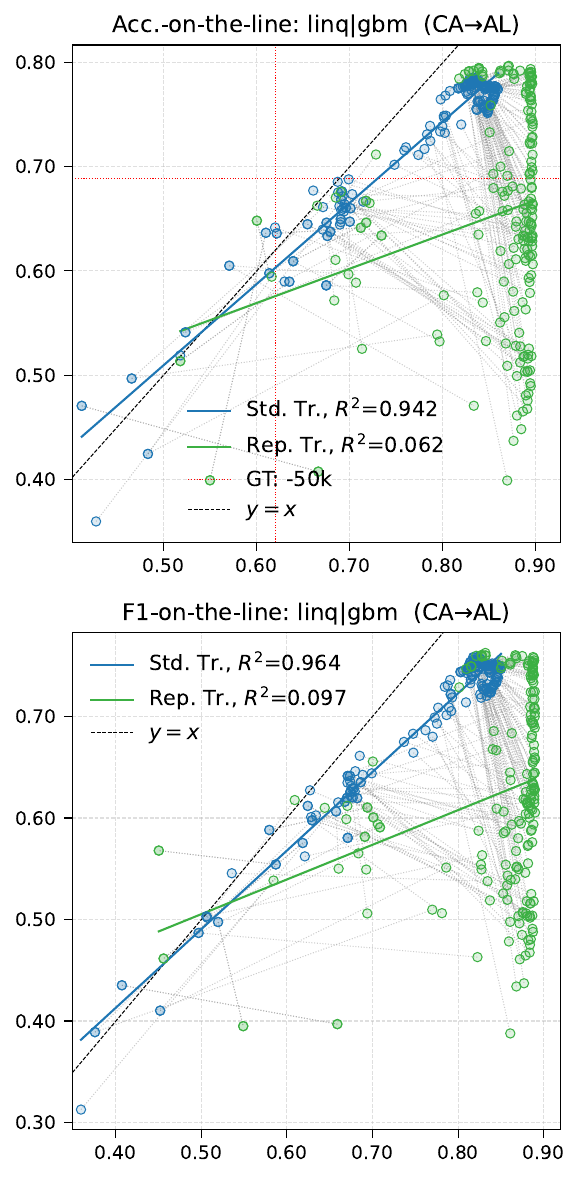}\label{fig:al-gbm-linq}}
\subfloat[\texttt{SFR}]{\includegraphics[width=0.25\textwidth]{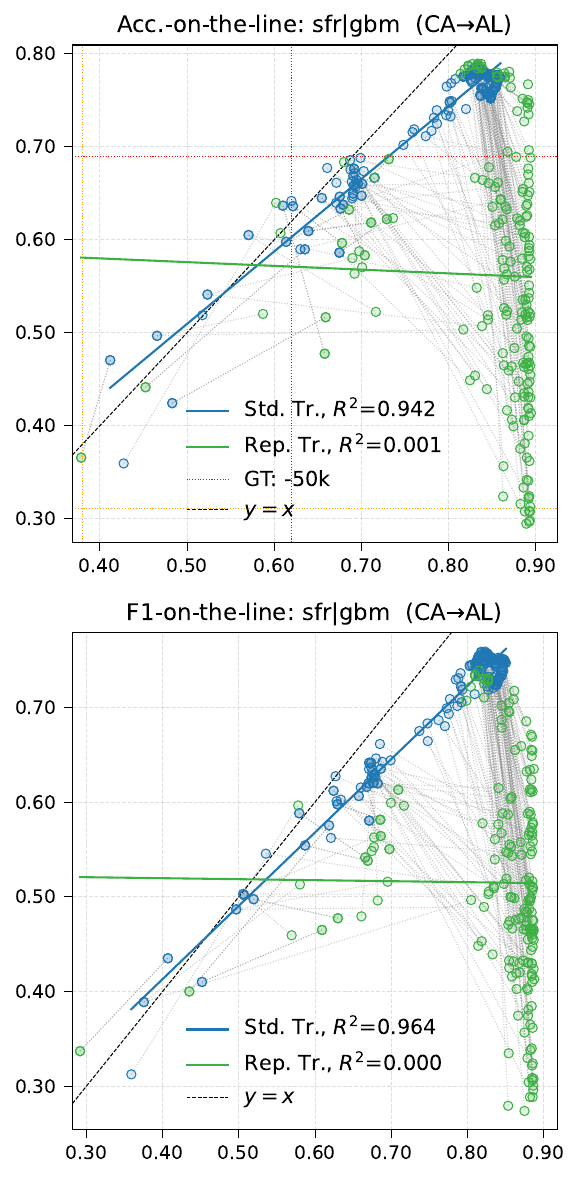}\label{fig:al-gbm-sfr}}
\subfloat[\texttt{Zeta}]{\includegraphics[width=0.25\textwidth]{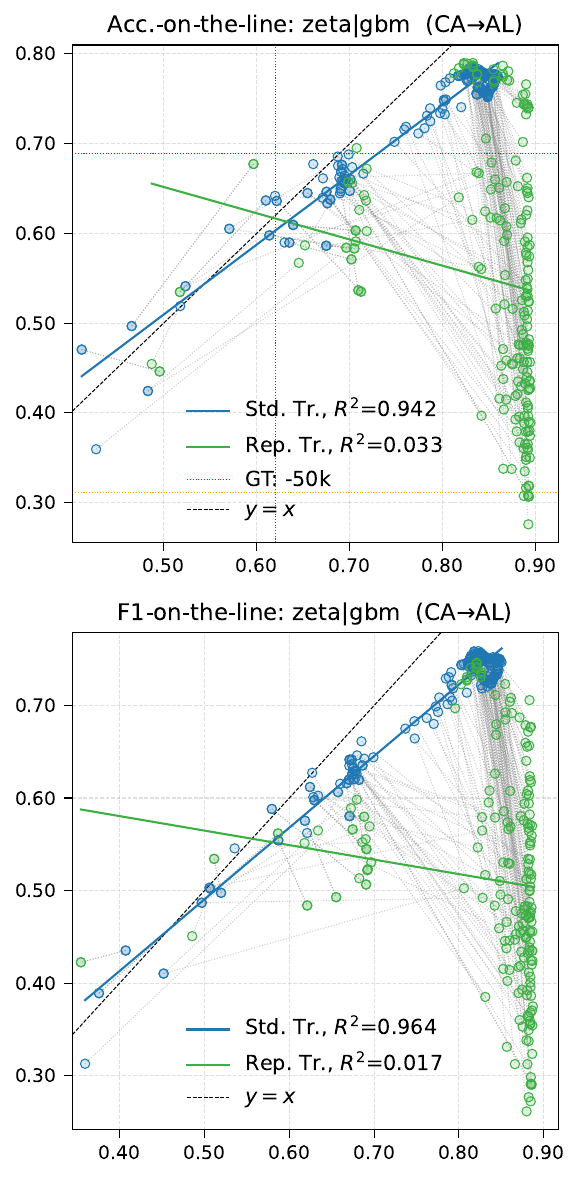}\label{fig:al-gbm-zeta}}
\caption{Pattern behaviour across different LLMs for GBM.}
\label{fig:al-gbm}
\end{figure}

\begin{figure}[bp]
\centering
\subfloat[\texttt{e5}]{\includegraphics[width=0.25\textwidth]{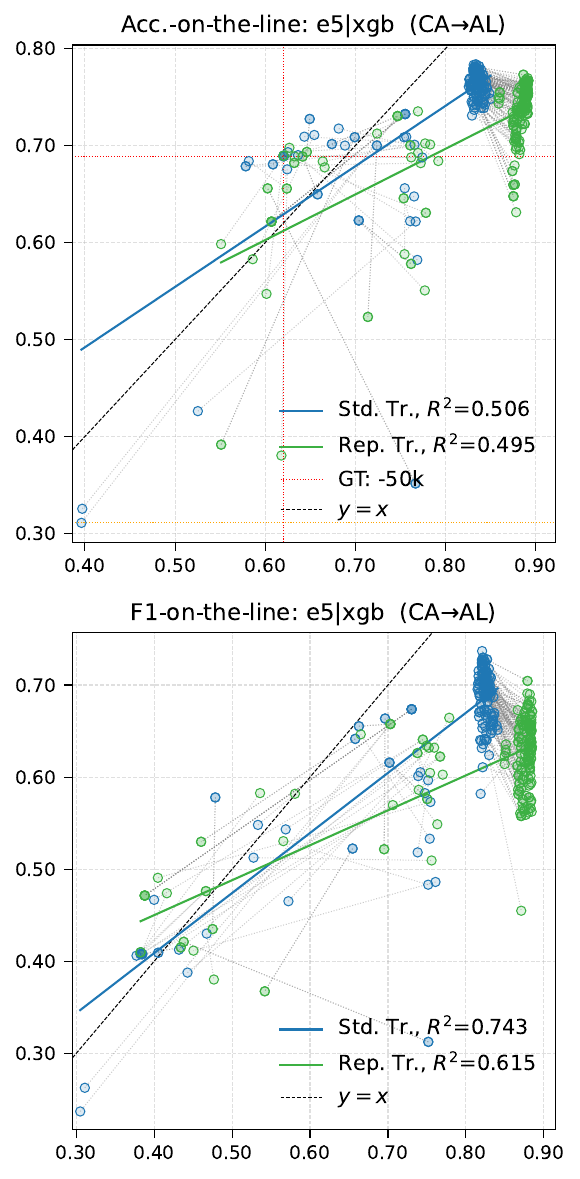}
\label{fig:al-xgb-e5}}
\subfloat[\texttt{Linq}]{\includegraphics[width=0.25\textwidth]{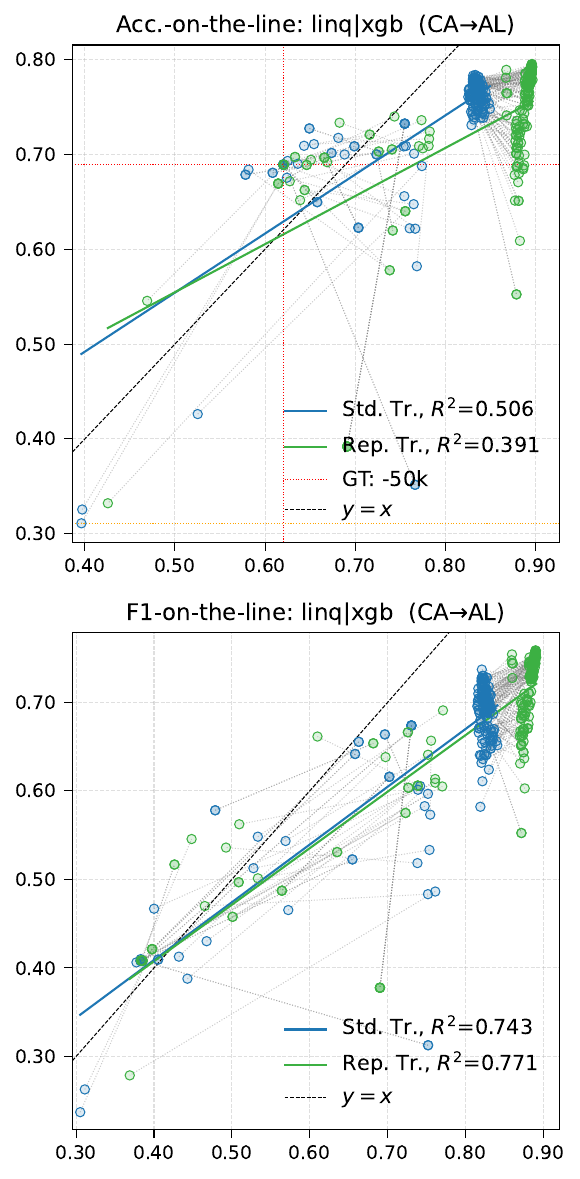}\label{fig:al-xgb-linq}}
\subfloat[\texttt{SFR}]{\includegraphics[width=0.25\textwidth]{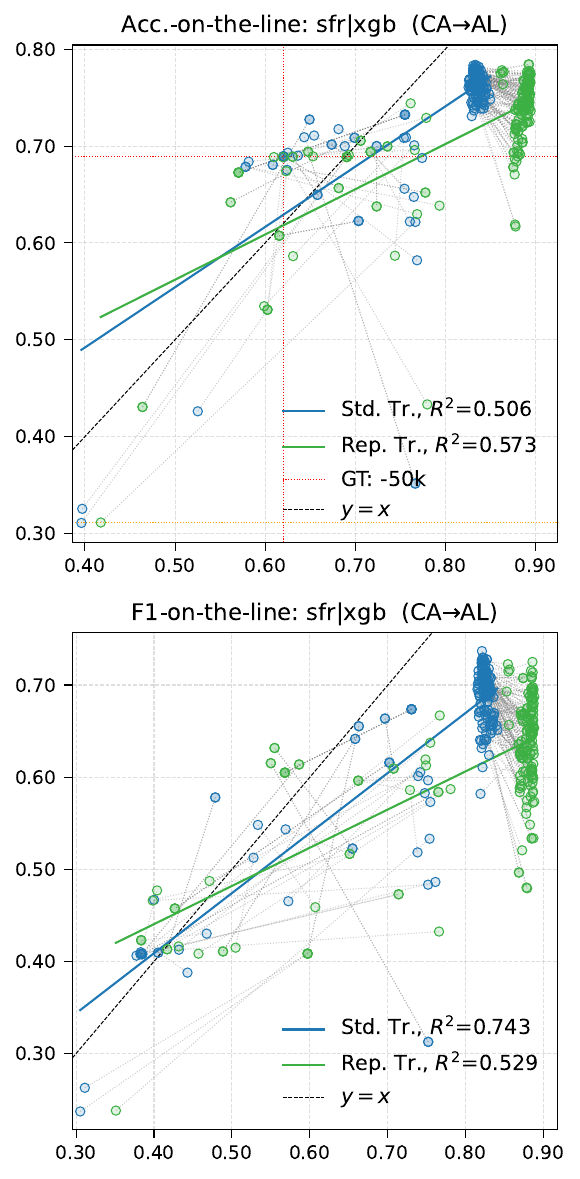}\label{fig:al-xgb-sfr}}
\subfloat[\texttt{Zeta}]{\includegraphics[width=0.25\textwidth]{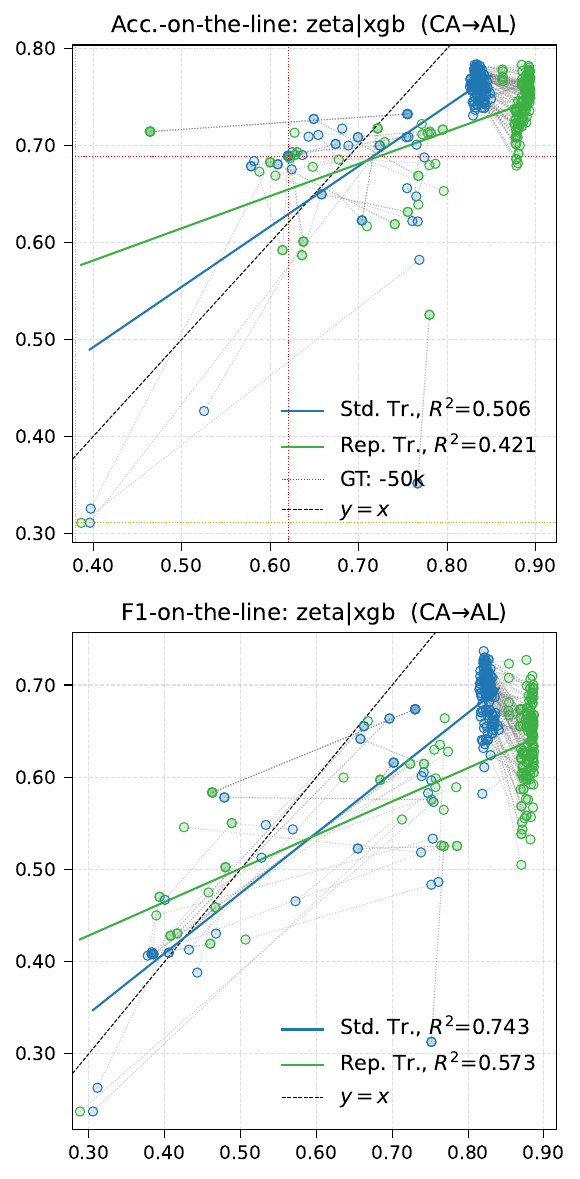}\label{fig:al-xgb-zeta}}
\caption{Pattern behaviour across different LLMs for XGB.}
\label{fig:al-xgb}
\end{figure}

\clearpage

\paragraph{Target State: Arizona} 

\begin{figure}[bp!]
\centering
\subfloat[\texttt{e5}]{\includegraphics[width=0.25\textwidth]{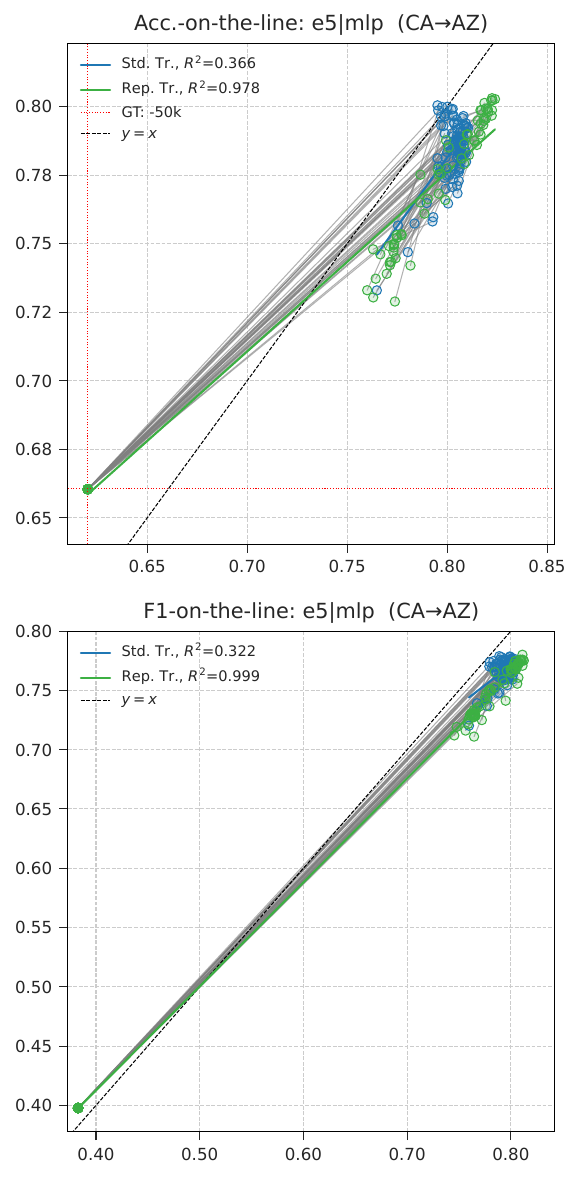}
\label{fig:az-mlp-e5}}
\subfloat[\texttt{Linq}]{\includegraphics[width=0.25\textwidth]{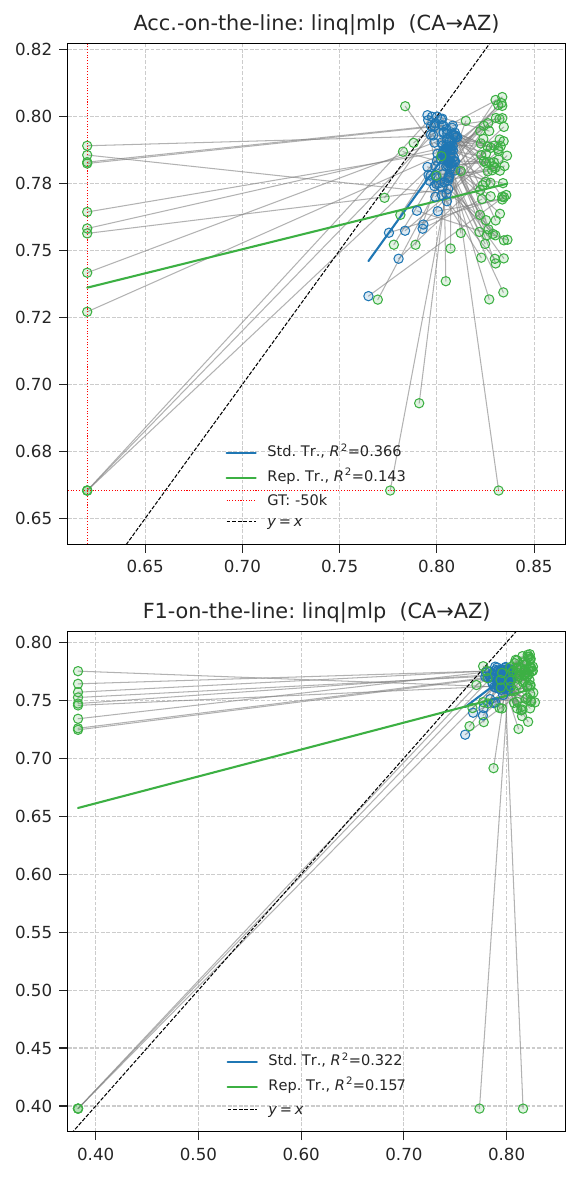}\label{fig:az-mlp-linq}}
\subfloat[\texttt{SFR}]{\includegraphics[width=0.25\textwidth]{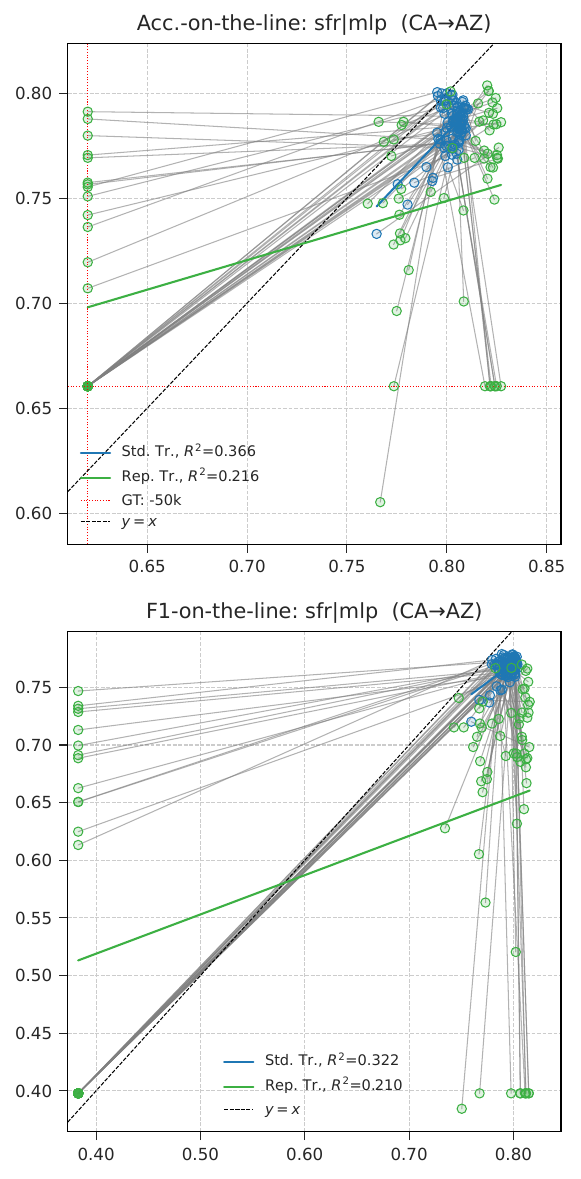}\label{fig:az-mlp-sfr}}
\subfloat[\texttt{Zeta}]{\includegraphics[width=0.25\textwidth]{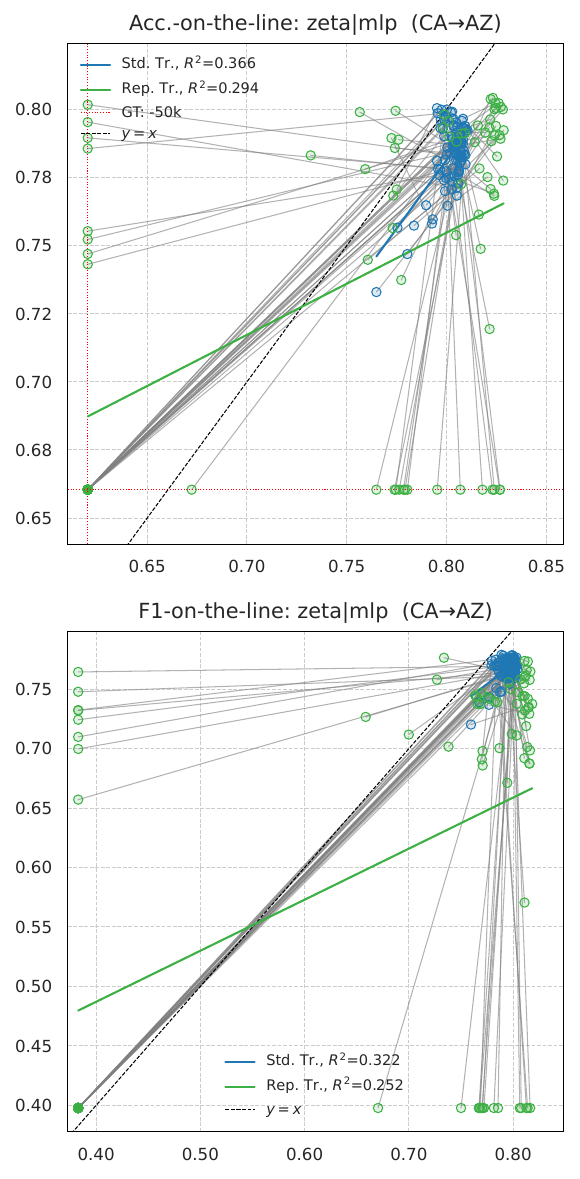}\label{fig:az-mlp-zeta}}
\caption{Pattern behaviour across different LLMs for MLP.}
\label{fig:az-mlp}
\end{figure}

\begin{figure}[bp!]
\centering
\subfloat[\texttt{e5}]{\includegraphics[width=0.25\textwidth]{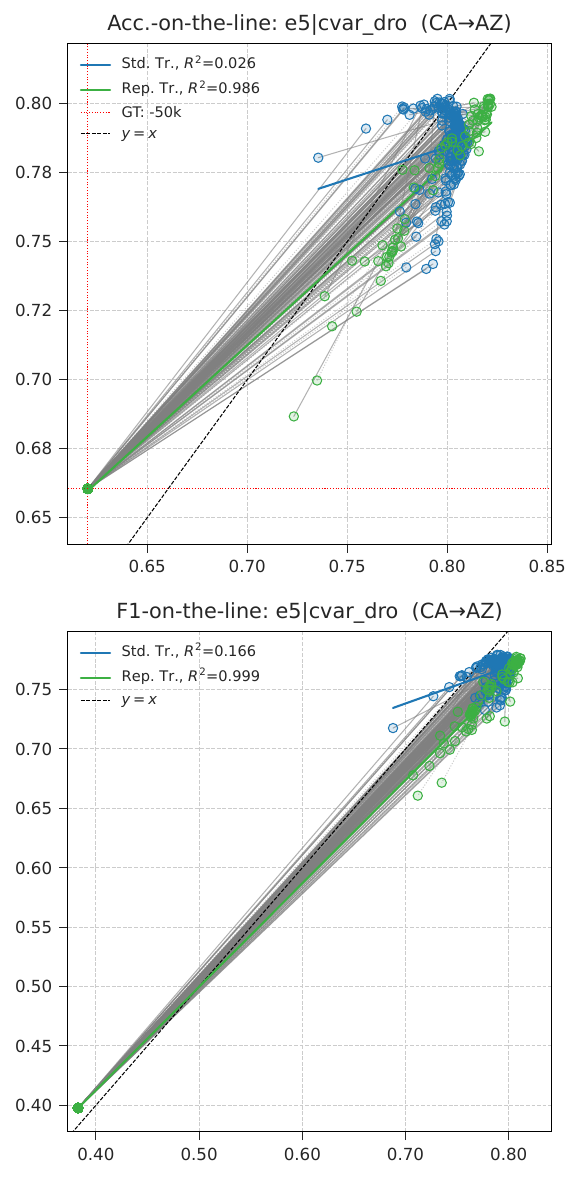}
\label{fig:az-cvar_dro-e5}}
\subfloat[\texttt{Linq}]{\includegraphics[width=0.25\textwidth]{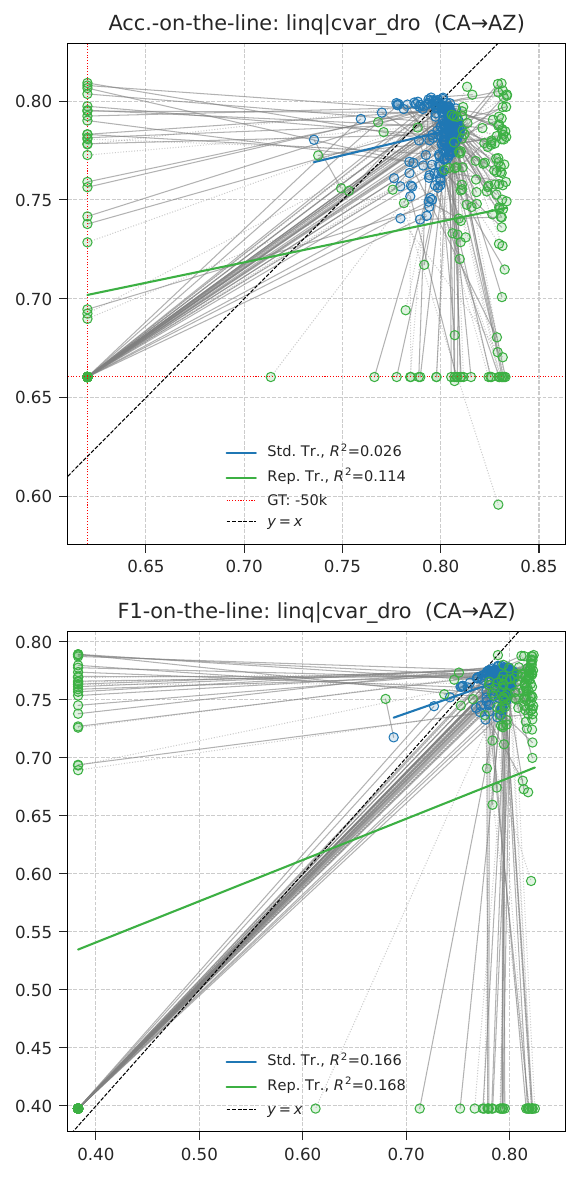}\label{fig:az-cvar_dro-linq}}
\subfloat[\texttt{SFR}]{\includegraphics[width=0.25\textwidth]{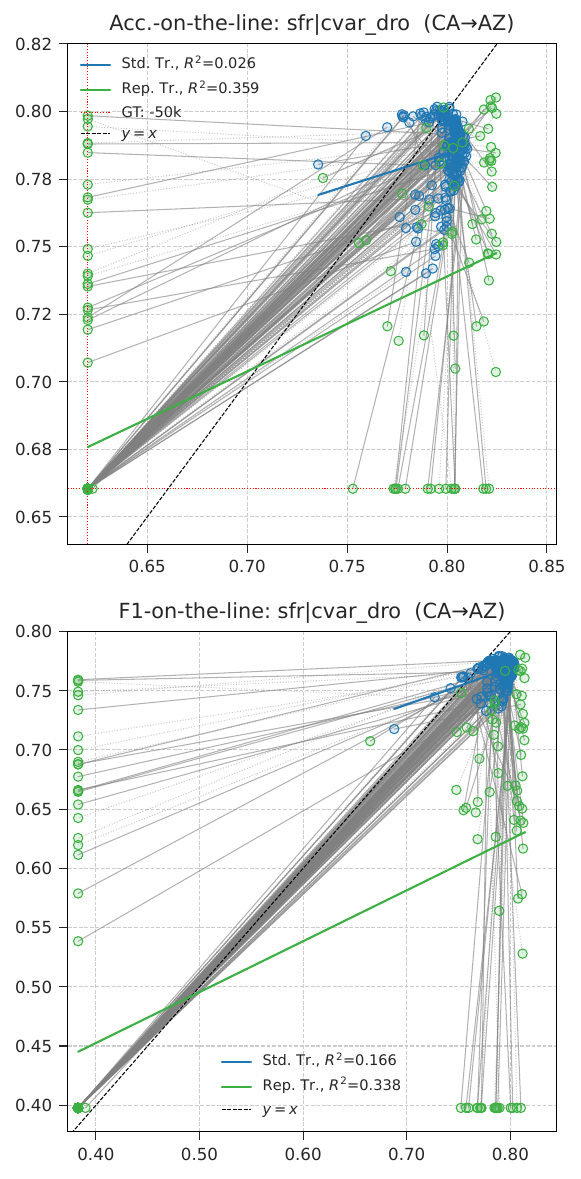}\label{fig:az-cvar_dro-sfr}}
\subfloat[\texttt{Zeta}]{\includegraphics[width=0.25\textwidth]{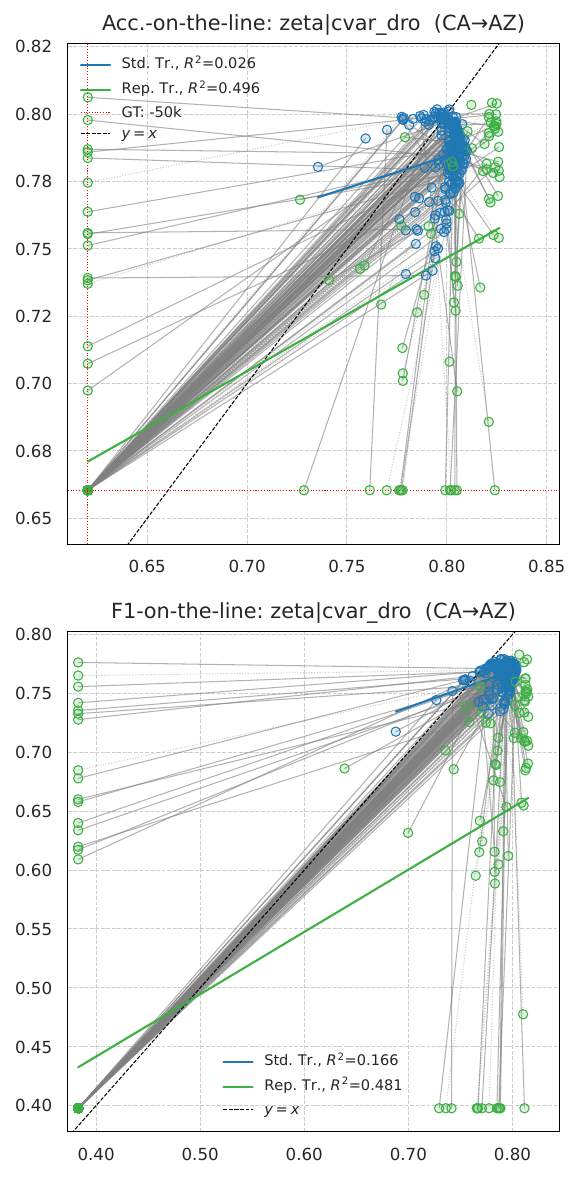}\label{fig:az-cvar_dro-zeta}}
\caption{Pattern behaviour across different LLMs for CVaR-DRO.}
\label{fig:az-cvar_dro}
\end{figure}

\begin{figure}[bp]
\centering
\subfloat[\texttt{e5}]{\includegraphics[width=0.25\textwidth]{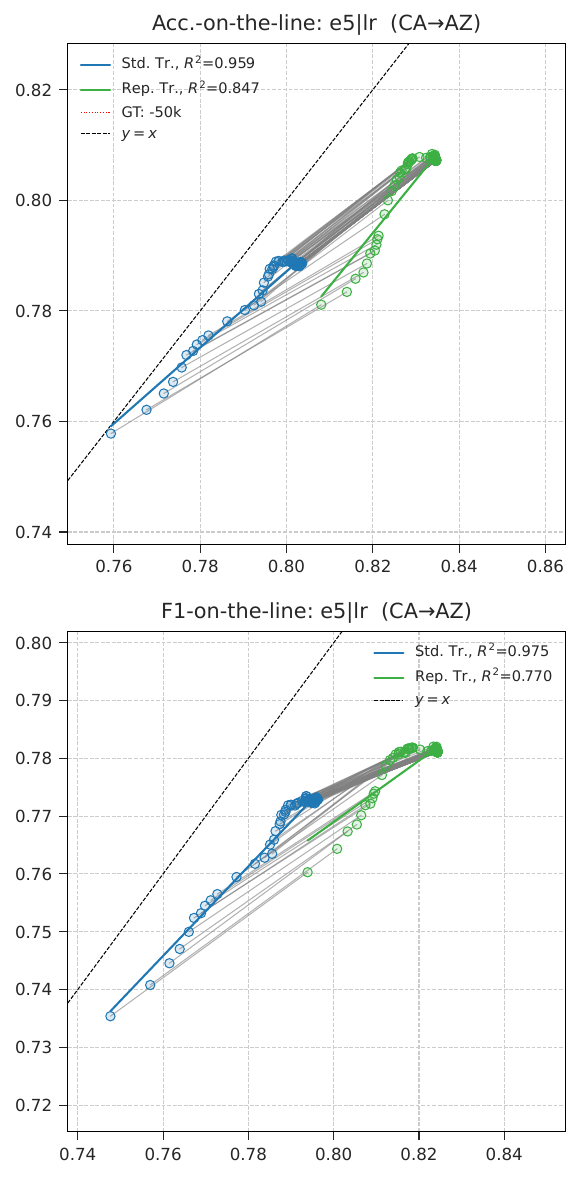}
\label{fig:az-lr-e5}}
\subfloat[\texttt{Linq}]{\includegraphics[width=0.25\textwidth]{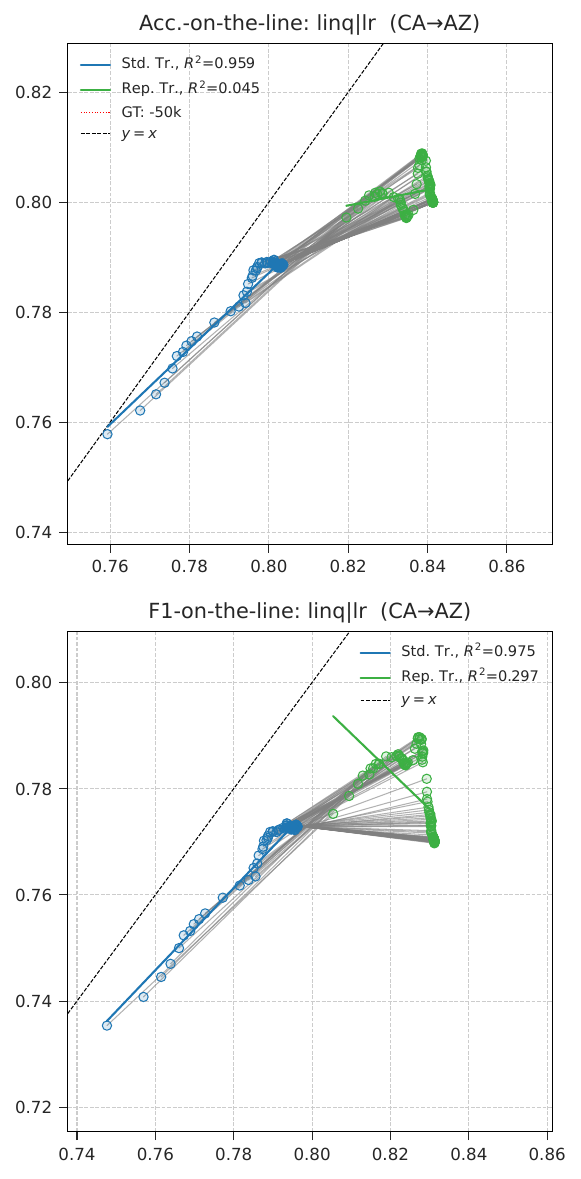}\label{fig:az-lr-linq}}
\subfloat[\texttt{SFR}]{\includegraphics[width=0.25\textwidth]{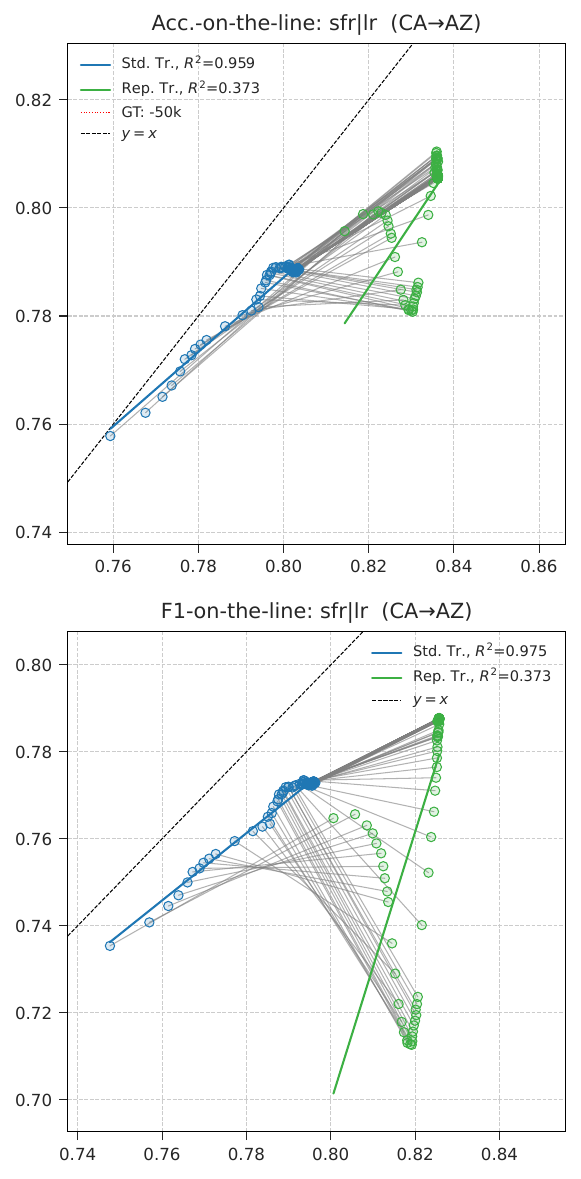}\label{fig:az-lr-sfr}}
\subfloat[\texttt{Zeta}]{\includegraphics[width=0.25\textwidth]{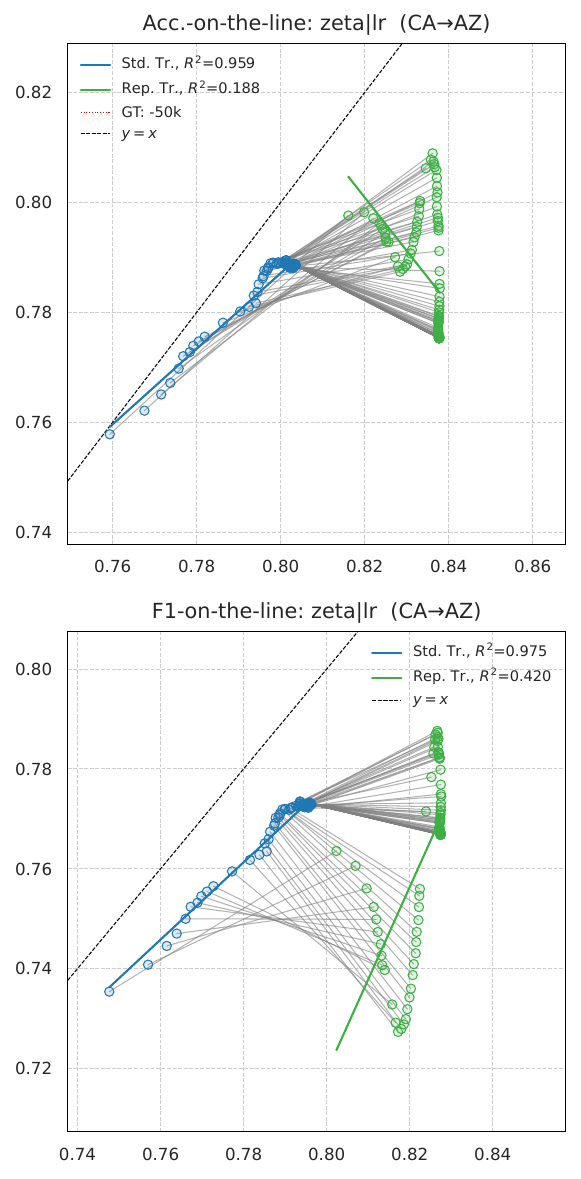}\label{fig:az-lr-zeta}}
\caption{Pattern behaviour across different LLMs for LR.}
\label{fig:az-lr}
\end{figure}

\begin{figure}[bp]
\centering
\subfloat[\texttt{e5}]{\includegraphics[width=0.25\textwidth]{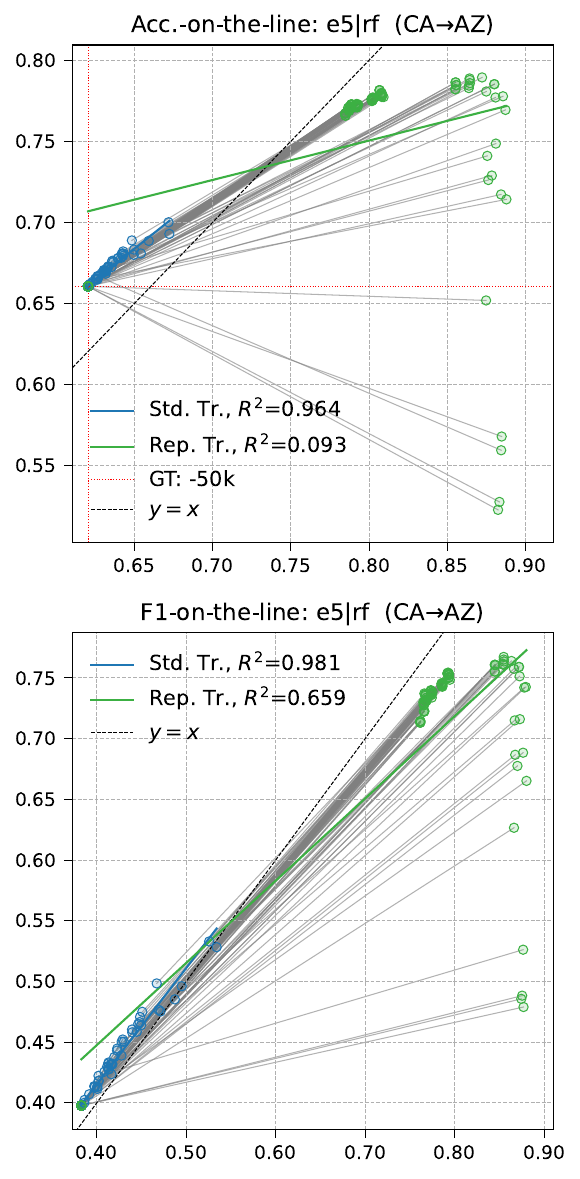}
\label{fig:az-rf-e5}}
\subfloat[\texttt{Linq}]{\includegraphics[width=0.25\textwidth]{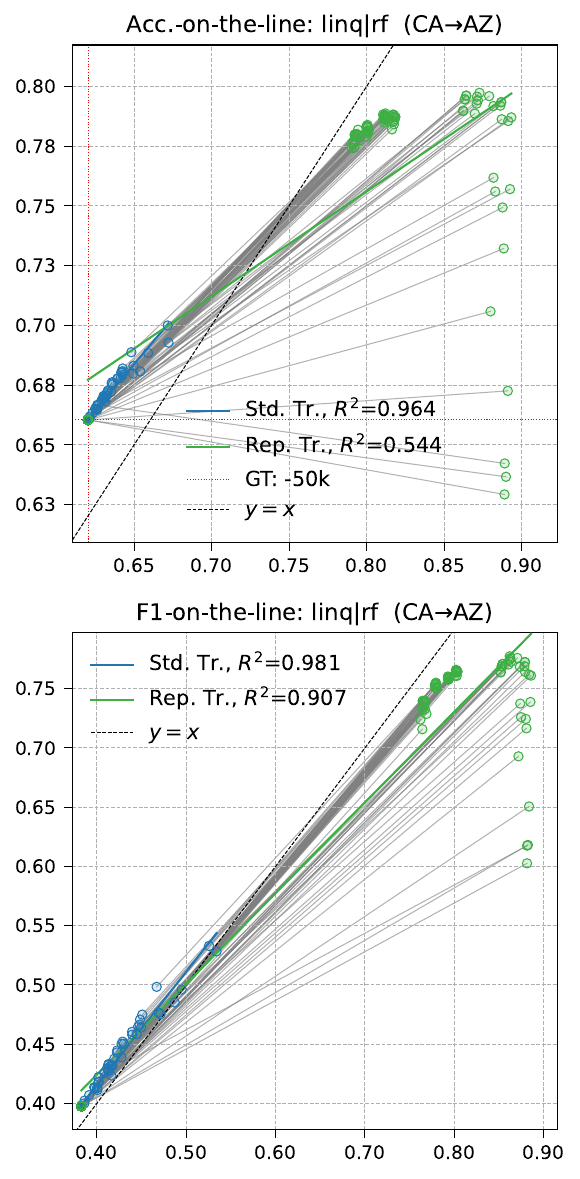}\label{fig:az-rf-linq}}
\subfloat[\texttt{SFR}]{\includegraphics[width=0.25\textwidth]{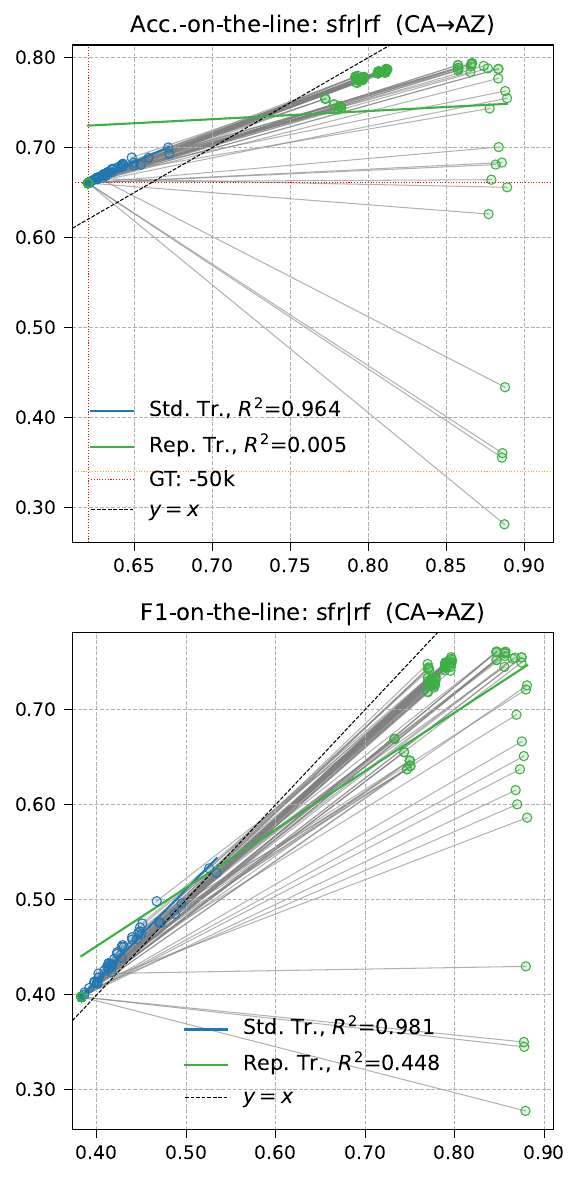}\label{fig:az-rf-sfr}}
\subfloat[\texttt{Zeta}]{\includegraphics[width=0.25\textwidth]{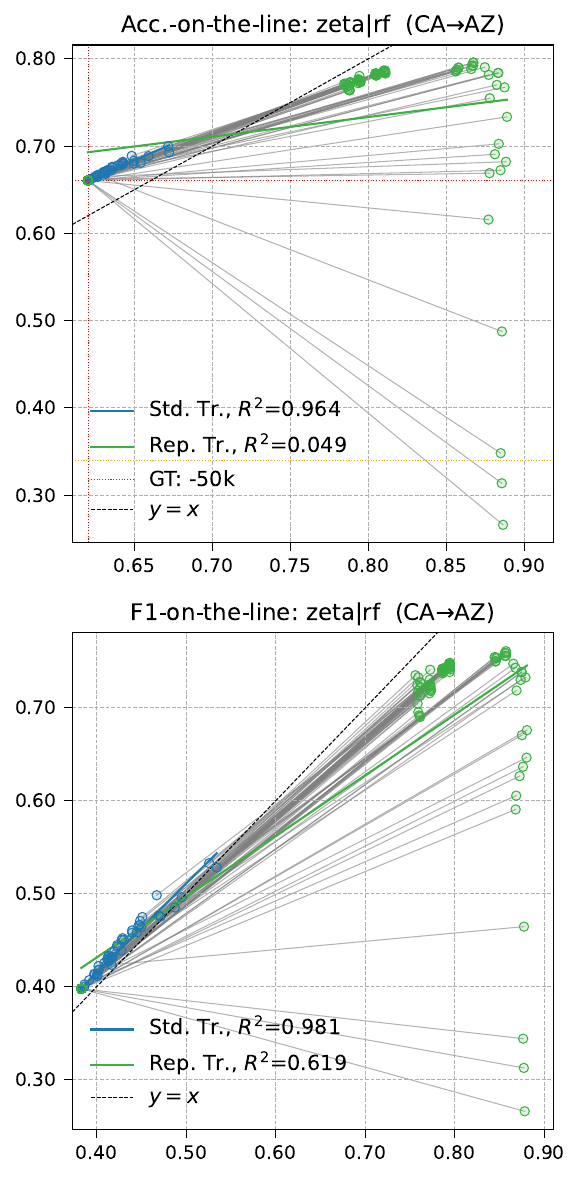}\label{fig:az-rf-zeta}}
\caption{Pattern behaviour across different LLMs for RF.}
\label{fig:az-rf}
\end{figure}

\begin{figure}[bp]
\centering
\subfloat[\texttt{e5}]{\includegraphics[width=0.25\textwidth]{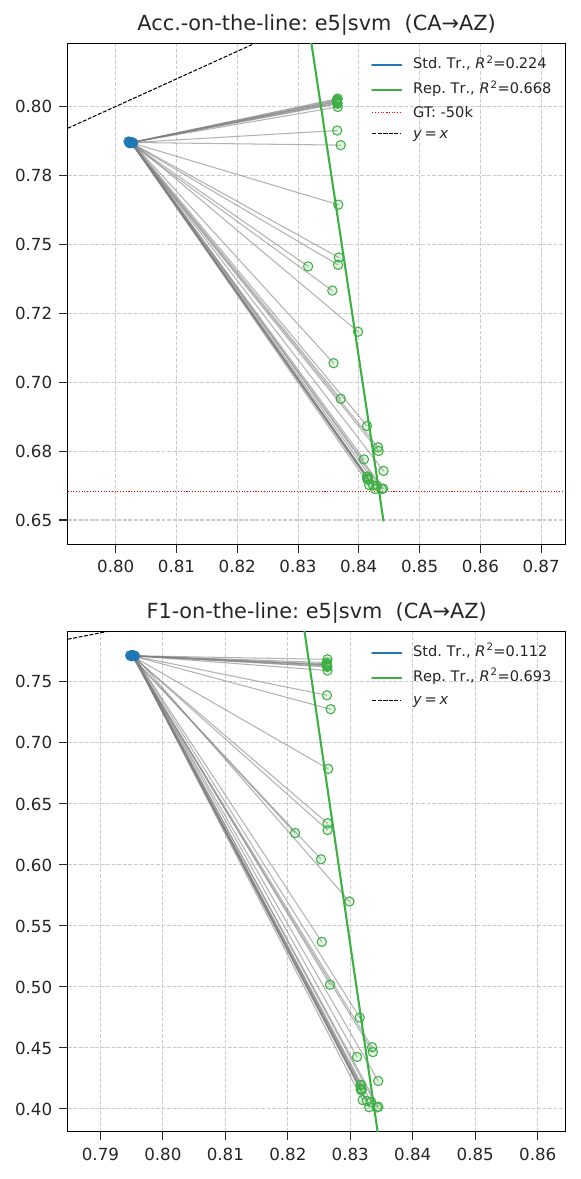}
\label{fig:az-svm-e5}}
\subfloat[\texttt{Linq}]{\includegraphics[width=0.25\textwidth]{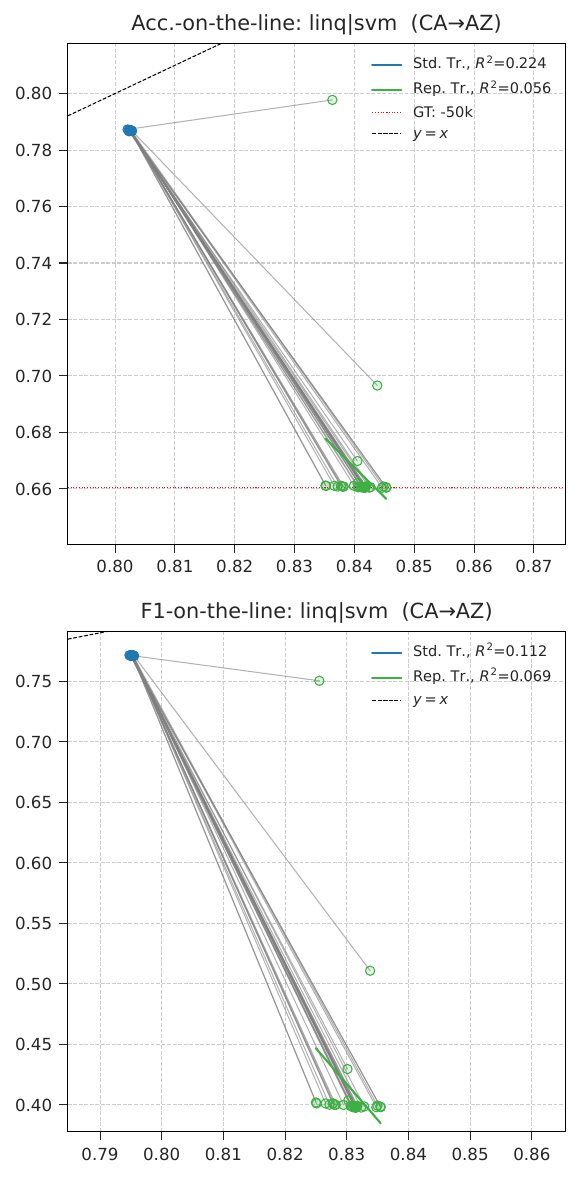}\label{fig:az-svm-linq}}
\subfloat[\texttt{SFR}]{\includegraphics[width=0.25\textwidth]{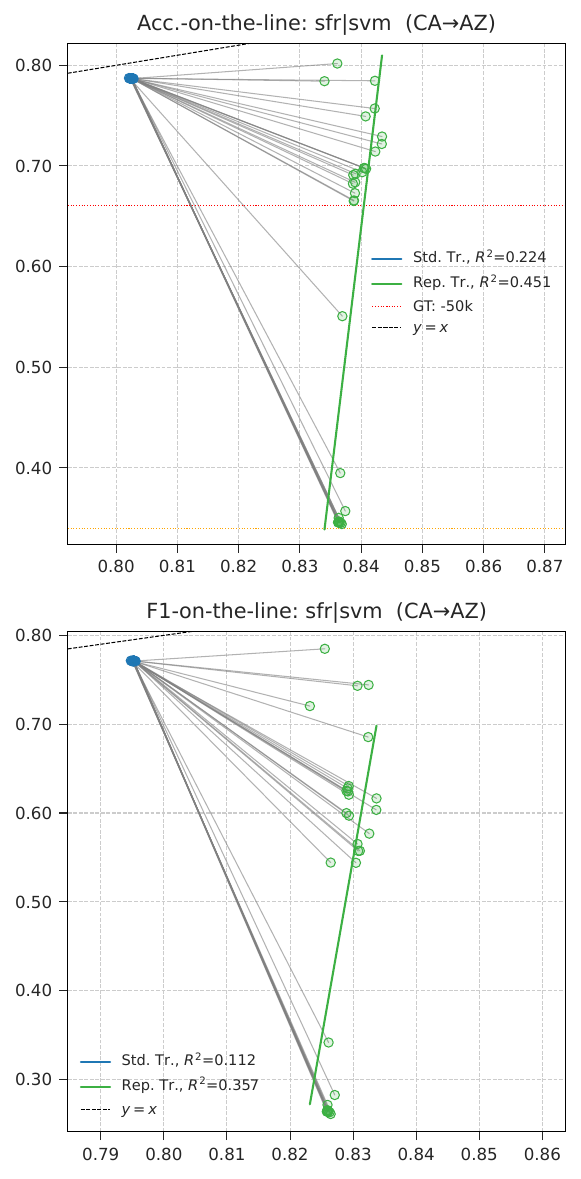}\label{fig:az-svm-sfr}}
\subfloat[\texttt{Zeta}]{\includegraphics[width=0.25\textwidth]{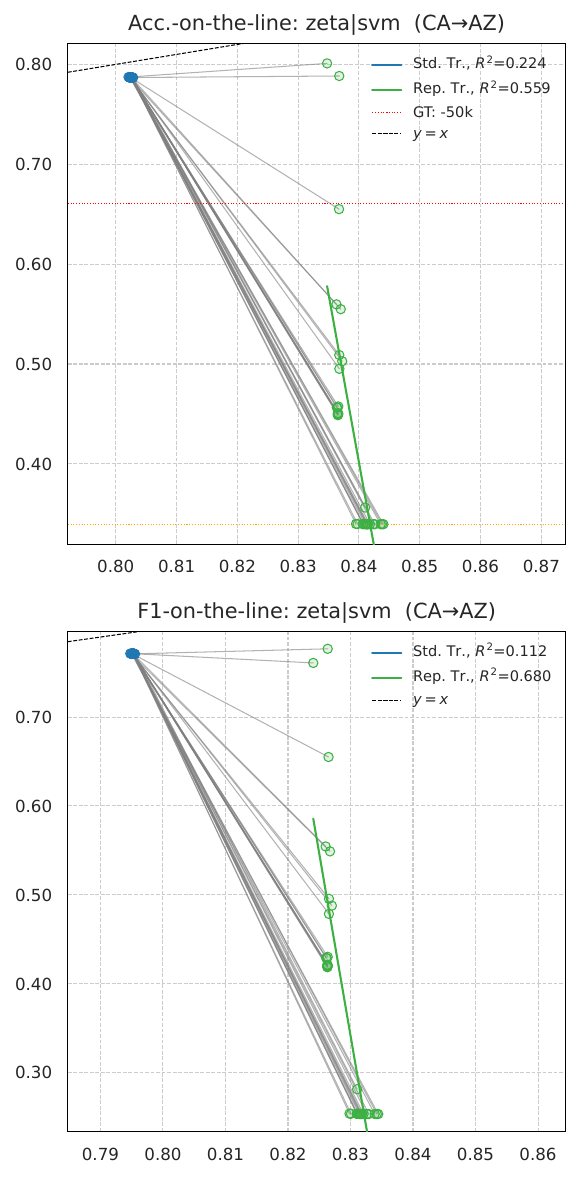}\label{fig:az-svm-zeta}}
\caption{Pattern behaviour across different LLMs for SVM.}
\label{fig:az-svm}
\end{figure}

\begin{figure}[bp]
\centering
\subfloat[\texttt{e5}]{\includegraphics[width=0.25\textwidth]{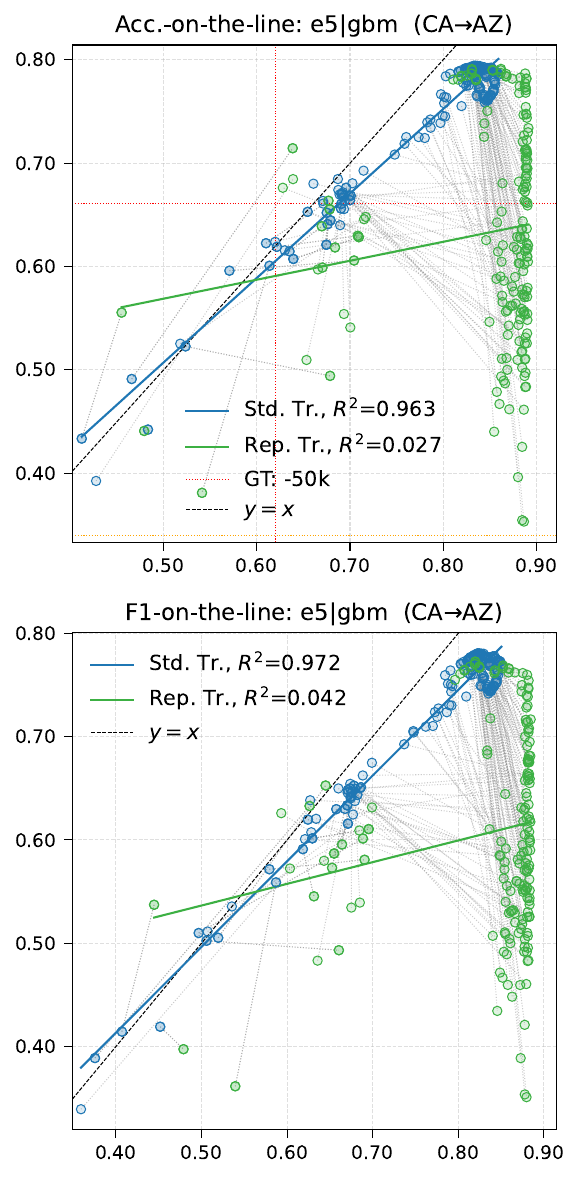}
\label{fig:az-gbm-e5}}
\subfloat[\texttt{Linq}]{\includegraphics[width=0.25\textwidth]{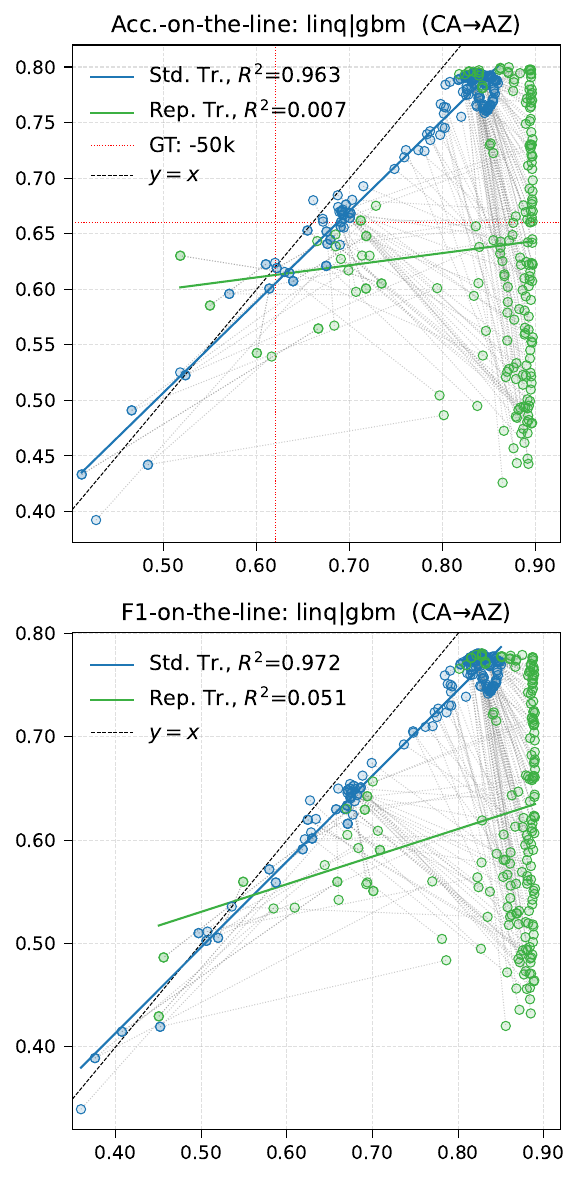}\label{fig:az-gbm-linq}}
\subfloat[\texttt{SFR}]{\includegraphics[width=0.25\textwidth]{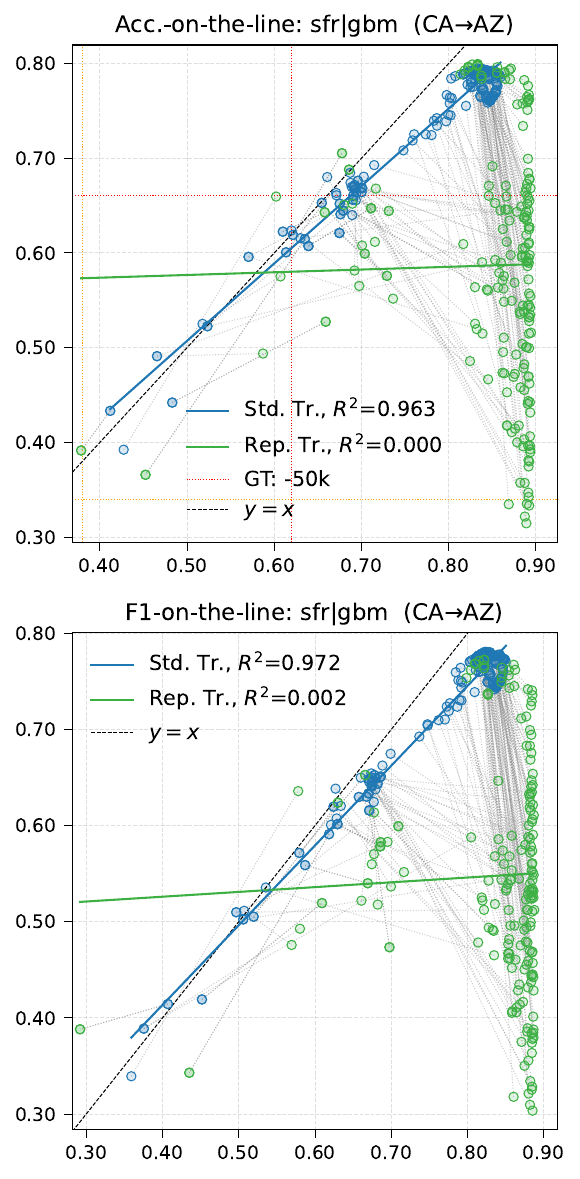}\label{fig:az-gbm-sfr}}
\subfloat[\texttt{Zeta}]{\includegraphics[width=0.25\textwidth]{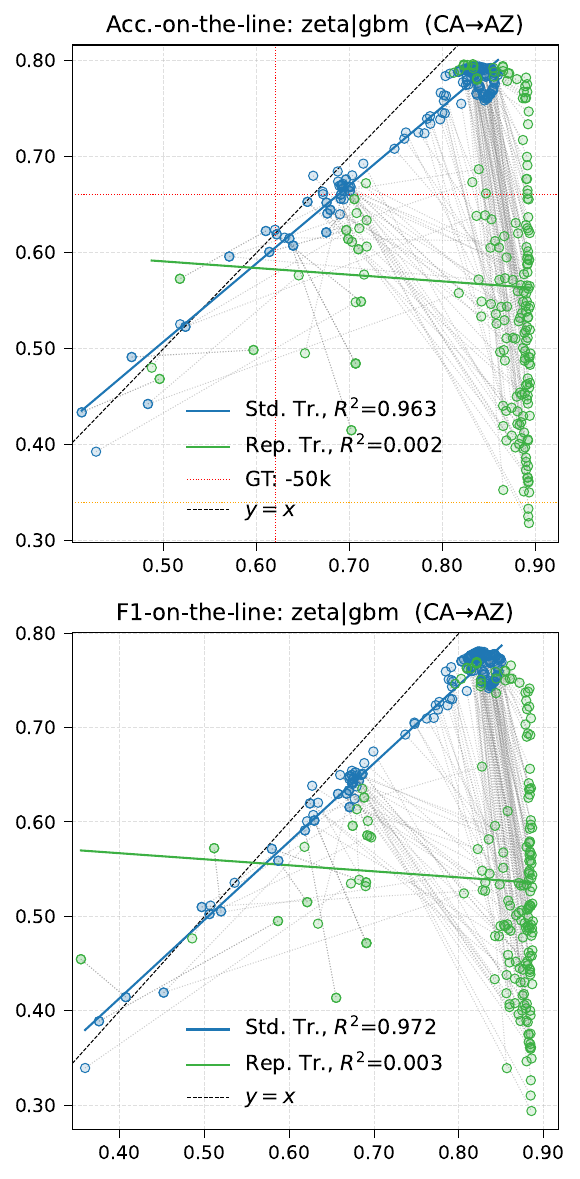}\label{fig:az-gbm-zeta}}
\caption{Pattern behaviour across different LLMs for GBM.}
\label{fig:az-gbm}
\end{figure}

\begin{figure}[bp]
\centering
\subfloat[\texttt{e5}]{\includegraphics[width=0.25\textwidth]{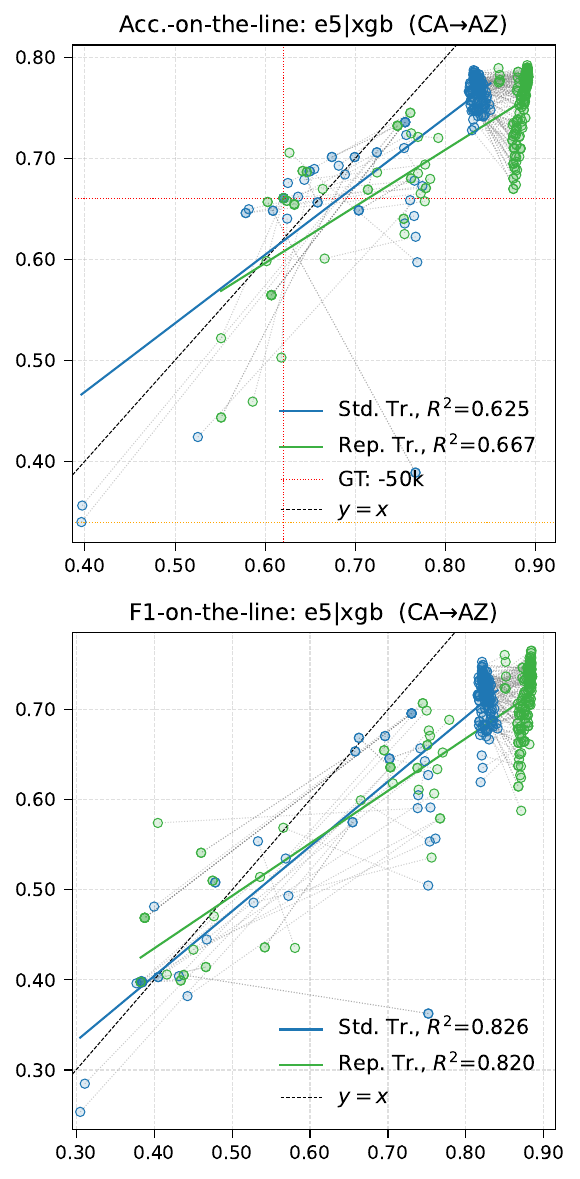}
\label{fig:az-xgb-e5}}
\subfloat[\texttt{Linq}]{\includegraphics[width=0.25\textwidth]{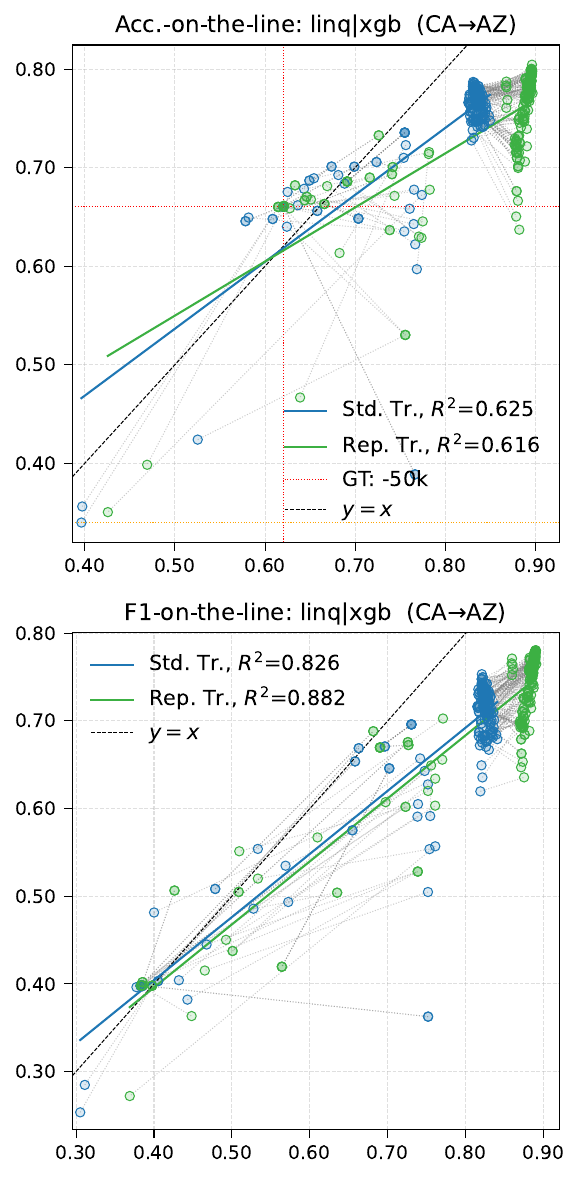}\label{fig:az-xgb-linq}}
\subfloat[\texttt{SFR}]{\includegraphics[width=0.25\textwidth]{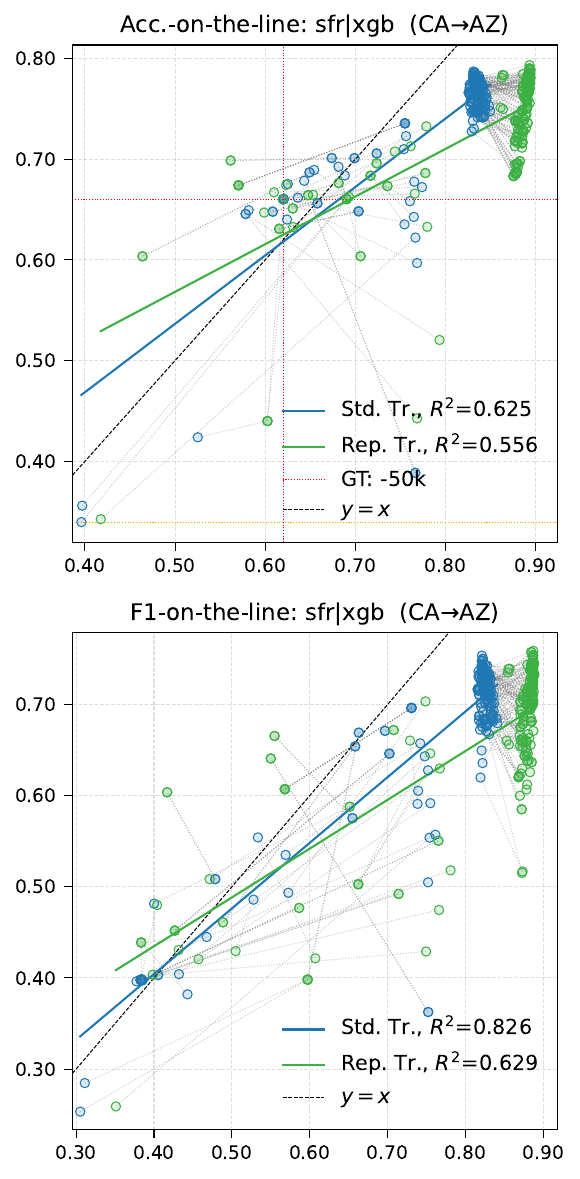}\label{fig:az-xgb-sfr}}
\subfloat[\texttt{Zeta}]{\includegraphics[width=0.25\textwidth]{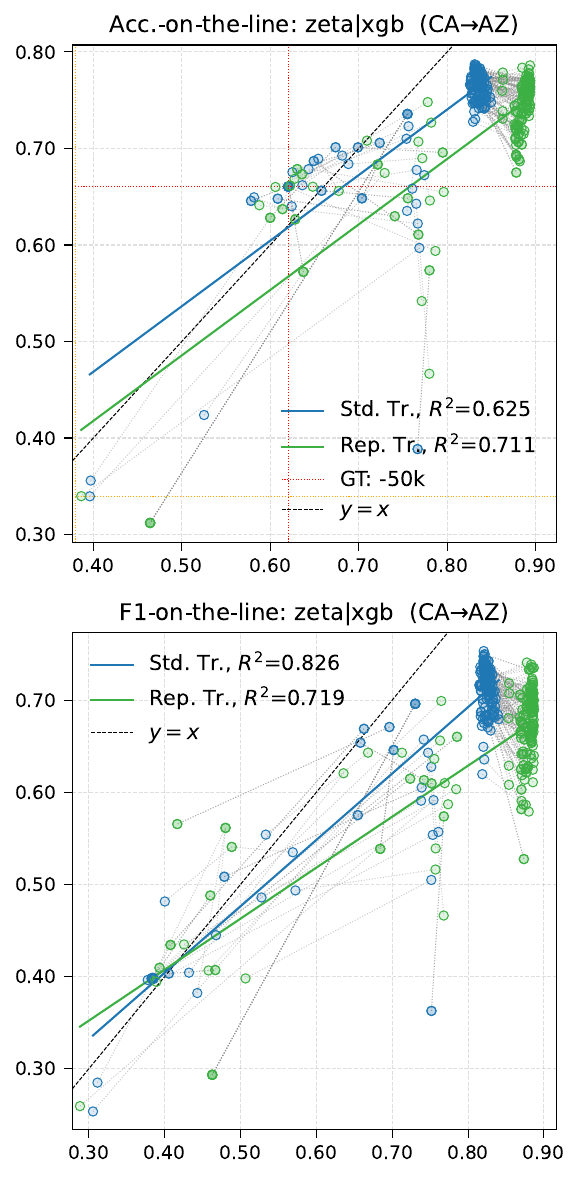}\label{fig:az-xgb-zeta}}
\caption{Pattern behaviour across different LLMs for XGB.}
\label{fig:az-xgb}
\end{figure}

\clearpage


\paragraph{Target State: Alaska} 

\begin{figure}[bp!]
\centering
\subfloat[\texttt{e5}]{\includegraphics[width=0.25\textwidth]{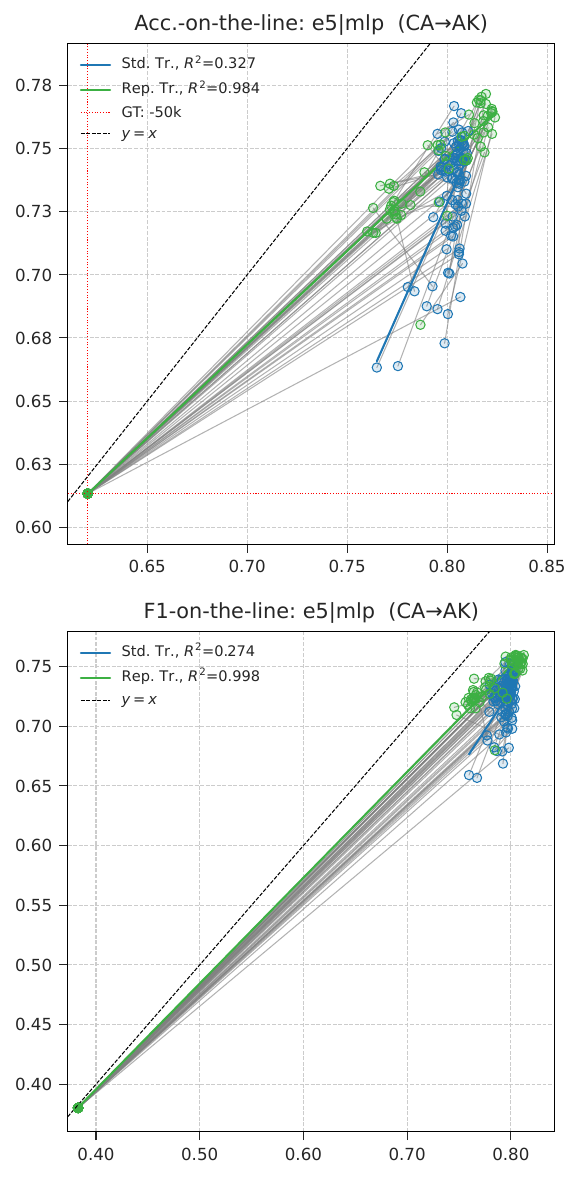}
\label{fig:ak-mlp-e5}}
\subfloat[\texttt{Linq}]{\includegraphics[width=0.25\textwidth]{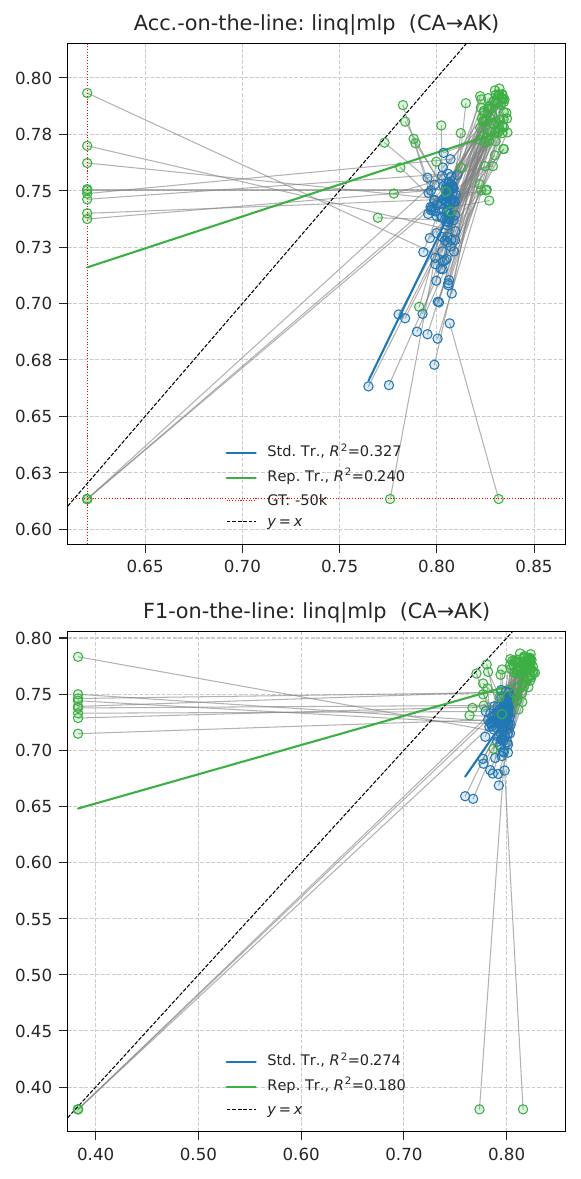}\label{fig:ak-mlp-linq}}
\subfloat[\texttt{SFR}]{\includegraphics[width=0.25\textwidth]{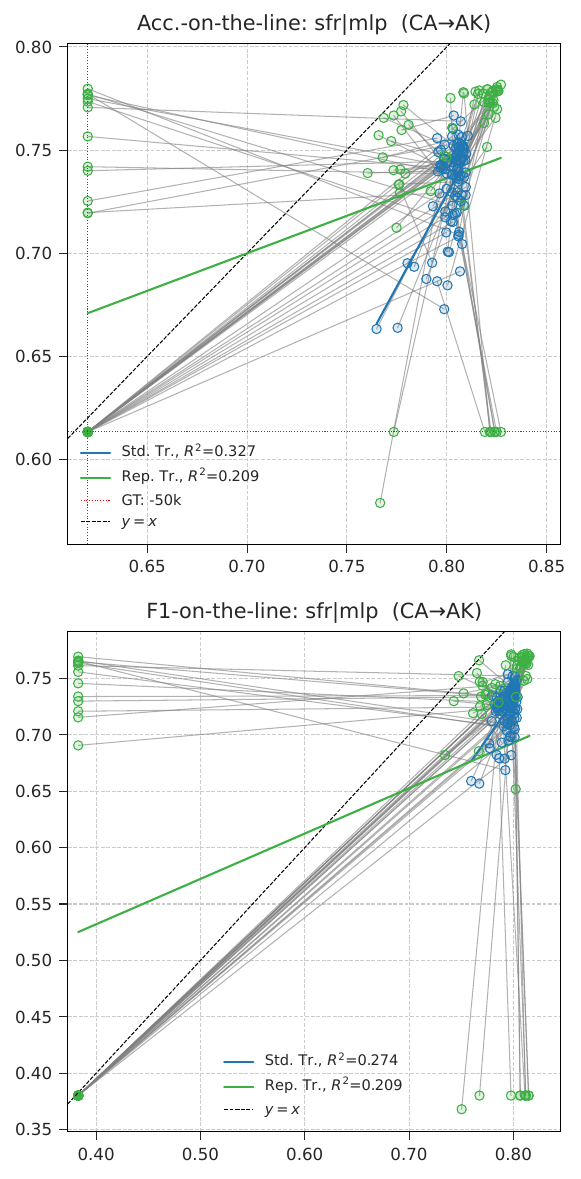}\label{fig:ak-mlp-sfr}}
\subfloat[\texttt{Zeta}]{\includegraphics[width=0.25\textwidth]{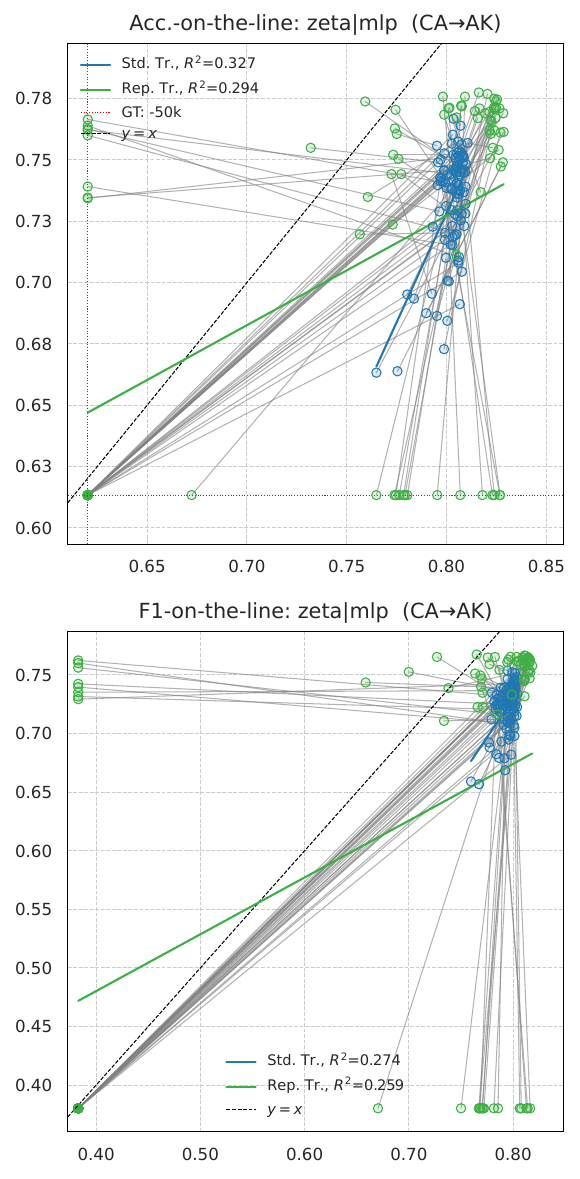}\label{fig:ak-mlp-zeta}}
\caption{Pattern behaviour across different LLMs for MLP.}
\label{fig:ak-mlp}
\end{figure}

\begin{figure}[bp!]
\centering
\subfloat[\texttt{e5}]{\includegraphics[width=0.25\textwidth]{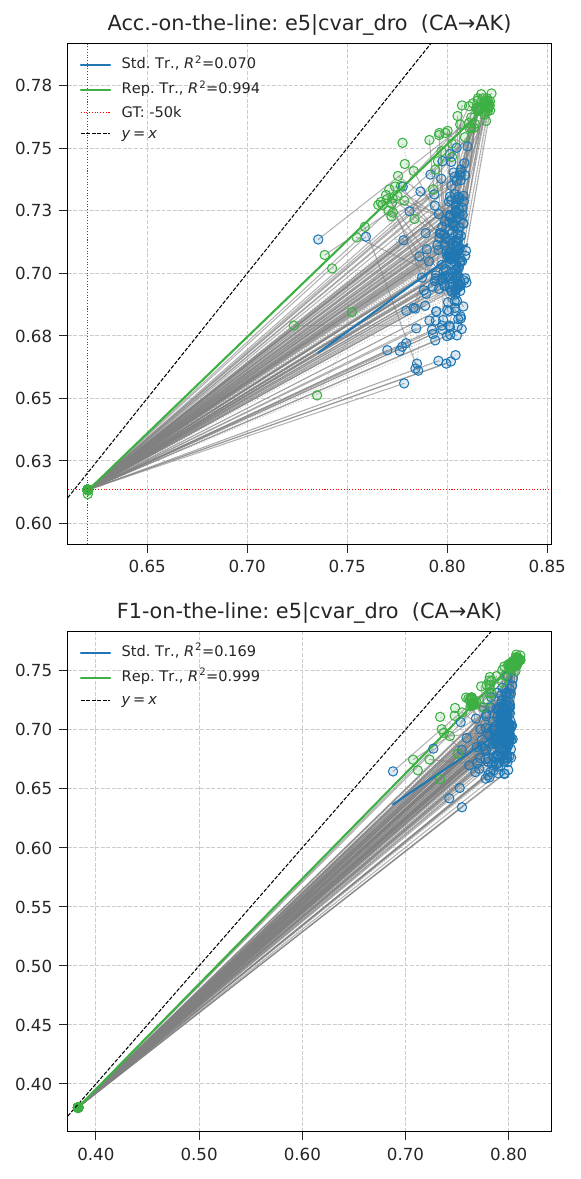}
\label{fig:ak-cvar_dro-e5}}
\subfloat[\texttt{Linq}]{\includegraphics[width=0.25\textwidth]{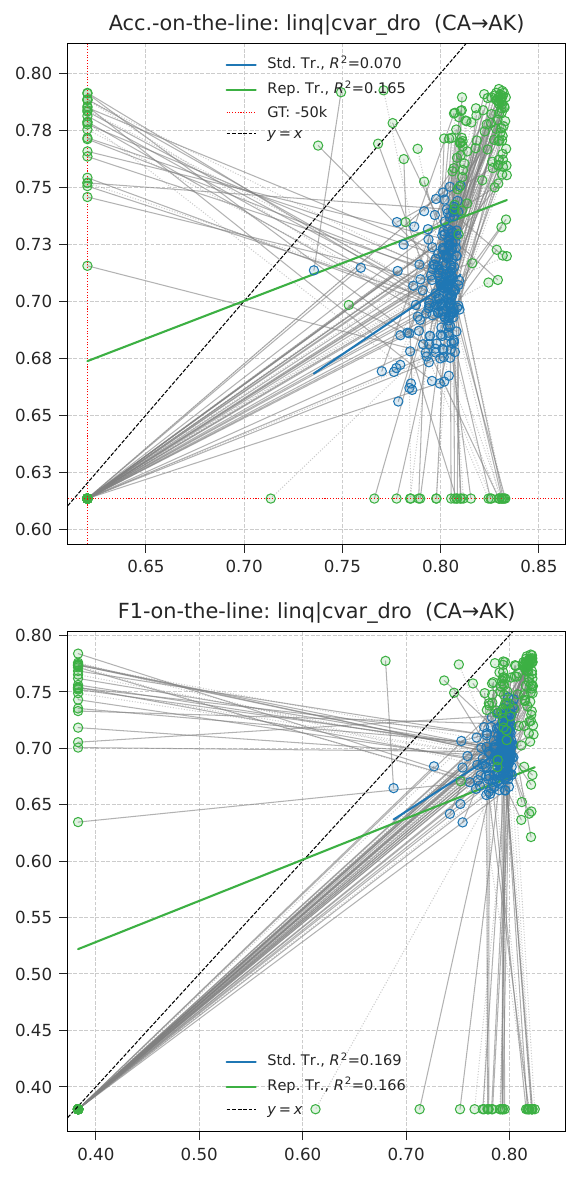}\label{fig:ak-cvar_dro-linq}}
\subfloat[\texttt{SFR}]{\includegraphics[width=0.25\textwidth]{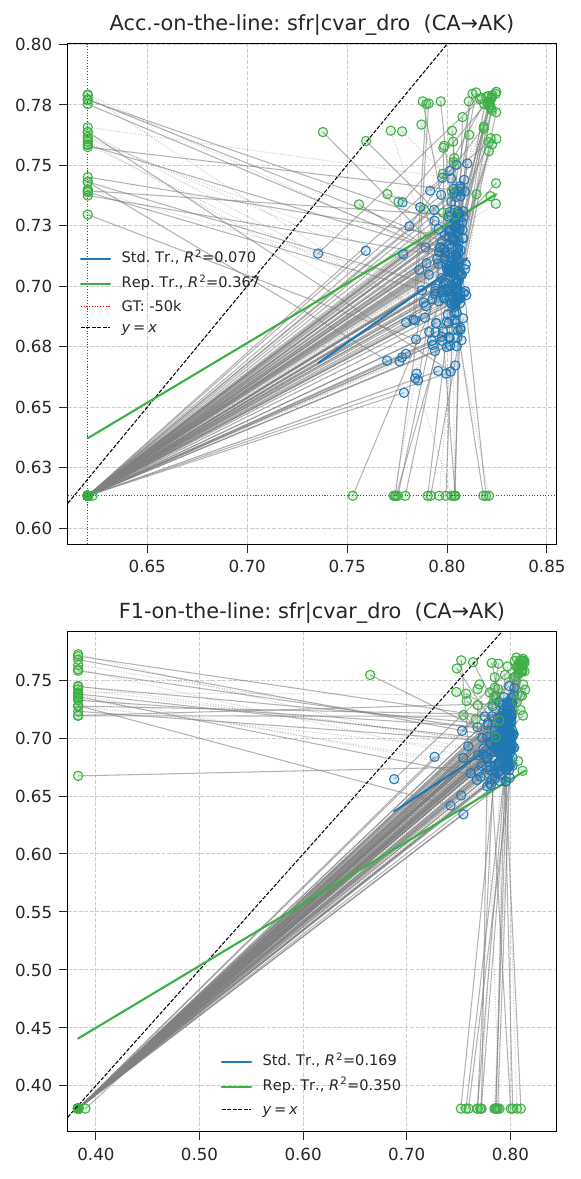}\label{fig:ak-cvar_dro-sfr}}
\subfloat[\texttt{Zeta}]{\includegraphics[width=0.25\textwidth]{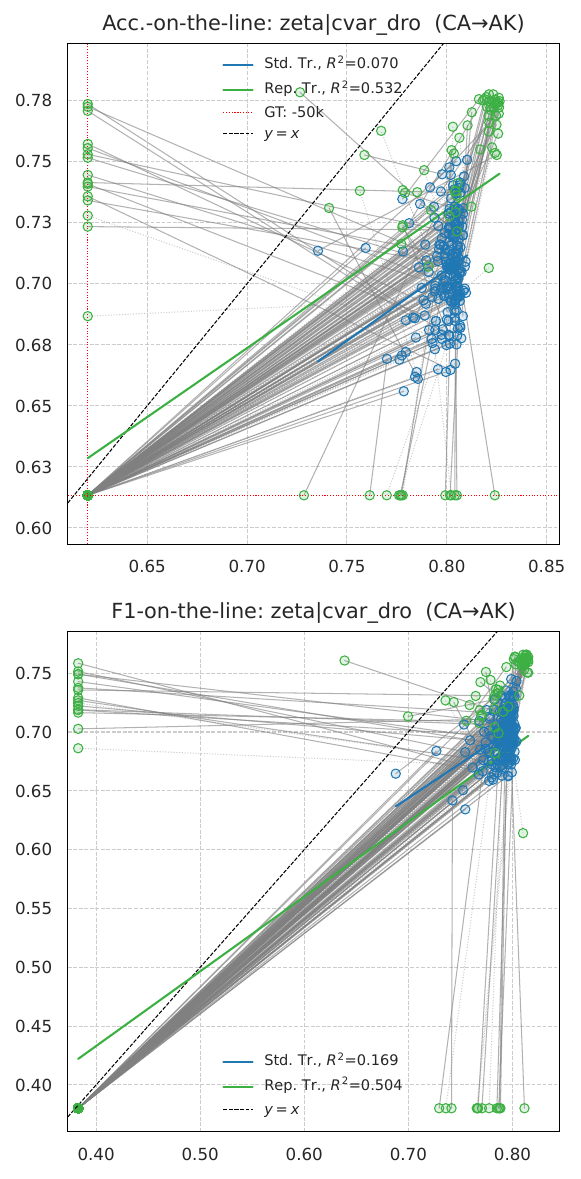}\label{fig:ak-cvar_dro-zeta}}
\caption{Pattern behaviour across different LLMs for CVaR-DRO.}
\label{fig:ak-cvar_dro}
\end{figure}

\begin{figure}[bp]
\centering
\subfloat[\texttt{e5}]{\includegraphics[width=0.25\textwidth]{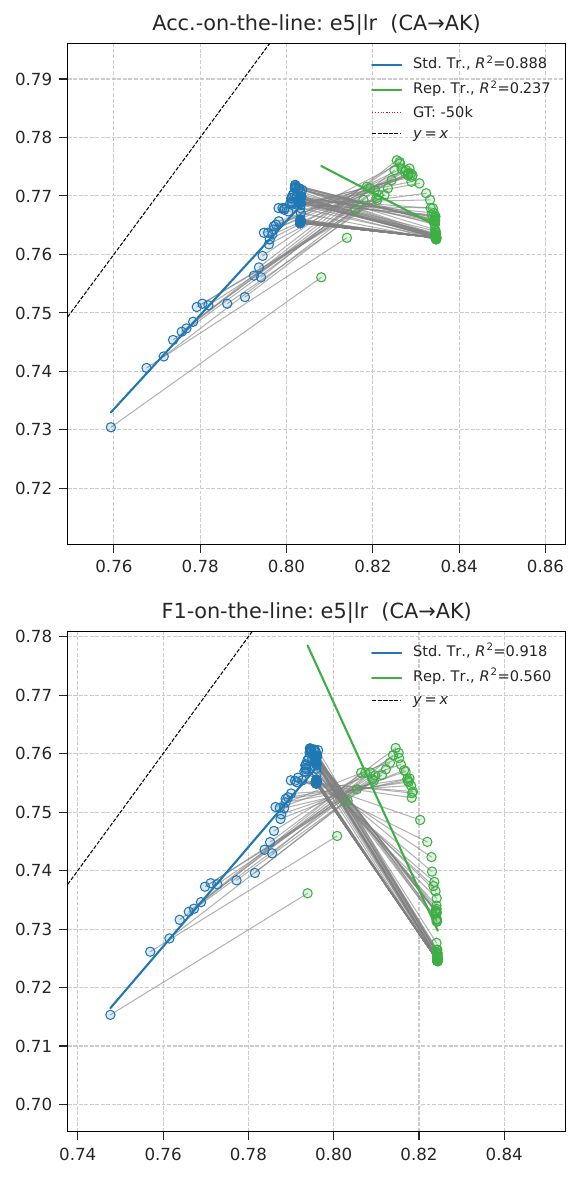}
\label{fig:ak-lr-e5}}
\subfloat[\texttt{Linq}]{\includegraphics[width=0.25\textwidth]{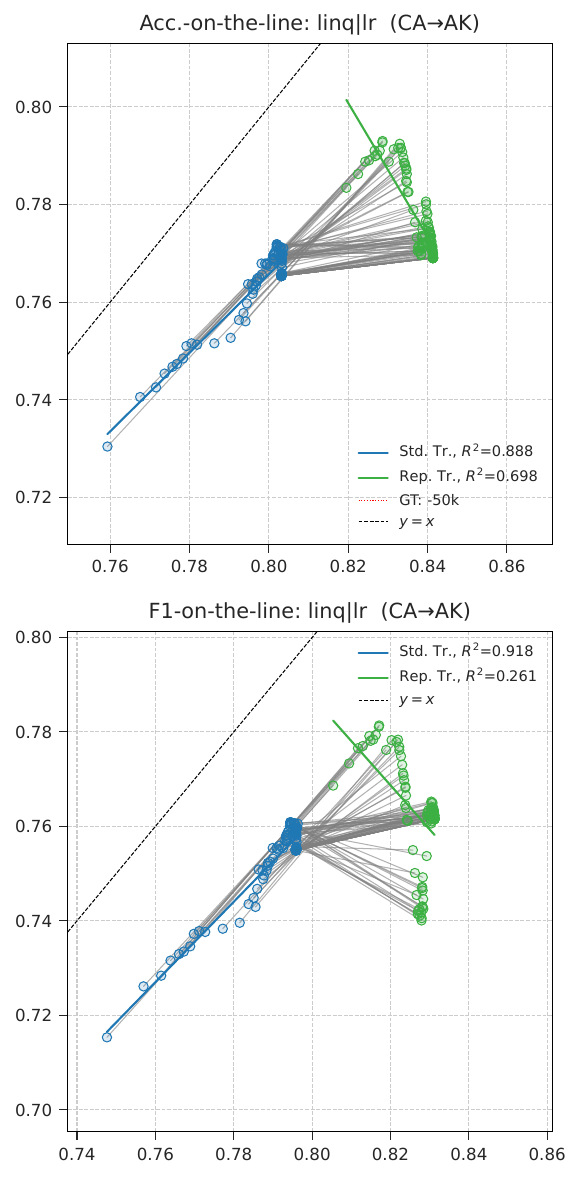}\label{fig:ak-lr-linq}}
\subfloat[\texttt{SFR}]{\includegraphics[width=0.25\textwidth]{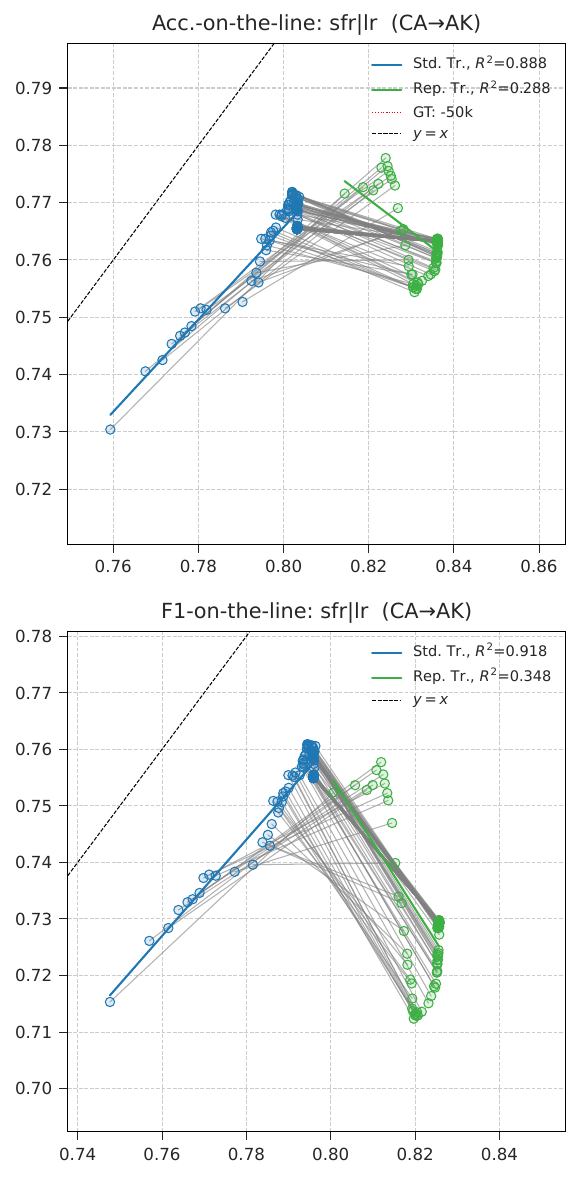}\label{fig:ak-lr-sfr}}
\subfloat[\texttt{Zeta}]{\includegraphics[width=0.25\textwidth]{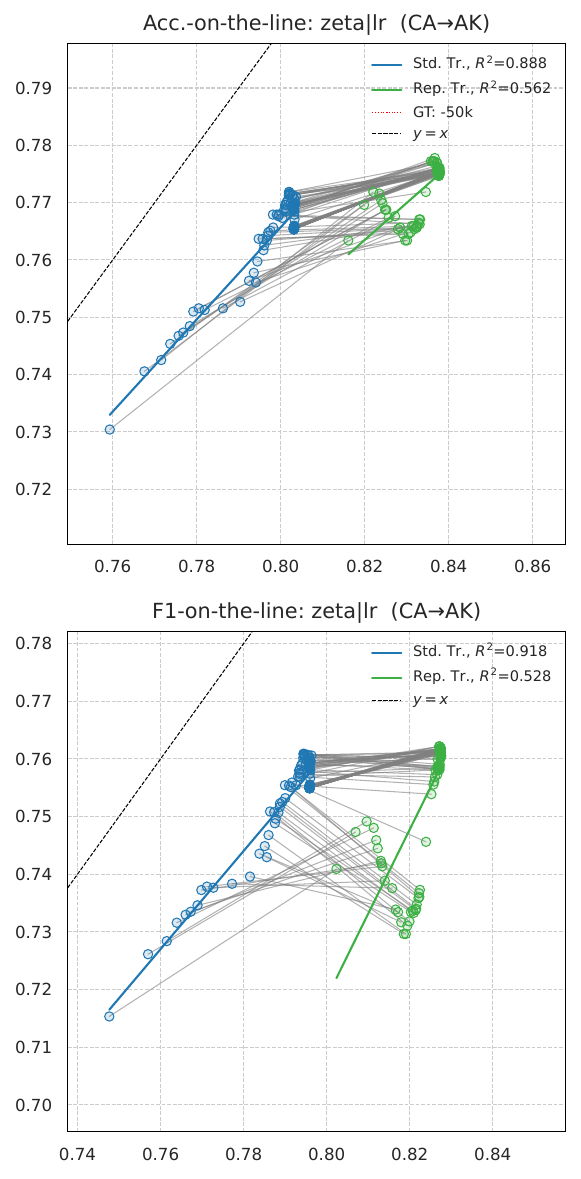}\label{fig:ak-lr-zeta}}
\caption{Pattern behaviour across different LLMs for LR.}
\label{fig:ak-lr}
\end{figure}

\begin{figure}[bp]
\centering
\subfloat[\texttt{e5}]{\includegraphics[width=0.25\textwidth]{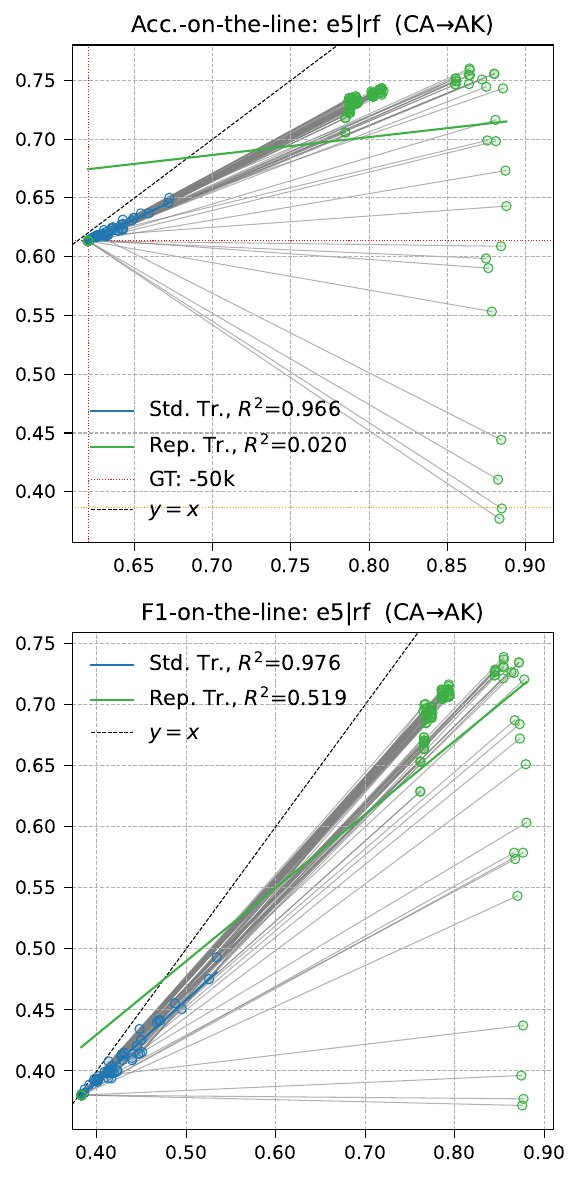}
\label{fig:ak-rf-e5}}
\subfloat[\texttt{Linq}]{\includegraphics[width=0.25\textwidth]{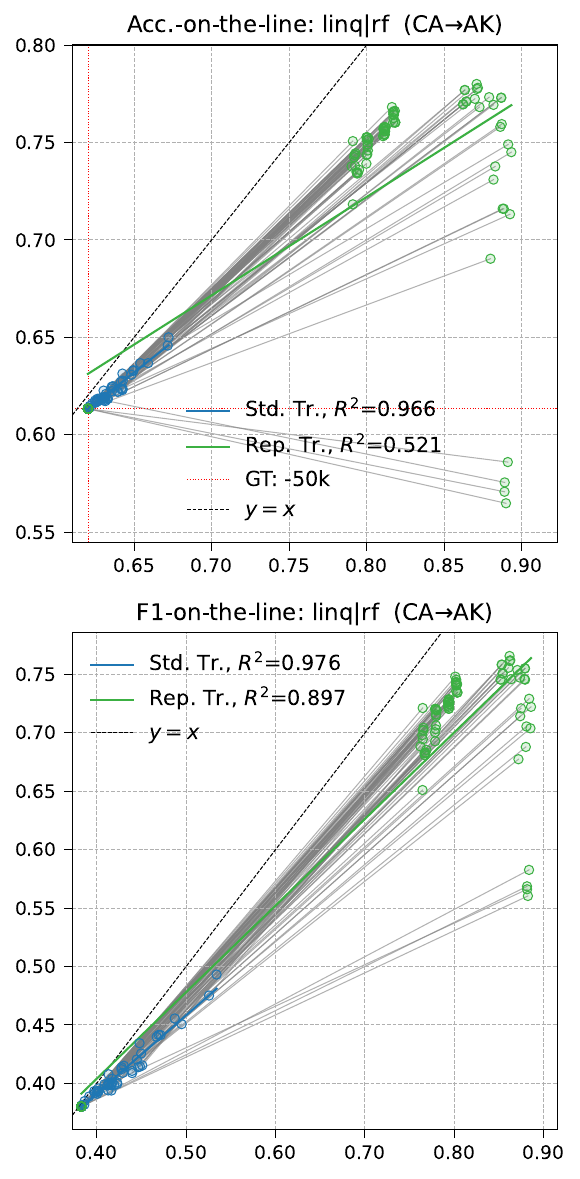}\label{fig:ak-rf-linq}}
\subfloat[\texttt{SFR}]{\includegraphics[width=0.25\textwidth]{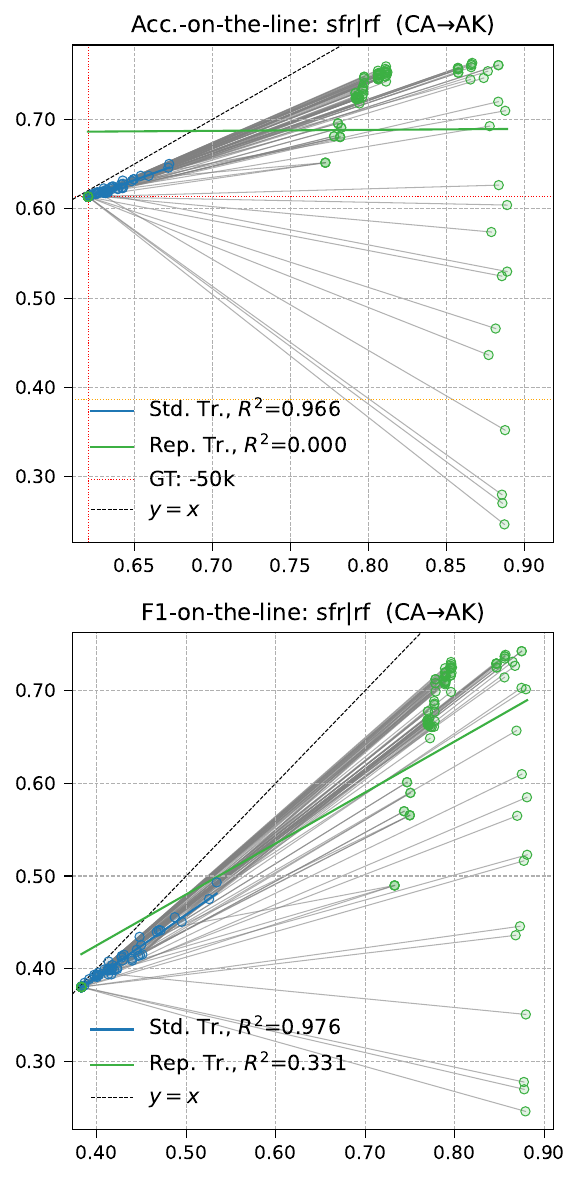}\label{fig:ak-rf-sfr}}
\subfloat[\texttt{Zeta}]{\includegraphics[width=0.25\textwidth]{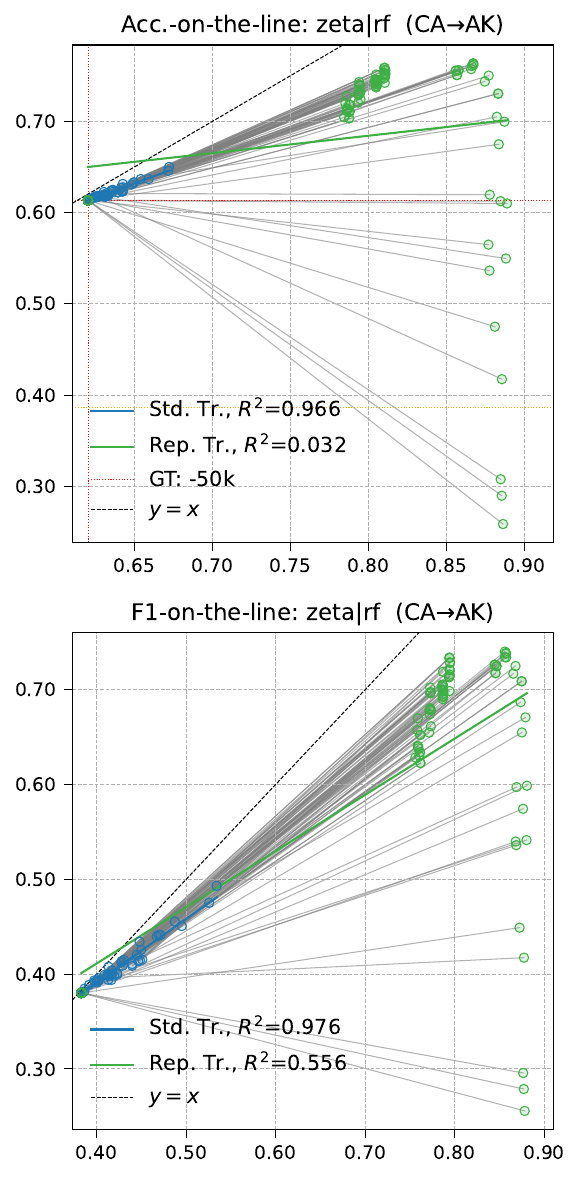}\label{fig:ak-rf-zeta}}
\caption{Pattern behaviour across different LLMs for RF.}
\label{fig:ak-rf}
\end{figure}

\begin{figure}[bp]
\centering
\subfloat[\texttt{e5}]{\includegraphics[width=0.25\textwidth]{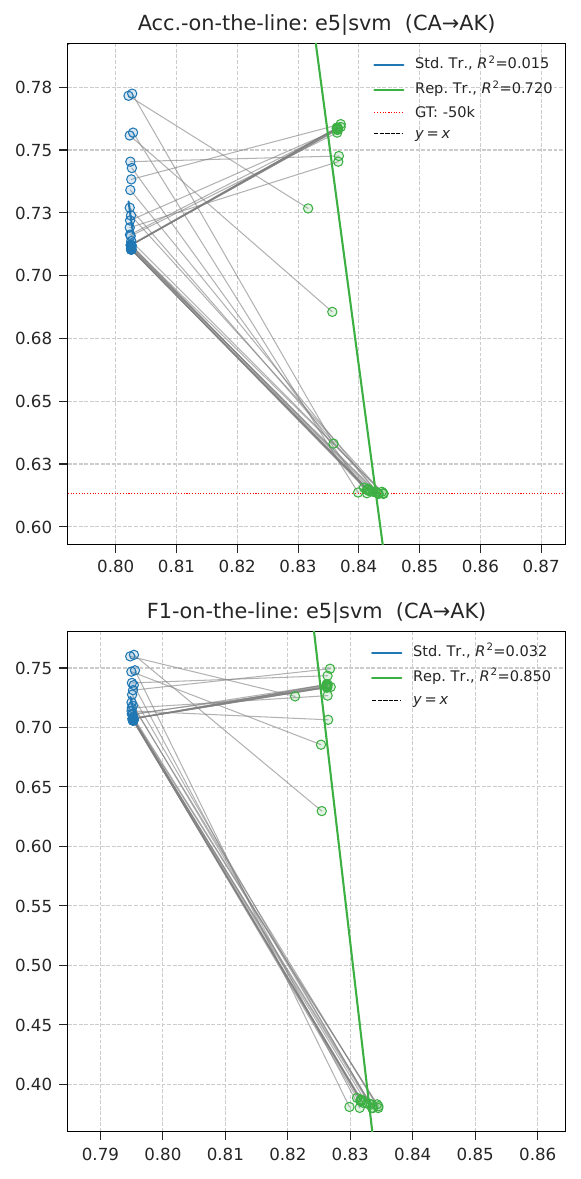}
\label{fig:ak-svm-e5}}
\subfloat[\texttt{Linq}]{\includegraphics[width=0.25\textwidth]{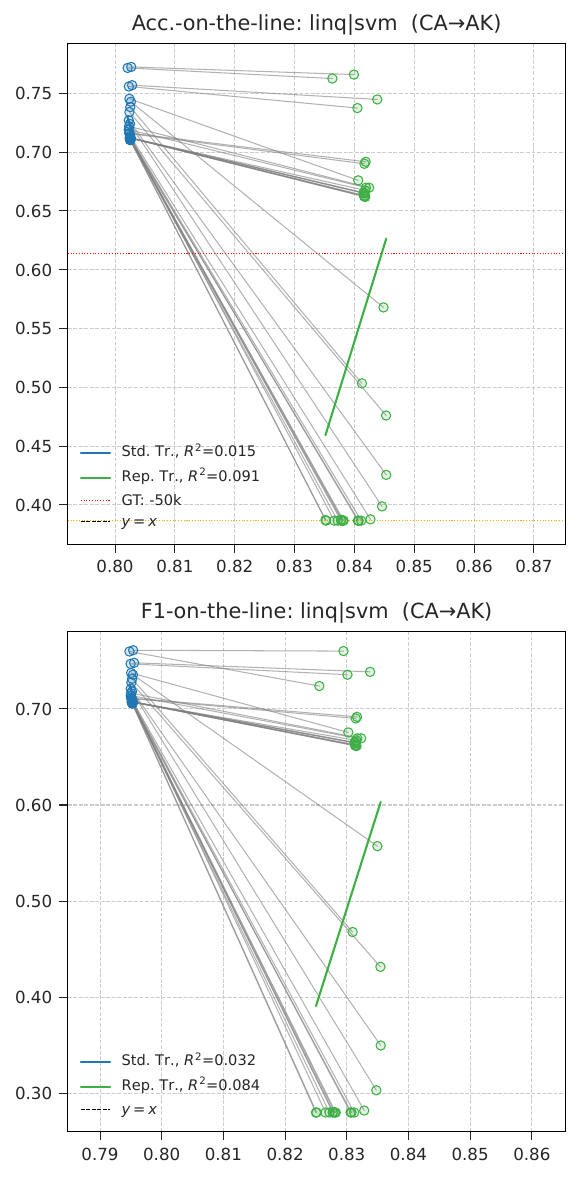}\label{fig:ak-svm-linq}}
\subfloat[\texttt{SFR}]{\includegraphics[width=0.25\textwidth]{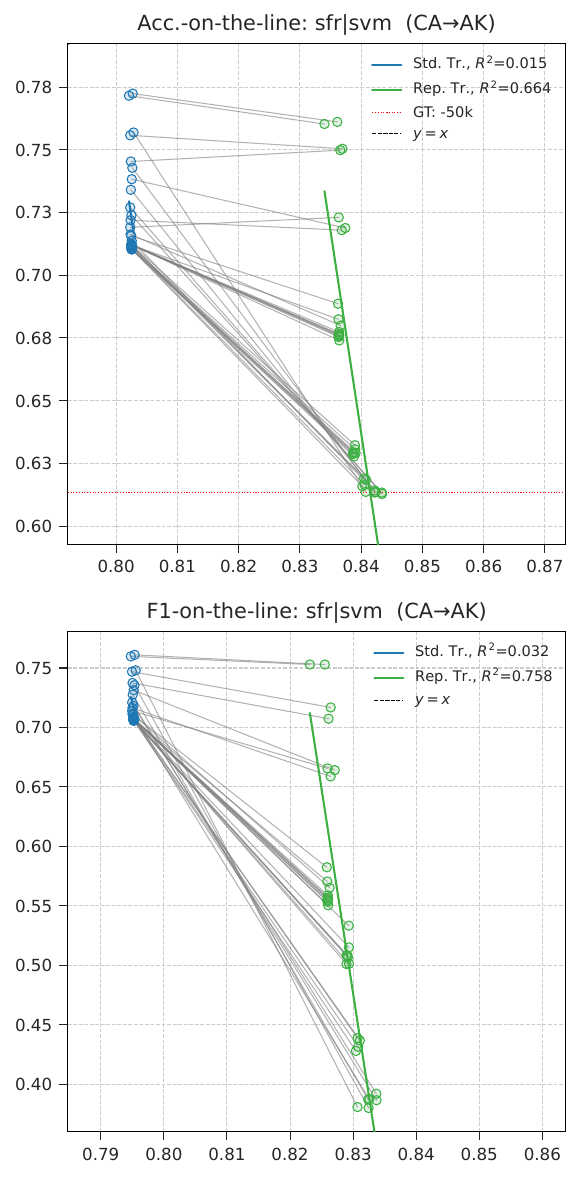}\label{fig:ak-svm-sfr}}
\subfloat[\texttt{Zeta}]{\includegraphics[width=0.25\textwidth]{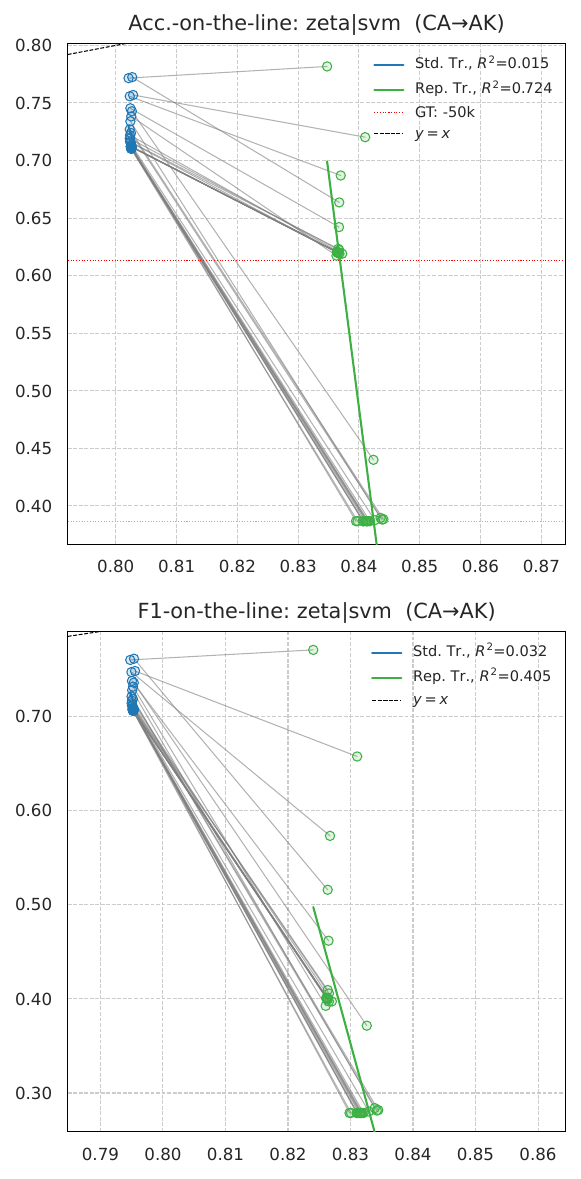}\label{fig:ak-svm-zeta}}
\caption{Pattern behaviour across different LLMs for SVM.}
\label{fig:ak-svm}
\end{figure}

\begin{figure}[bp]
\centering
\subfloat[\texttt{e5}]{\includegraphics[width=0.25\textwidth]{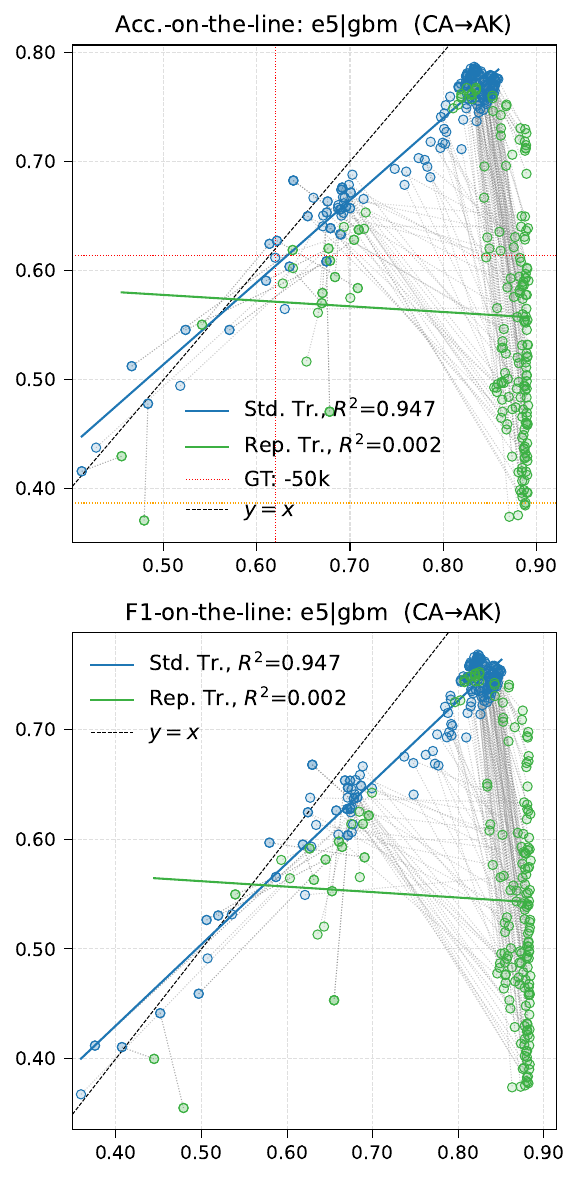}
\label{fig:ak-gbm-e5}}
\subfloat[\texttt{Linq}]{\includegraphics[width=0.25\textwidth]{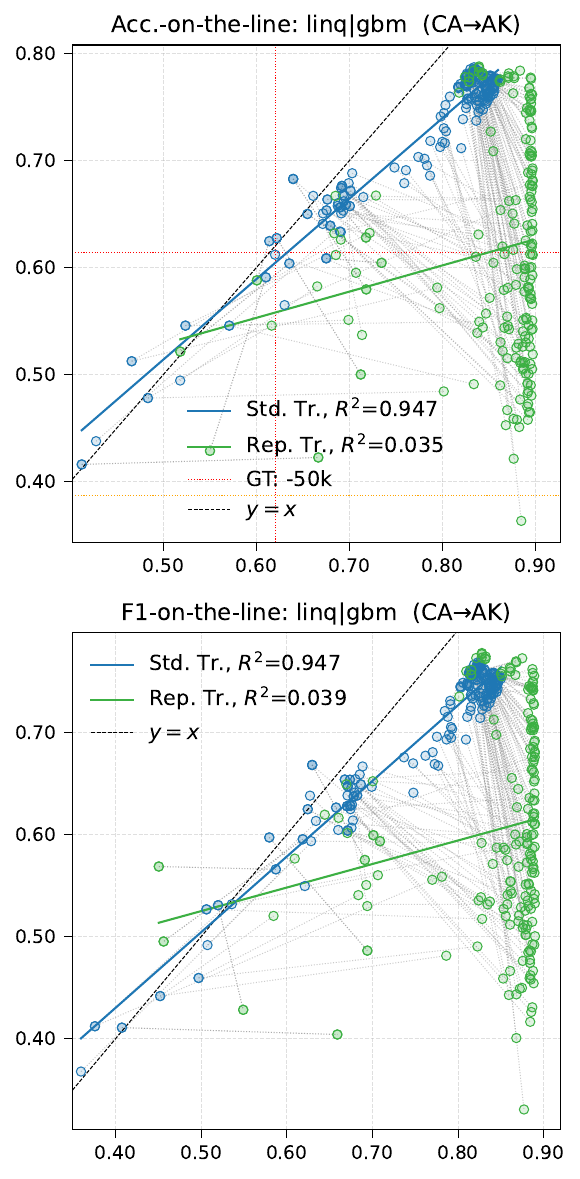}\label{fig:ak-gbm-linq}}
\subfloat[\texttt{SFR}]{\includegraphics[width=0.25\textwidth]{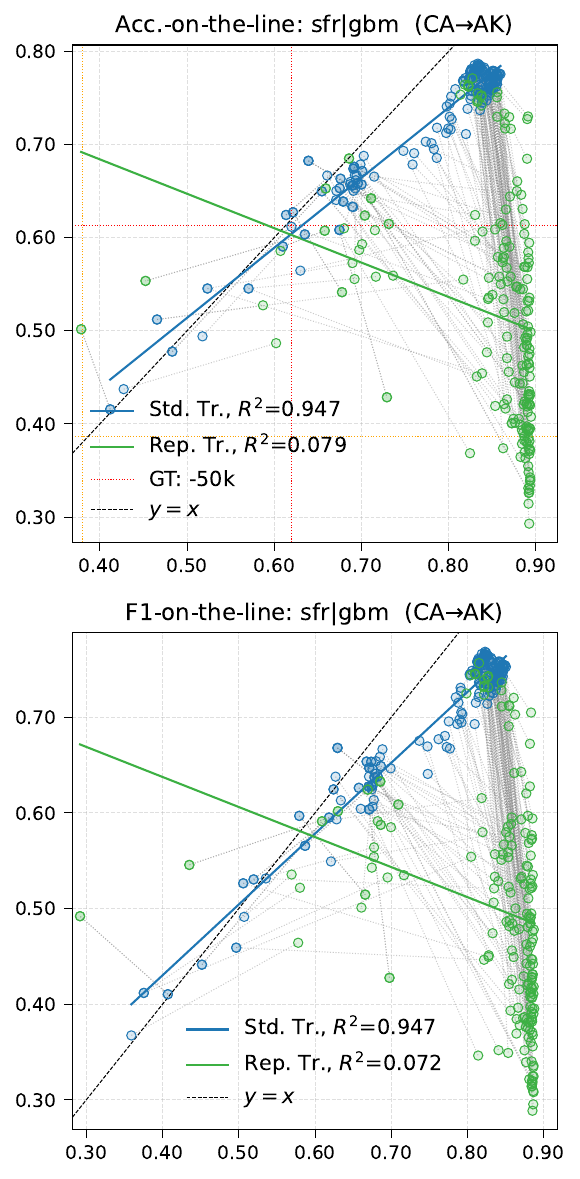}\label{fig:ak-gbm-sfr}}
\subfloat[\texttt{Zeta}]{\includegraphics[width=0.25\textwidth]{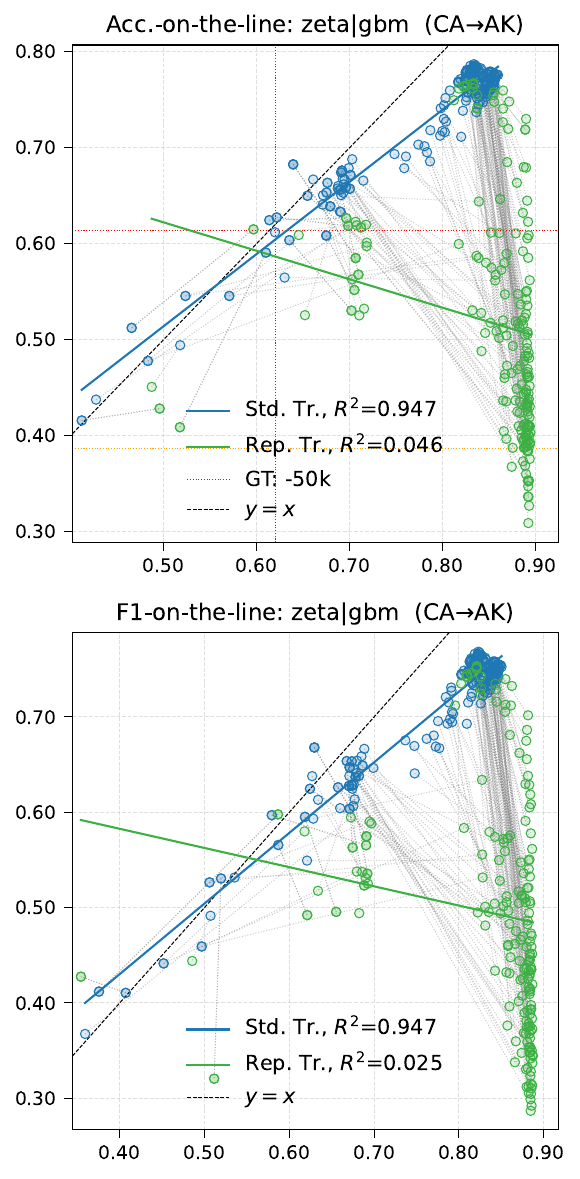}\label{fig:ak-gbm-zeta}}
\caption{Pattern behaviour across different LLMs for GBM.}
\label{fig:ak-gbm}
\end{figure}

\begin{figure}[bp]
\centering
\subfloat[\texttt{e5}]{\includegraphics[width=0.25\textwidth]{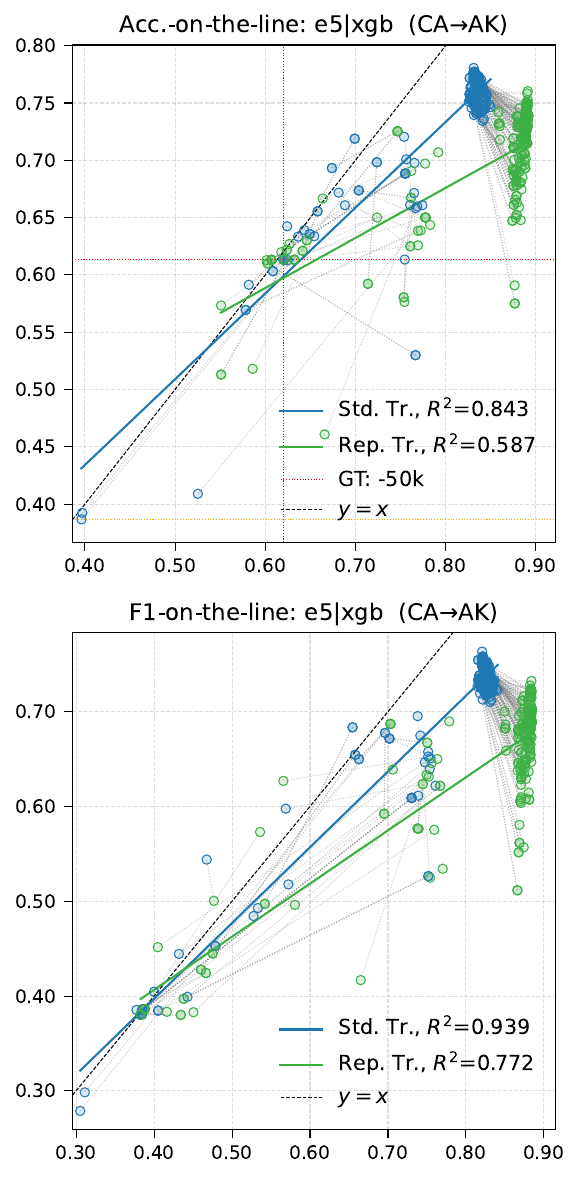}
\label{fig:ak-xgb-e5}}
\subfloat[\texttt{Linq}]{\includegraphics[width=0.25\textwidth]{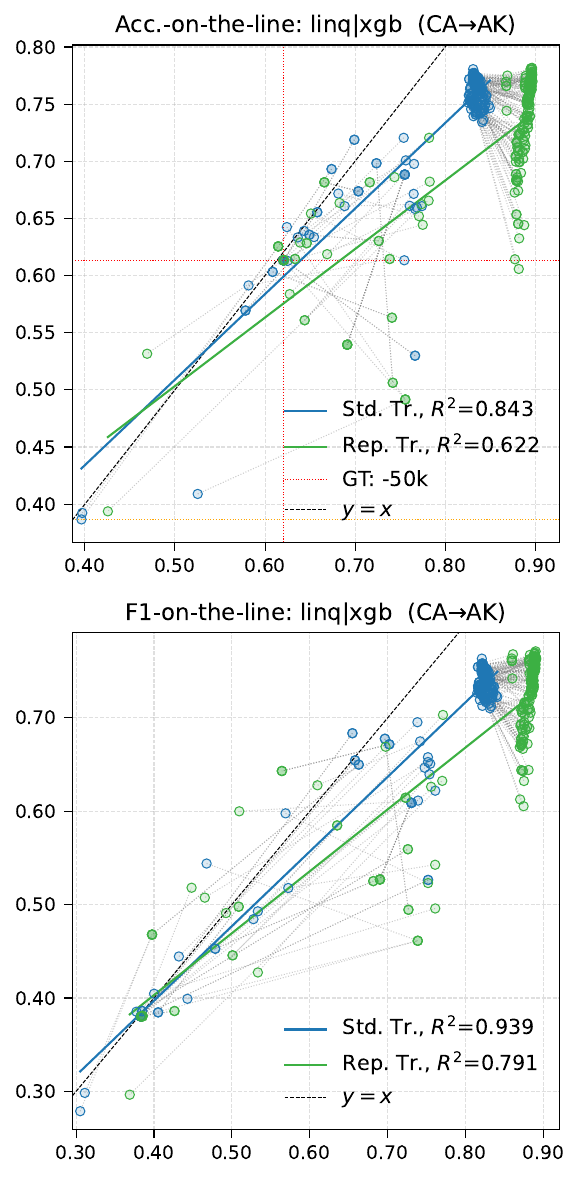}\label{fig:ak-xgb-linq}}
\subfloat[\texttt{SFR}]{\includegraphics[width=0.25\textwidth]{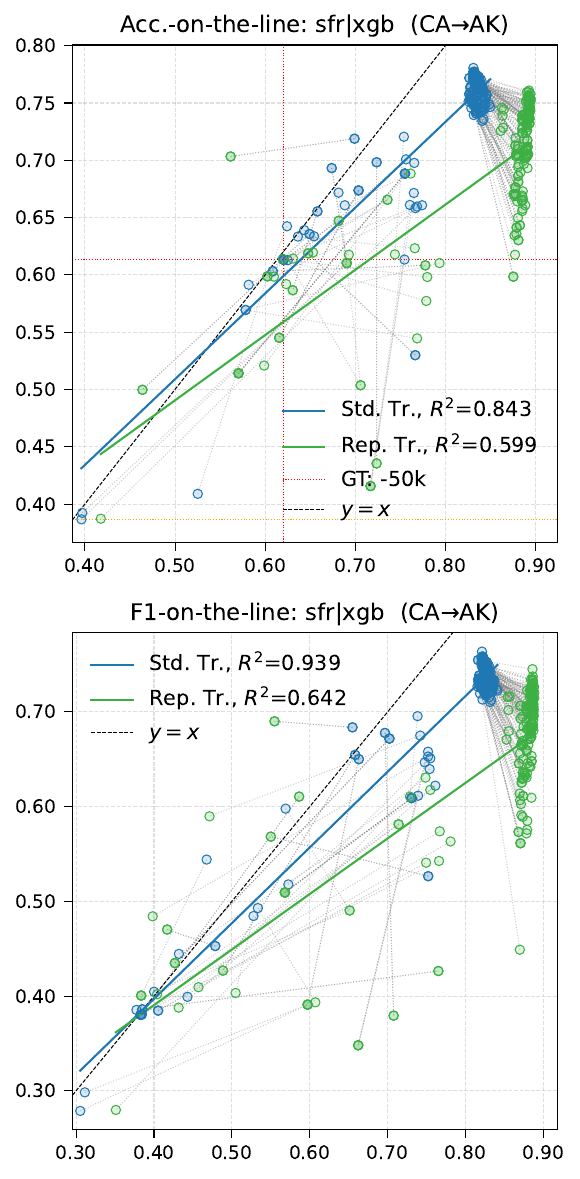}\label{fig:ak-xgb-sfr}}
\subfloat[\texttt{Zeta}]{\includegraphics[width=0.25\textwidth]{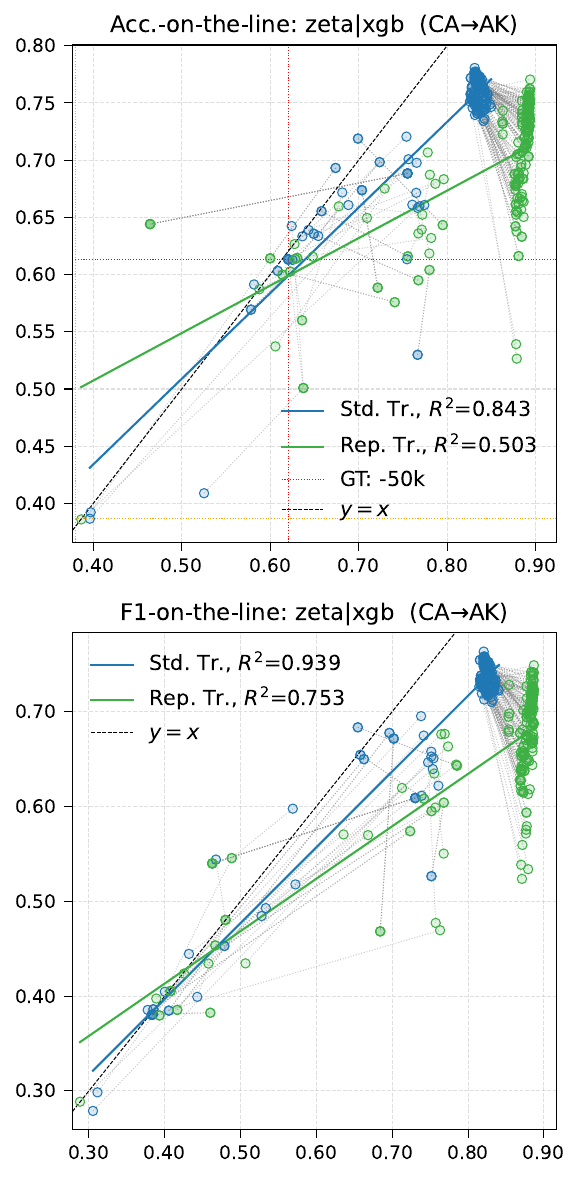}\label{fig:ak-xgb-zeta}}
\caption{Pattern behaviour across different LLMs for XGB.}
\label{fig:ak-xgb}
\end{figure}

\clearpage

\paragraph{Target State: Arkansas} 

\begin{figure}[bp!]
\centering
\subfloat[\texttt{e5}]{\includegraphics[width=0.25\textwidth]{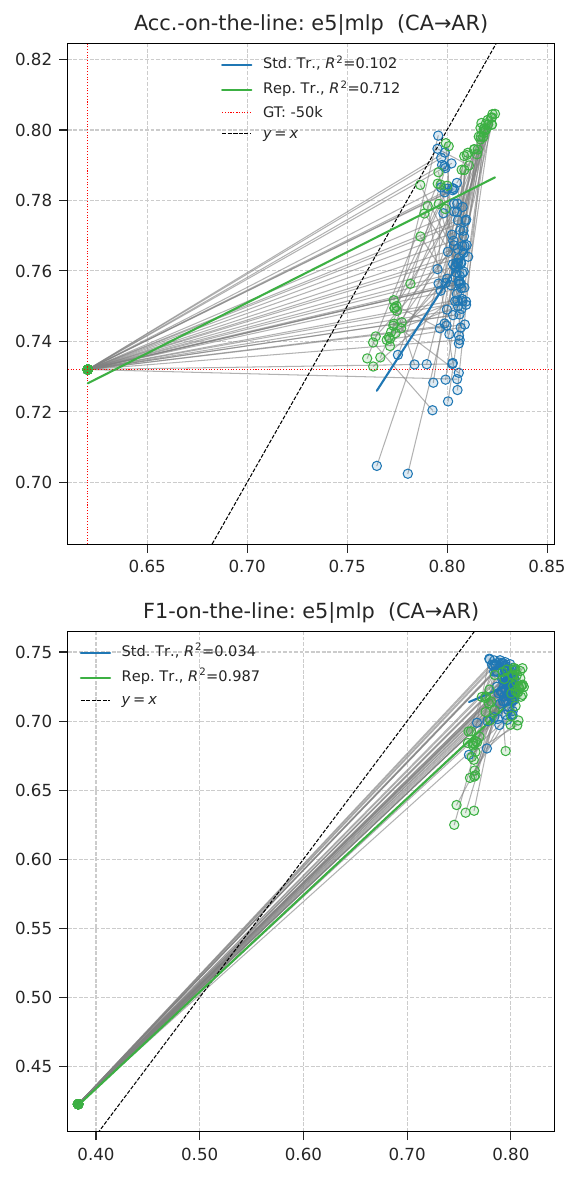}
\label{fig:ar-mlp-e5}}
\subfloat[\texttt{Linq}]{\includegraphics[width=0.25\textwidth]{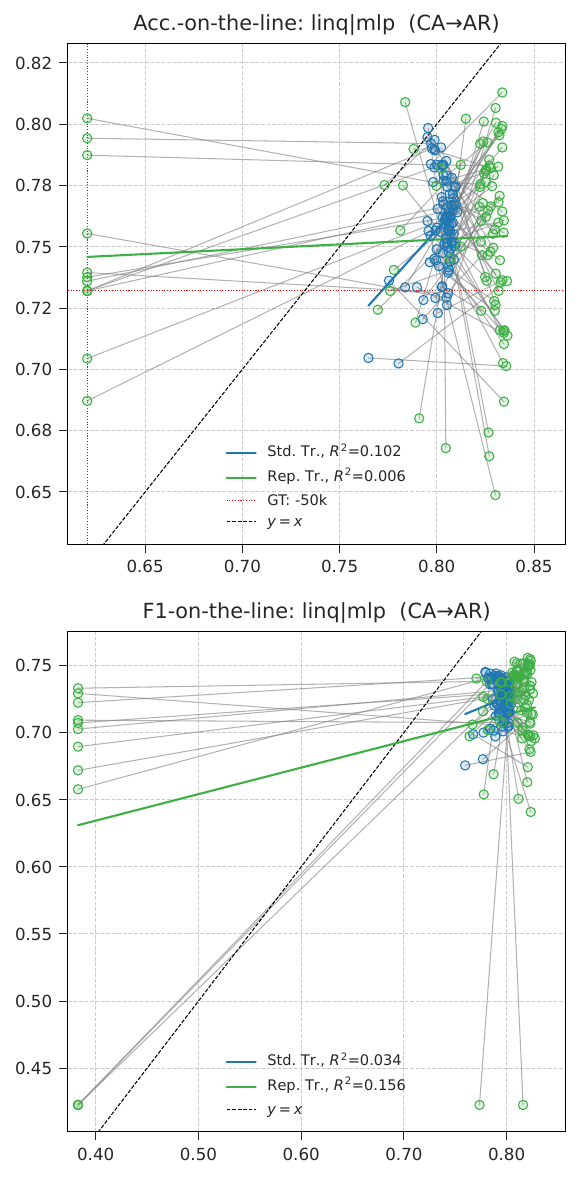}\label{fig:ar-mlp-linq}}
\subfloat[\texttt{SFR}]{\includegraphics[width=0.25\textwidth]{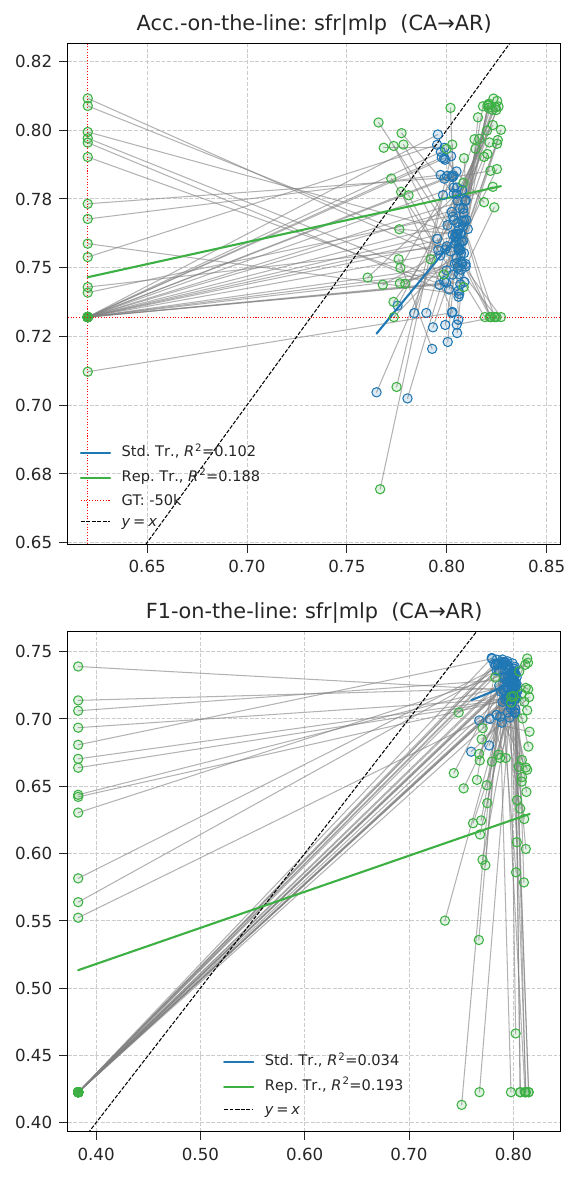}\label{fig:ar-mlp-sfr}}
\subfloat[\texttt{Zeta}]{\includegraphics[width=0.25\textwidth]{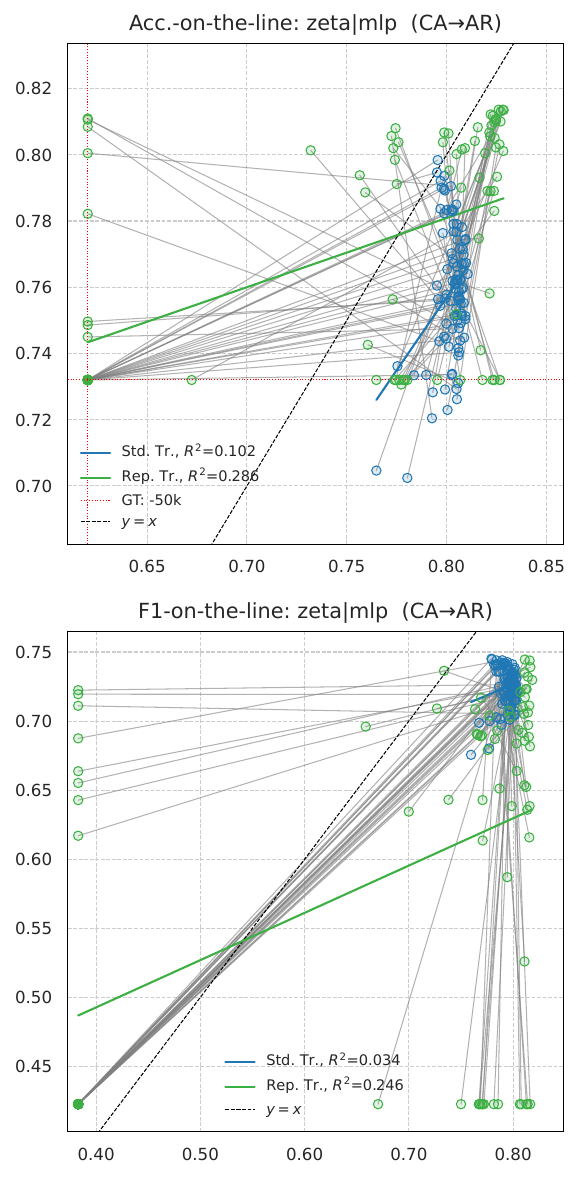}\label{fig:ar-mlp-zeta}}
\caption{Pattern behaviour across different LLMs for MLP.}
\label{fig:ar-mlp}
\end{figure}

\begin{figure}[bp!]
\centering
\subfloat[\texttt{e5}]{\includegraphics[width=0.25\textwidth]{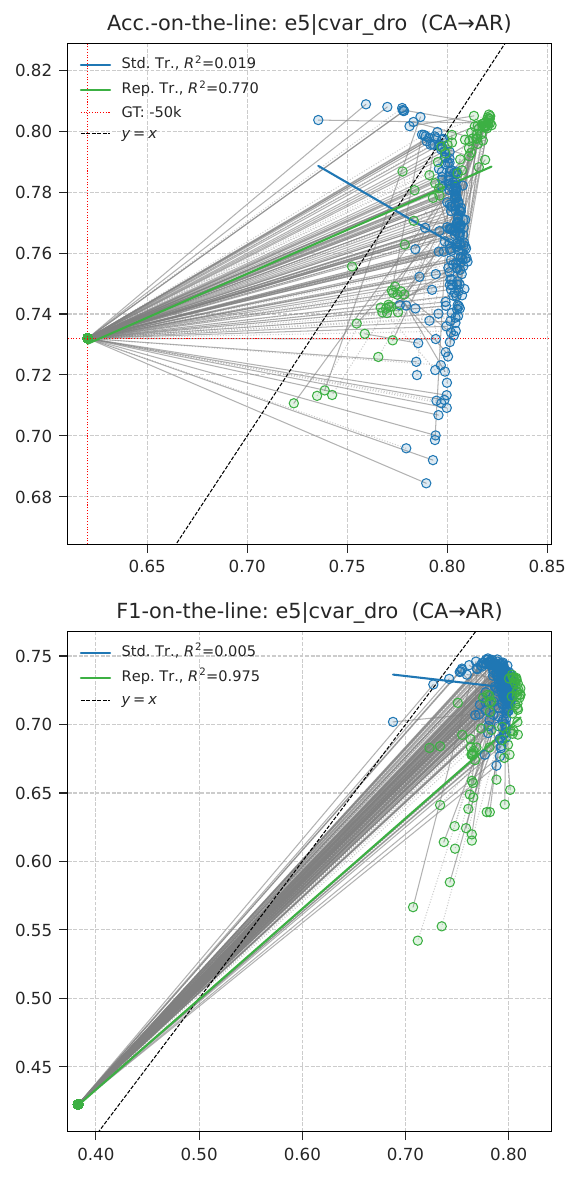}
\label{fig:ar-cvar_dro-e5}}
\subfloat[\texttt{Linq}]{\includegraphics[width=0.25\textwidth]{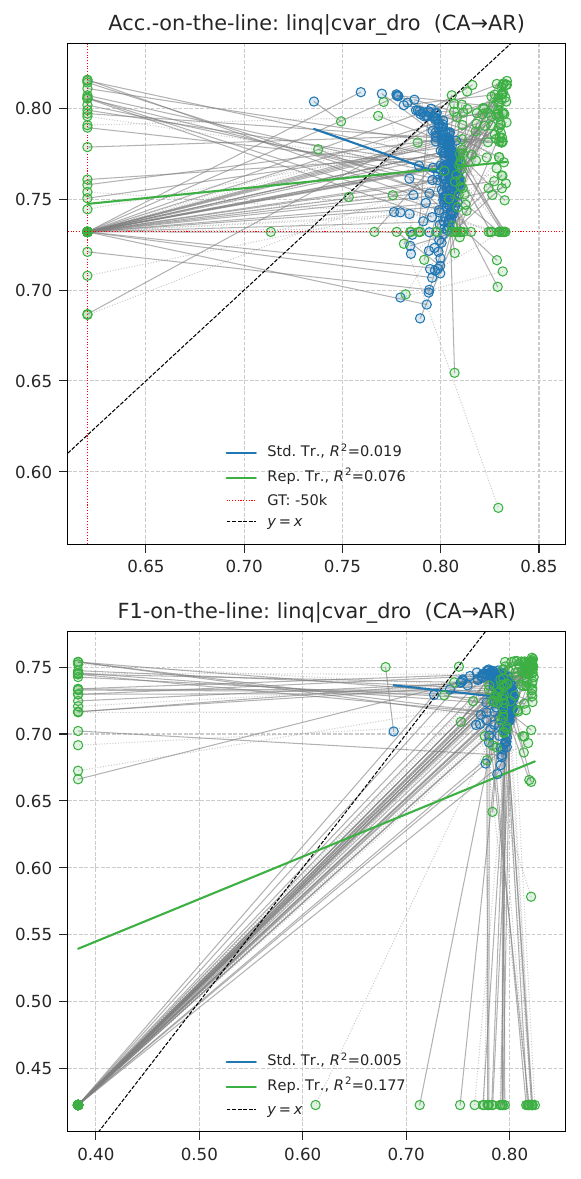}\label{fig:ar-cvar_dro-linq}}
\subfloat[\texttt{SFR}]{\includegraphics[width=0.25\textwidth]{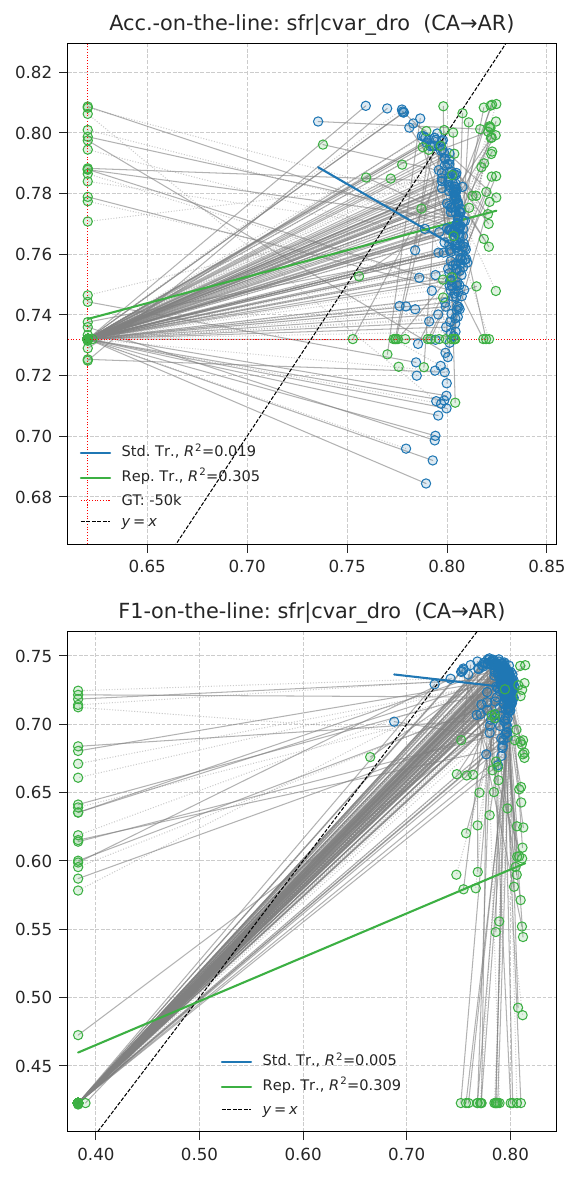}\label{fig:ar-cvar_dro-sfr}}
\subfloat[\texttt{Zeta}]{\includegraphics[width=0.25\textwidth]{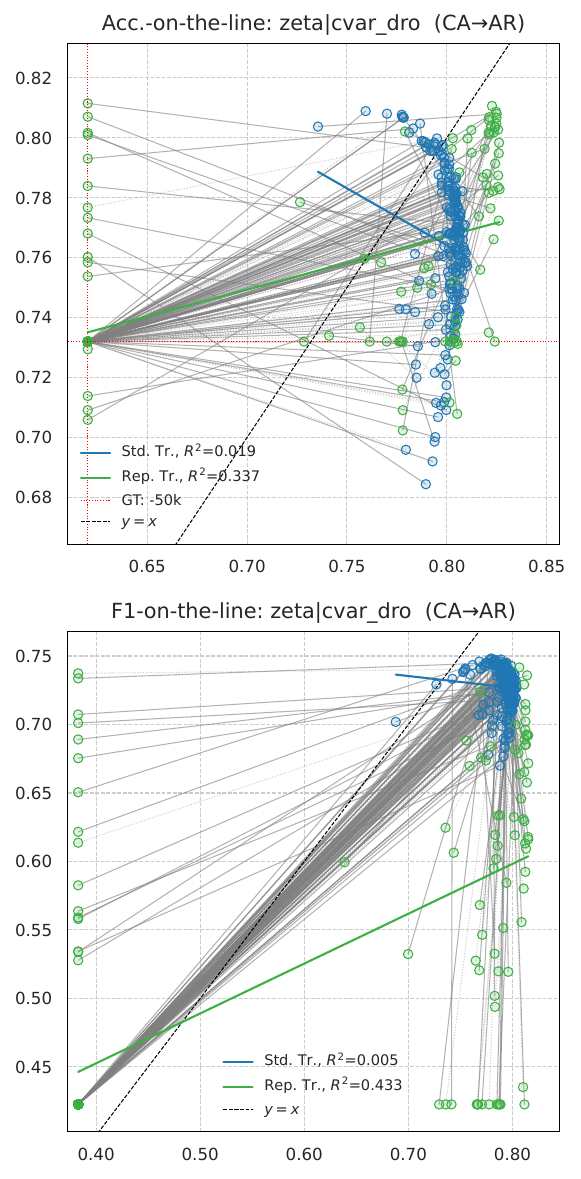}\label{fig:ar-cvar_dro-zeta}}
\caption{Pattern behaviour across different LLMs for CVaR-DRO.}
\label{fig:ar-cvar_dro}
\end{figure}

\begin{figure}[bp]
\centering
\subfloat[\texttt{e5}]{\includegraphics[width=0.25\textwidth]{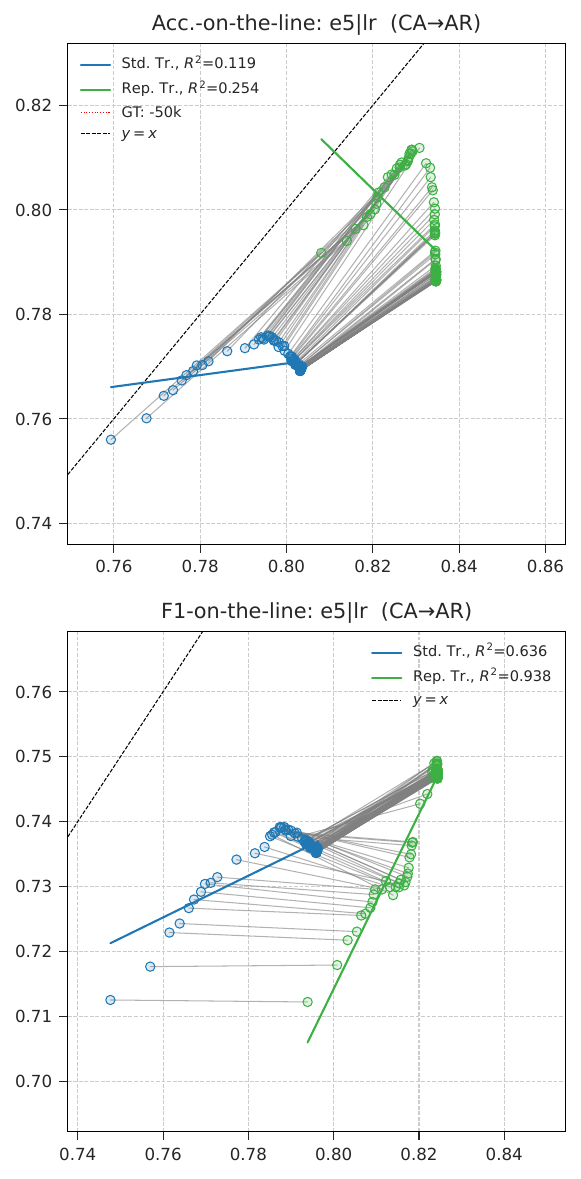}
\label{fig:ar-lr-e5}}
\subfloat[\texttt{Linq}]{\includegraphics[width=0.25\textwidth]{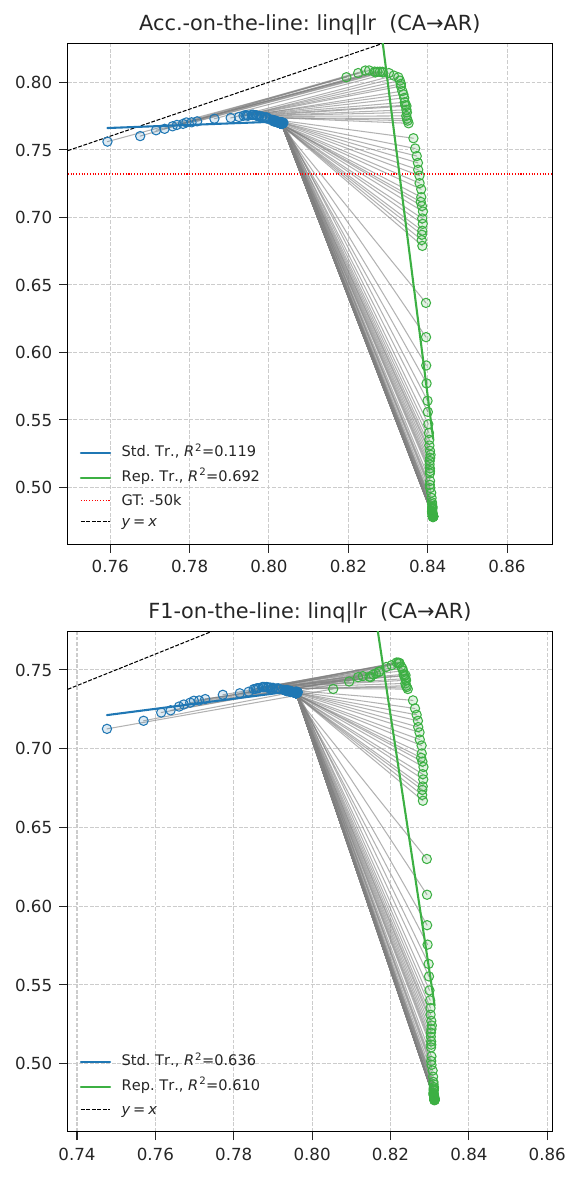}\label{fig:ar-lr-linq}}
\subfloat[\texttt{SFR}]{\includegraphics[width=0.25\textwidth]{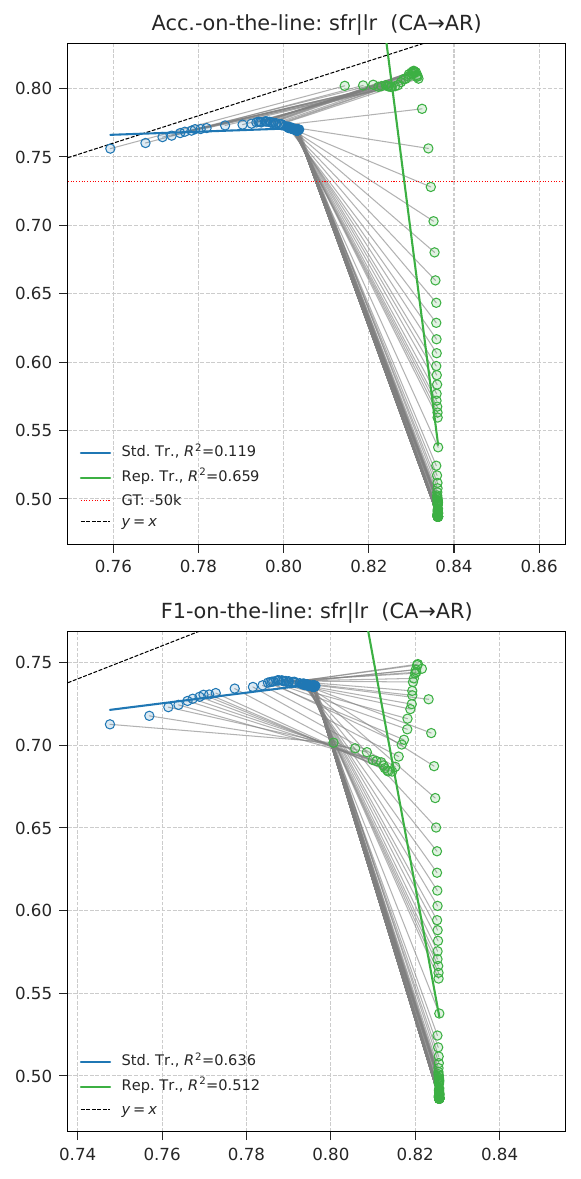}\label{fig:ar-lr-sfr}}
\subfloat[\texttt{Zeta}]{\includegraphics[width=0.25\textwidth]{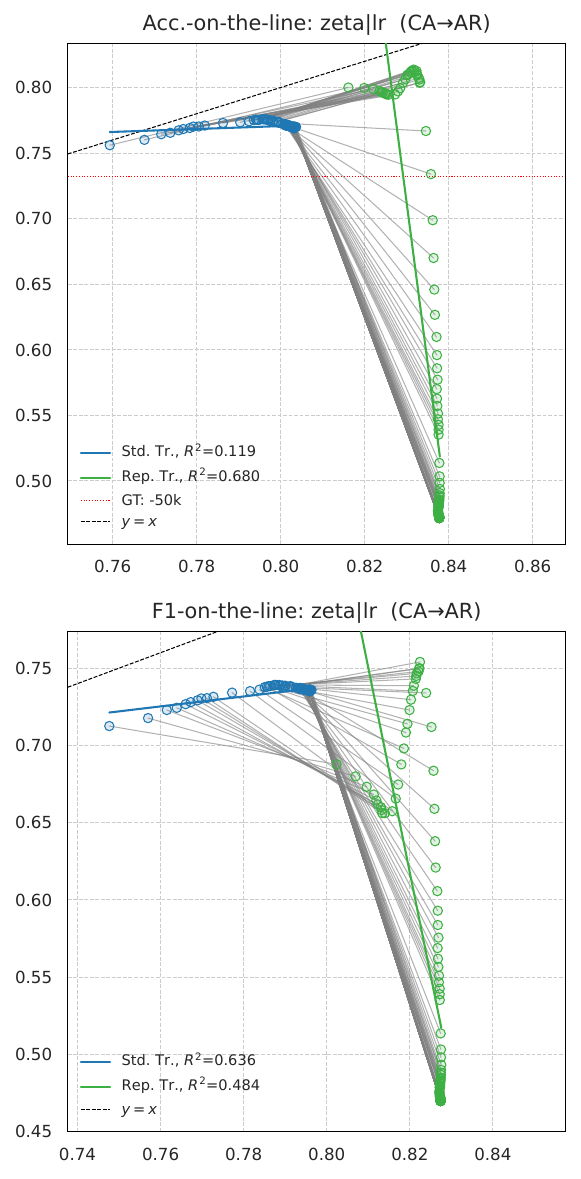}\label{fig:ar-lr-zeta}}
\caption{Pattern behaviour across different LLMs for LR.}
\label{fig:ar-lr}
\end{figure}

\begin{figure}[bp]
\centering
\subfloat[\texttt{e5}]{\includegraphics[width=0.25\textwidth]{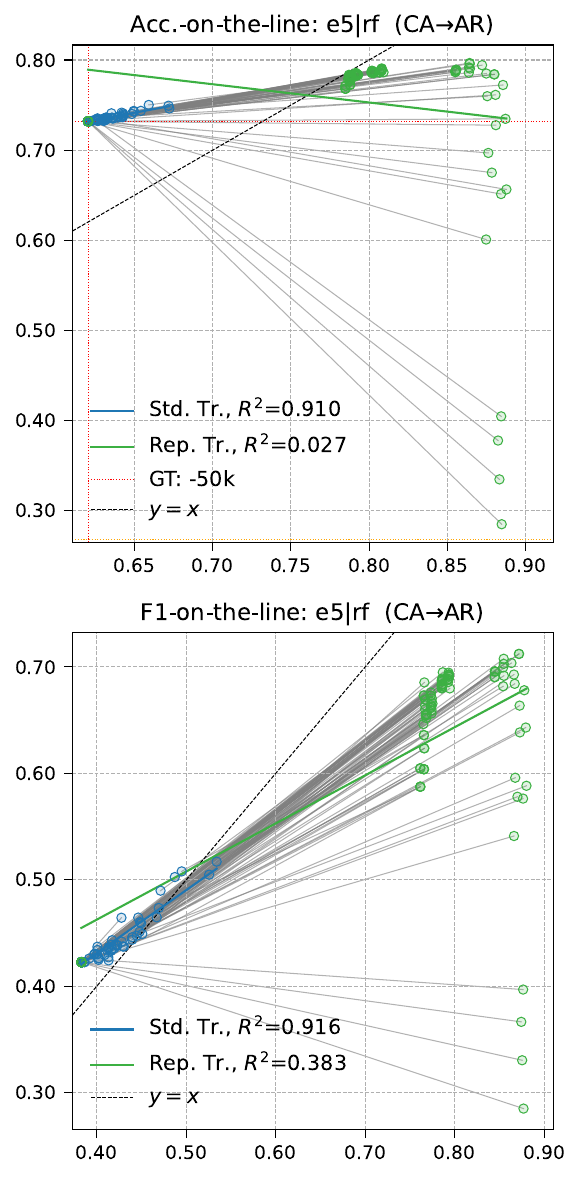}
\label{fig:ar-rf-e5}}
\subfloat[\texttt{Linq}]{\includegraphics[width=0.25\textwidth]{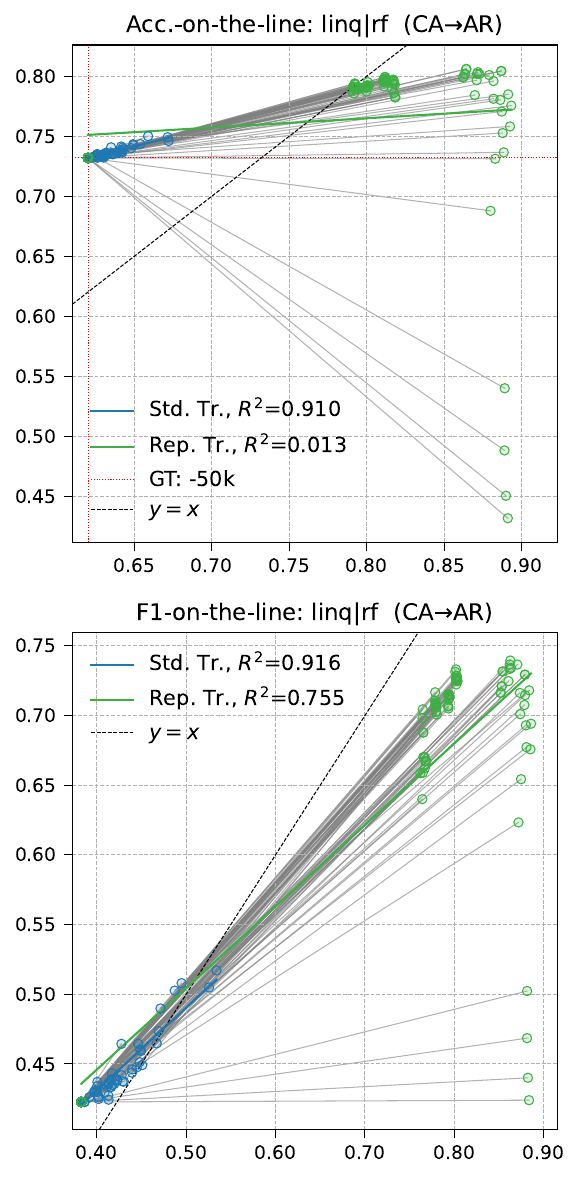}\label{fig:ar-rf-linq}}
\subfloat[\texttt{SFR}]{\includegraphics[width=0.25\textwidth]{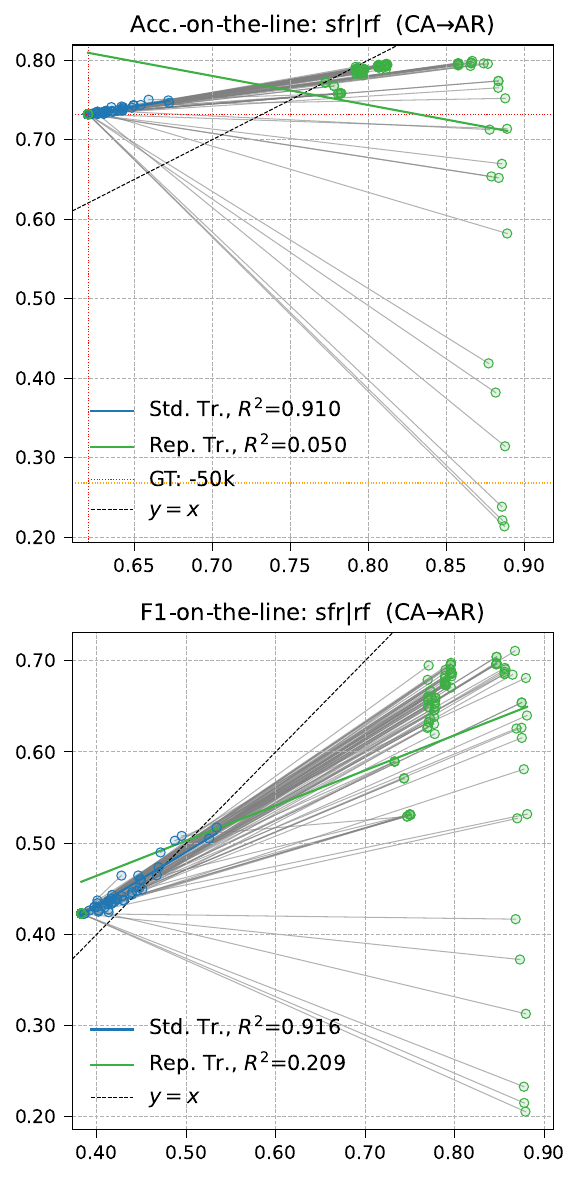}\label{fig:ar-rf-sfr}}
\subfloat[\texttt{Zeta}]{\includegraphics[width=0.25\textwidth]{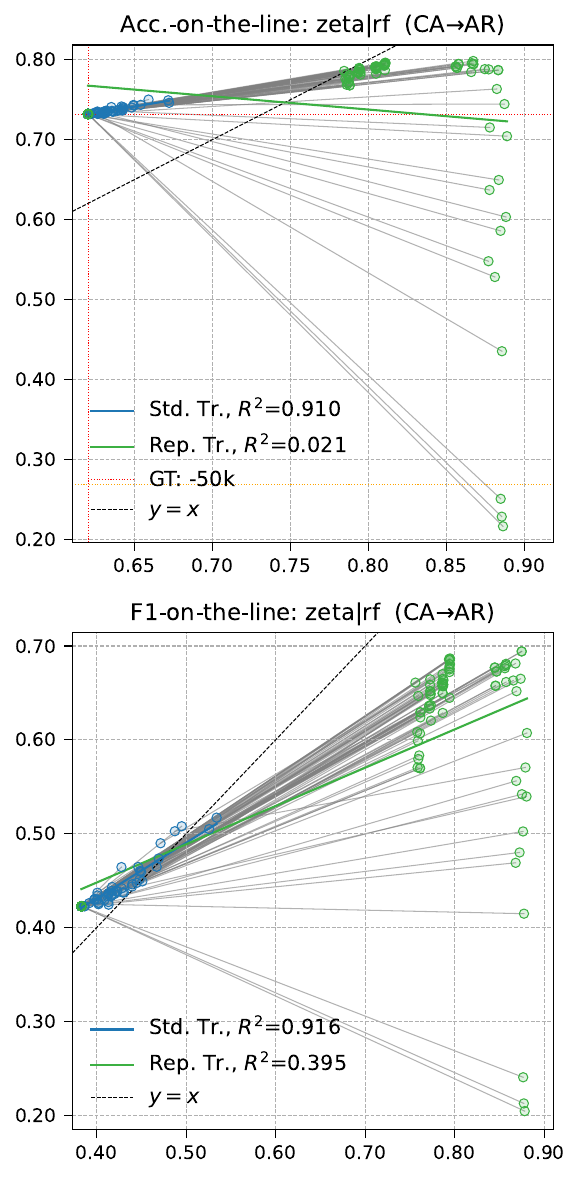}\label{fig:ar-rf-zeta}}
\caption{Pattern behaviour across different LLMs for RF.}
\label{fig:ar-rf}
\end{figure}

\begin{figure}[bp]
\centering
\subfloat[\texttt{e5}]{\includegraphics[width=0.25\textwidth]{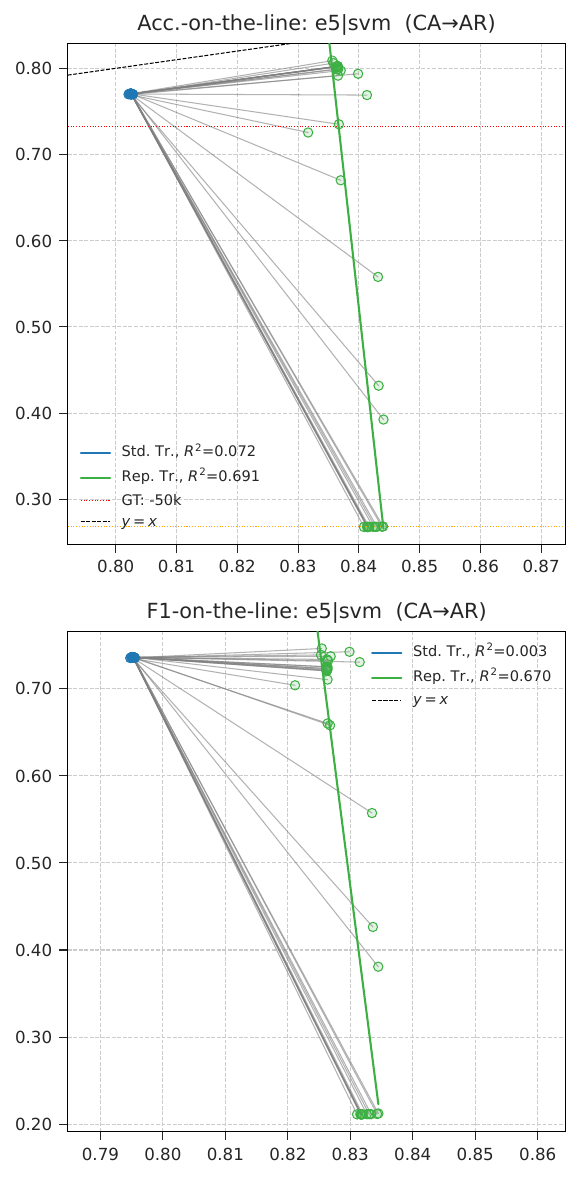}
\label{fig:ar-svm-e5}}
\subfloat[\texttt{Linq}]{\includegraphics[width=0.25\textwidth]{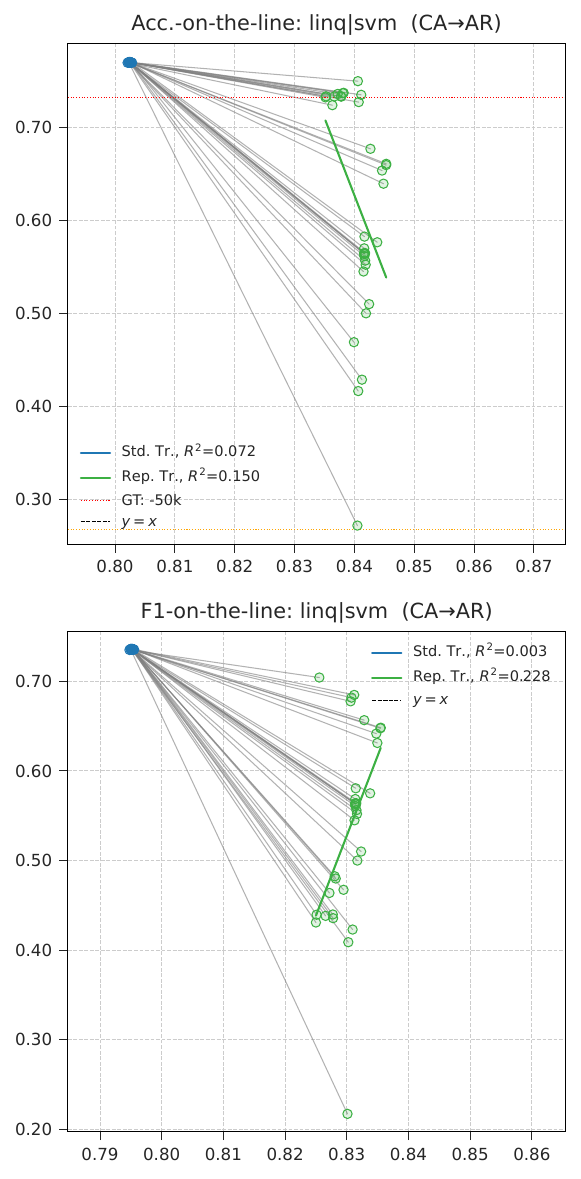}\label{fig:ar-svm-linq}}
\subfloat[\texttt{SFR}]{\includegraphics[width=0.25\textwidth]{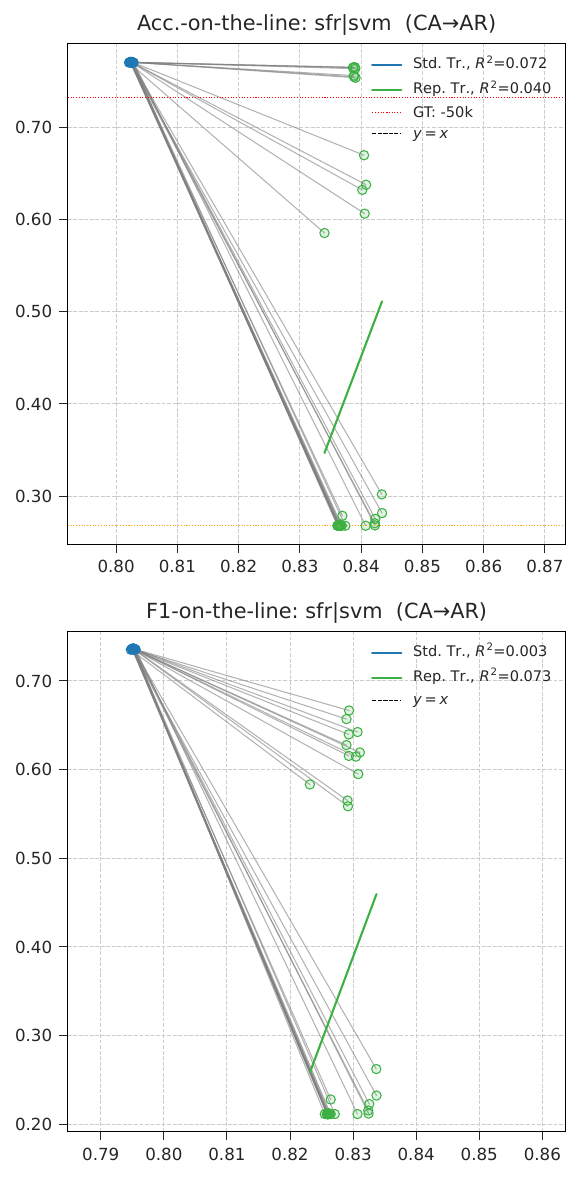}\label{fig:ar-svm-sfr}}
\subfloat[\texttt{Zeta}]{\includegraphics[width=0.25\textwidth]{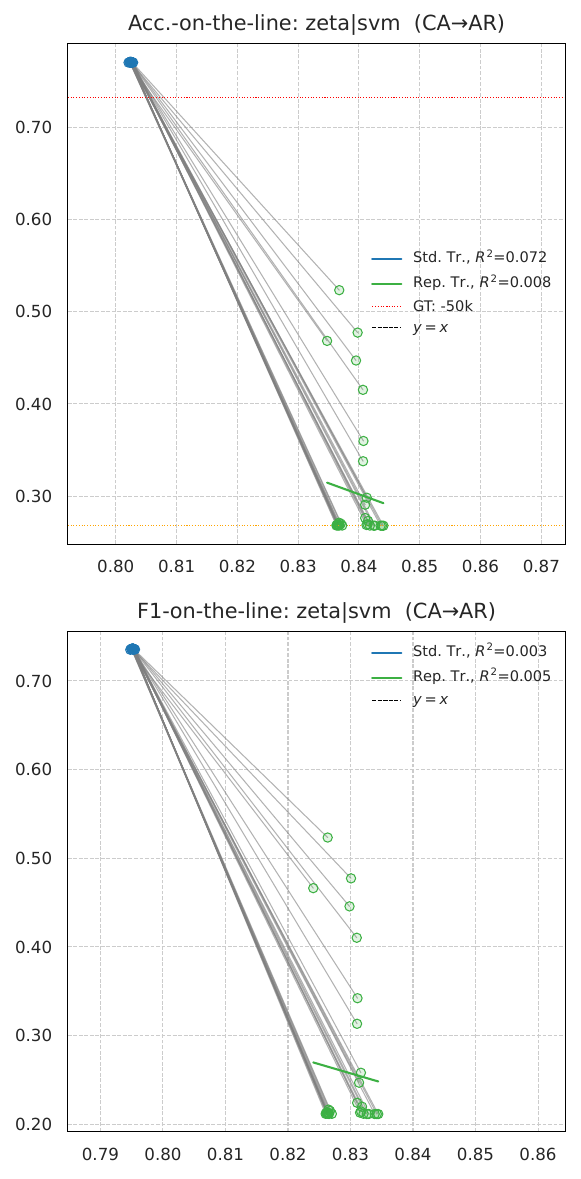}\label{fig:ar-svm-zeta}}
\caption{Pattern behaviour across different LLMs for SVM.}
\label{fig:ar-svm}
\end{figure}

\begin{figure}[bp]
\centering
\subfloat[\texttt{e5}]{\includegraphics[width=0.25\textwidth]{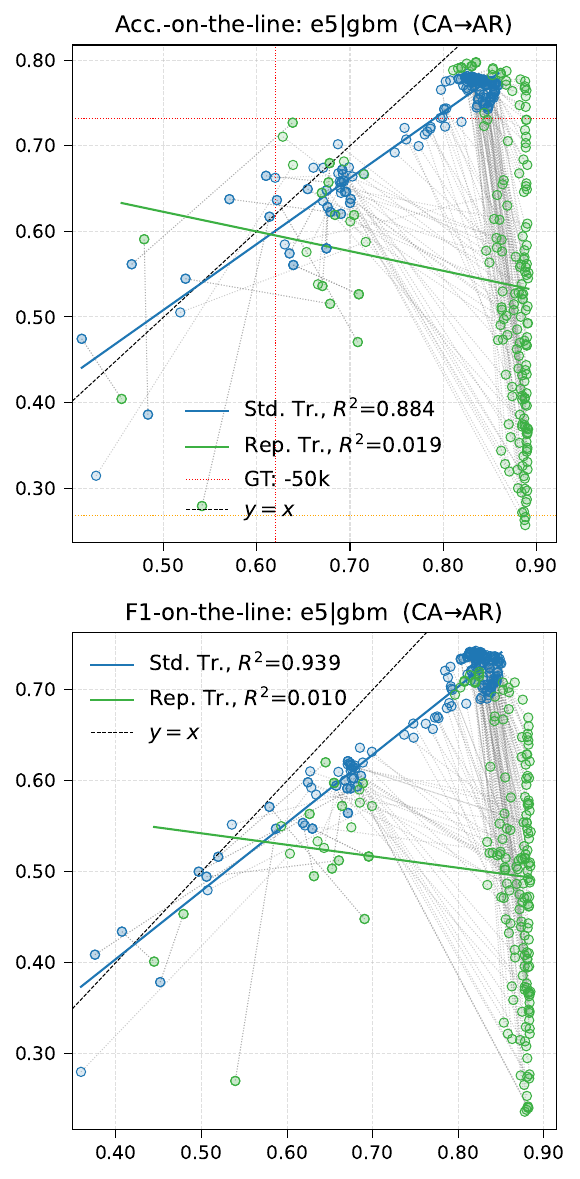}
\label{fig:ar-gbm-e5}}
\subfloat[\texttt{Linq}]{\includegraphics[width=0.25\textwidth]{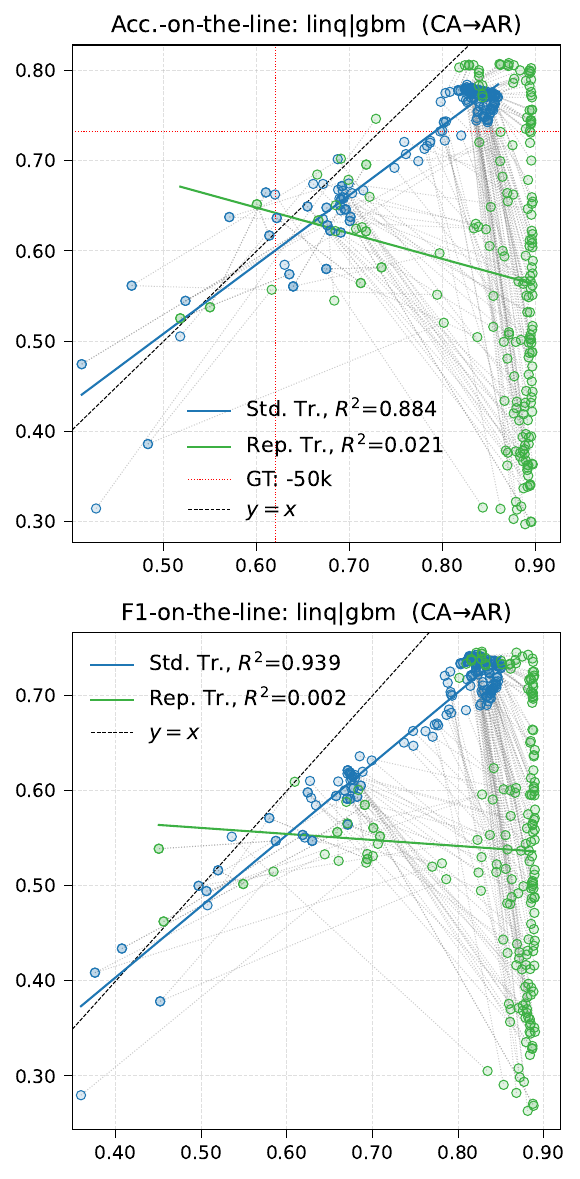}\label{fig:ar-gbm-linq}}
\subfloat[\texttt{SFR}]{\includegraphics[width=0.25\textwidth]{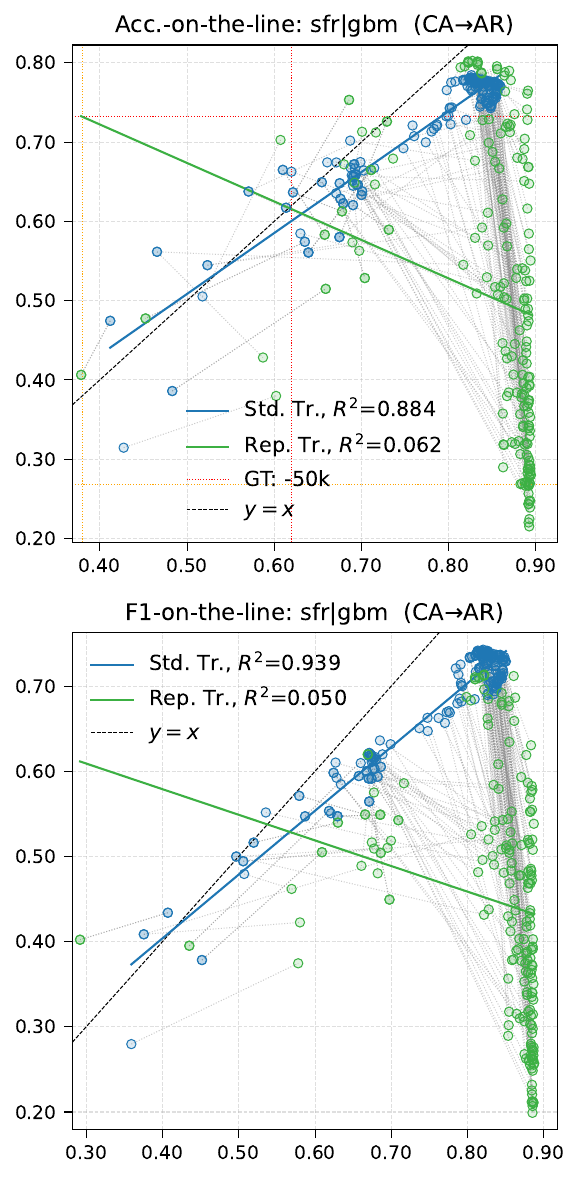}\label{fig:ar-gbm-sfr}}
\subfloat[\texttt{Zeta}]{\includegraphics[width=0.25\textwidth]{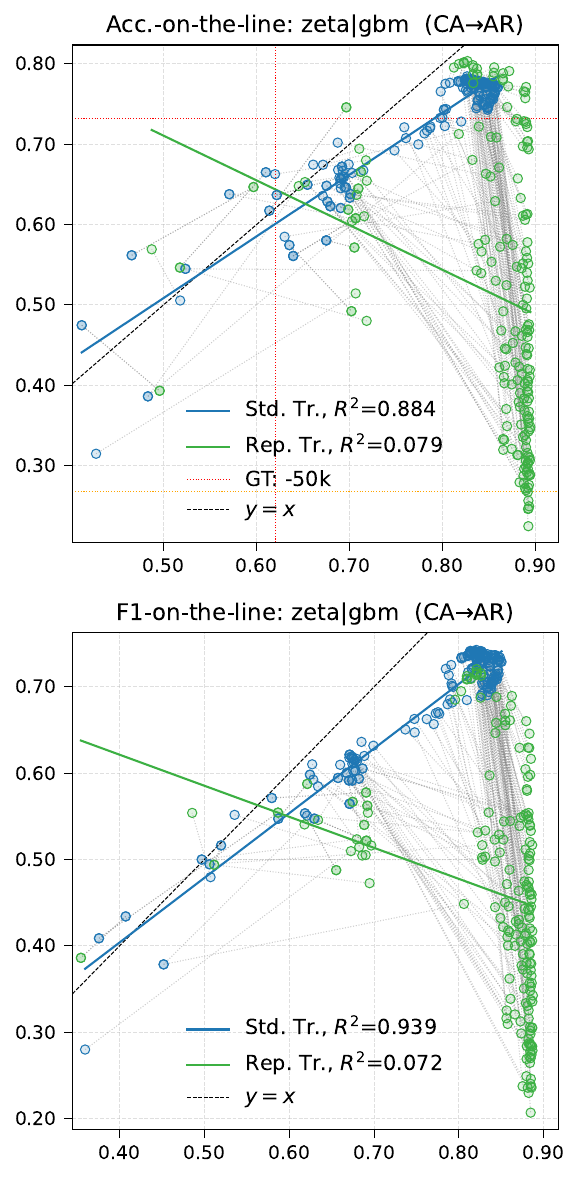}\label{fig:ar-gbm-zeta}}
\caption{Pattern behaviour across different LLMs for GBM.}
\label{fig:ar-gbm}
\end{figure}

\begin{figure}[bp]
\centering
\subfloat[\texttt{e5}]{\includegraphics[width=0.25\textwidth]{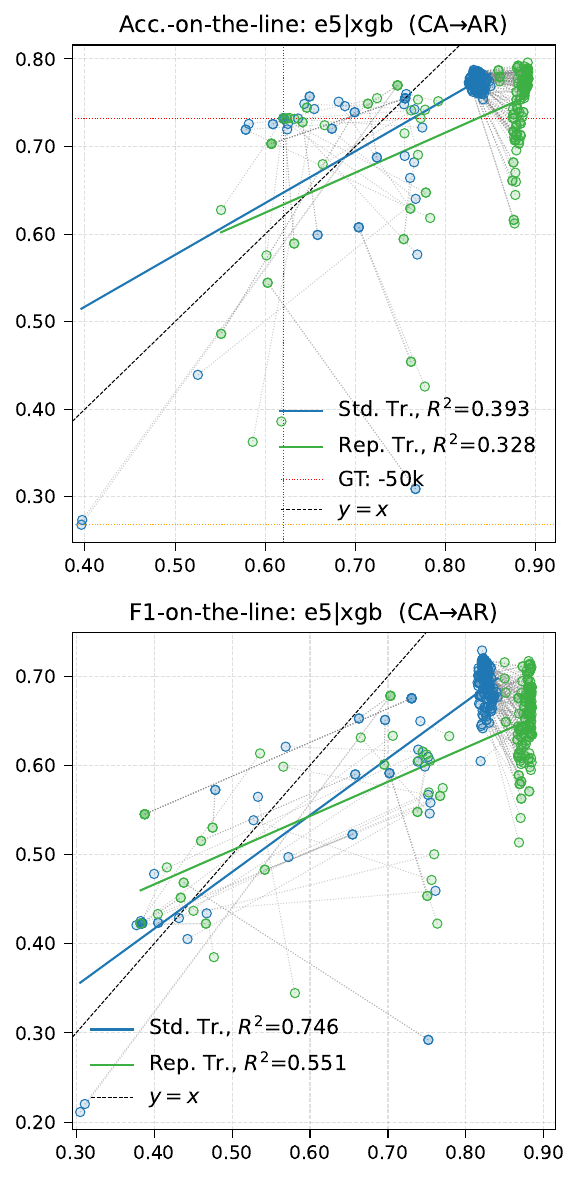}
\label{fig:ar-xgb-e5}}
\subfloat[\texttt{Linq}]{\includegraphics[width=0.25\textwidth]{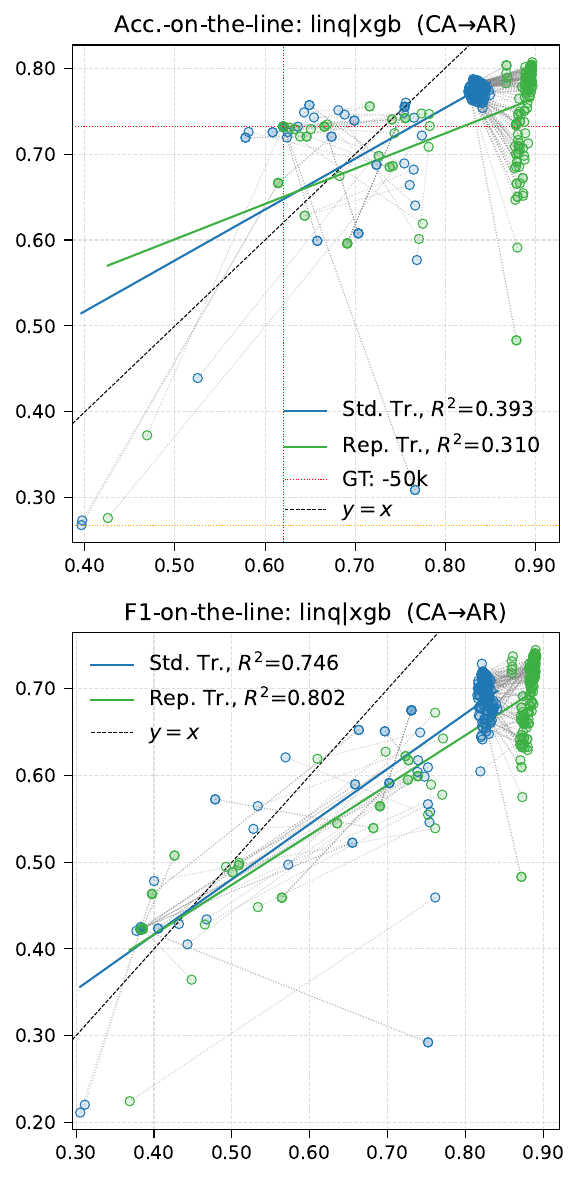}\label{fig:ar-xgb-linq}}
\subfloat[\texttt{SFR}]{\includegraphics[width=0.25\textwidth]{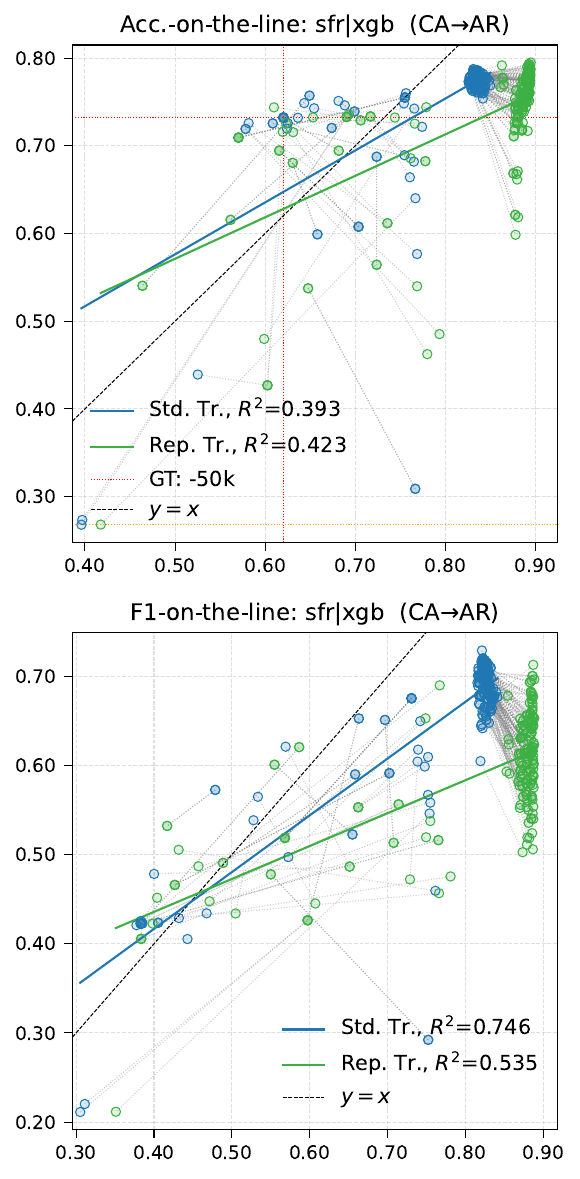}\label{fig:ar-xgb-sfr}}
\subfloat[\texttt{Zeta}]{\includegraphics[width=0.25\textwidth]{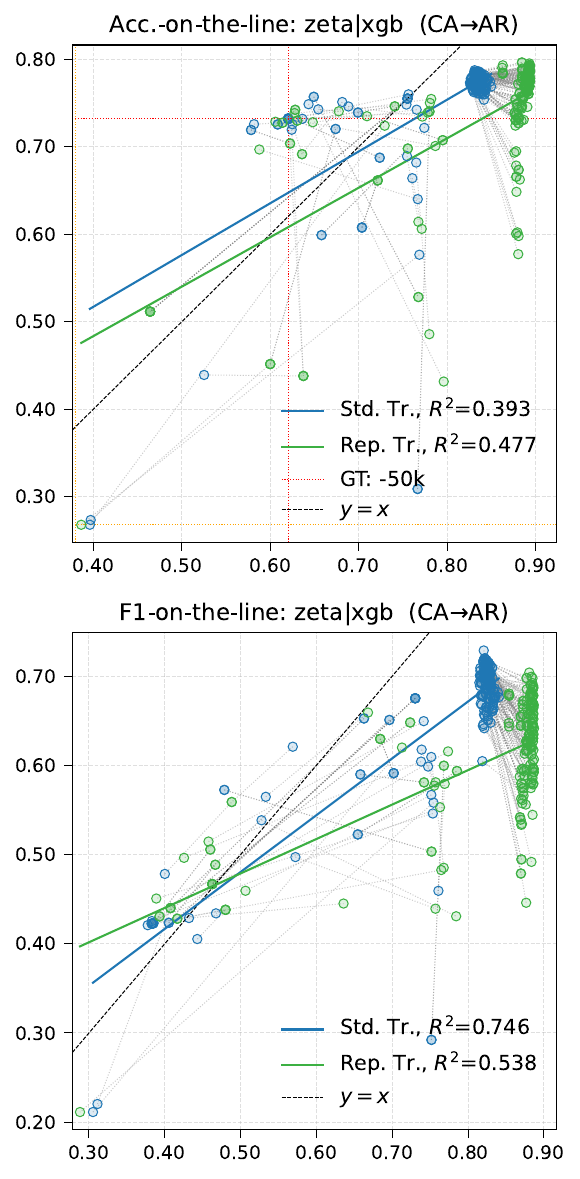}\label{fig:ar-xgb-zeta}}
\caption{Pattern behaviour across different LLMs for XGB.}
\label{fig:ar-xgb}
\end{figure}